\documentclass[sn-mathphys]{sn-jnl}

\usepackage{amsmath,amssymb,amsfonts}
\usepackage{graphicx}
\usepackage{textcomp}
\usepackage{wrapfig}
\usepackage{caption}
\usepackage{subcaption}
\usepackage{float}
\usepackage{multirow}
\usepackage{ragged2e}
\usepackage{bbm}
\usepackage{amssymb}
\usepackage{color}
\usepackage{enumerate}
\usepackage{amsmath}
\usepackage{amsfonts}
\usepackage{booktabs}

\jyear{2022}%

\theoremstyle{thmstyleone}%
\theoremstyle{thmstyletwo}%

\theoremstyle{thmstylethree}%

\begin{document}

\title[CenterLoc3D: Monocular 3D Vehicle Localization Network for Roadside Surveillance Cameras]{CenterLoc3D: Monocular 3D Vehicle Localization Network for Roadside Surveillance Cameras}


\author[1]{\fnm{Xinyao} \sur{Tang}}\email{andy19966212@126.com}

\author*[1]{\fnm{Wei} \sur{Wang}}\email{wangwei\_211@chd.edu.cn}

\author[1]{\fnm{Huansheng} \sur{Song}}\email{hshsong@chd.edu.cn}

\author[1]{\fnm{Chunhui} \sur{Zhao}}\email{zch9426@qq.com}
%


\abstract{Monocular 3D vehicle localization is an important task for vehicle behaviour analysis, traffic flow parameter estimation and autonomous driving in Intelligent Transportation System (ITS) and Cooperative Vehicle Infrastructure System (CVIS), which is usually achieved by monocular 3D vehicle detection. However, monocular cameras cannot obtain depth information directly due to the inherent imaging mechanism, resulting in more challenging monocular 3D tasks. Currently, most of the monocular 3D vehicle detection methods still rely on 2D detectors and additional geometric constraint modules to recover 3D vehicle information, which reduces the efficiency. At the same time, most of the research is based on datasets of onboard scenes, instead of roadside perspective, which is limited in large-scale 3D perception. Therefore, we focus on 3D vehicle detection without 2D detectors in roadside scenes. We propose a 3D vehicle localization network CenterLoc3D for roadside monocular cameras, which directly predicts centroid and eight vertexes in image space, and the dimension of 3D bounding boxes without 2D detectors. To improve the precision of 3D vehicle localization, we propose a multi-scale weighted-fusion module and a loss with spatial constraints embedded in CenterLoc3D. Firstly, the transformation matrix between 2D image space and 3D world space is solved by camera calibration. Secondly, vehicle type, centroid, eight vertexes, and the dimension of 3D vehicle bounding boxes are obtained by CenterLoc3D. Finally, centroid in 3D world space can be obtained by camera calibration and CenterLoc3D for 3D vehicle localization. To the best of our knowledge, this is the first application of 3D vehicle localization for roadside monocular cameras. Hence, we also propose a benchmark for this application including a dataset (SVLD-3D), an annotation tool (LabelImg-3D), and evaluation metrics. Through experimental validation, the proposed method achieves high accuracy with $A{P_{3D}}$ of 51.30\%, average 3D localization precision of 98\%, average 3D dimension precision of 85\% and real-time performance with FPS of 41.18.}




\maketitle

\section{Introduction}
\label{section:introduction}
In recent years, 3D vehicle localization is becoming an essential component in CVIS and ITS. Vehicle behaviors \cite{2019behaviour, 2020behaviour} and traffic flow statistics \cite{2020vehiclecount} can be analyzed and used for traffic state estimation \cite{2020Vehicle} and traffic management \cite{2020trafficcontrol}, which is of great research significance and practical value. In practical applications, obtaining vehicle localization and dimension in 3D space is more important than vehicle information in 2D images. Different from sensors in autonomous vehicles, roadside sensors are usually installed on higher to obtain full-road vehicle information easily. In CVIS, by estimating 3D vehicle localization, vehicle locations can be obtained and sent to each vehicle for accurate path planning to avoid collisions.

The most common sensors currently used for 3D vehicle localization are lidar \cite{2019pointrcnn, 2020pvrcnn, 2019pointpillars, 2018voxelnet} and stereo vision systems represented by binocular and RGB-D depth cameras \cite{2017stereo, 2018rgbd, 2019stereorcnn}. We can obtain accurate 3D location provided by point cloud data directly. However, they are usually expensive and environmentally demanding. In contrast, monocular RGB cameras \cite{2019monogrnet, 2019mono, 2020SMOKE, 2020d4lcn, 2021MonoFlex, 2021fcos3d} are cost-effective due to widespread deployment and fast data processing speed, which is more suitable for large-scale traffic scenes. Currently, monocular 3D vehicle localization is achieved by monocular 3D vehicle detection methods, which are focused on the onboard view. Differently, the roadside view is fixed with more geometric priors. At the same time, the roadside view is higher and wider than the onboard, which is more suitable for large-area perception. However, due to missing depth information, challenges like occlusion and congestion still exist in monocular roadside surveillance scenes.

Monocular 3D vehicle detection methods can be divided into the following two categories: (1) geometric constraint-based methods \cite{2017deep3dbox, 2017deepmanta, 2019cubeslam, 2019gs3d}. (2) 3D feature estimation based methods \cite{20193dbox, 2020keypoint, 2020rtm3d, 2020perspective, 2021litefpn, 2021km3d}. Methods of the first category are implemented based on mature 2D detectors, which use 3D bounding boxes or CAD models to represent vehicles. Then, 3D vehicle models are solved by establishing constraints of 3D bounding boxes fitting closely to 2D boxes. However, this geometric constraint is not enough for generating unambiguous results. Sub-networks will also reduce the processing efficiency. Compared with the first category, 3D vehicle information like keypoint, orientation, and dimension is directly extracted without 2D detectors. Due to missing depth information in monocular images, additional geometric inference modules are needed, which will also increase processing time to some extent.

To solve the above problems, we propose a monocular 3D vehicle localization method for surveillance cameras in traffic scenes. Firstly, the transformation matrix between 2D image space and 3D world space is solved by camera calibration, which is further used for 3D vehicle localization. Secondly, a one-stage 3D vehicle localization network CenterLoc3D is proposed, which contains three modules: backbone, multi-scale feature fusion, and multi-task detection head. In multi-scale feature fusion, we propose a weighted-fusion module to fuse five feature maps containing multi-scale information with different weights for multi-scale vehicle detection. In multi-task detection head, the multi-scale feature map is used as the input to obtain outputs. Outputs contain four branches: vehicle type, centroid, eight vertexes, and the dimension of 3D bounding boxes. To improve the precision of 3D vehicle localization without sacrificing efficiency, a loss with spatial constraints embedding is proposed, including reprojection constraint of 2D-3D transformation obtained by camera calibration and IoU constraint of 3D box projection. Finally, we also propose a benchmark including a dataset, an annotation tool, and evaluation metrics for 3D vehicle localization in roadside monocular traffic scenes. Through experimental validation, it can be proved that our method is efficient and robust.

The main contributions of this paper are summarized as follows:

\begin{itemize}
	\item A monocular 3D vehicle localization network CenterLoc3D for roadside surveillance cameras in traffic scenes is proposed, which directly predicts accurate 3D vehicle projection vertexes and dimensions.
	
	\item A weighted-fusion module is proposed in multi-scale feature fusion, which further enhances feature extraction capability.
	
	\item A loss with spatial constraints embedding is proposed, which can effectively improve the accuracy of 3D vehicle localization.
	
	\item A benchmark including a dataset, an annotation tool, and evaluation metrics is proposed for experimental validation, which is helpful for the development of monocular 3D vehicle localization in roadside monocular traffic scenes.
\end{itemize}


\section{Related Work}
\label{section:related work}
Deep learning-based monocular 3D vehicle detection methods can be divided into the following two categories: geometric constraint-based and 3D feature estimation-based.

\textbf{Geometric Constraint-Based Methods.}
Due to the continuous development of convolutional neural networks (CNNs), many excellent 2D object detection methods \cite{2017fasterrcnn, 2016ssd, 2020yolov4, 2017retinanet, 2019centernet} have emerged. Some methods \cite{2016mono3ddet, 2019deepfit} directly apply the regional proposal in 2D object detection to 3D object detection. Mono3D \cite{2016mono3ddet} introduces sliding windows for 3D vehicle detection. Firstly, a series of candidate bounding boxes in 3D space are generated and projected to the image plane by camera calibration. Secondly, priori information like vehicle segmentation contours is combined to further obtain vehicle areas and 3D bounding boxes. The best results are finally selected by non-maximum suppression (NMS). However, the search range in 3D space is much larger than that in 2D space while obtaining priori information is time-consuming, which greatly reduces the efficiency. Currently, 2D object detection methods are still used in most 3D vehicle detection methods \cite{2019monogrnet, 2019mono, 2017deep3dbox, 2017deepmanta, 2019cubeslam, 2019gs3d} to obtain vehicle location, dimension and orientation. To improve efficiency, generating candidate boxes in 3D space is replaced by using geometric constraints of 3D bounding boxes fitting closely to 2D boxes. Deep3DBox \cite{2017deep3dbox} obtains 3D vehicle bounding boxes and orientation by constructing 2D-3D box constraints and multi-bin loss function. Deep MANTA \cite{2017deepmanta}, a multi-task network, uses 3D CAD models and region proposal network (RPN) to obtain vehicle type, 2D bounding boxes, location, visibility, and similarity to CAD model, which is used to select the best 3D CAD model. CubeSLAM \cite{2019cubeslam} obtains three orthogonal vanishing points by extracting straight line segments of vehicles and constructs 3D vehicle bounding boxes by geometric constraints between vanishing points and 2D bounding boxes. GS3D \cite{2019gs3d} is able to correct 3D bounding boxes, which contains two sub-networks: 2D+O and 3D sub-network. The 2D+O sub-network adds a vehicle orientation regression branch to Faster R-CNN \cite{2017fasterrcnn}, which is used to obtain 2D vehicle bounding boxes and orientation simultaneously. With camera calibration and the above information, a coarse 3D bounding box called guidance can be obtained. The 3D sub-network is used to extract visible 3D features of the guidance and complete the correction to obtain more accurate detection results. The above methods based on geometric constraints usually require 2D bounding boxes and time-consuming priori information extraction. The constraints that 2D bounding boxes can provide are limited, which leads to ambiguous results. At the same time, sub-networks will additionally increase the processing time.

\textbf{3D Feature Estimation-Based Methods.}
To avoid ambiguous results from geometric constraints and further improve accuracy and efficiency, many new methods \cite{20193dbox, 2020keypoint, 2020rtm3d, 2020perspective, 2020SMOKE, 2021MonoFlex, 2021fcos3d, 2021litefpn, 2021km3d} focus on direct 3D information extraction in images which can be used for 3D bounding box regression by CNNs. Mono3DBox \cite{20193dbox} proposes an end-to-end detector that can directly predict the center point of the vehicle bottom in the image. Then, the point is converted into 3D space by a look-up table to realize 3D vehicle detection and localization. MonoGRK \cite{2020keypoint} proposes an end-to-end keypoint-based 3D vehicle detection and localization network, using ResNet-101 \cite{2016resnet} as the backbone. The detection head contains three sub-networks: 2D vehicle detection, 2D keypoint regression, and 3D vehicle dimension regression. Besides, 3D CAD models are also used like DeepMANTA \cite{2017deepmanta}, but only 14 keypoints are labelled for each model, which describes the CAD model coarsely. 2D-3D space is linked in the geometric inference module, which enables the model end-to-end training. Transformer3D \cite{2020perspective} proposes a roadside 3D vehicle detection method using CNNs. Automatic camera calibration \cite{2017autocalibimprove} is firstly used to obtain three orthogonal vanishing points and scale factors for perspective transformation. Then, the transformed image is fed into RetinaNet \cite{2017retinanet} to obtain 2D bounding boxes and conversion parameters for 3D boxes. Finally, 3D bounding boxes are recovered by camera calibration results and the conversion parameters for speed measurement. The pipeline is simple, but 3D bounding boxes are not directly evaluated in the experimental results. RTM3D \cite{2020rtm3d} views 3D vehicle detection as keypoint detection without leveraging 2D detection results. The network is constructed based on ResNet-18 \cite{2016resnet} and DLA-34 \cite{2018dla}, which directly predicts center points and eight vertexes of 3D bounding boxes in the image. Then, vehicle location, dimension, and orientation can be solved by minimizing the reprojection error of 2D-3D bounding boxes. Lite-FPN \cite{2021litefpn} proposes a light-weight keypoint-based feature pyramid network, which contains three parts: backbone, detection head and post-processing. Top-k operation is used in the detection head to connect keypoints and regression sub-network. The attention module is added in keypoint detection loss to effectively solve the problem of mismatch between keypoint confidence and location. KM3D \cite{2021km3d} proposes a 3D vehicle detection network that contains keypoint detection and geometric inference modules. Outputs include center point, eight vertexes, orientation, dimension, and confidence. In the network, keypoints are used to further guide vehicle orientation regression and 3D IoU strategy is used to help confidence branch training. The above method based on 3D feature estimation can improve the accuracy of 3D vehicle detection by designing a geometric inference module in the network. However, this module will increase model complexity and inference time.

Among the reviewed literature, geometric constraint-based methods leverage 2D detectors and additional geometric constraint modules to obtain 3D vehicle information, which reduces the efficiency. 3D feature estimation-based methods embed geometric inference modules into networks without 2D detectors, which increases the complexity. Moreover, almost all the reviewed research uses datasets of onboard scenes, instead of roadside perspective, which is limited in large-scale 3D perception.

In this paper, we focus on an efficient 3D vehicle detection structure for roadside surveillance cameras to directly obtain 3D vehicle information without 2D detectors and geometry inference modules, to tackle the above-mentioned research gaps and weaknesses.

\section{CenterLoc3D for 3D vehicle localization}
\label{section:methods}

\subsection{Framework}
\label{subsec:framework}
\begin{figure}[t]
	\centering
	\includegraphics[width=0.7\linewidth]{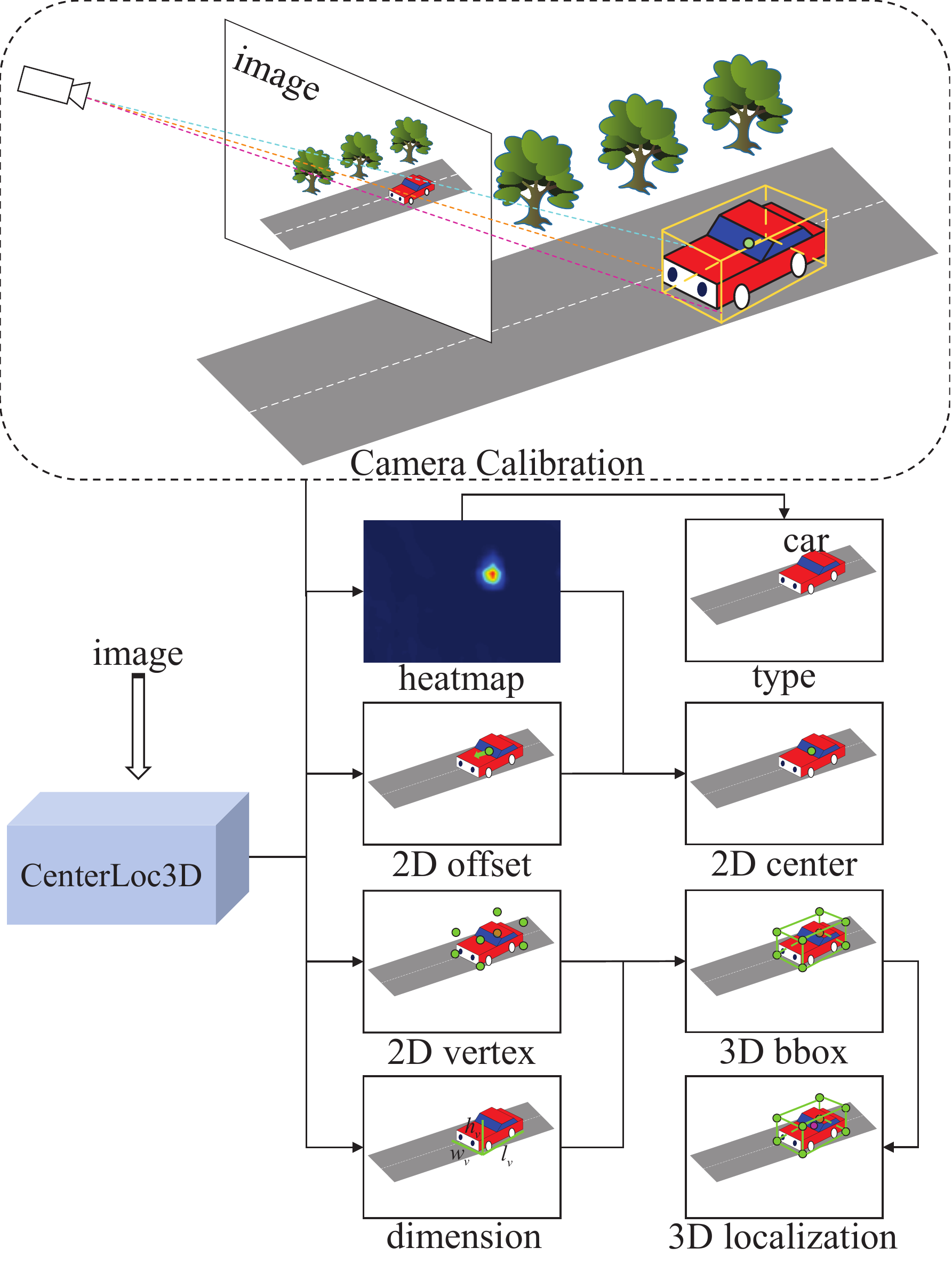}
	\caption{\leftskip=0pt \rightskip=0pt plus 0cm The overall framework of the proposed method. The method consists of two parts: camera calibration and 3D vehicle localization. A single image is the input of CenterLoc3D to obtain vehicle type, centroid, eight vertexes, and dimensions of 3D bounding boxes. Combined with camera calibration, 3D vehicle localization results can be further obtained.}
	\label{fig:fig_framework}
\end{figure}

The overall framework of the proposed method is shown in Figure \ref{fig:fig_framework}, which consists of two components: camera calibration and 3D vehicle localization. Firstly, the transformation matrix between 2D image space and 3D world space is solved by camera calibration. Secondly, vehicle type, centroid, eight vertexes, and dimensions of 3D bounding boxes are obtained by CenterLoc3D. Finally, 3D vehicle centroids can be obtained by camera calibration for 3D vehicle localization.
\subsection{Camera Calibration}
\label{subsec:camera calibration}
To complete 3D vehicle localization in traffic scenes, the transformation matrix between 2D image space and 3D world space must be derived through camera calibration. We refer to the previous work \cite{2010Taxonomy,2019Automatic} to define coordinate systems, establish the camera calibration model, and choose the single vanishing point-based calibration method VWL (One Vanishing Point, Known Width and Length) \cite{2010Taxonomy} to complete camera calibration.

\begin{figure}[t]
	\centering
	\includegraphics[width=0.7\linewidth]{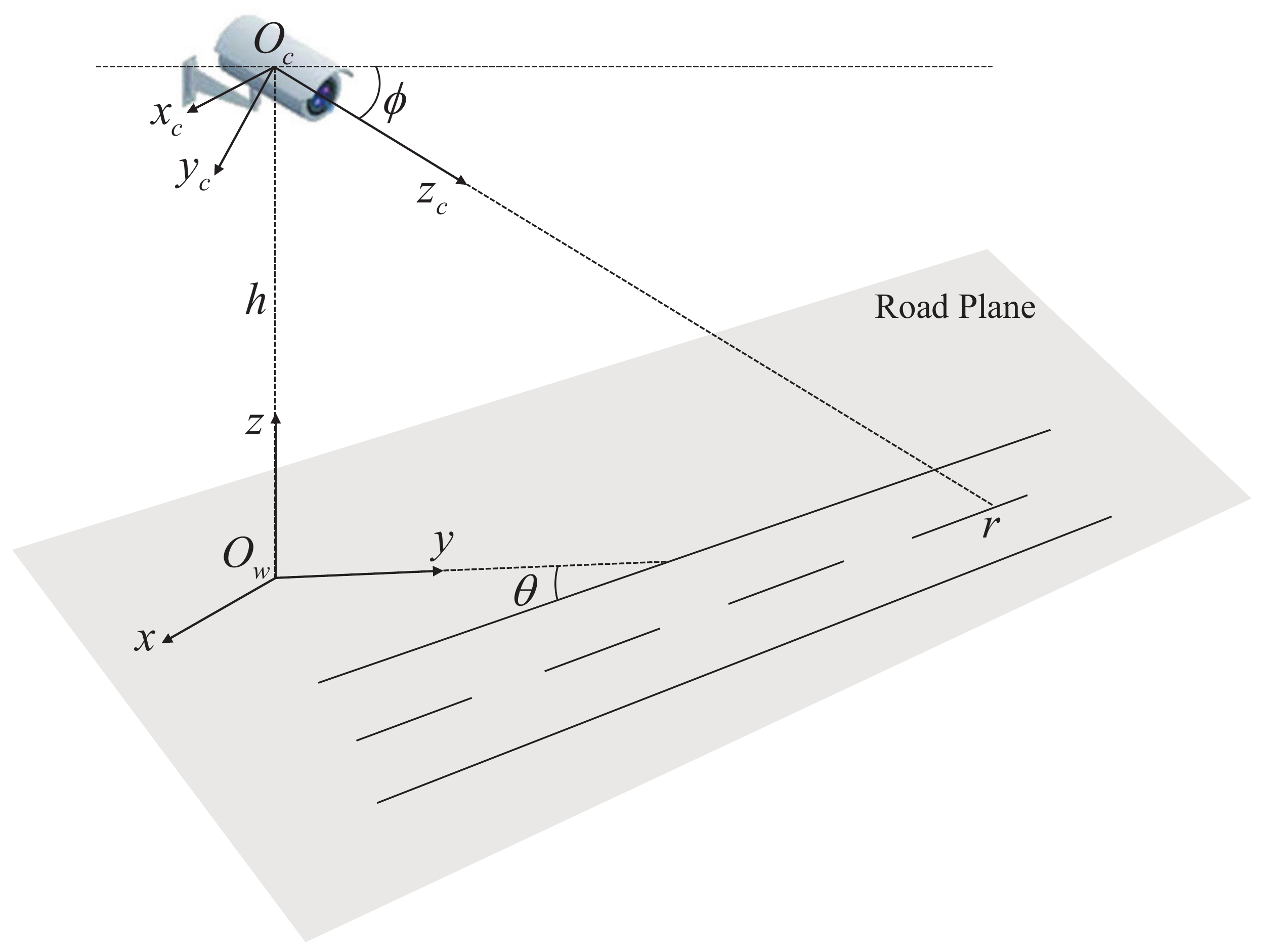}
	\caption{\leftskip=0pt \rightskip=0pt plus 0cm Schematic diagram of coordinate systems and camera calibration model. The world coordinate system is defined as ${O_w} - x,{\rm{ }}y,{\rm{ }}z$, camera coordinate system ${O_c} - {x_c},{\rm{ }}{y_c},{\rm{ }}{z_c}$, image coordinate system ${O_i} - u,{\rm{ }}v$. Parameters of calibration model include camera focal length $f$, camera height from the ground $h$, camera tilt angle $\phi$ and camera pan angle $\theta$.}
	\label{fig:camera_model}
\end{figure}

Schematic diagram of coordinate systems and camera calibration model is shown in Figure \ref{fig:camera_model}. Three coordinate systems are defined, all of which are right-handed. The world coordinate system is defined by $x,{\rm{ }}y,{\rm{ }}z$ axis. The origin ${O_w}$ is located at the projection point of the camera on the road plane, whereas $z$ is perpendicular to the road plane upwards. The camera coordinate system is defined by ${x_{\rm{c}}},{\rm{ }}{y_{\rm{c}}},{\rm{ }}{z_{\rm{c}}}$ axis. The origin ${O_c}$ is located at the camera optical center. ${x_{\rm{c}}}$ is parallel to $x$. ${z_{\rm{c}}}$ points to the ground along the camera optical axis. ${y_{\rm{c}}}$ is perpendicular to the plane ${x_c}{O_c}{z_c}$. The image coordinate system is defined by $u,{\rm{ }}v$ axis. The origin ${O_i}$ is located at image center. In the image coordinate system, $u$ is horizontal right and $v$ is vertical downward. ${z_{\rm{c}}}$ intersects the road plane at $r = ({c_x},{c_y})$ in the image coordinate system, which is called the principal point and the default location is at image center. ${c_x}$ and ${c_y}$ represent half of the image width and height, respectively.

In addition to the above parameters, another parameter is the roll angle, which can be expressed by a simple image rotation and has no effect on calibration results. Therefore, it is not considered. In this paper, VWL and road marks \cite{2019Automatic} are used to solve and optimize calibration parameters $f,h,\phi ,\theta $. The vanishing point $VP = ({u_0},{v_0})$ along the direction of traffic flow is extracted by road edge lines. To illustrate the road space in a straightforward way, ${O_i}$ and $y$ axis are adjusted. Firstly, ${O_i}$ is moved to the upper left corner of the image, corresponding to the intrinsic parameter matrix $K$:
\begin{equation}
	\label{equation_K}
	K = \left[ {\begin{array}{*{20}{c}}
			f&0&{{c_x}}\\
			0&f&{{c_y}}\\
			0&0&1
	\end{array}} \right]
\end{equation}

Then, $y$ axis is adjusted to the direction along traffic flow. The rotation matrix $R$ contains a rotation of $\phi  + {\pi  \mathord{\left/
		{\vphantom {\pi  2}} \right.
		\kern-\nulldelimiterspace} 2}$ about $x$ axis and $\theta $ about $z$ axis, which can be defined as:
\begin{equation}
	\label{equation_R}
	\begin{aligned}
		R &= {R_x}(\phi  + {\pi  \mathord{\left/
				{\vphantom {\pi  2}} \right.
				\kern-\nulldelimiterspace} 2}){R_z}(\theta )\\
		&= \left[ {\begin{array}{*{20}{r}}
				{\cos \theta }&{ - \sin \theta }&0\\
				{ - \sin \phi \sin \theta }&{ - \sin \phi \cos \theta }&{ - \cos \phi }\\
				{\cos \phi \sin \theta }&{\cos \phi \cos \theta }&{ - \sin \phi }
		\end{array}} \right]
	\end{aligned}
\end{equation}

The translation matrix $T$ is:
\begin{equation}
	\label{equation_T}
	T = \left[ {\begin{array}{*{20}{c}}
			1&0&0&0\\
			0&1&0&0\\
			0&0&1&{ - h}
	\end{array}} \right]
\end{equation}

The adjusted transformation formula from world point $(x,y,z)$ to image point $(u,v)$ is:
\begin{equation}
	\label{equation_H}
	s\left[ {\begin{array}{*{20}{c}}
			u\\
			v\\
			1
	\end{array}} \right] = KRT\left[ {\begin{array}{*{20}{c}}
			x\\
			y\\
			z\\
			1
	\end{array}} \right] = H\left[ {\begin{array}{*{20}{c}}
			x\\
			y\\
			z\\
			1
	\end{array}} \right]
\end{equation}
where $H = \left[ {{h_{ij}}} \right],i = 1,2,3;j = 1,2,3,4$ is the $3 \times 4$ projection matrix from world to image coordinate. $s$ is the scale factor.

Finally, according to the above derivation, the adjusted transformation between world and image can be described as follows:
\begin{equation}
	\label{equa_xyz2uv}
	\left\{ \begin{array}{l}
		u = \frac{{{h_{11}}x + {h_{12}}y + {h_{13}}z + {h_{14}}}}{{{h_{31}}x + {h_{32}}y + {h_{33}}z + {h_{34}}}}\\
		v = \frac{{{h_{21}}x + {h_{22}}y + {h_{23}}z + {h_{24}}}}{{{h_{31}}x + {h_{32}}y + {h_{33}}z + {h_{34}}}}
	\end{array} \right.
\end{equation}
\begin{equation}
	\label{equa_uv2xyz}
	\left\{ \begin{array}{l}
		x = \frac{{{b_1}({h_{22}} - {h_{32}}v) - {b_2}({h_{12}} - {h_{32}}u)}}{{({h_{11}} - {h_{31}}u)({h_{22}} - {h_{32}}v) - ({h_{12}} - {h_{32}}u)({h_{21}} - {h_{31}}v)}}\\
		y = \frac{{ - {b_1}({h_{21}} - {h_{31}}v) + {b_2}({h_{11}} - {h_{31}}u)}}{{({h_{11}} - {h_{31}}u)({h_{22}} - {h_{32}}v) - ({h_{12}} - {h_{32}}u)({h_{21}} - {h_{31}}v)}}
	\end{array} \right.
\end{equation}
where $\left\{ \begin{array}{l}
	{b_1} = u({h_{33}}z + {h_{34}}) - ({h_{13}}z + {h_{14}})\\
	{b_2} = v({h_{33}}z + {h_{34}}) - ({h_{23}}z + {h_{24}})
\end{array} \right.$.

\subsection{CenterLoc3D}
\label{subsec:centerloc3d}
In the roadside camera view, 3D vehicle localization can be described in terms of 3D vehicle detection. We propose a 3D vehicle localization network CenterLoc3D for roadside cameras, which uses a single RGB image as the input and the outputs include vehicle type, centroid, vertexes, and dimensions of 3D bounding boxes. Combined with camera calibration in Section \ref{subsec:camera calibration}, 3D vehicle localization results can be calculated.

\subsubsection{Network Architecture}
\label{subsubsec:network arch}
\begin{figure*}[htbp]
	\centering
	\includegraphics[width=1.0\linewidth]{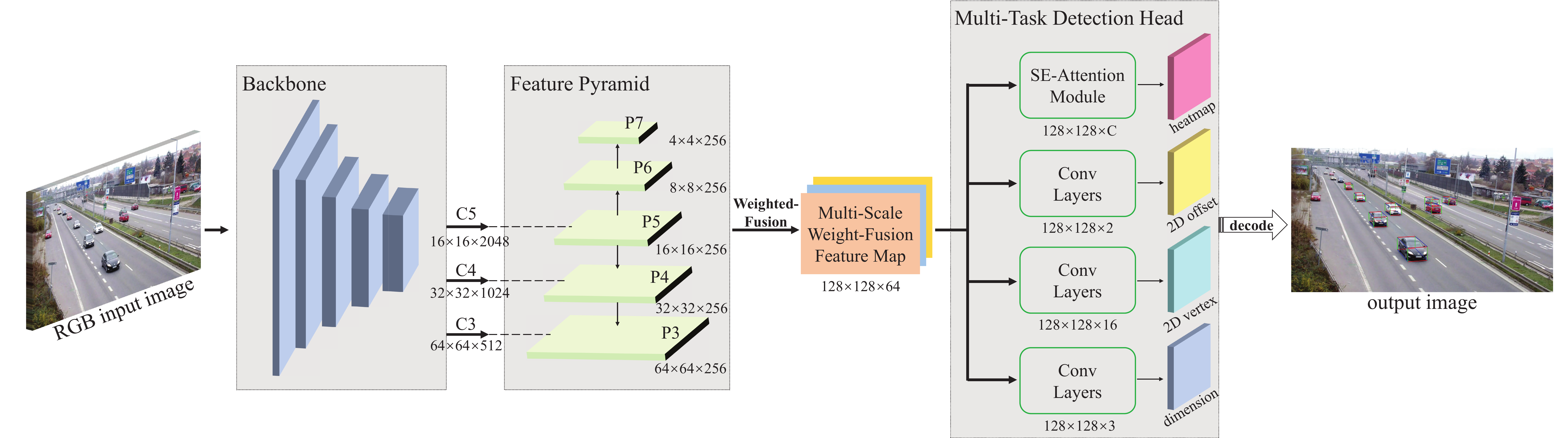}
	\caption{\leftskip=0pt \rightskip=0pt plus 0cm The architecture of CenterLoc3D. The network mainly consists of three parts: backbone, multi-scale feature fusion, and multi-task detection head. Multi-scale weighted-fusion feature map is obtained in the first two parts. 3D bounding box regression includes vehicle type, centroid, eight vertexes, and dimensions, which are decoded to final outputs.}
	\label{fig:arch_centerloc}
\end{figure*}

The overall architecture of the network is shown in Figure \ref{fig:arch_centerloc}, which contains three parts: backbone, multi-scale feature fusion and multi-task detection head. Firstly, the RGB image is scaled to $I \in { \mathbb{R} ^{H \times W \times 3}}$ as input, where $H = W = 512$. Then, the image is downsampled with a factor of $S = 4$.

\textbf{Backbone.} To make a trade-off between accuracy and efficiency, we use ResNet-50 \cite{2016resnet} as our backbone, containing residual and inverted bottleneck structures, which have good feature extraction capability. We extract the last three features with channels of 512, 1024, and 2048 for feature fusion.

\begin{figure}
	\centering
	\includegraphics[width=1.0\linewidth]{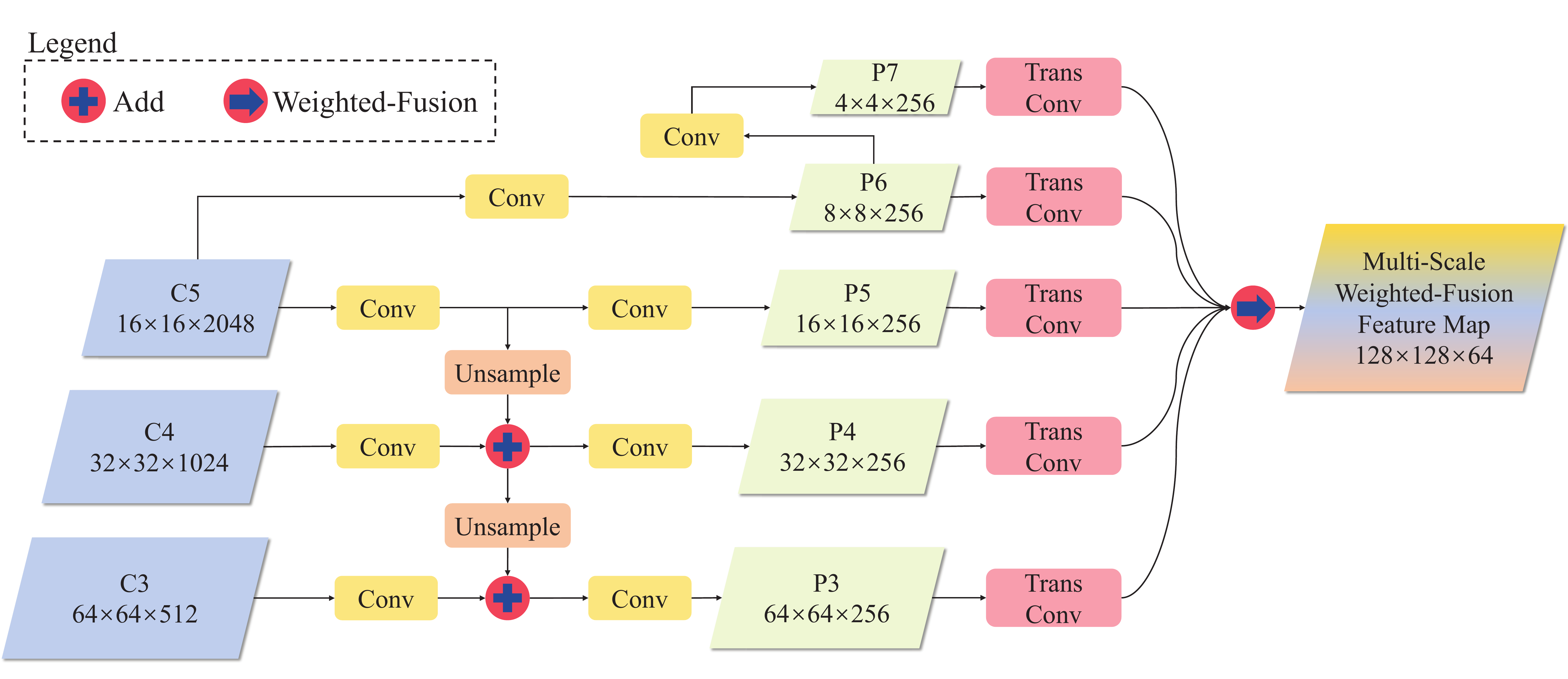}
	\caption{Schematic diagram of weighted-fusion module.}
	\label{fig:fig_weighted-fusion}
\end{figure}

\textbf{Multi-Scale Feature Fusion.} The series of feature maps obtained by the backbone are hierarchical. The high-resolution feature map retains more accurate local features and is suitable for small object detection. The low-resolution feature map contains higher-level semantic information and is suitable for large object detection. In monocular roadside traffic scenes, vehicles are distributed in different locations of different sizes. To make the network adaptive to vehicles of different sizes, a weighted-fusion module is proposed in multi-scale feature fusion based on RetinaNet \cite{2017retinanet}. Figure \ref{fig:fig_weighted-fusion} shows the schematic diagram of weighted-fusion module. The feature maps ${C_3},{C_4},{C_5}$ of size $64 \times 64$, $32 \times 32$, and $16 \times 16$ are extracted by backbone and used to construct feature pyramid ${F_p} = \left\{ {{P_3},{P_4},{P_5},{P_6},{P_7}} \right\}$. Different from YOLOv4 \cite{2020yolov4}, which directly retains the feature pyramid, we use deconvolution and unsample \cite{2017fpn} to unify the five feature maps of different sizes in ${F_p}$ to the same size $\overline {{F_p}}  = \left\{ {\overline {{P_3}} ,\overline {{P_4}} ,\overline {{P_5}} ,\overline {{P_6}} ,\overline {{P_7}} } \right\}$. Then, weighted-fusion module is added by the fusion strategy shown in Equation \ref{equa_feature_fusion}. Finally, a fused feature map $F \in {\mathbb{R}^{\frac{H}{S} \times \frac{W}{S} \times 64}}$ containing multi-scale information of vehicles is obtained. This feature fusion module not only generates multi-scale feature maps, but also reduces computational effort in the prediction process and improve the overall efficiency.
\begin{equation}
	\label{equa_feature_fusion}
	F = \sum\limits_{i = 3}^7 {{w_i} \times \overline {{P_i}} }
\end{equation}
where ${w_i}$ denotes the weight of the feature map with $\sum\limits_{i = 3}^7 {{w_i}}  = 1$, which can be set according to the importance degree. We set 0.5, 0.2, 0.1, 0.1 and 0.1 respectively in the experiment.

\textbf{Multi-Task Detection Head.} Based on the multi-scale fusion feature map, a multi-task detection head is designed based on the actual demand for 3D vehicle localization in roadside scenes. The detection head is comprised of four branches: vehicle type, centroid, eight vertexes, and dimensions of 3D bounding boxes. To further improve the ability to distinguish different types of vehicles, the type branch is implemented by fully convolutional layers and attention module \cite{2018senet}. The remaining branches are implemented by fully convolutional layers only. Inspired by CenterNet \cite{2019centernet}, vehicle type is defined as centroid heatmap ${M_c} \in {[0,1]^{\frac{H}{S} \times \frac{W}{S} \times C}}$, where $C$ denotes the number of vehicle types. The remaining branches are defined as centroid offset ${M_{co}} \in {[0,1]^{\frac{H}{S} \times \frac{W}{S} \times 2}}$, vertexes ${M_v} \in {\mathbb{R}^{\frac{H}{S} \times \frac{W}{S} \times 16}}$ and dimension ${M_s} \in {\mathbb{R}^{\frac{H}{S} \times \frac{W}{S} \times 3}}$. To ensure the stability during training, we normalize centroids and vertexes to the fusion feature map with size ${H \mathord{\left/
		{\vphantom {H S}} \right.
		\kern-\nulldelimiterspace} S},{W \mathord{\left/
		{\vphantom {W S}} \right.
		\kern-\nulldelimiterspace} S}$.

\subsubsection{Loss Function}
\label{subsubsec:loss func}
In the training process, the loss function contains six components: vehicle classification, centroid offset, vertexes, dimension, and a loss with spatial constraints embedding for improving the precision of 3D vehicle localization. The spatial constraints include reprojection constraint of 2D-3D transformation obtained by camera calibration and IoU constraint of 3D box projection.

\textbf{Vehicle Classification Loss.} To solve the problem of imbalanced positive and negative samples, we use focal loss \cite{2017retinanet} as vehicle classification loss:

\begin{equation}
	\label{equa_focal_loss}
	{L_c} =  - \frac{1}{N}\sum\limits_{k = 1}^C {\sum\limits_{i = 1}^{{W \mathord{\left/
					{\vphantom {W S}} \right.
					\kern-\nulldelimiterspace} S}} {\sum\limits_{j = 1}^{{H \mathord{\left/
						{\vphantom {H S}} \right.
						\kern-\nulldelimiterspace} S}} {\left\{ {\begin{array}{*{20}{c}}
						{{{(1 - {{\hat p}_{cij}})}^\alpha }\log ({{\hat p}_{cij}})}&{if{\rm{ }}{p_{cij}} = 1}\\
						{\hat p_{cij}^\alpha {{(1 - {p_{cij}})}^\beta }\log (1 - {{\hat p}_{cij}})}&{if{\rm{ }}{p_{cij}} < 1}
				\end{array}} \right.} } }
\end{equation}

\noindent where $N$ is the number of positive samples, $\alpha $ and $\beta $ are hyper-parameters used to adjust loss weights of positive and negative samples, which are usually set to 2 and 4. ${p_{cij}}$ is the response value of each ground truth vehicle in the heatmap described by the Gaussian kernel function ${e^{ - \frac{{{{(x - {p_{cijx}})}^2} + {{(y - {p_{cijy}})}^2}}}{{2{\sigma ^2}}}}}$. $\sigma $ is the standard deviation calculated from the ground truth vehicle dimension in image space \cite{2019centernet}.

\textbf{3D Vehicle Information Regression Loss.} We use L1 regression loss for vehicle centroid offset, vertex, and dimension loss:
\begin{equation}
	\label{equa_lco}
	{L_{co}} = \frac{1}{N}\sum\limits_{i = 1}^{{W \mathord{\left/
				{\vphantom {W S}} \right.
				\kern-\nulldelimiterspace} S}} {\sum\limits_{j = 1}^{{H \mathord{\left/
					{\vphantom {H S}} \right.
					\kern-\nulldelimiterspace} S}} {\mathbbm{1}_{ij}^{obj}\lvert {M_{co}^{ij} - ({{{p_{center}}} \mathord{\left/
						{\vphantom {{{p_{center}}} S}} \right.
						\kern-\nulldelimiterspace} S} - {{\tilde p}_{center}})} \rvert} }
\end{equation}
\begin{equation}
	\label{equa_lv}
	{L_v} = \frac{1}{N}\sum\limits_{i = 1}^{{W \mathord{\left/
				{\vphantom {W S}} \right.
				\kern-\nulldelimiterspace} S}} {\sum\limits_{j = 1}^{{H \mathord{\left/
					{\vphantom {H S}} \right.
					\kern-\nulldelimiterspace} S}} {\mathbbm{1}_{ij}^{obj}\lvert {M_v^{ij} - {{{p_{vertex}}} \mathord{\left/
						{\vphantom {{{p_{vertex}}} S}} \right.
						\kern-\nulldelimiterspace} S}} \rvert} }
\end{equation}
\begin{equation}
	\label{equa_ls}
	{L_s} = \frac{1}{N}\sum\limits_{i = 1}^{{W \mathord{\left/
				{\vphantom {W S}} \right.
				\kern-\nulldelimiterspace} S}} {\sum\limits_{j = 1}^{{H \mathord{\left/
					{\vphantom {H S}} \right.
					\kern-\nulldelimiterspace} S}} {\mathbbm{1}_{ij}^{obj}\lvert {M_s^{ij} - \tilde M_s^{ij}} \rvert} }
\end{equation}
where $\mathbbm{1}_{ij}^{obj}$ denotes whether centroid appears at $i,j$. ${p_{center}}$ and ${p_{vertex}}$ denote the ground truth centroids and vertexes of 3D bounding boxes in input image with size $H,W$. $\tilde p_{center}^{}$ denotes the ground truth centroids in the fusion feature map with size ${H \mathord{\left/
		{\vphantom {H S}} \right.
		\kern-\nulldelimiterspace} S},{W \mathord{\left/
		{\vphantom {W S}} \right.
		\kern-\nulldelimiterspace} S}$. ${\tilde M_s}$ denotes the ground truth feature map of 3D vehicle dimension.

\textbf{Loss with spatial constraints embedding.} To further improve the precision of 3D vehicle localization, loss functions with spatial constraints embedding are designed, including reprojection constraints using camera calibration and vehicle IoU constraints.

\begin{figure}[htbp]
	\centering
	\subcaptionbox{\centering Reprojection constraint\label{subfig:projection_constraint}}
	{%
		\includegraphics[height=3.0cm]{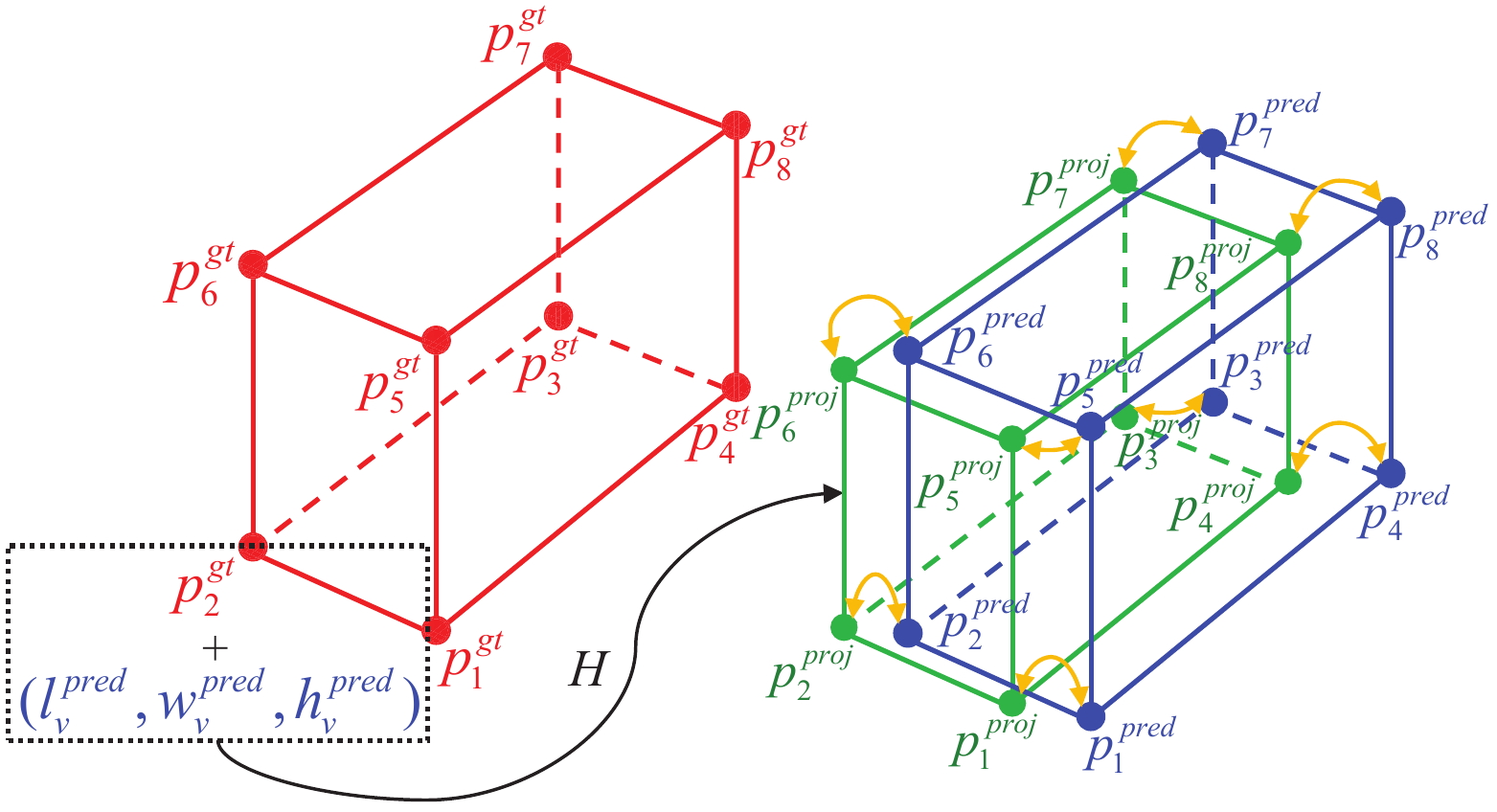}%
	}
	\subcaptionbox{\centering IoU constraint\label{subfig:iou_constraint}}
	{%
		\includegraphics[height=3.0cm]{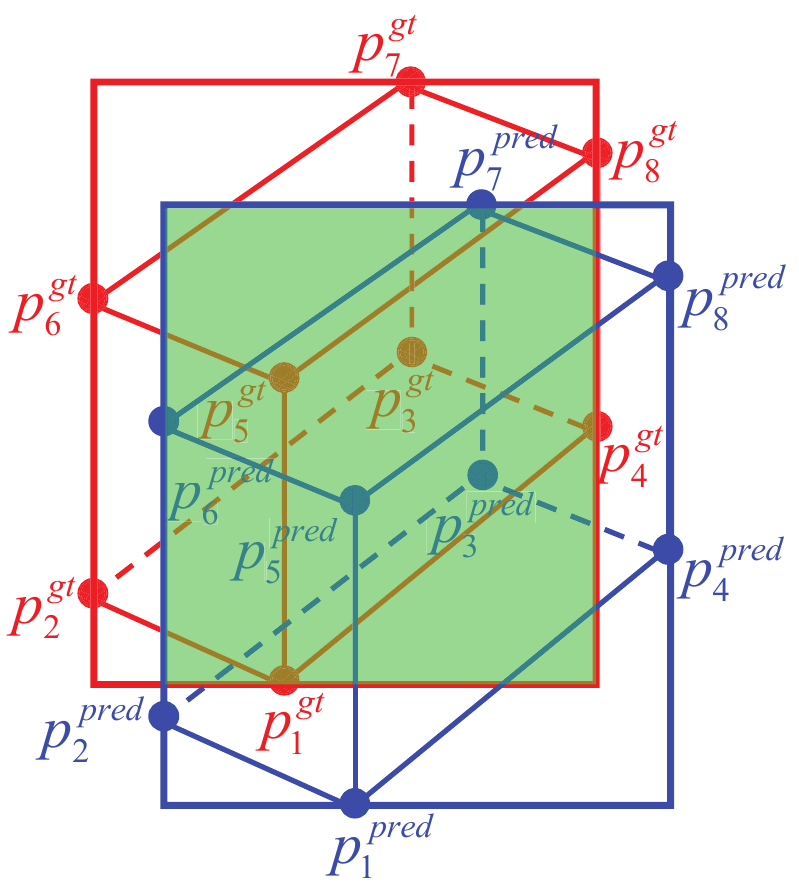}%
	}
	\caption{Schematic diagram of spatial constraints.}
	\label{fig:fig_spatial_constraint}
\end{figure}

The schematic diagram of reprojection constraint and vehicle IoU constraint is shown in Figure \ref{fig:fig_spatial_constraint}. We define the predicted and ground truth vertexes in image as $p_i^{pred} = (u_i^{pred},v_i^{pred})$ and $p_i^{gt} = (u_i^{gt},v_i^{gt})$. The corresponding vertexes in world are $P_i^{pred} = (x_i^{pred},y_i^{pred},z_i^{pred})$ and $P_i^{gt} = (x_i^{gt},y_i^{gt},z_i^{gt})$. The projection of predicted vertexes in world and image are defined as $P_i^{proj} = (x_i^{proj},y_i^{proj},z_i^{proj})$ and $p_i^{proj} = (u_i^{proj},v_i^{proj})$, $i = 1,2, \cdots ,8$. The predicted dimension is $D_v^{pred} = (l_v^{pred},w_v^{pred},h_v^{pred})$. Based on $p_i^{pred}$ from CenterLoc3D, $P_i^{proj}$ can be obtained by Equation \ref{equa_uv2xyz} and Table \ref{tab:table_3d_proj_vertexes}. Then, $p_i^{proj}$ can be obtained by Equation \ref{equa_xyz2uv}. As shown in Figure \ref{subfig:projection_constraint}, $p_i^{proj}$ and $p_i^{pred}$ constitute the projection constraints, where predicted, ground truth, and projection boxes are represented in blue, red, and green, respectively.

The minimum external rectangles of predicted and ground truth vertexes are $v_{rec}^{pred}$ and $v_{rec}^{gt}$. As shown in Figure \ref{subfig:iou_constraint}, blue indicates the predicted box while red is the ground truth. The green area indicates the overlap between the two boxes.

\begin{table}[htbp]
	\centering
	\caption{Calculation of 3D bounding box projection vertexes in world space.}
	\label{tab:table_3d_proj_vertexes}
	\begin{tabular}{cc}
		\toprule
		Vertex & World coordinate \\
		\midrule
		$P_1^{proj}$& $(x_2^{gt} + w_v^{pred},y_2^{gt},z_2^{gt})$ \\
		$P_2^{proj}$& $(x_2^{gt},y_2^{gt},z_2^{gt})$ \\
		$P_3^{proj}$& $(x_2^{gt},y_2^{gt} + l_v^{pred},z_2^{gt})$ \\
		$P_4^{proj}$& $(x_2^{gt} + w_v^{pred},y_2^{gt} + l_v^{pred},z_2^{gt})$ \\
		$P_5^{proj}$& $(x_2^{gt} + w_v^{pred},y_2^{gt},z_2^{gt} + h_v^{pred})$ \\
		$P_6^{proj}$& $(x_2^{gt},y_2^{gt},z_2^{gt} + h_v^{pred})$ \\
		$P_7^{proj}$& $(x_2^{gt},y_2^{gt} + l_v^{pred},z_2^{gt} + h_v^{pred})$ \\
		$P_8^{proj}$& $(x_2^{gt} + w_v^{pred},y_2^{gt} + l_v^{pred},z_2^{gt} + h_v^{pred})$ \\
		\bottomrule
	\end{tabular}
\end{table}

The loss comprised of two constraints shown in Figure \ref{fig:fig_spatial_constraint} is defined as follows:
\begin{equation}
	\label{equa_lproj}
	{L_{proj}} = \frac{1}{N}\sum\limits_{i = 1}^{{W \mathord{\left/
				{\vphantom {W S}} \right.
				\kern-\nulldelimiterspace} S}} {\sum\limits_{j = 1}^{{H \mathord{\left/
					{\vphantom {H S}} \right.
					\kern-\nulldelimiterspace} S}} {\mathbbm{1}_{ij}^{obj}\lvert {M_{proj}^{ij}(H,M_s^{ij}) - {\bar M_v^{ij}} } \rvert} }
\end{equation}
\begin{equation}
	\label{equa_liou}
	{L_{iou}} = \frac{1}{N}\sum\limits_{i = 1}^{{W \mathord{\left/
				{\vphantom {W S}} \right.
				\kern-\nulldelimiterspace} S}} {\sum\limits_{j = 1}^{{H \mathord{\left/
					{\vphantom {H S}} \right.
					\kern-\nulldelimiterspace} S}} {\mathbbm{1}_{ij}^{obj} \cdot IoU(R_{bbox}^{ij} - \tilde R_{bbox}^{ij})} }
\end{equation}
where $M_{proj}^{}(H,M_s^{ij})$ denotes the feature map of projection vertexes calculated by camera calibration matrix $H$ in Section \ref{subsec:camera calibration} and vehicle dimension feature map ${M_s}$ predicted by the network. $\bar {M_v^{}} $ denotes predicted 3D bounding boxes vertexes in the original image. $R_{bbox}^{ij}$ and $\tilde R_{bbox}^{ij}$ denote the predicted and ground truth minimum external rectangles. $IoU$ denotes the IoU loss strategy and we use CIoU loss \cite{2020ciou} in the experiment.

With the above six loss functions, the multi-task loss function can be defined as follows:
\begin{equation}
	\label{equa_lossfunc}
	L = {\lambda _c}{L_c} + {\lambda _{co}}{L_{co}} + {\lambda _v}{L_v} + {\lambda _s}{L_s} + {\lambda _{proj}}{L_{proj}} + {\lambda _{iou}}{L_{iou}}
\end{equation}
where $\lambda $ is the weight to balance the loss of each component. We set ${\lambda _c} = 1$, ${\lambda _{co}} = 1$ , ${\lambda _v} = 0.1$ , ${\lambda _s} = 0.1$ , ${\lambda _{proj}} = 0.1$ and ${\lambda _{iou}} = 1$ in the experiment.

\section{Dataset of 3D Vehicle Localization}
\label{sec:datasets}
Most currently available 3D vehicle localization datasets are based on onboard views \cite{2012kitti, 2020waymo, 2020nuscenes} instead of roadside, which is difficult for large-scale 3D perception and validation of roadside vehicle perception methods. Therefore, we propose a 3D vehicle localization dataset (SVLD-3D) for roadside surveillance cameras and an annotation tool (LabelImg-3D) for experimental validation.

\subsection{Dataset Composition}

\begin{figure}[htbp]
	\centering
	\subcaptionbox{\centering Scene A\label{subfig:dataset_samples_a}}
	{%
		\includegraphics[width=0.3\linewidth]{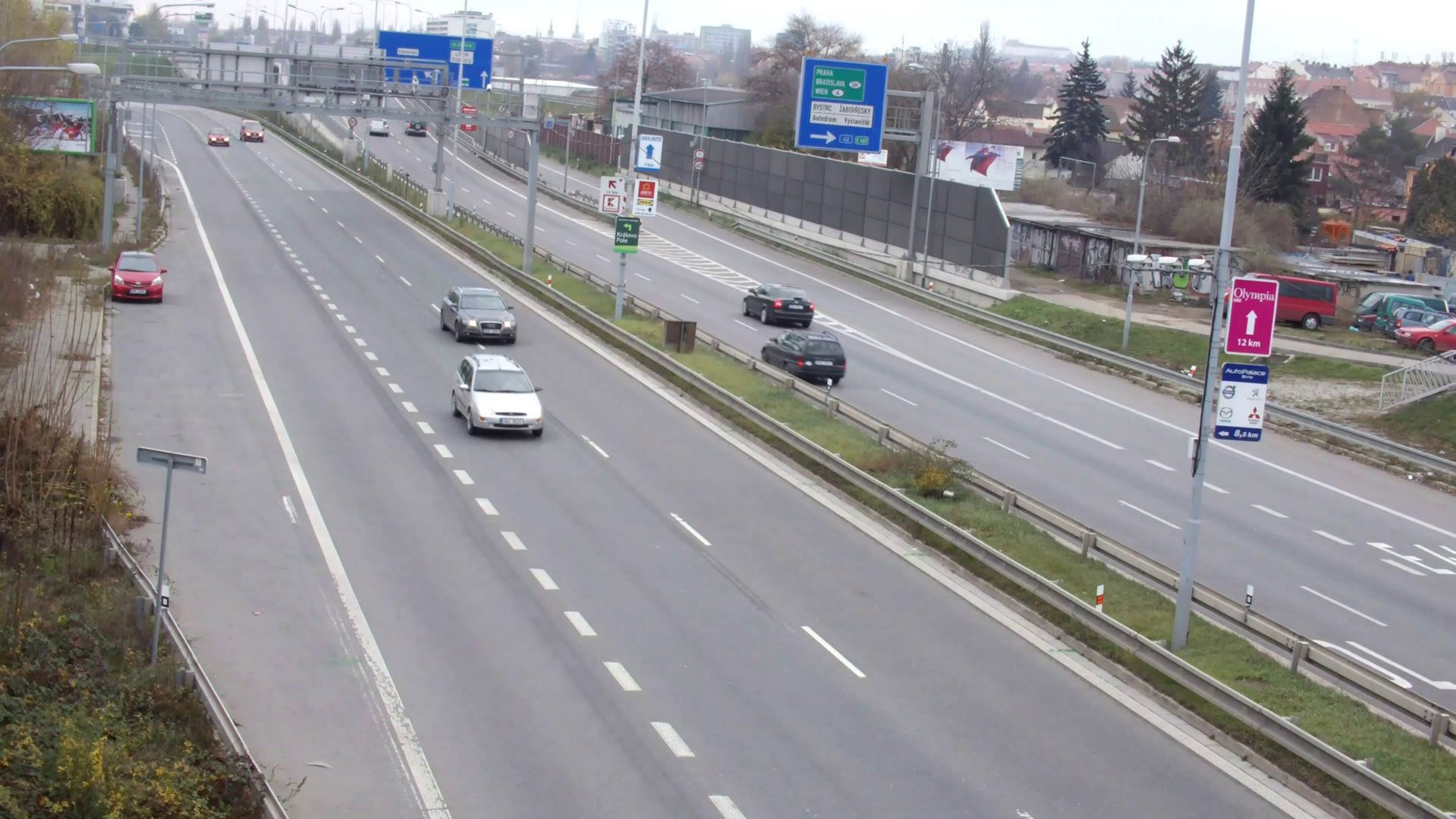}\quad
		\includegraphics[width=0.3\linewidth]{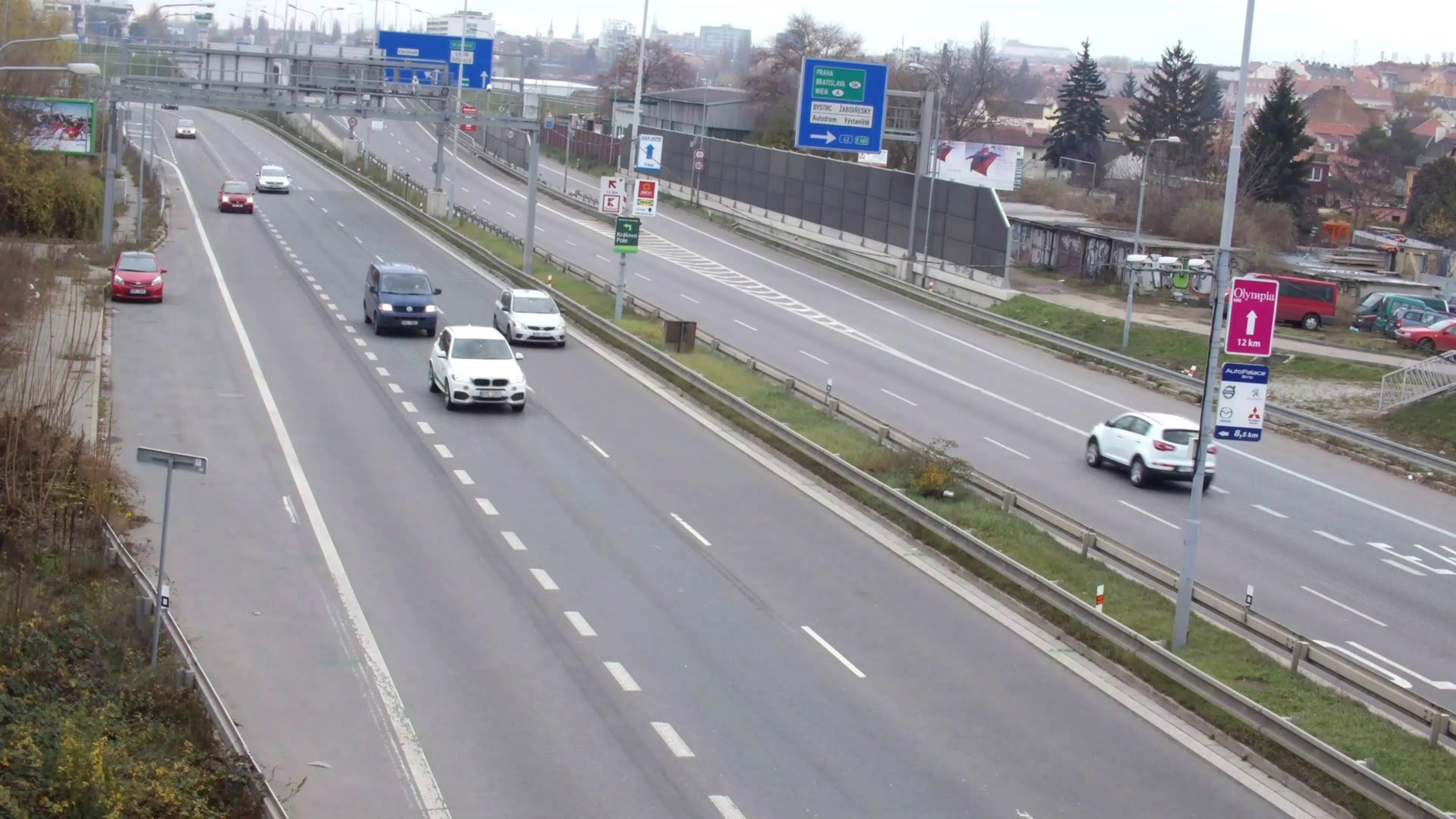}\quad
		\includegraphics[width=0.3\linewidth]{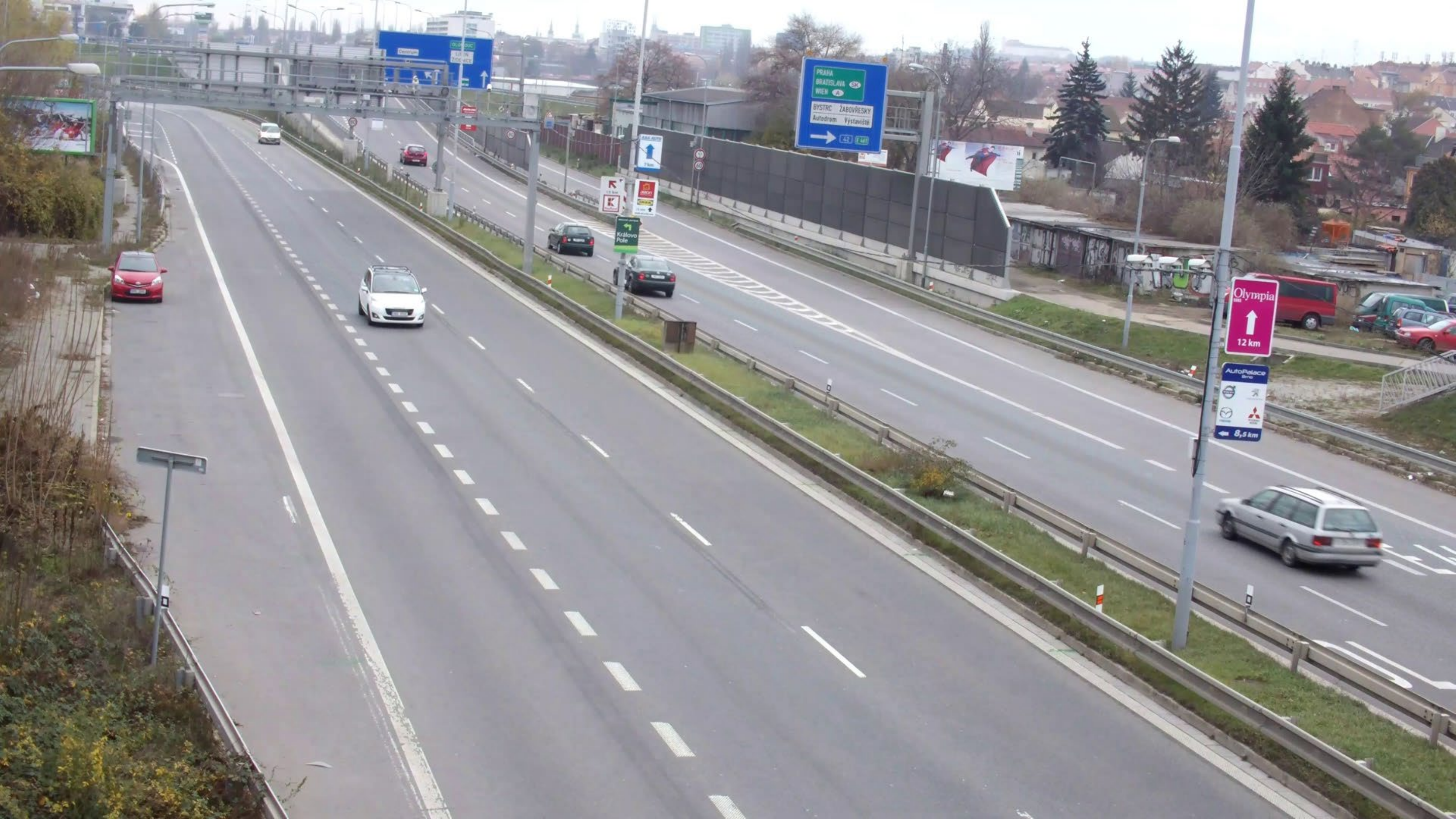}
	}
	\subcaptionbox{\centering Scene B\label{subfig:dataset_samples_b}}
	{%
		\includegraphics[width=0.3\linewidth]{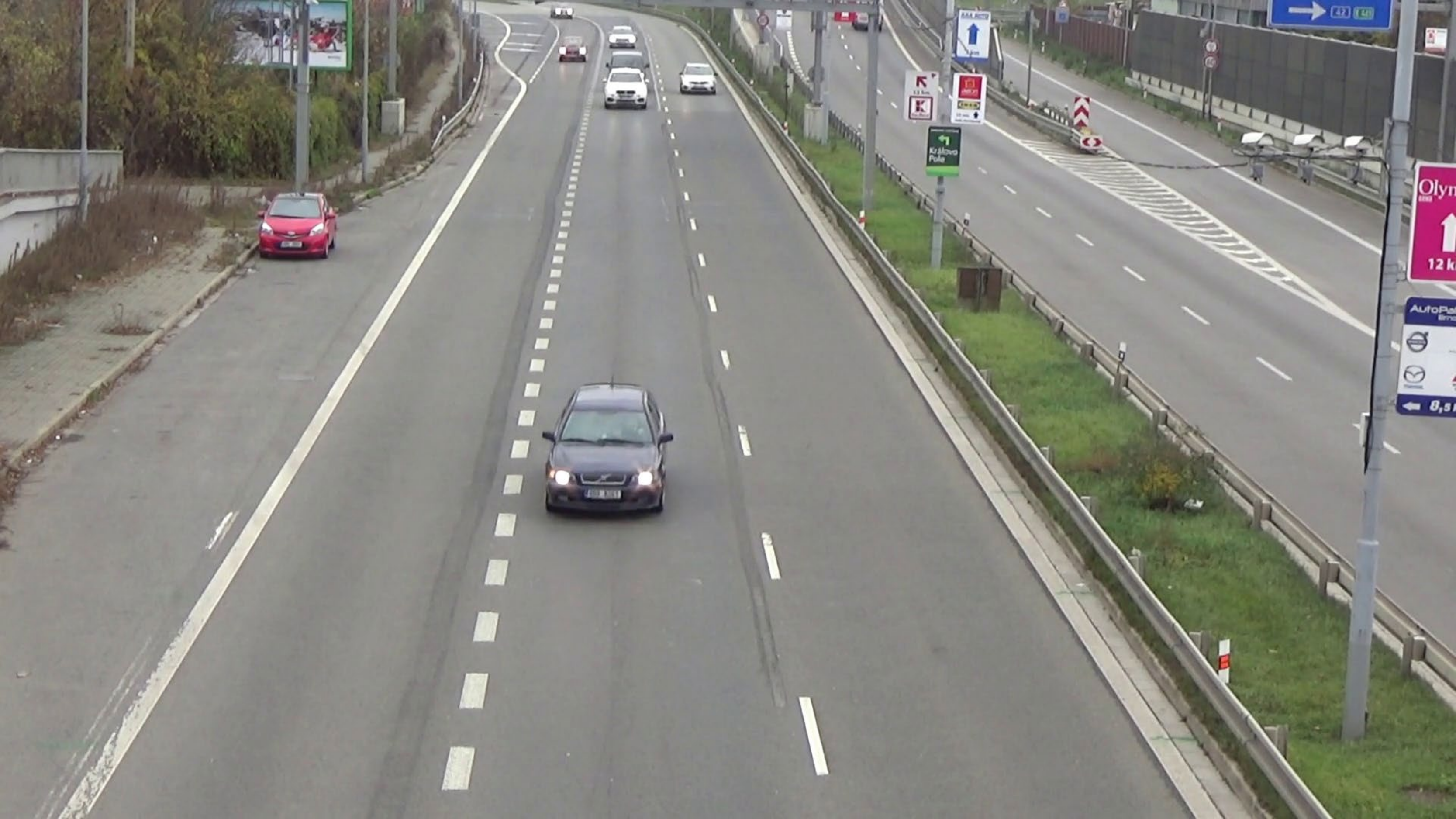}\quad
		\includegraphics[width=0.3\linewidth]{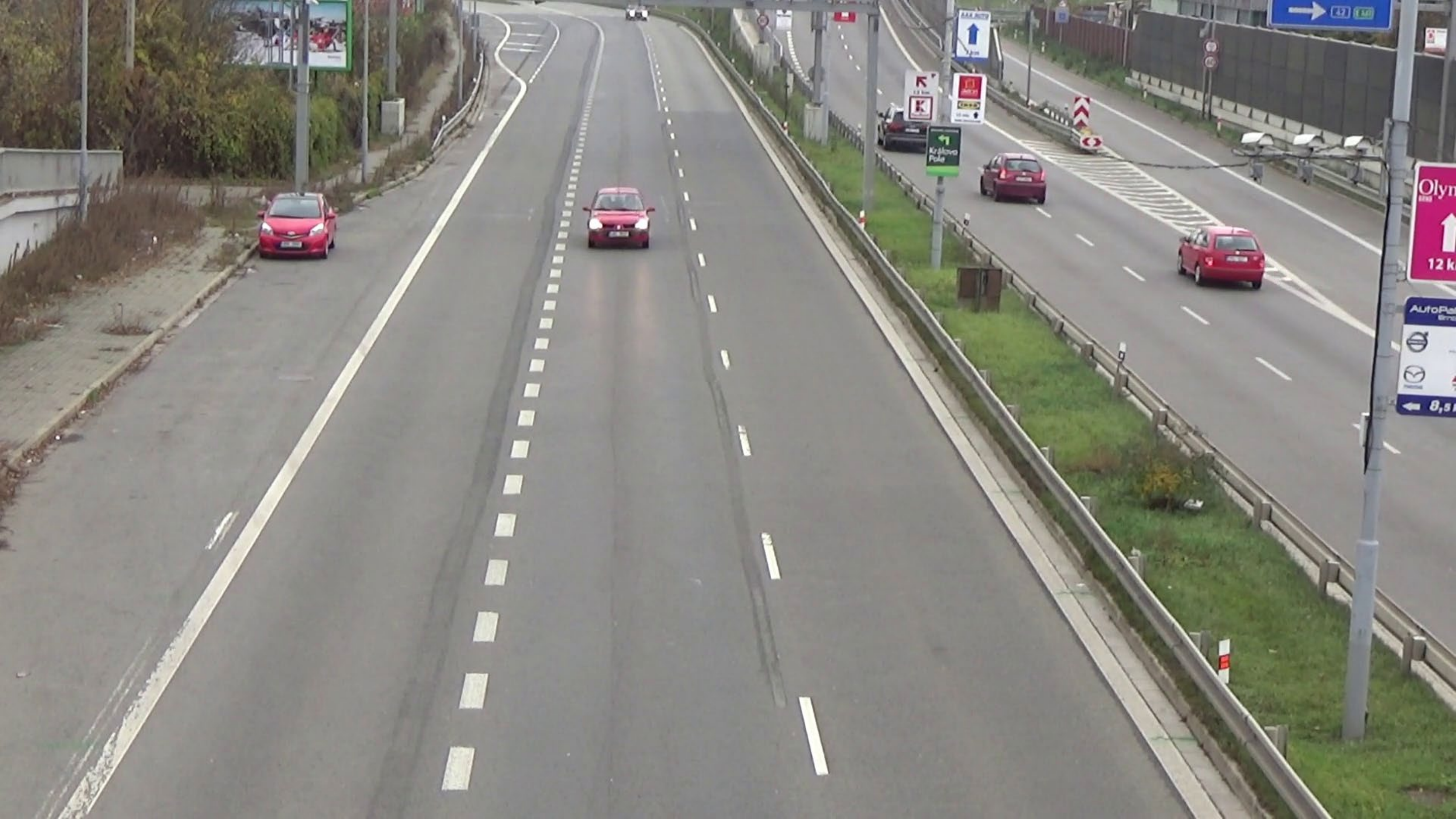}\quad
		\includegraphics[width=0.3\linewidth]{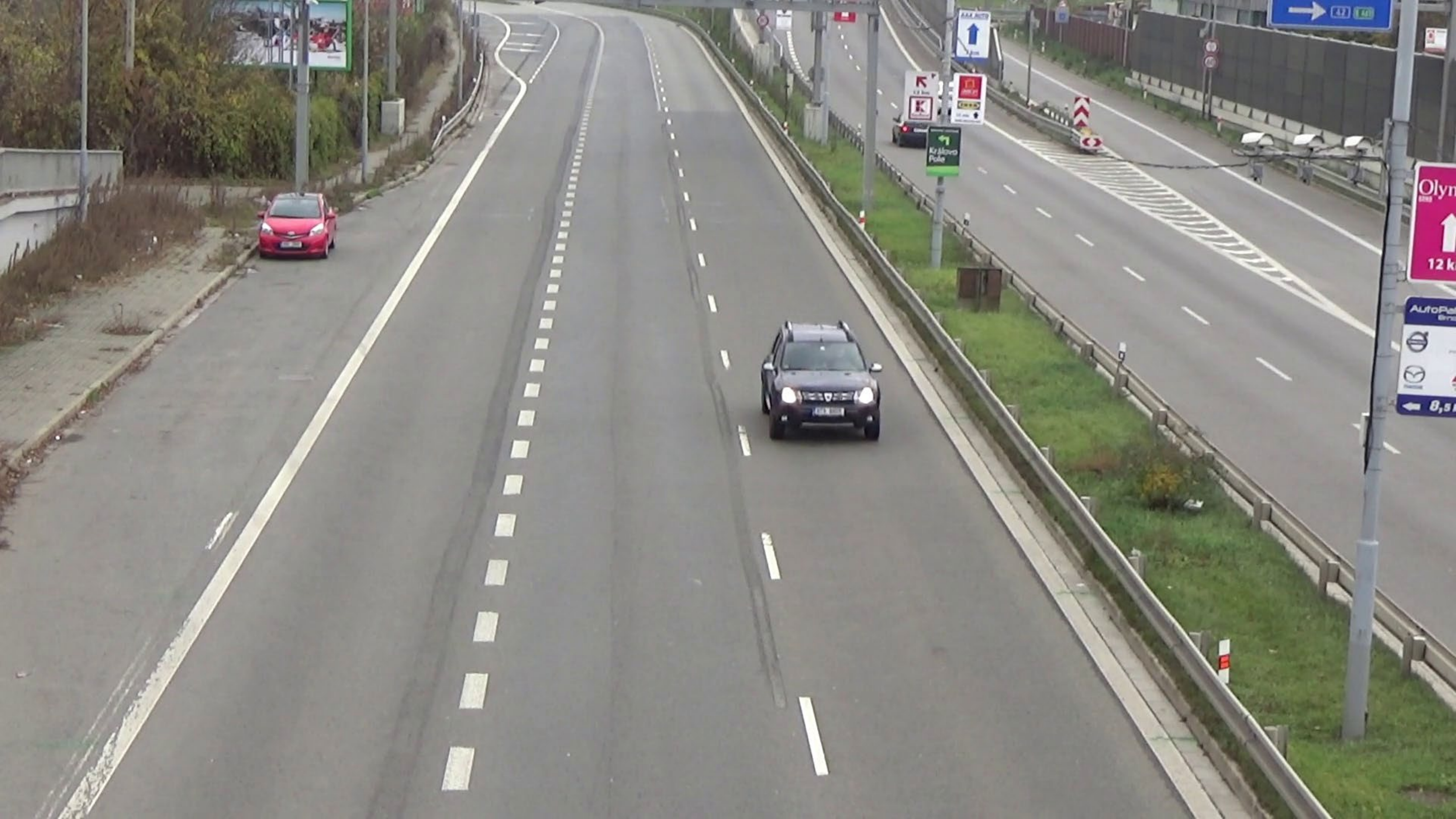}
	}
	\subcaptionbox{\centering Scene C\label{subfig:dataset_samples_c}}
	{%
		\includegraphics[width=0.3\linewidth]{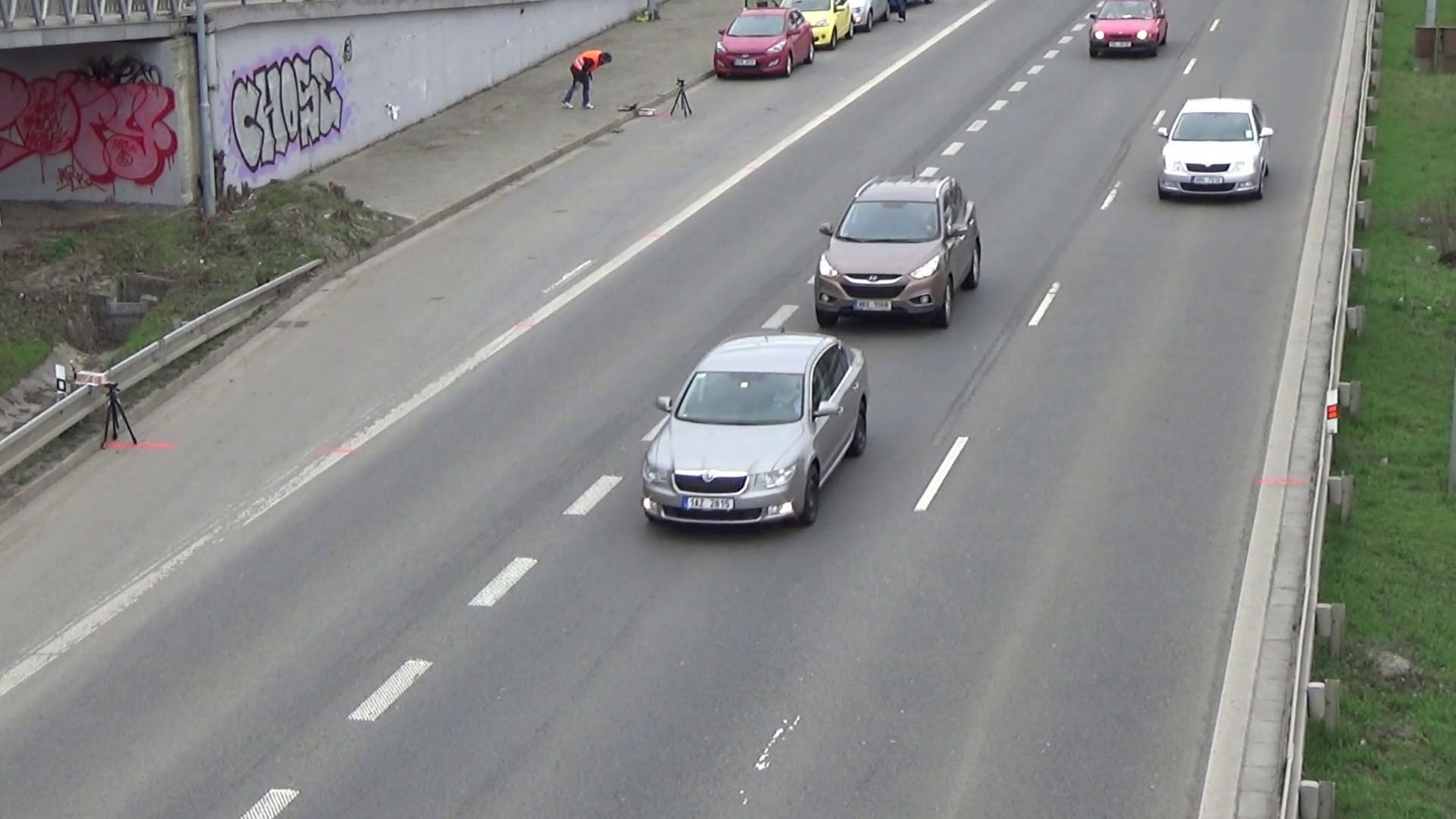}\quad
		\includegraphics[width=0.3\linewidth]{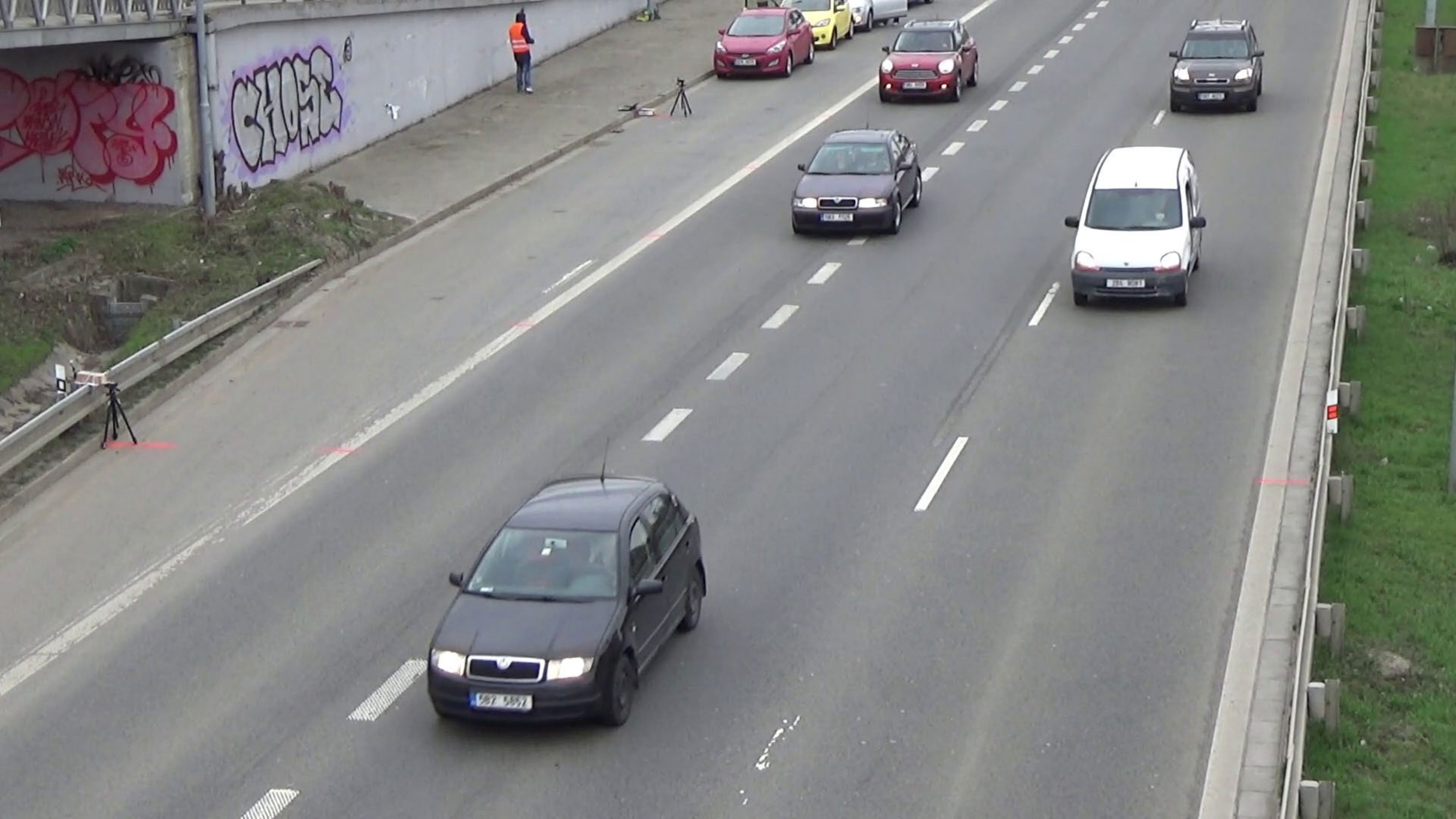}\quad
		\includegraphics[width=0.3\linewidth]{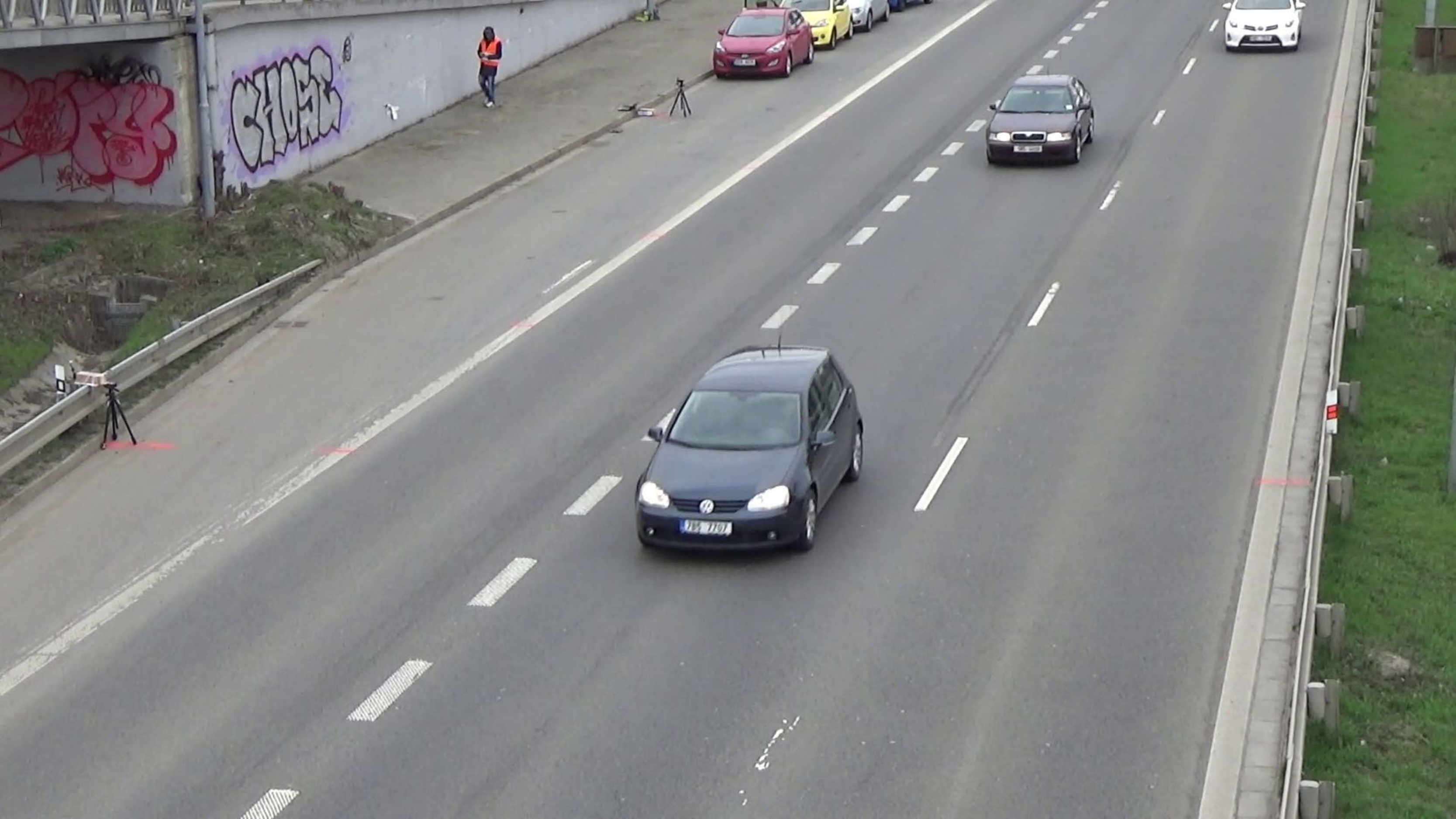}
	}
	\subcaptionbox{\centering Scene
		D\label{subfig:dataset_samples_d}}
	{%
		\includegraphics[width=0.3\linewidth]{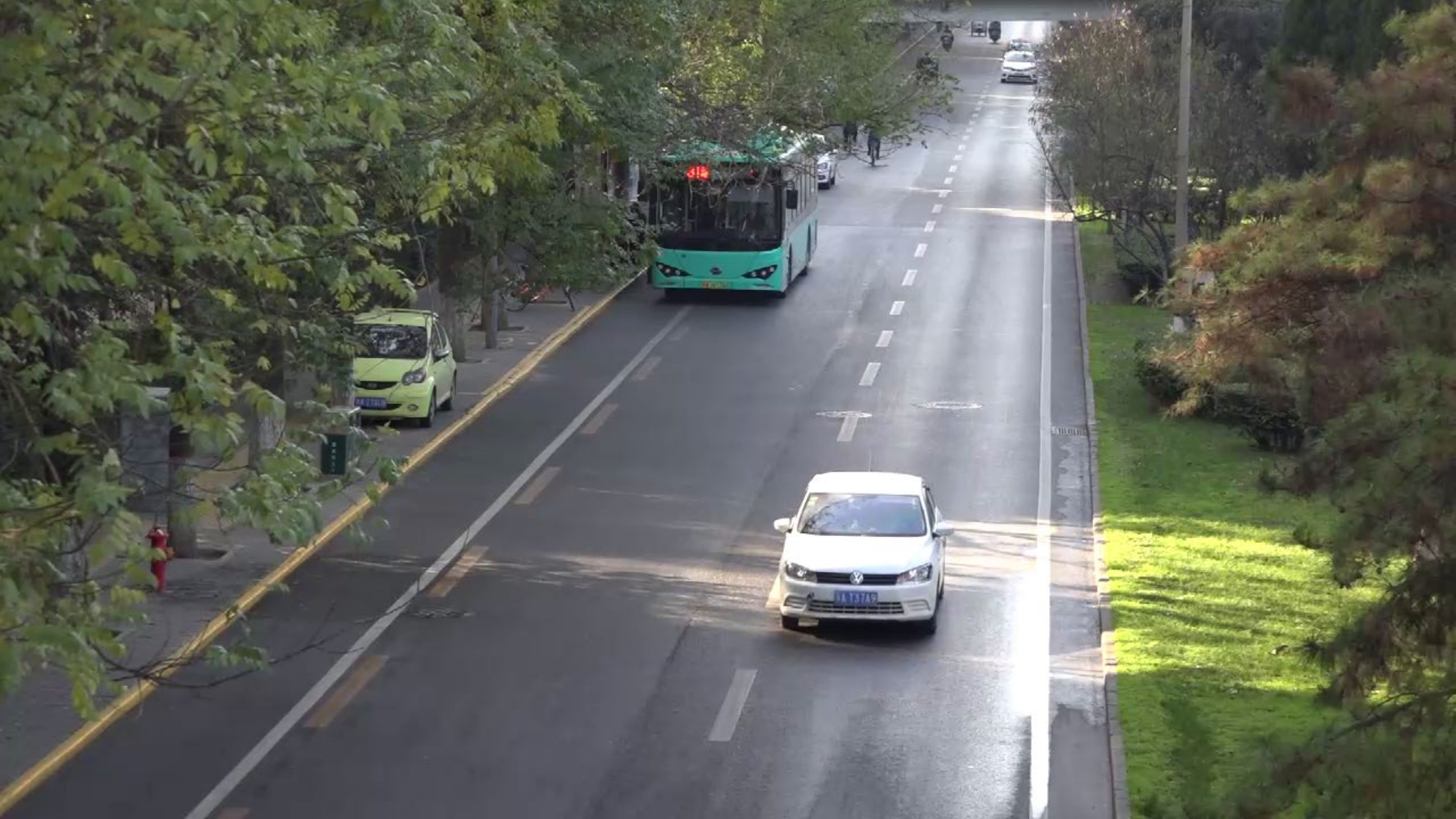}\quad
		\includegraphics[width=0.3\linewidth]{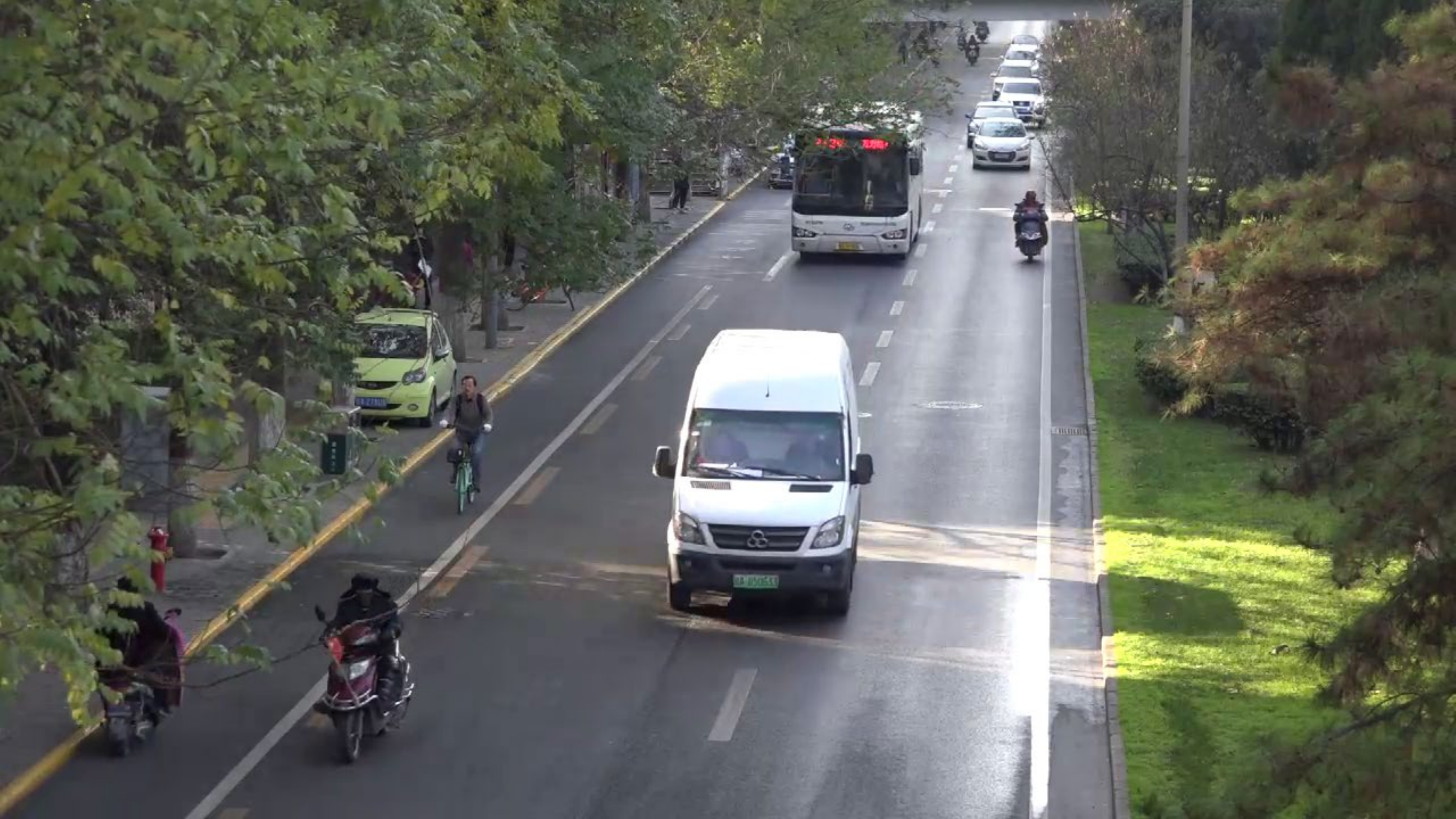}\quad
		\includegraphics[width=0.3\linewidth]{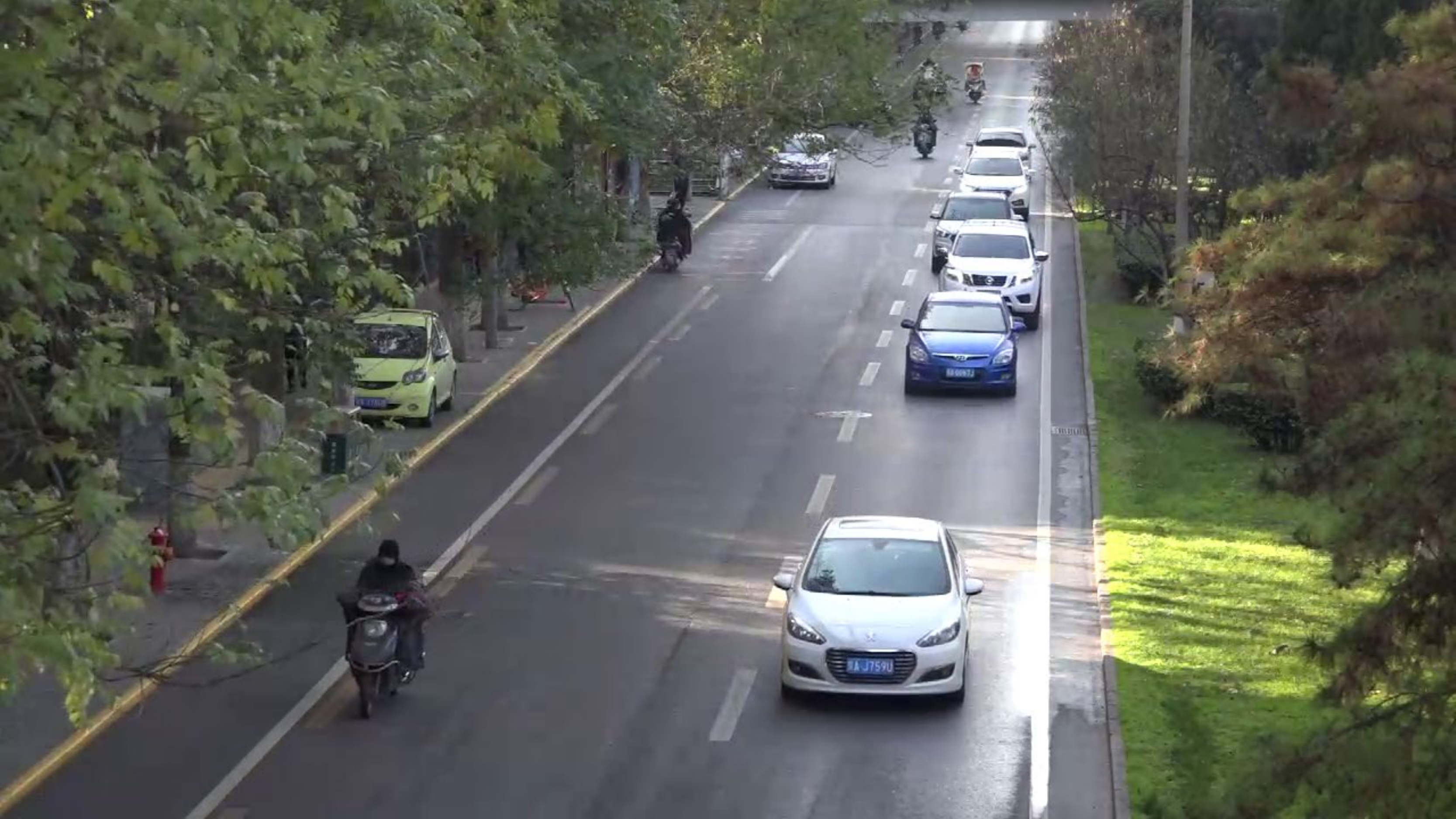}
	}
	\subcaptionbox{\centering Scene
		E\label{subfig:dataset_samples_e}}
	{%
		\includegraphics[width=0.3\linewidth]{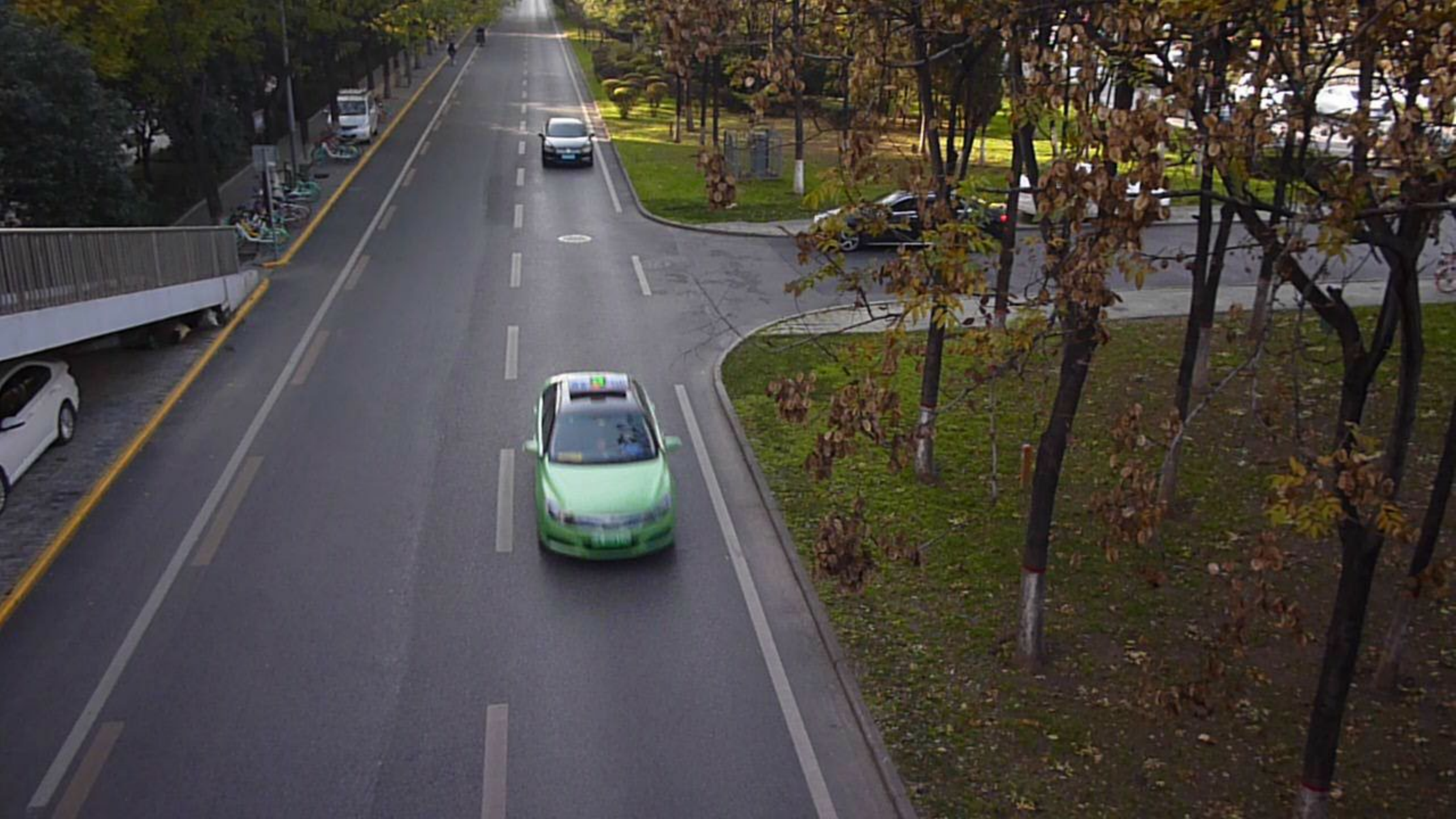}\quad
		\includegraphics[width=0.3\linewidth]{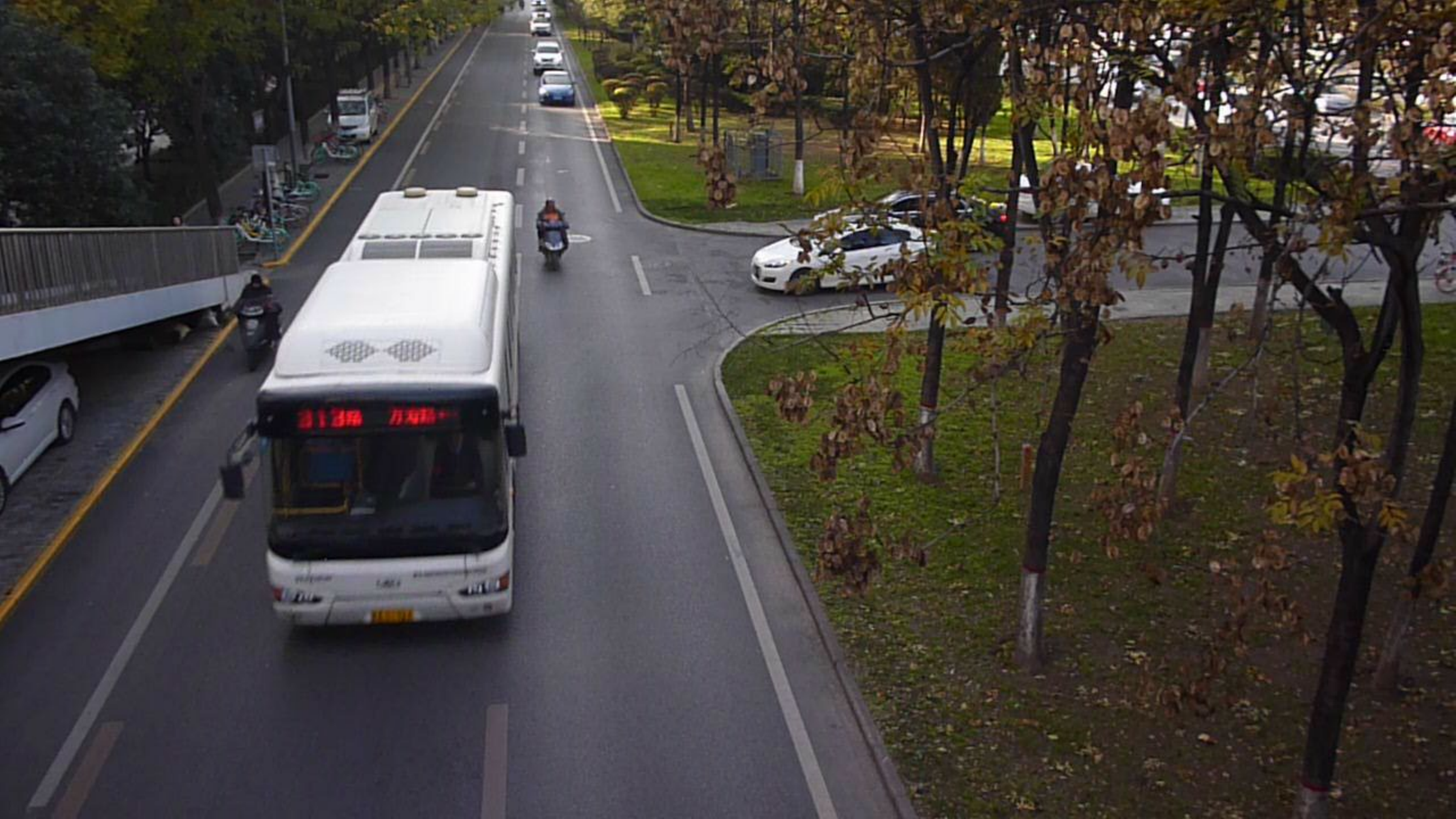}\quad
		\includegraphics[width=0.3\linewidth]{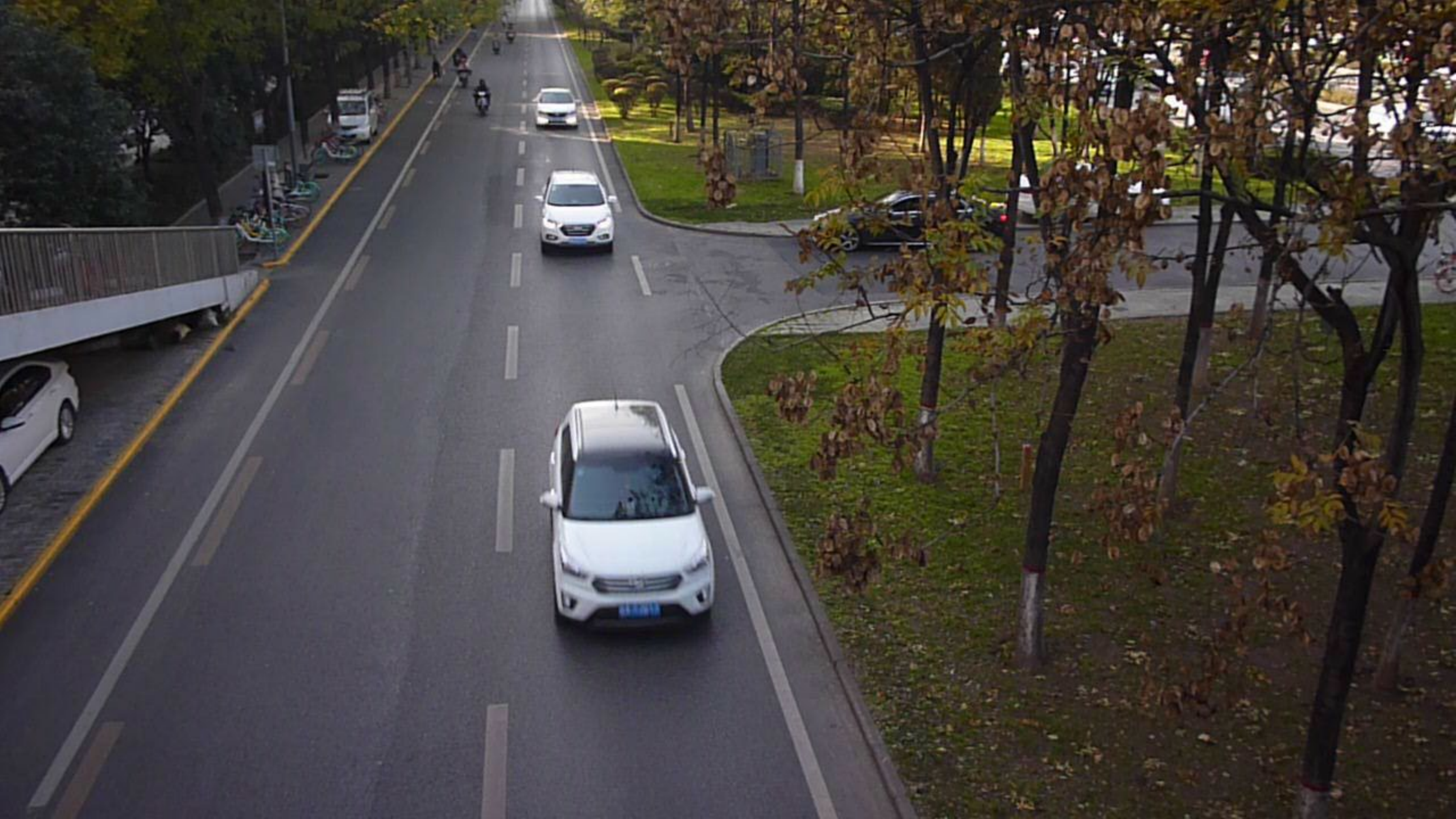}
	}
	\caption{\leftskip=0pt \rightskip=0pt plus 0cm Samples in SVLD-3D. Five different scenes are listed from the first to the fifth row.\label{fig:dataset_samples}}
\end{figure}

Scenes in SVLD-3D dataset are from BrnoCompSpeed \cite{2019brnocompspeed} and self-collected urban scenes with resolution of ${\rm{1920}} \times {\rm{1080}}$ and ${\rm{1080}} \times {\rm{720}}$, respectively. BrnoCompSpeed is public and provided by Brno University of Technology, which contains six highway scenes with three views (left, center, and right). Each frame contains ground truth road marks and vanishing points (used for camera calibration (section \ref{subsec:camera calibration})). Each vehicle contains ground truth location and speed collected by Lidar and GPS (used for vehicle dimension annotation (section \ref{subsec:label proc}) and training loss function (section \ref{subsubsec:loss func})). However, only highway scenes with low traffic volumes and single-vehicle types exist in BrnoCompSpeed. To further expand dataset diversity, urban scenes with more vehicle types, congestion, and occlusion are also included in SVLD-3D.

SVLD-3D contains five typical scenes with three vehicle types (car, truck, and bus), with a total of 14593 images in the training and validation dataset and 2273 images in the test dataset. Some samples in SVLD-3D are shown in Figure \ref{fig:dataset_samples}. Table \ref{tab:table_calib_results} shows detailed information of different scenes in the dataset, where the effective field of view ${D_r} = ({D_{ry}},{D_{rx}})$ is the maximum distance that the roadside camera can perceive along and perpendicular to the road direction.

\begin{table}[htbp]
	\centering
	\caption{Details of different scenes in SVLD-3D.}
	\label{tab:table_calib_results}
	\begin{tabular}{ccccccc}
		\toprule
		\multirow{2}{*}{Scene} & \multirow{2}{*}{${D_{ry}}$} & \multirow{2}{*}{${D_{rx}}$} & \multicolumn{4}{c}{Camera Calibration Parameters} \\ \cline{4-7}
		&                             &                             & $f$      & $\phi /rad$  & $\theta /rad$  & $h/mm$  \\ \midrule
		A                      & 120                         & 25                          & 2878.13  & 0.17874      & 0.26604        & 10119.08 \\
		B                      & 120                         & 25                          & 3994.17  & 0.15717      & 0.35346        & 8071.00   \\
		C                      & 60                          & 15                          & 3384.25  & 0.26295      & -0.24869        & 8126.49 \\
		D                      & 80                          & 10                          & 3743.78  & 0.11225      & -0.07516        & 7353.40 \\
		E                      & 60                          & 10                          & 1142.26  & 0.33372      & 0.14387        & 7166.44 \\
		\bottomrule
	\end{tabular}
\end{table}

\subsection{Label Process}
\label{subsec:label proc}
Since the original data in SVLD-3D only contains images and 3D vehicle location provided by Lidar, we develop an annotation tool LabelImg-3D and relabel vehicles to obtain ground truth vehicle type, centroid, vertexes, and dimensions of 3D bounding boxes. The ground truth 3D and 2D vehicle centroid, and dimension are denoted as $P_{cen}^{gt} = (x_{cen}^{gt},y_{cen}^{gt},z_{cen}^{gt})$, $p_{cen}^{gt} = (u_{cen}^{gt},v_{cen}^{gt})$ and $D_v^{gt} = (l_v^{gt},w_v^{gt},h_v^{gt})$.

Compared with variable views of autonomous vehicles, the roadside camera usually has certain installation standards and the pan angle is relatively fixed. Therefore, only the scenes with typical pan angles are selected. To reduce label effort, previous work \cite{2021Spatial} and 2D bounding boxes provided by YOLOv4 \cite{2020yolov4} are used as guidance. The specific label steps are as follows:
\begin{enumerate}[(1)]
	\item Vehicle type. If the vehicle type obtained by YOLOv4 is consistent with the ground truth, label this type as ground truth. Otherwise, adjust to the correct type in LabelImg-3D.
	\item Vehicle dimension. Firstly, YOLOv4 is used to obtain 2D vehicle bounding boxes as guidance. Secondly, the geometric constraint of 3D boxes fitting closely to 2D is used to adjust vehicle dimension by camera calibration. At the same time, vehicle dimensions can also be obtained by observing vehicle models and referring to relevant documents during the labeling process. These two sub-steps can be helpful to obtain more accurate vehicle dimensions $D_v^{gt}$.
	\item Vehicle centroid. $p_{cen}^{gt}$ can be obtained by Equation \ref{equa_xyz2uv} when $P_{cen}^{gt}$ is known by lidar.
	\item Vehicle vertexes. Based on $P_{cen}^{gt}$ and $D_v^{gt}$ in step (2), $p_i^{gt}$ can be obtained by Table \ref{tab:table_3d_gt_vertexes} and Equation \ref{equa_xyz2uv}.
\end{enumerate}

\begin{table}[htbp]
	\centering
	\caption{Calculation of 3D bounding box ground truth vertexes in world space.}
	\label{tab:table_3d_gt_vertexes}
	\begin{tabular}{cc}
		\toprule
		Vertex & World coordinate \\
		\midrule
		$P_1^{gt}$& $(x_{cen}^{gt} + {{w_v^{gt}} \mathord{\left/
				{\vphantom {{w_v^{gt}} 2}} \right.
				\kern-\nulldelimiterspace} 2},y_{cen}^{gt} - {{l_v^{gt}} \mathord{\left/
				{\vphantom {{l_v^{gt}} 2}} \right.
				\kern-\nulldelimiterspace} 2},z_{cen}^{gt} - {{h_v^{gt}} \mathord{\left/
				{\vphantom {{h_v^{gt}} 2}} \right.
				\kern-\nulldelimiterspace} 2})$ \\
		$P_2^{gt}$& $(x_{cen}^{gt} - {{w_v^{gt}} \mathord{\left/
				{\vphantom {{w_v^{gt}} 2}} \right.
				\kern-\nulldelimiterspace} 2},y_{cen}^{gt} - {{l_v^{gt}} \mathord{\left/
				{\vphantom {{l_v^{gt}} 2}} \right.
				\kern-\nulldelimiterspace} 2},z_{cen}^{gt} - {{h_v^{gt}} \mathord{\left/
				{\vphantom {{h_v^{gt}} 2}} \right.
				\kern-\nulldelimiterspace} 2})$ \\
		$P_3^{gt}$& $(x_{cen}^{gt} - {{w_v^{gt}} \mathord{\left/
				{\vphantom {{w_v^{gt}} 2}} \right.
				\kern-\nulldelimiterspace} 2},y_{cen}^{gt} + {{l_v^{gt}} \mathord{\left/
				{\vphantom {{l_v^{gt}} 2}} \right.
				\kern-\nulldelimiterspace} 2},z_{cen}^{gt} - {{h_v^{gt}} \mathord{\left/
				{\vphantom {{h_v^{gt}} 2}} \right.
				\kern-\nulldelimiterspace} 2})$ \\
		$P_4^{gt}$& $(x_{cen}^{gt} + {{w_v^{gt}} \mathord{\left/
				{\vphantom {{w_v^{gt}} 2}} \right.
				\kern-\nulldelimiterspace} 2},y_{cen}^{gt} + {{l_v^{gt}} \mathord{\left/
				{\vphantom {{l_v^{gt}} 2}} \right.
				\kern-\nulldelimiterspace} 2},z_{cen}^{gt} - {{h_v^{gt}} \mathord{\left/
				{\vphantom {{h_v^{gt}} 2}} \right.
				\kern-\nulldelimiterspace} 2})$ \\
		$P_5^{gt}$& $(x_{cen}^{gt} + {{w_v^{gt}} \mathord{\left/
				{\vphantom {{w_v^{gt}} 2}} \right.
				\kern-\nulldelimiterspace} 2},y_{cen}^{gt} - {{l_v^{gt}} \mathord{\left/
				{\vphantom {{l_v^{gt}} 2}} \right.
				\kern-\nulldelimiterspace} 2},z_{cen}^{gt} + {{h_v^{gt}} \mathord{\left/
				{\vphantom {{h_v^{gt}} 2}} \right.
				\kern-\nulldelimiterspace} 2})$ \\
		$P_6^{gt}$& $(x_{cen}^{gt} - {{w_v^{gt}} \mathord{\left/
				{\vphantom {{w_v^{gt}} 2}} \right.
				\kern-\nulldelimiterspace} 2},y_{cen}^{gt} - {{l_v^{gt}} \mathord{\left/
				{\vphantom {{l_v^{gt}} 2}} \right.
				\kern-\nulldelimiterspace} 2},z_{cen}^{gt} + {{h_v^{gt}} \mathord{\left/
				{\vphantom {{h_v^{gt}} 2}} \right.
				\kern-\nulldelimiterspace} 2})$ \\
		$P_7^{gt}$& $(x_{cen}^{gt} - {{w_v^{gt}} \mathord{\left/
				{\vphantom {{w_v^{gt}} 2}} \right.
				\kern-\nulldelimiterspace} 2},y_{cen}^{gt} + {{l_v^{gt}} \mathord{\left/
				{\vphantom {{l_v^{gt}} 2}} \right.
				\kern-\nulldelimiterspace} 2},z_{cen}^{gt} + {{h_v^{gt}} \mathord{\left/
				{\vphantom {{h_v^{gt}} 2}} \right.
				\kern-\nulldelimiterspace} 2})$ \\
		$P_8^{gt}$& $(x_{cen}^{gt} + {{w_v^{gt}} \mathord{\left/
				{\vphantom {{w_v^{gt}} 2}} \right.
				\kern-\nulldelimiterspace} 2},y_{cen}^{gt} + {{l_v^{gt}} \mathord{\left/
				{\vphantom {{l_v^{gt}} 2}} \right.
				\kern-\nulldelimiterspace} 2},z_{cen}^{gt} + {{h_v^{gt}} \mathord{\left/
				{\vphantom {{h_v^{gt}} 2}} \right.
				\kern-\nulldelimiterspace} 2})$ \\
		\bottomrule
	\end{tabular}
\end{table}

\begin{figure}[htbp]
	\centering
	\subcaptionbox{\centering Flowchart of label process\label{subfig:flowchart_label_process}}
	{%
		\includegraphics[width=0.5\linewidth]{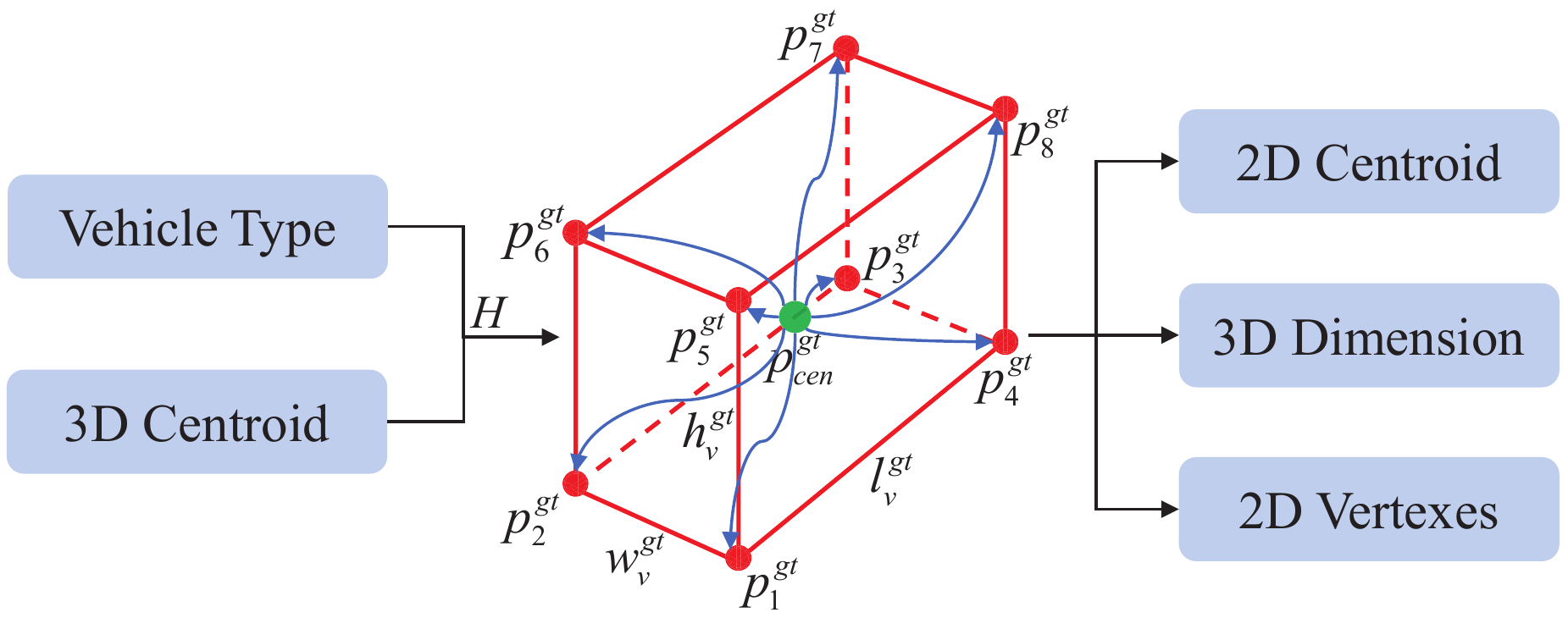}%
	}
	\subcaptionbox{\centering Visualization of annotations\label{subfig:vis_annotation}}
	{%
		\includegraphics[width=0.38\linewidth]{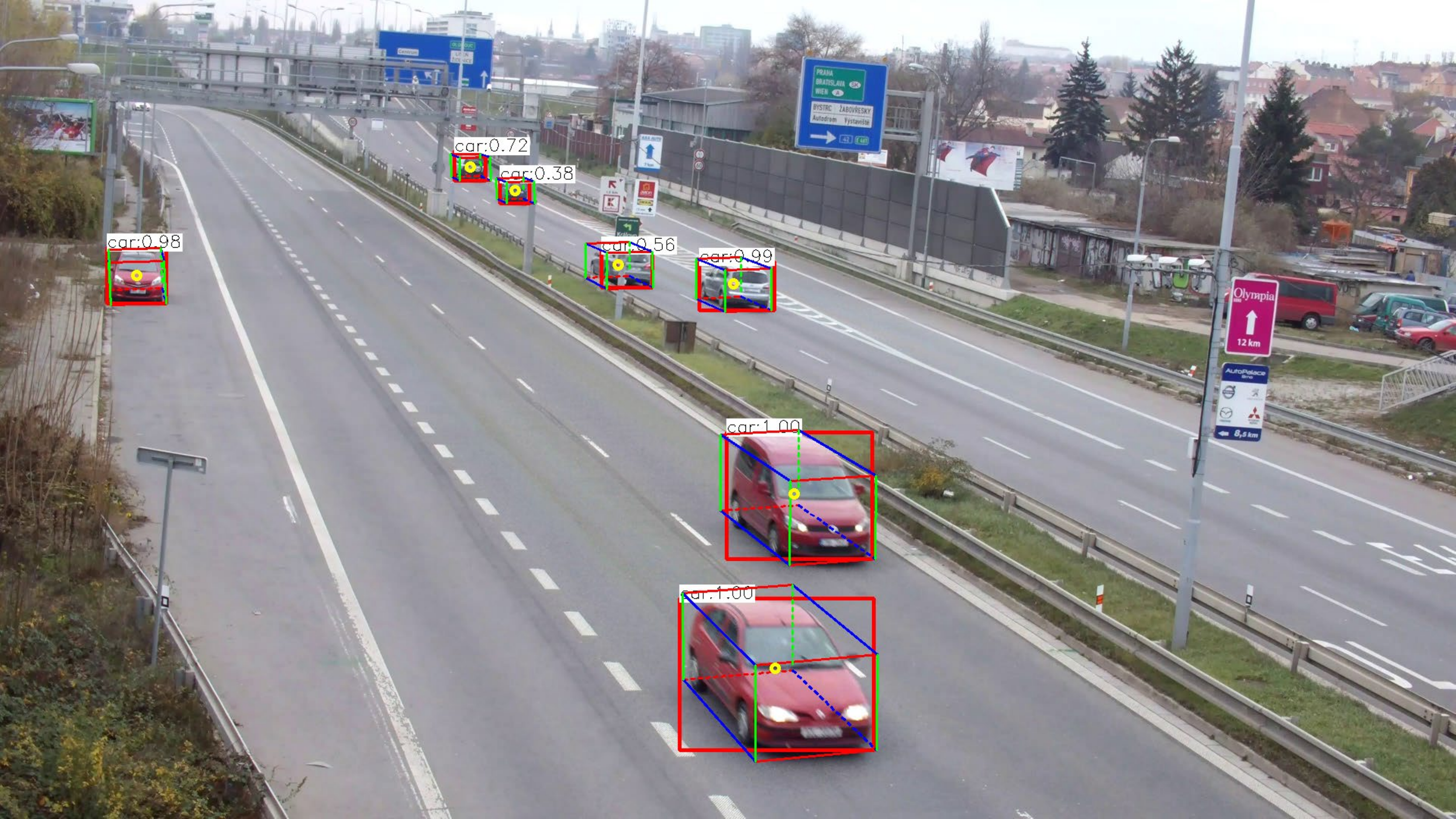}%
	}
	\caption{Label process and visualization.}
	\label{fig:fig_draw_bbox_result}
\end{figure}

Finally, vehicle type, centroid, vertexes, and dimensions are recorded in the annotation file. Figure \ref{fig:fig_draw_bbox_result} shows the flowchart of label process and visualization of annotations.

\section{Experimental Protocols}
\label{section:experimental protocols}

In this section, we introduce implementation details and evaluation metrics for our experiments.

\subsection{Implementation Details}
\label{subsec:implem datails}
We implement our network using PyTorch platform with Core i7-8700 CPU and one GTX 1080Ti GPU. The original image is scaled to $512 \times 512$ for training and testing. We split the dataset into the training set and validation set with a ratio of 9:1. We use Adam optimizer with a base learning rate of 0.001 for 100 epochs. The learning rate reduces by a factor of 10 when validation loss no longer decreases for three continuous epochs. The pretrained model on ImageNet is used for fine-tuning. We train our network using weights freezing in the first 60 epochs with a batch size of 16. The batch size drops $2 \times$ in the rest epochs. When validation loss no longer decreases for 7 continuous epochs, the training process will be stopped by early stopping.

\begin{figure}[htbp]
	\centering
	\subfloat[\centering ]{\includegraphics[width=0.22\linewidth]{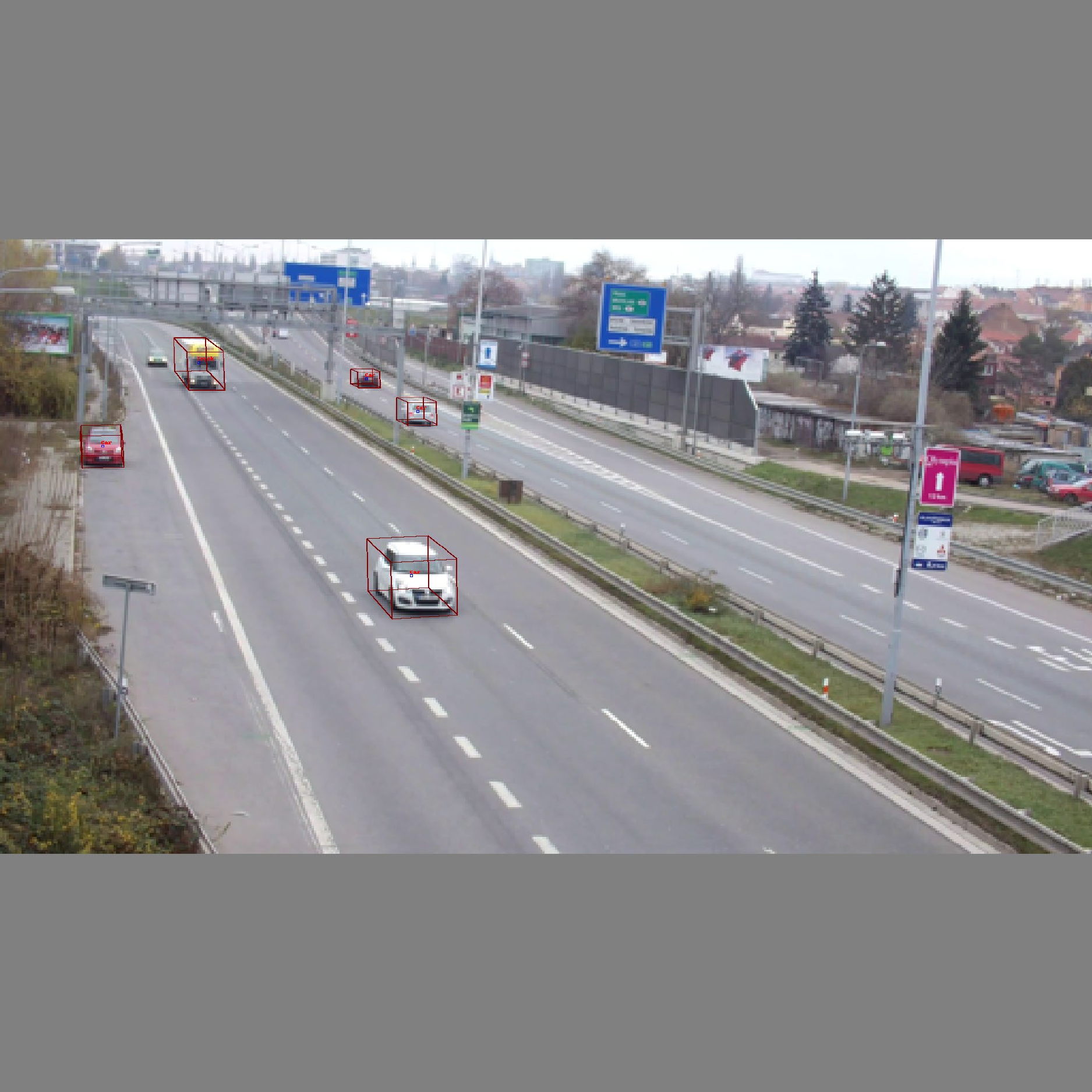}%
		\label{fig:data_aug_a}}
	\hfil
	\subfloat[\centering ]{\includegraphics[width=0.22\linewidth]{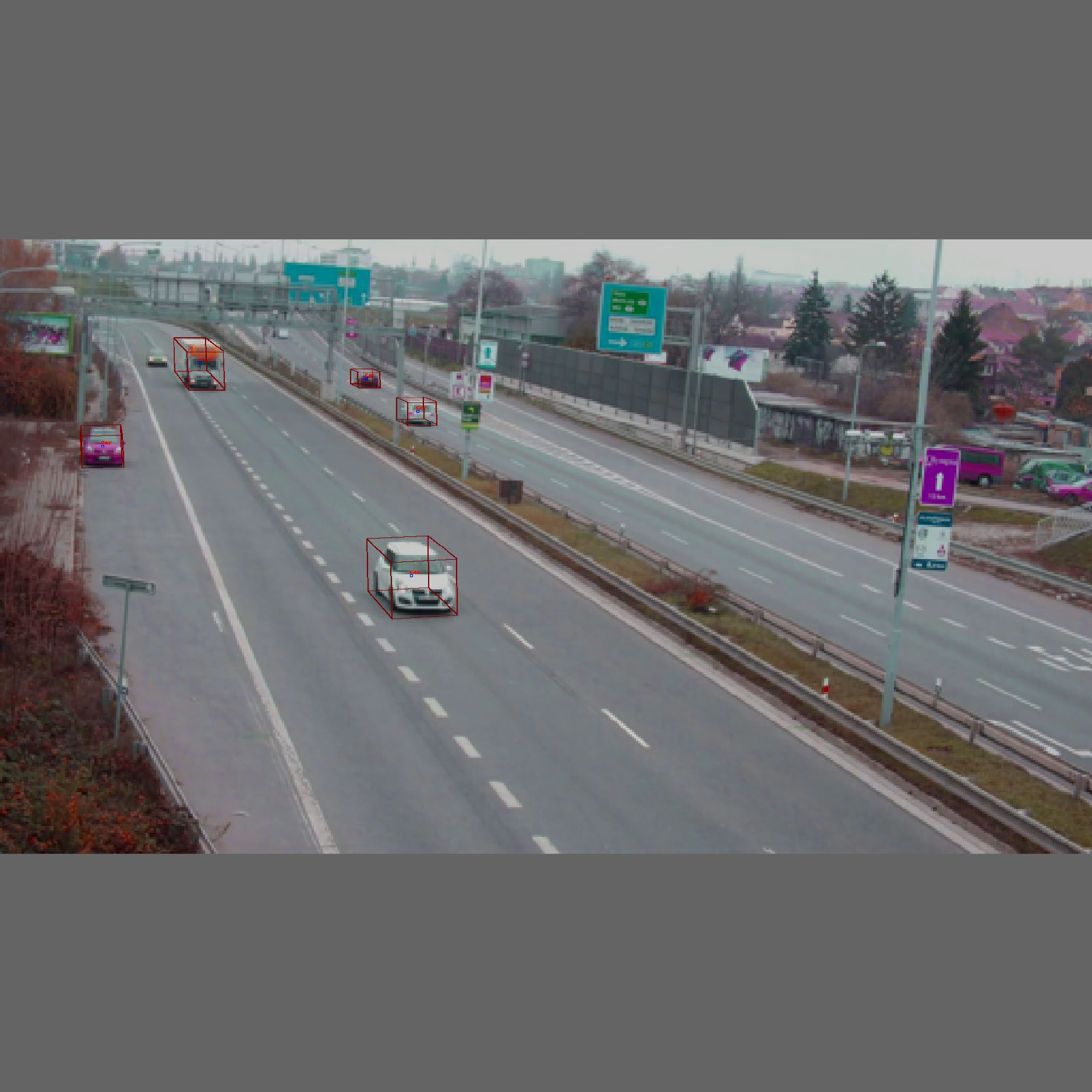}%
		\label{fig:data_aug_b}}
	\hfil
	\subfloat[\centering ]{\includegraphics[width=0.22\linewidth]{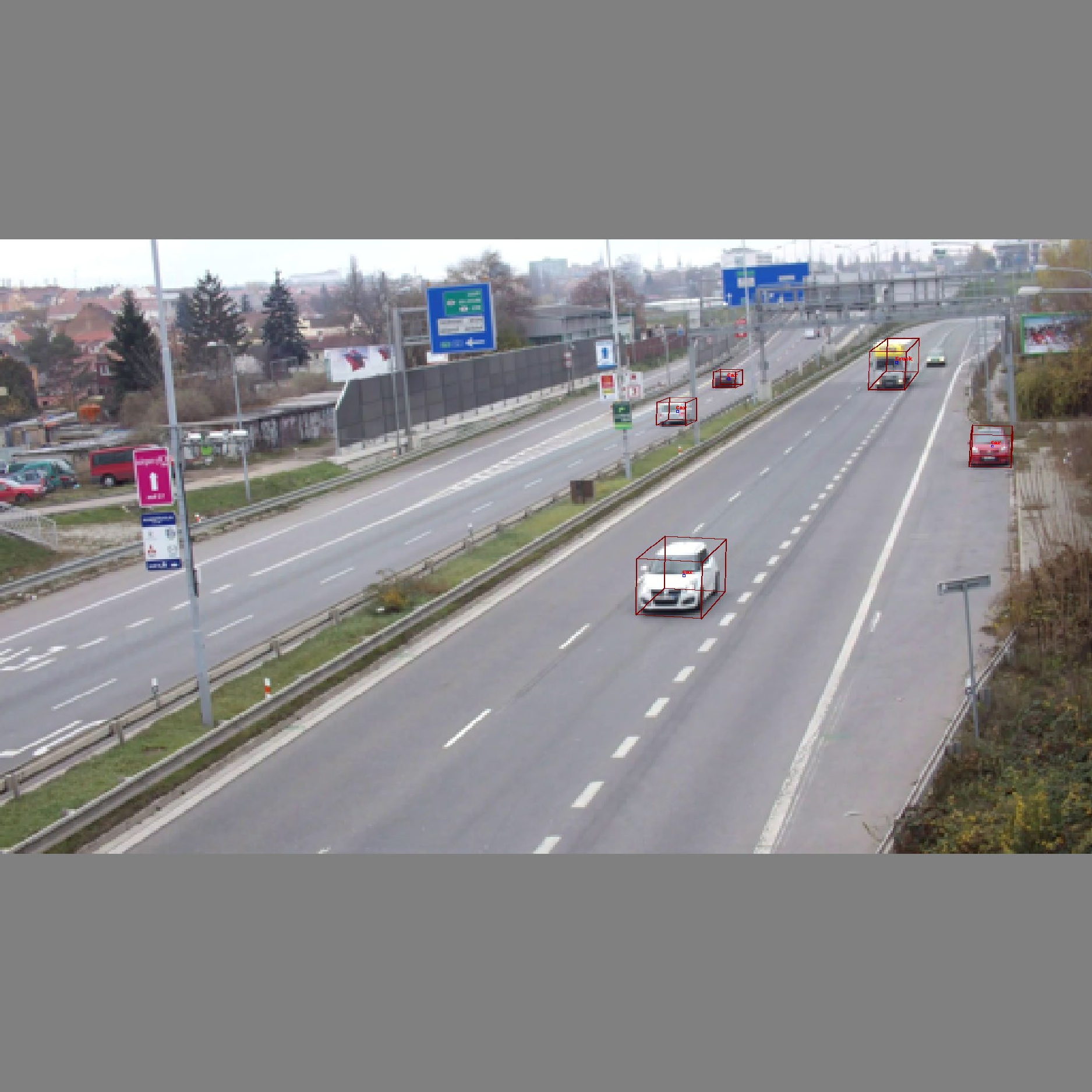}%
		\label{fig:data_aug_c}}
	\hfil
	\subfloat[\centering ]{\includegraphics[width=0.22\linewidth]{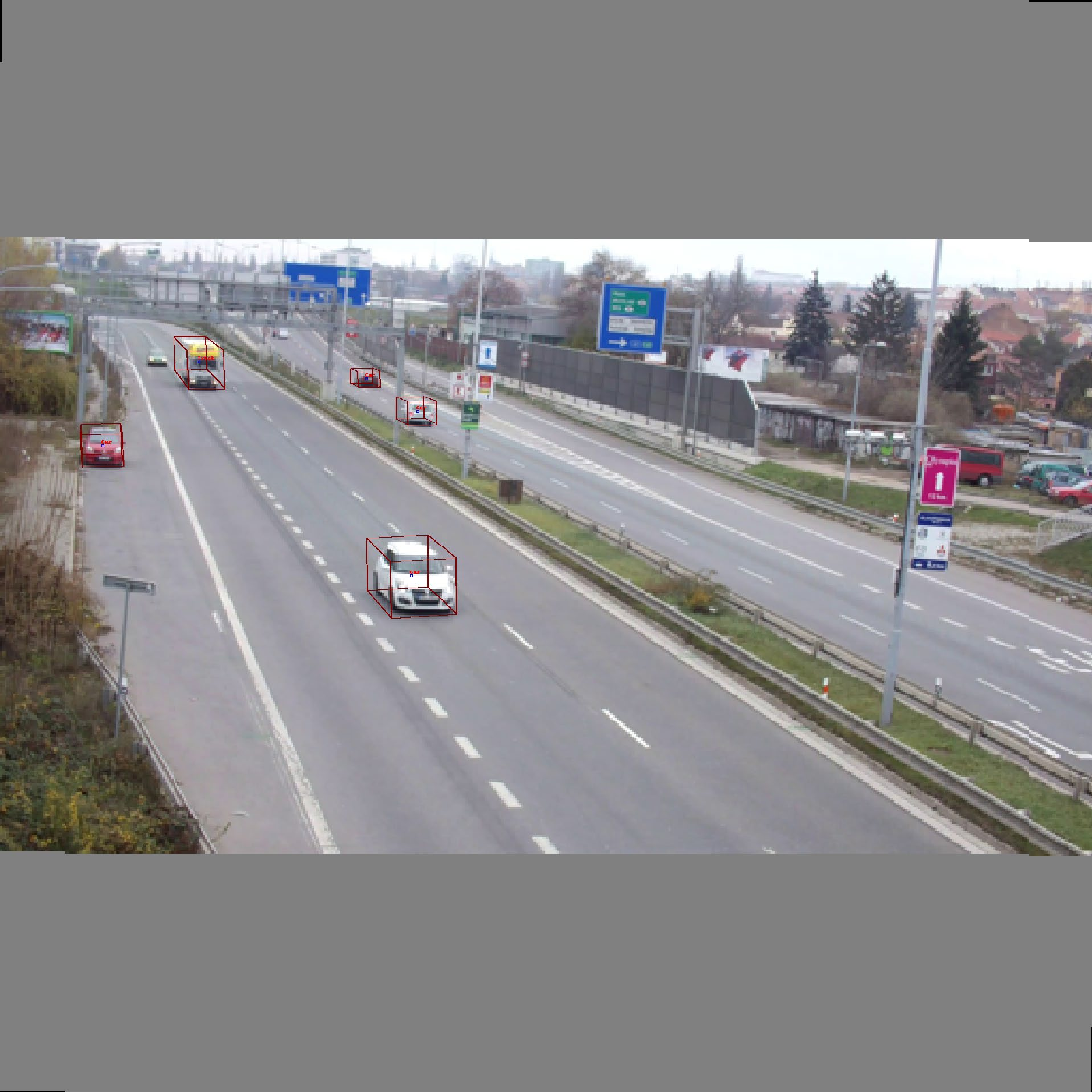}%
		\label{fig:data_aug_d}}
	\newline
	\subfloat[\centering ]{\includegraphics[width=0.22\linewidth]{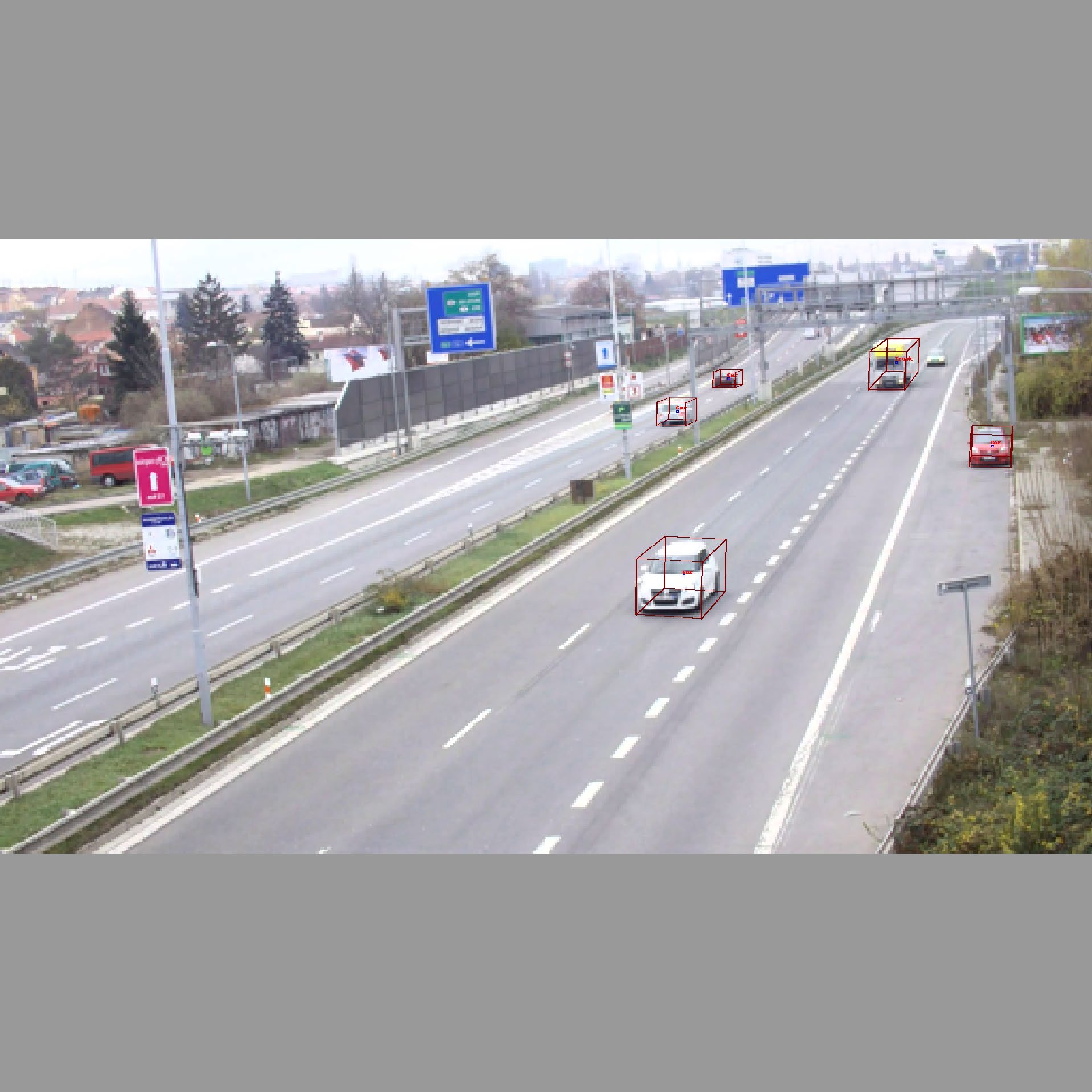}%
		\label{fig:data_aug_e}}
	\hfil
	\subfloat[\centering ]{\includegraphics[width=0.22\linewidth]{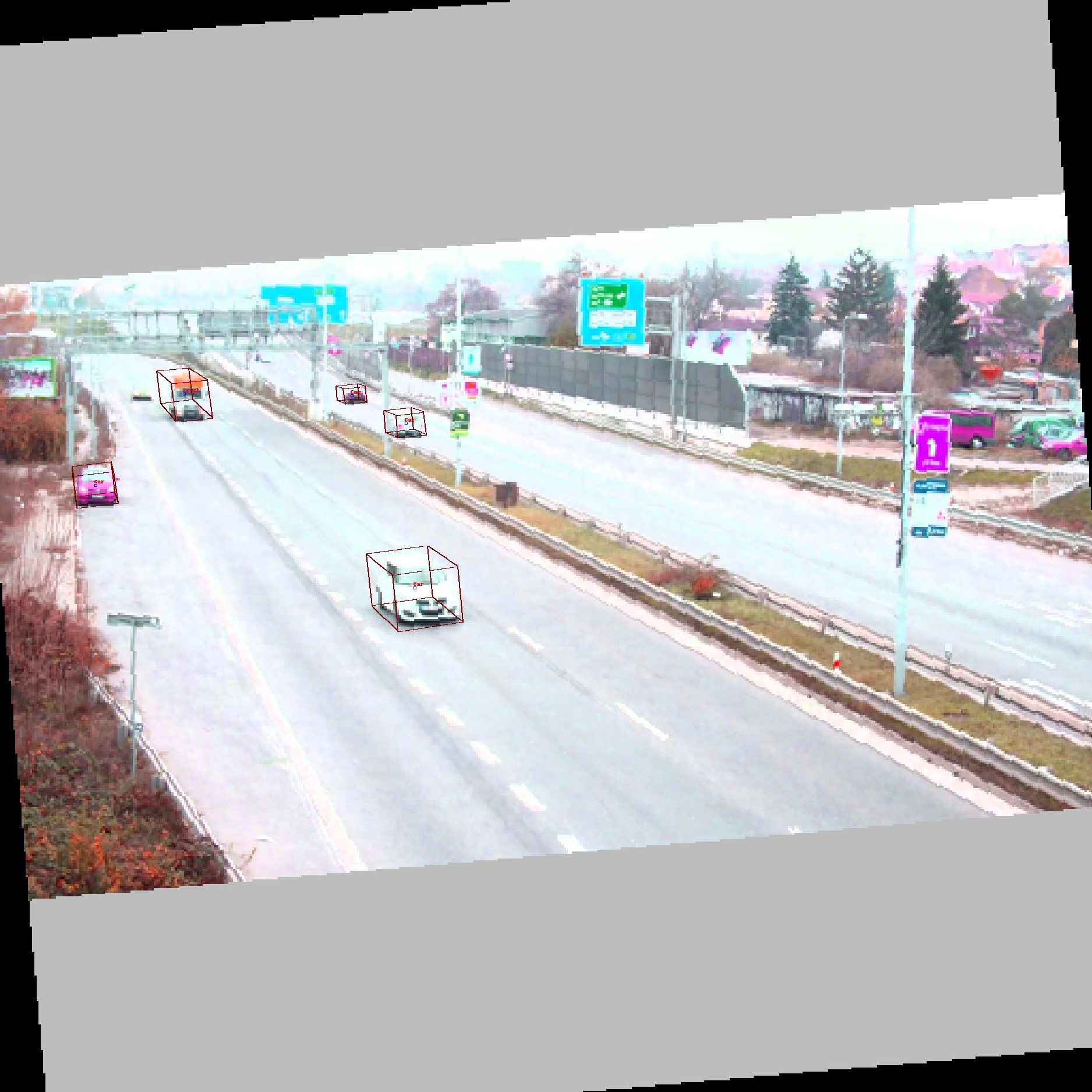}%
		\label{fig:data_aug_f}}
	\hfil
	\subfloat[\centering ]{\includegraphics[width=0.22\linewidth]{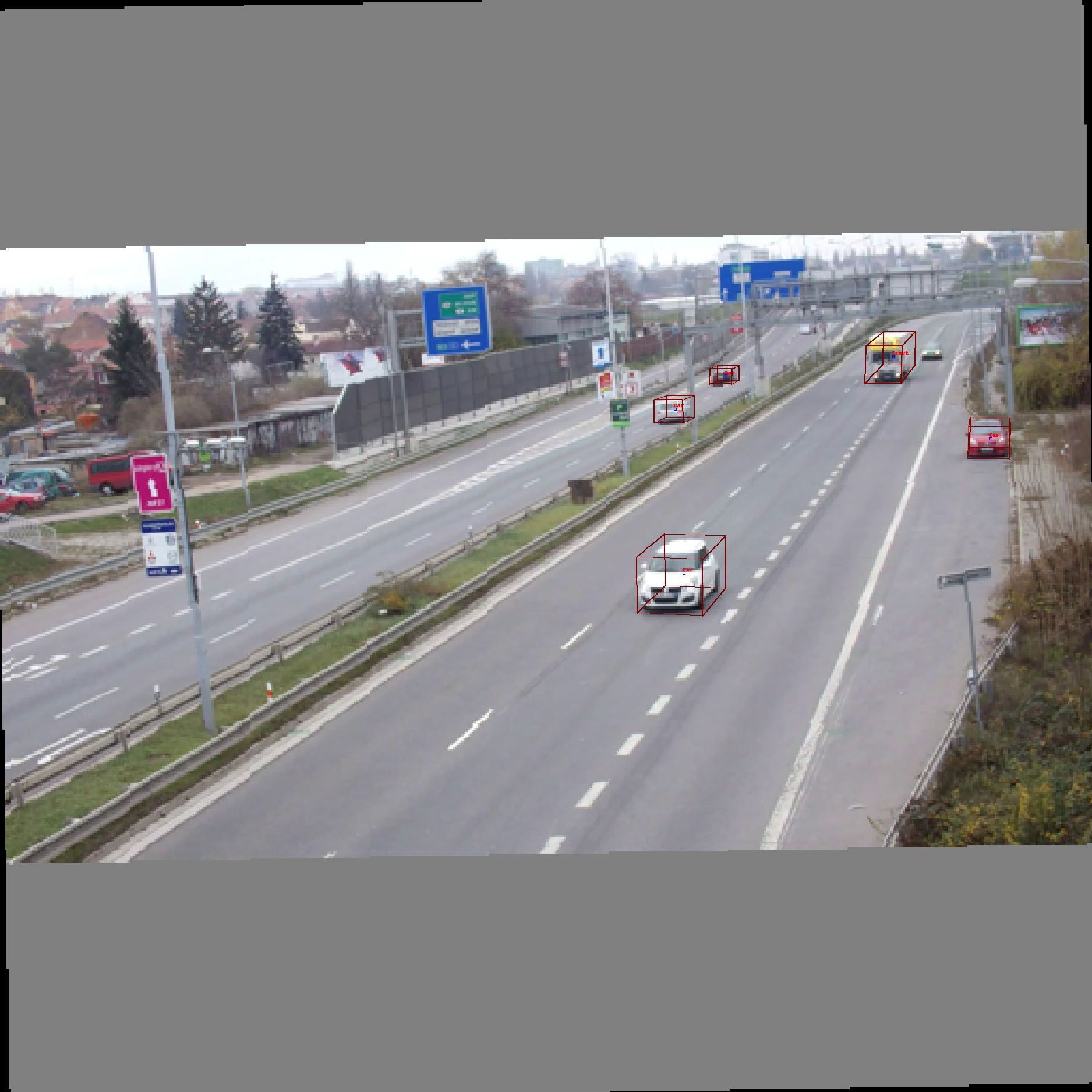}%
		\label{fig:data_aug_g}}
	\hfil
	\subfloat[\centering ]{\includegraphics[width=0.22\linewidth]{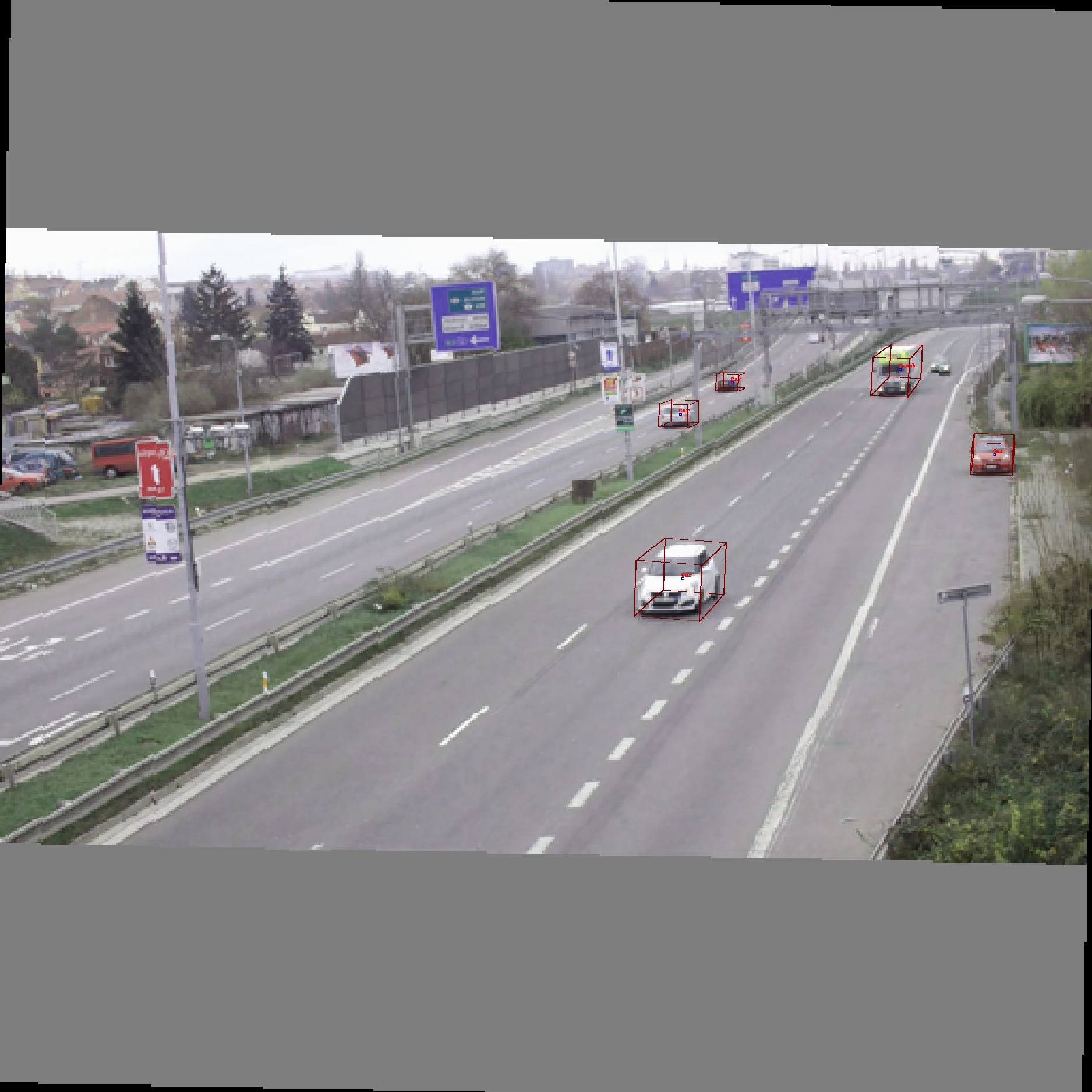}%
		\label{fig:data_aug_h}}
	\newline
	\caption{\leftskip=0pt \rightskip=0pt plus 0cm Schematic diagram of data augmentation. (a) No augmentation. (b) Color jitter (CJ). (c) Horizontal flip (HF). (d) Perspective transformation (PT). (e) CJ + HF. (f) CJ + PT. (g) HF + PT. (h) CJ + HF + PT.}
	\label{fig:data_aug}
\end{figure}

We use random color jitter, horizontal flip, and perspective transformation as image augmentation. As shown in Figure \ref{fig:data_aug}, perspective transformation is used as the simulation of roadside view change according to the camera imaging principle. After data augmentation, the training and validation set can be expanded to 58372 images, which quadruples the original volume.

\subsection{Evaluation Metrics}
\label{subsec:eval metrics}

Evaluation metrics include average precision (AP), frame per second (FPS), precision and error of 3D vehicle localization and dimension prediction.

\subsubsection{Average Precision and Speed}
\label{subsubsec:ap and fps}

Referring to evaluation metrics of existing 3D vehicle detection datasets, we use $A{P_{3D}}$ \cite{2016mono3ddet} to evaluate 3D average precision and FPS to evaluate speed.

$A{P_{3D}}$ is similar to $A{P_{2D}}$ in 2D detection provided by VOC dataset \cite{2010pascalvoc}. Differently, 3D IoU is used instead of 2D in calculating $A{P_{3D}}$. The equation of $AP$ can be expressed as:
\begin{equation}
	\label{equa_ap}
	\begin{array}{l}
		AP = \frac{1}{{11}}\sum\limits_{r \in \{ 0,0.1, \ldots ,1\} } {{p_{{\rm{interp}}}}(r)} \\
		{p_{{\rm{interp}}}}(r) = \mathop {\max }\limits_{\tilde r:\tilde r \ge r} p(\tilde r)
	\end{array}
\end{equation}
The precision at each recall level $r$ is interpolated by taking the maximum precision measured for a method for which the corresponding recall exceeds $r$, $p(\tilde r)$ is the measured precision at recall $\tilde r$.

The equation of FPS is defined as:
\begin{equation}
	\label{equa_fps}
	FPS = {1 \mathord{\left/
			{\vphantom {1 {{t_{proc}}}}} \right.
			\kern-\nulldelimiterspace} {{t_{proc}}}}
\end{equation}
where $t_{proc}$ represents the processing time (measured in second) of the network for a single frame.

\subsubsection{3D Vehicle Localization Precision and Error}
\label{subsubsec:3dlocpe}

Combined with camera calibration in Section \ref{subsec:camera calibration}, 3D vehicle centroids can be further obtained by Equation \ref{equa_uv2xyz} and used for 3D vehicle localization.

The predicted and ground-truth 3D vehicle centroid are denoted as $P_{cen}^{pred} = (x_{cen}^{pred},y_{cen}^{pred},z_{cen}^{pred})$ and $P_{cen}^{gt} = (x_{cen}^{gt},y_{cen}^{gt},z_{cen}^{gt})$.

3D vehicle localization precision and error can be defined as follows:
\begin{equation}
	\label{equa_ploc}
	{P_{loc}} = (1 - \sum\limits_{k \in \{ x,y\} } {\frac{{\lvert {k_{cen}^{pred} - k_{cen}^{gt}} \rvert}}{{{{{D_{rk}}} \mathord{\left/
						{\vphantom {{{D_{rk}}} 2}} \right.
						\kern-\nulldelimiterspace} 2}}}} ) \times 100\%
\end{equation}

\begin{equation}
	\label{equa_eloc}
	{E_{loc}} = \sum\limits_{k \in \{ x,y\} } {\lvert {k_{cen}^{pred} - k_{cen}^{gt}} \rvert}
\end{equation}
where ${D_r}$ can be found in Table \ref{tab:table_calib_results}.

\subsubsection{3D Vehicle Dimension Precision and Error}
\label{subsubsec:3dsizepe}

We design a 3D vehicle dimension regression branch in the network, which can be used for 3D vehicle dimension prediction, including length, width, and height in meters. 

The predicted and ground-truth 3D dimension are denoted as $D_v^{pred} = (l_v^{pred},w_v^{pred},h_v^{pred})$ and $D_v^{gt} = (l_v^{gt},w_v^{gt},h_v^{gt})$.

3D vehicle dimension precision and error can be defined as follows:
\begin{equation}
	\label{equa_pdim}
	{P_{dim}} = (1 - \sum\limits_{k \in \{ {l_v},{w_v},{h_w}\} } {\frac{{\lvert {{k^{pred}} - {k^{gt}}} \rvert}}{{{k^{gt}}}}} ) \times 100\%
\end{equation}

\begin{equation}
	\label{equa_edim}
	{E_{dim}} = \sum\limits_{k \in \{ {l_v},{w_v},{h_w}\} } {\lvert {{k^{pred}} - {k^{gt}}} \rvert}
\end{equation}

\section{Results and Discussions}
\label{section:results and discussions}

In this section, we provide experimental results, an ablation study and discussions to demonstrate the effectiveness of CenterLoc3D.

\subsection{Average Precision and Speed of CenterLoc3D}
\label{subsec:ap and fps of centerloc3d}

$A{P_{3D}}$ and FPS of different monocular 3D vehicle detection methods on validation and test set are compared in Table \ref{tab:table_compare_kitti}, which are calculated by Equation \ref{equa_ap} and Equation \ref{equa_fps}. In addition, FLOPS and parameters are also important metrics, with 28.61GFlops and 34.95M for our network. KITTI validation and test set are used in onboard scenes at easy, moderate, and hard settings, which are determined by vehicle size in image space. We use the proposed SVLD-3D dataset for experimental validation. The IoU thresholds are 0.5 and 0.7 respectively. SVLD-3D dataset has only one validation set without difficulty settings.

\begin{table*}[htbp]
	\centering
	\caption{Comparison of $A{P_{3D}}$ and FPS of different monocular 3D vehicle detection methods.}
	\label{tab:table_compare_kitti}
	\resizebox{\textwidth}{!}{%
		\begin{tabular}{ccccccccccc}
			\toprule
			\multirow{2}{*}{Method}  & \multirow{2}{*}{Scene}    & \multirow{2}{*}{Backbone} & \multirow{2}{*}{GPU}          & \multicolumn{3}{c}{\begin{tabular}[c]{@{}c@{}}$A{P_{3D}}(IOU > 0.5)$ \\ $[va{l_1}/va{l_2}]$ \end{tabular}} & \multicolumn{3}{c}{\begin{tabular}[c]{@{}c@{}}$A{P_{3D}}(IOU > 0.7)$\\ $[va{l_1}/va{l_2}/test]$\end{tabular}} & \multirow{2}{*}{FPS} \\ \cline{5-10}
			&                           &                           &                               & Easy                                         & Moderate                                     & Hard                                         & Easy                                           & Moderate                                       & Hard                                          &                      \\
			\midrule
			MonoGRNet \cite{2019monogrnet}                & onboard                   & VGG-16                    & GTX Titan X                   & 50.51/54.21                                  & 36.97/39.69                                  & 30.82/33.06                                  & 13.88 / 24.97 / -                              & 10.19 / 19.44 / -                              & 7.62 / 16.30 / -                              & 16.7                 \\ 
			Deep3DBox \cite{2017deep3dbox}                & onboard                   & VGG-16                    & -                             & 27.04 / -                                    & 20.55 / -                                    & 15.88 / -                                    & 5.85 / - / -                                   & 4.10 / - / -                                   & 3.84 / - / -                                  & -                    \\ 
			GS3D \cite{2019gs3d}                    & onboard                   & VGG-16                    & -                             & 32.15/30.60                                  & 29.89/26.40                                  & 26.19/22.89                                  & 13.46/11.63/7.69                               & 10.97/10.51/6.29                               & 10.38/10.51/6.16                              & 0.4                  \\ 
			\multirow{2}{*}{RTM3D \cite{2020rtm3d}}   & \multirow{2}{*}{onboard}  & ResNet-18                  & \multirow{2}{*}{GTX 1080Ti×2} & 47.43/46.52                                  & 33.86/32.61                                  & 31.04/30.95                                  & 18.13 / 18.38 / -                              & 14.14 / 14.66 / -                              & 13.33 / 12.35 / -                             & 28.6                 \\ 
			&                           & DLA-34                     &                               & 54.36/52.59                                  & 41.90/40.96                                  & 35.84/34.95                                  & 20.77/19.47/13.61                              & 16.86/16.29/10.09                              & 16.63/15.57/8.18                              & 18.2                 \\
			SMOKE \cite{2020SMOKE}                   & onboard                   & DLA-34                     & GTX TITAN X×4                 & \multicolumn{3}{c}{-}                                                                                                                     & 14.76/19.99/14.03                              & 12.85/15.61/9.76                               & 11.50/15.28/7.84                              & 33.3                 \\
			\multirow{2}{*}{KM3D \cite{2021km3d}}    & \multirow{2}{*}{onboard}  & ResNet-18                  & \multirow{2}{*}{GTX 1080Ti}   & 47.23/47.13                                  & 34.12/33.31                                  & 31.51/25.84                                  & 19.48/18.34/12.65                              & 15.32/14.91/8.39                               & 13.88/12.58/7.12                              & 47.6                 \\ 
			&                           & DLA-34                     &                               & 56.02/54.09                                  & 43.13/43.07                                  & 36.77/37.56                                  & 22.50/22.71/16.73                              & 19.60/17.71/11.45                              & 17.12/16.15/9.92                              & 25.0                 \\
			\multirow{3}{*}{Lite-FPN \cite{2021litefpn}} & \multirow{3}{*}{onboard}  & ResNet-18                  & \multirow{3}{*}{GTX 2080Ti}   & \multicolumn{3}{c}{\multirow{3}{*}{-}}                                                                                                    & 17.04 / - / -                                  & 14.02 / - / -                                  & 12.23 / - / -                                 & 88.57                \\
			&                           & ResNet-34                  &                               & \multicolumn{3}{c}{}                                                                                                                      & 18.01 / - / 15.32                              & 15.29 / - / 10.64                              & 14.28 / - / 8.59                              & 71.32                \\
			&                           & DLA-34                     &                               & \multicolumn{3}{c}{}                                                                                                                      & 19.31 / - / -                                  & 16.19 / - / -                                  & 15.47 / - / -                                 & 42.37  \\
			Ours    & roadside & ResNet-50                  & GTX 1080Ti   & \multicolumn{3}{c}{91.34 / -}                                                                                                             & \multicolumn{3}{c}{79.36 / - / 51.30}                                                                                                          & 41.18                 \\
			\bottomrule
		\end{tabular}%
	}
\end{table*}

\begin{figure*}[!h]
	\centering
	\subcaptionbox{\centering Scene A\label{subfig:vis_results_all_a}}
	{%
		\includegraphics[height=2.1cm]{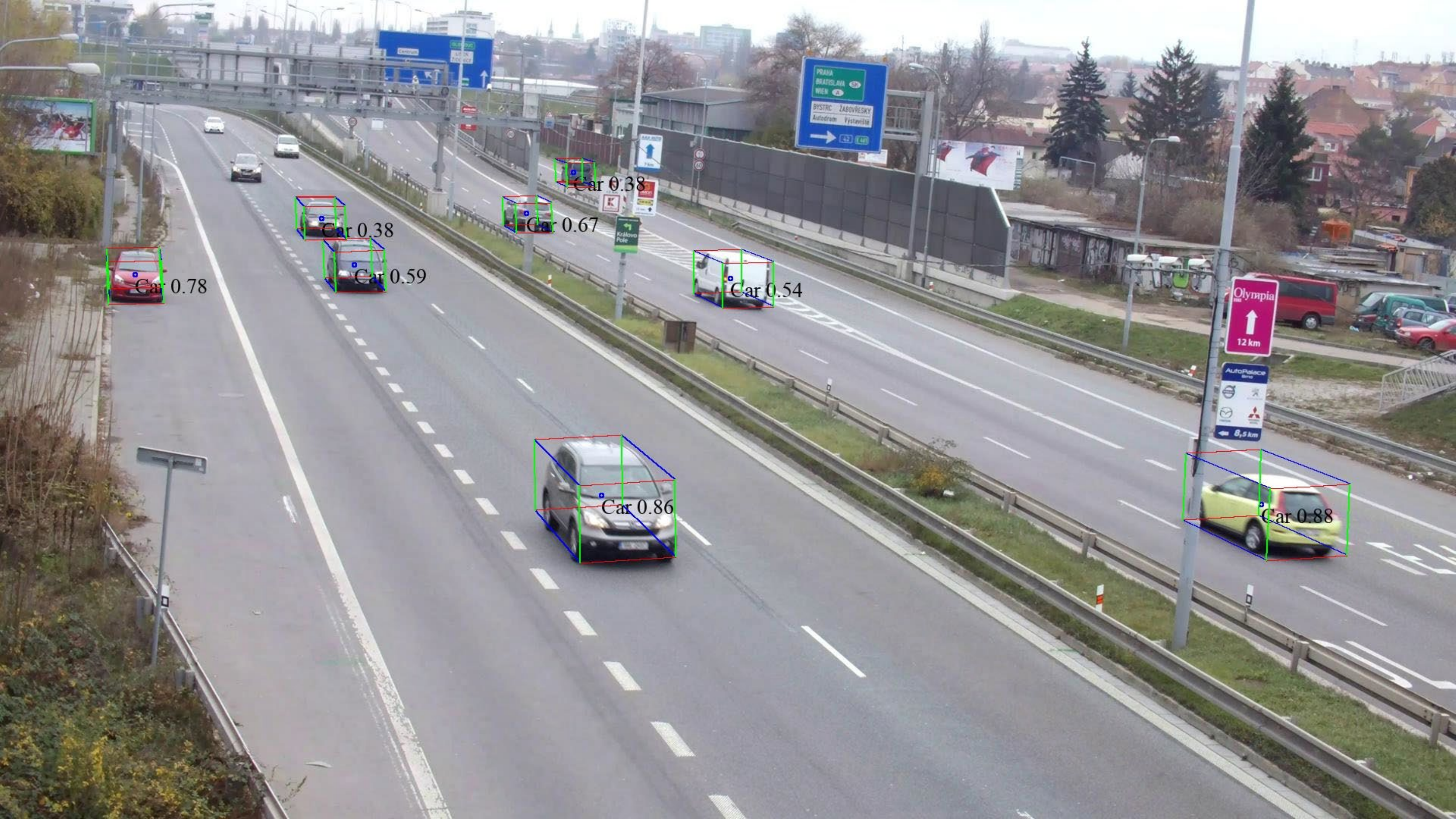}%
		\includegraphics[height=2.1cm]{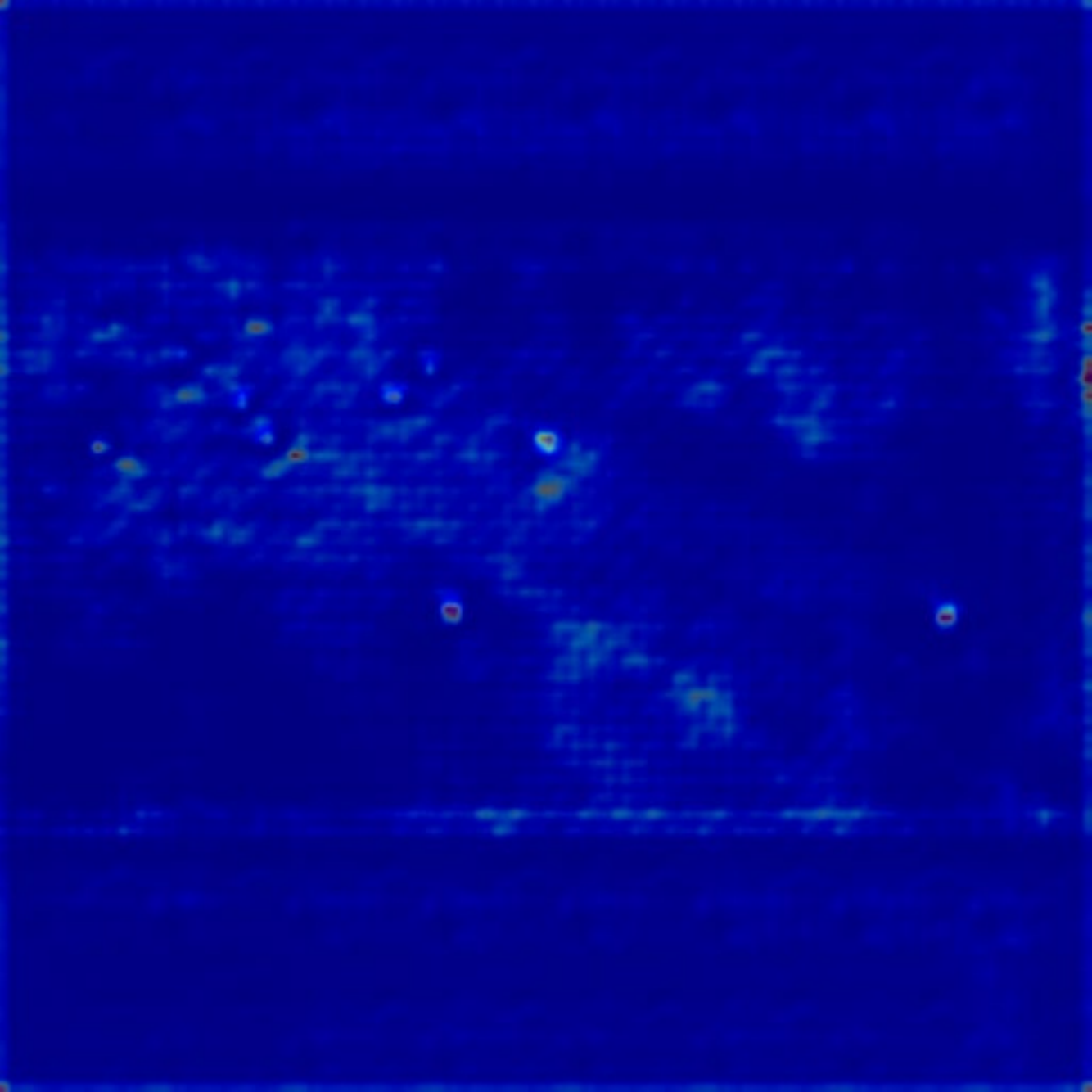}\quad
		\includegraphics[height=2.1cm]{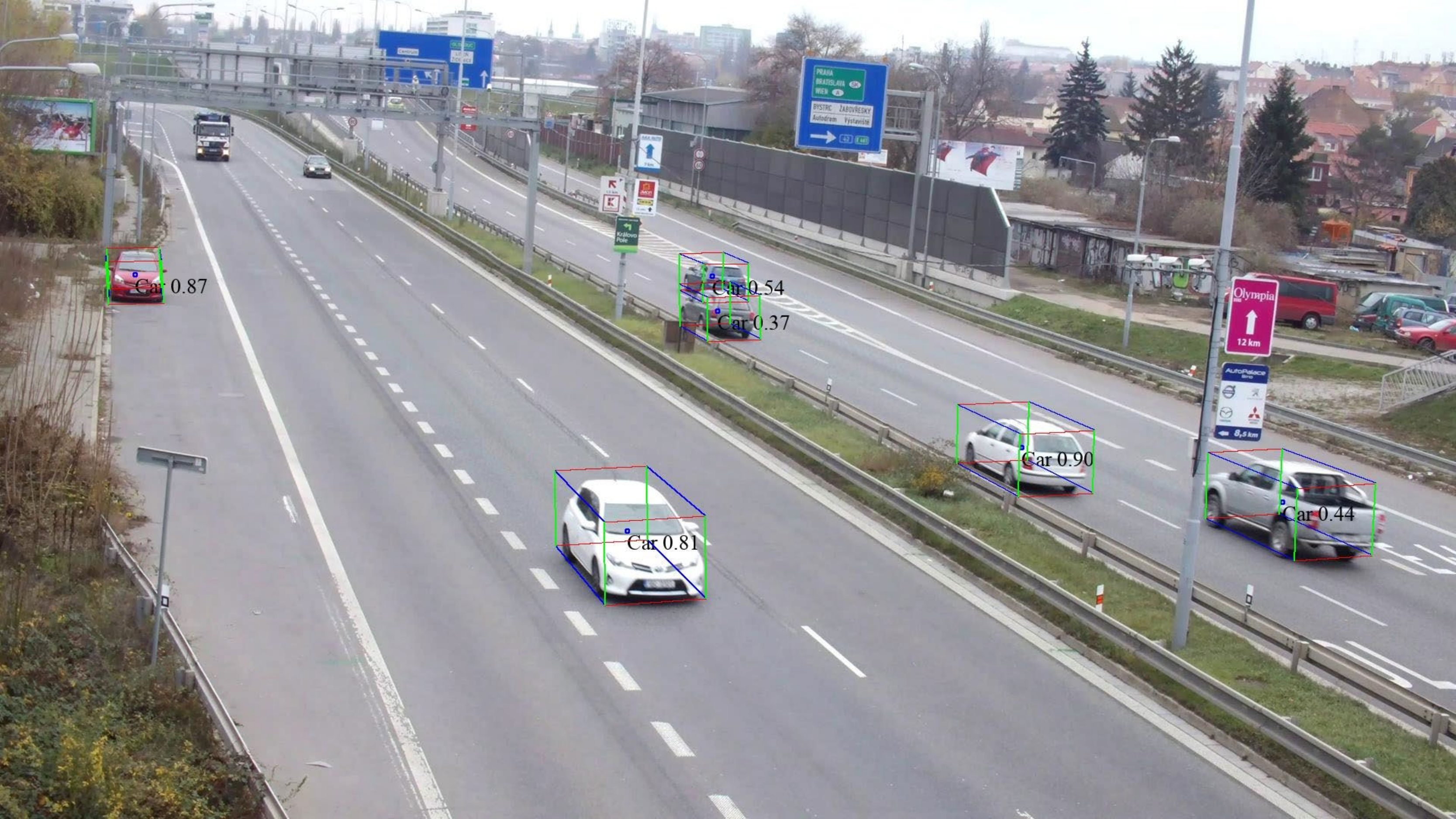}%
		\includegraphics[height=2.1cm]{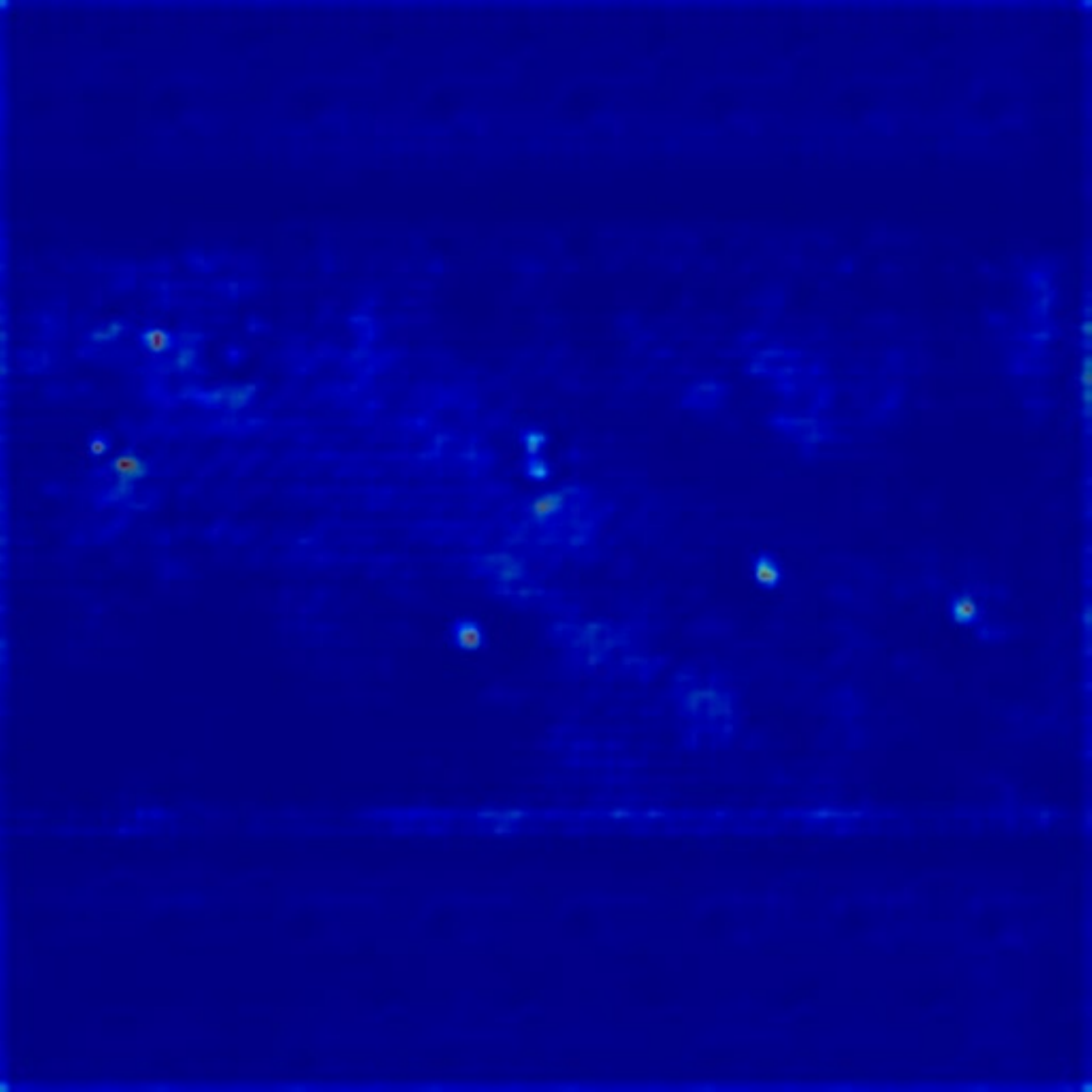}%
	}
	\subcaptionbox{\centering Scene B\label{subfig:vis_results_all_b}}
	{%
		\includegraphics[height=2.1cm]{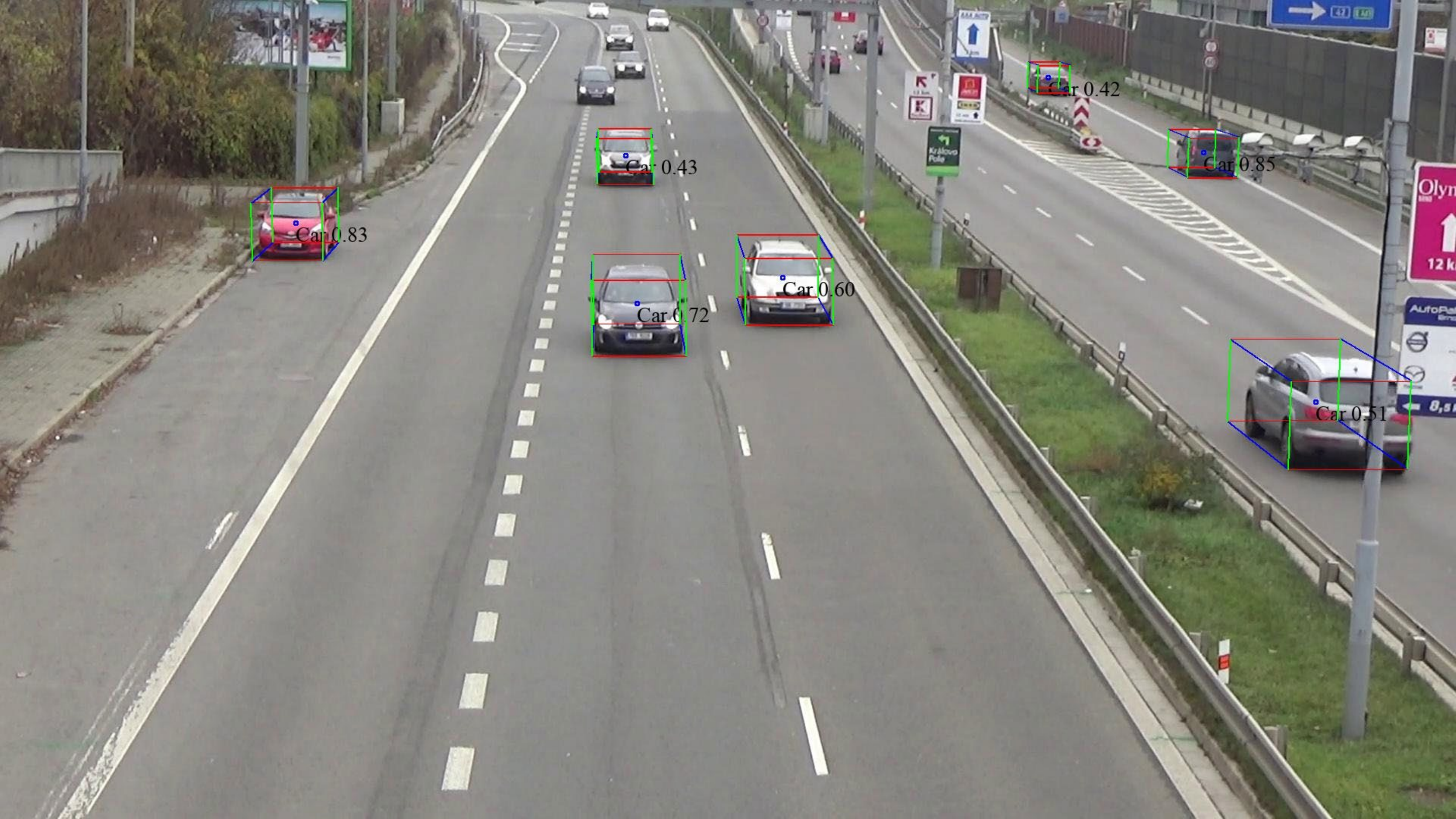}%
		\includegraphics[height=2.1cm]{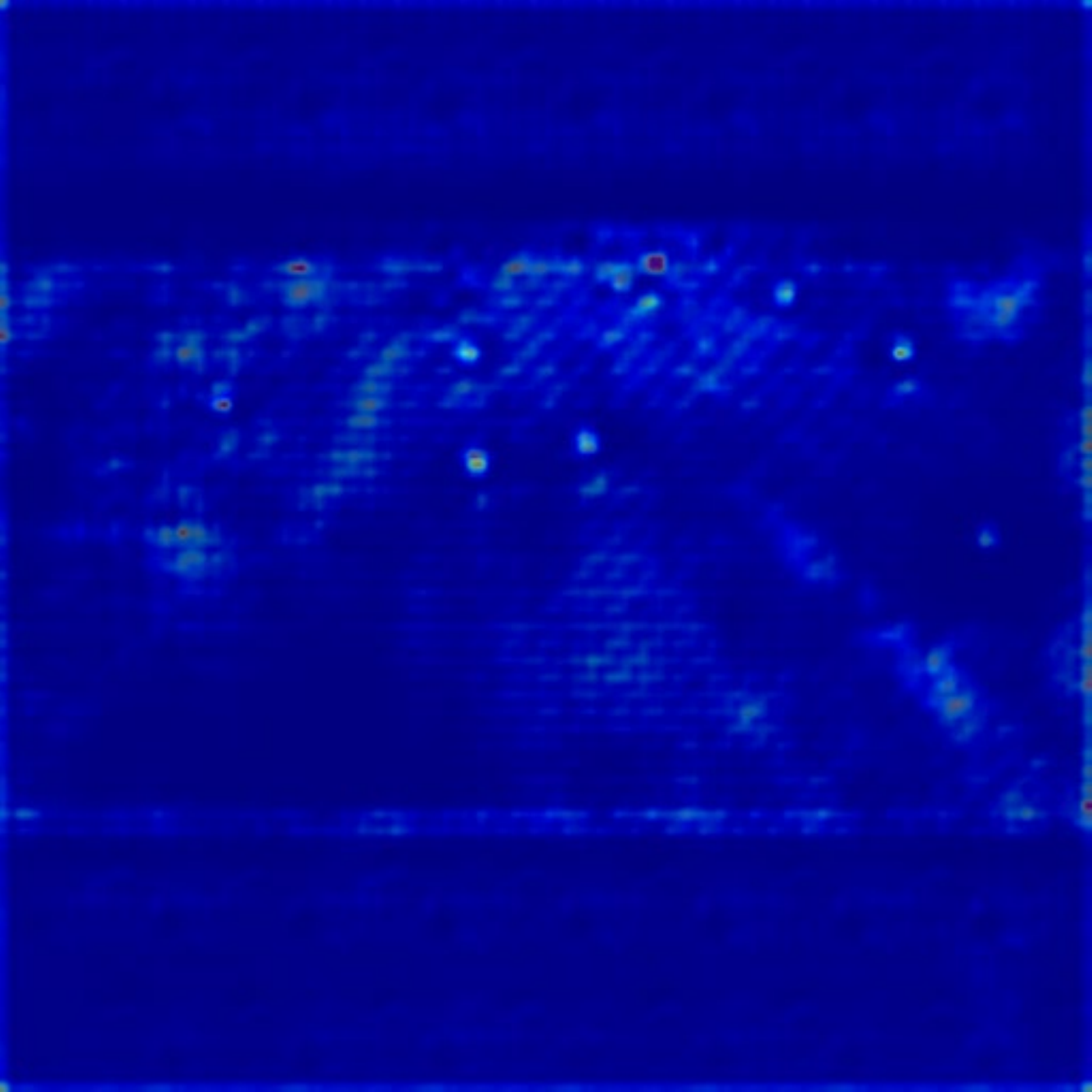}\quad
		\includegraphics[height=2.1cm]{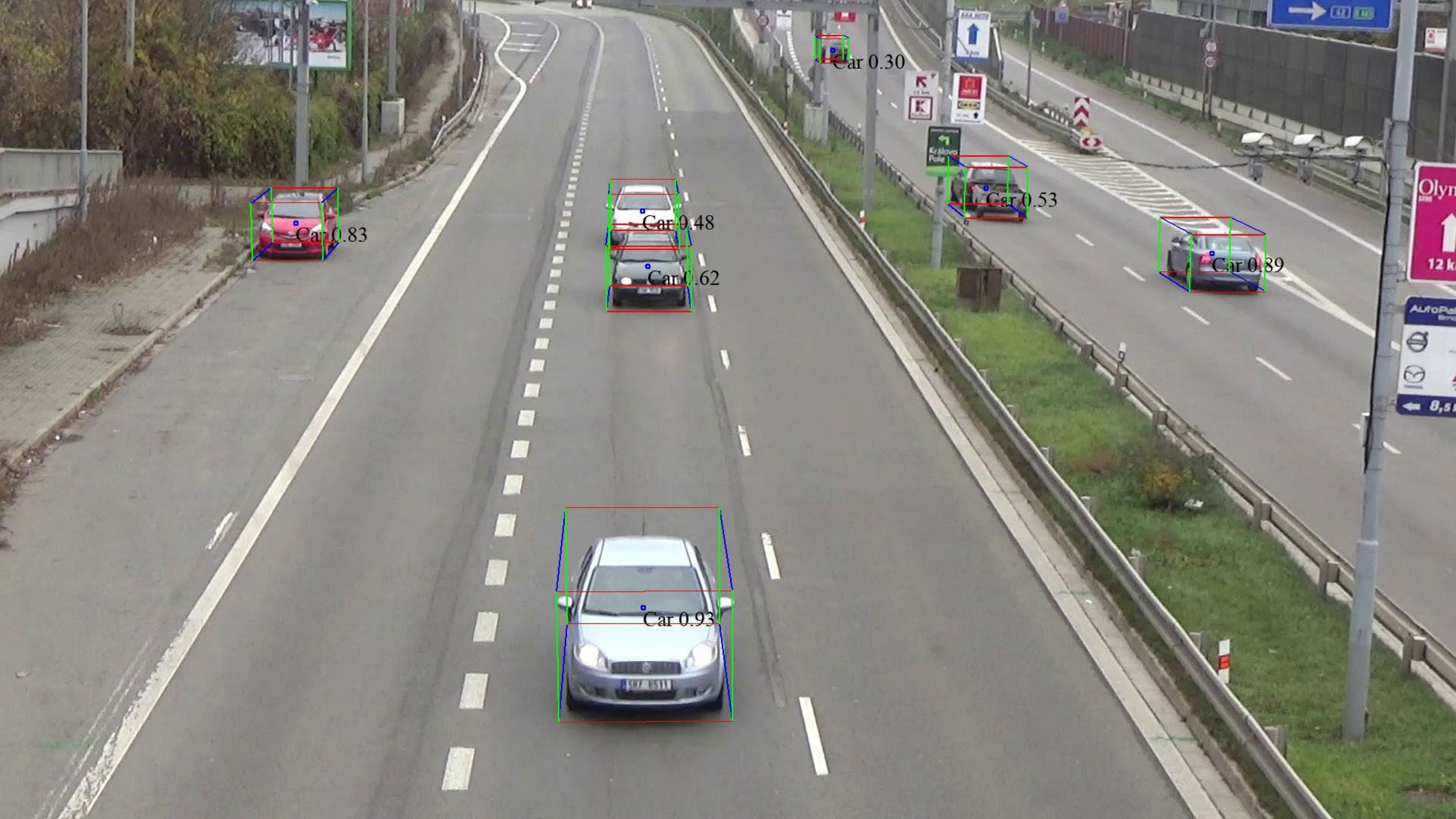}%
		\includegraphics[height=2.1cm]{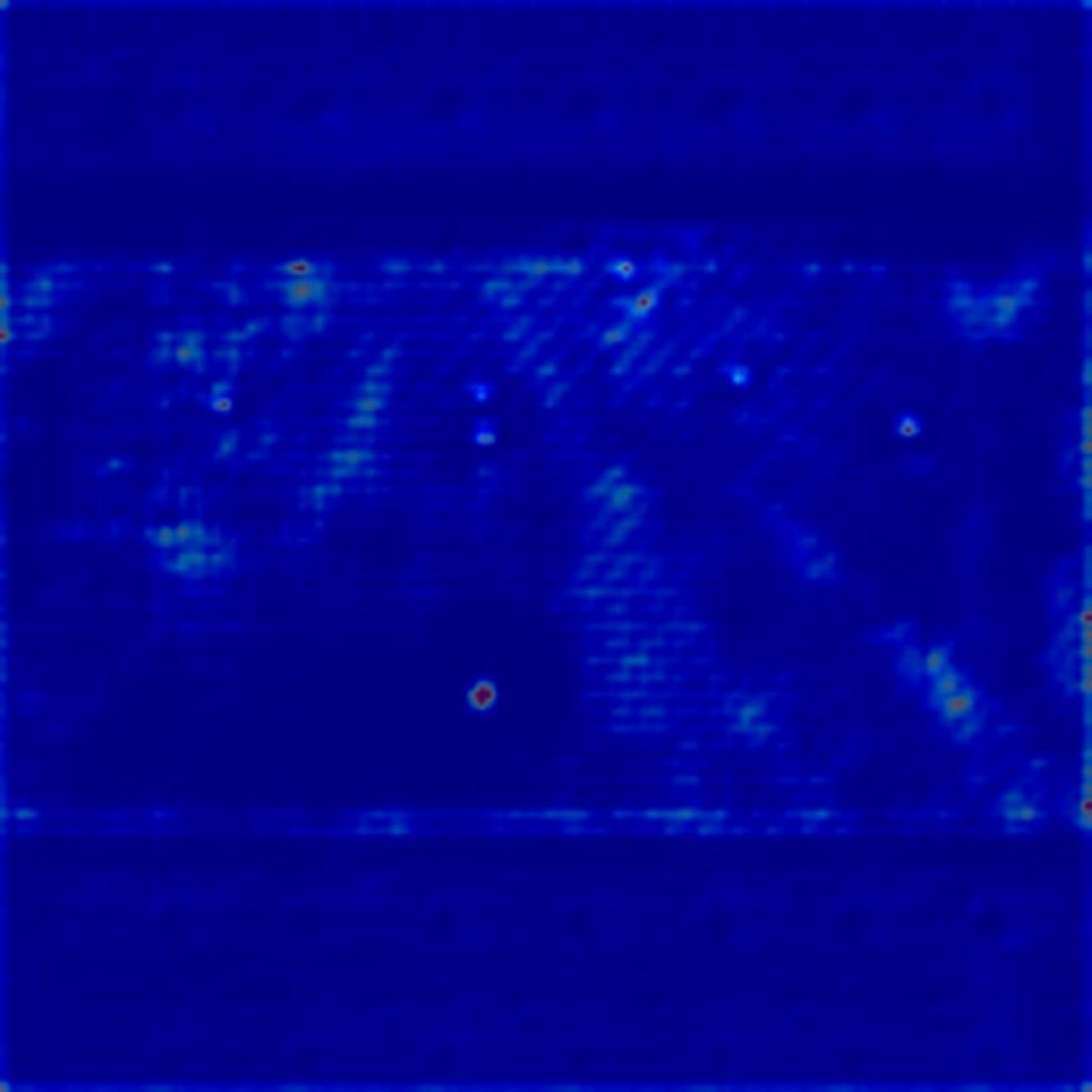}%
	}
	\subcaptionbox{\centering Scene C\label{subfig:vis_results_all_c}}
	{%
		\includegraphics[height=2.1cm]{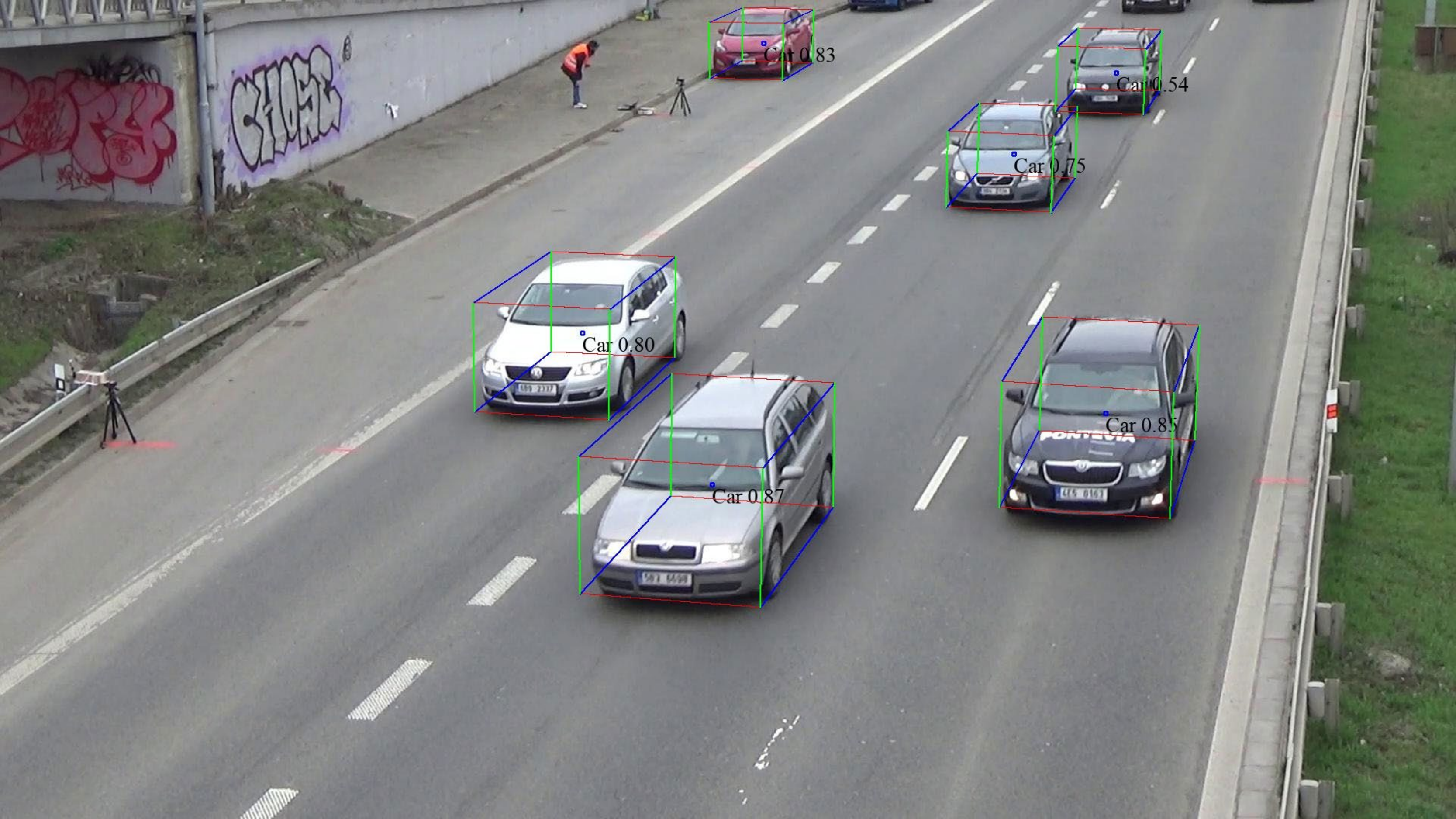}%
		\includegraphics[height=2.1cm]{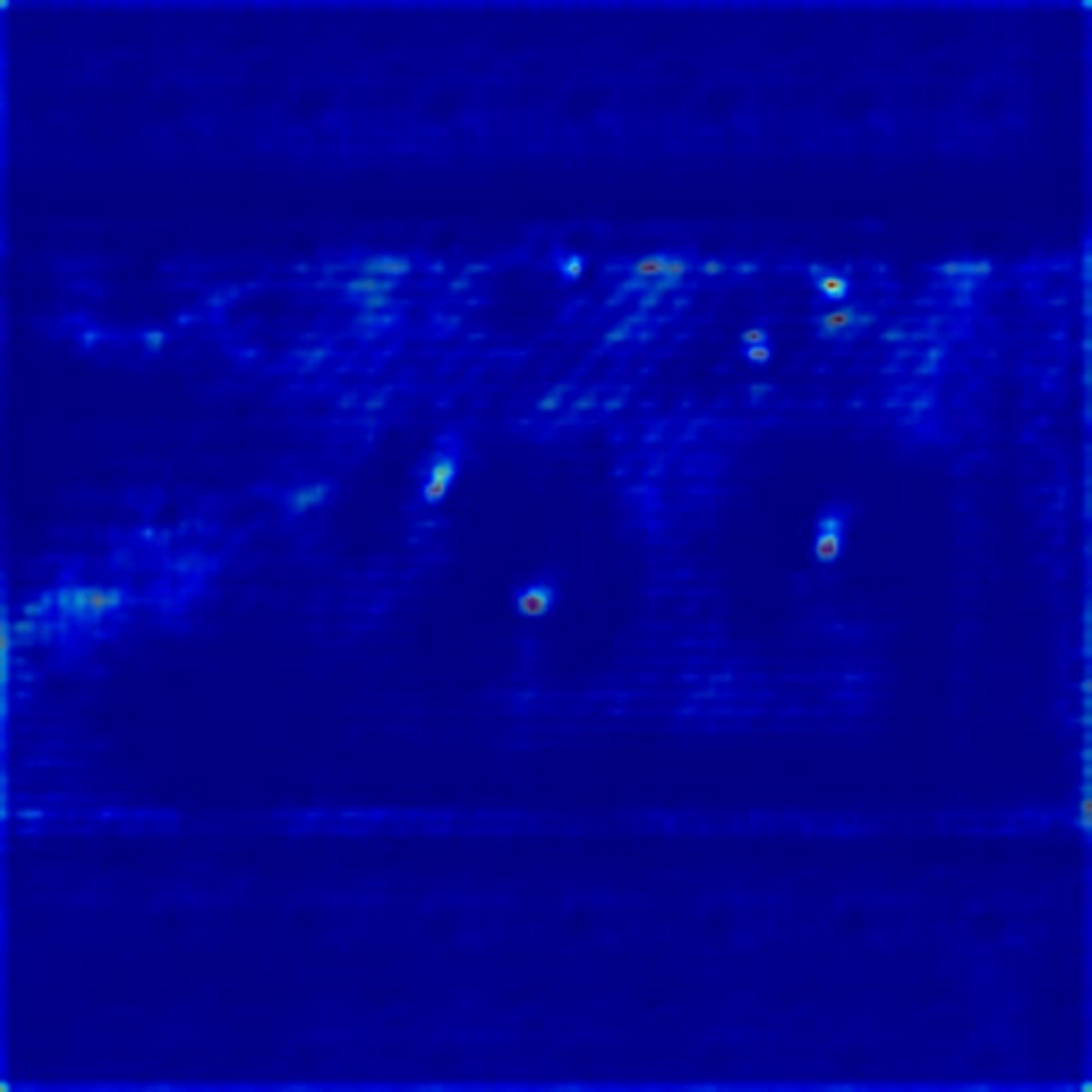}\quad
		\includegraphics[height=2.1cm]{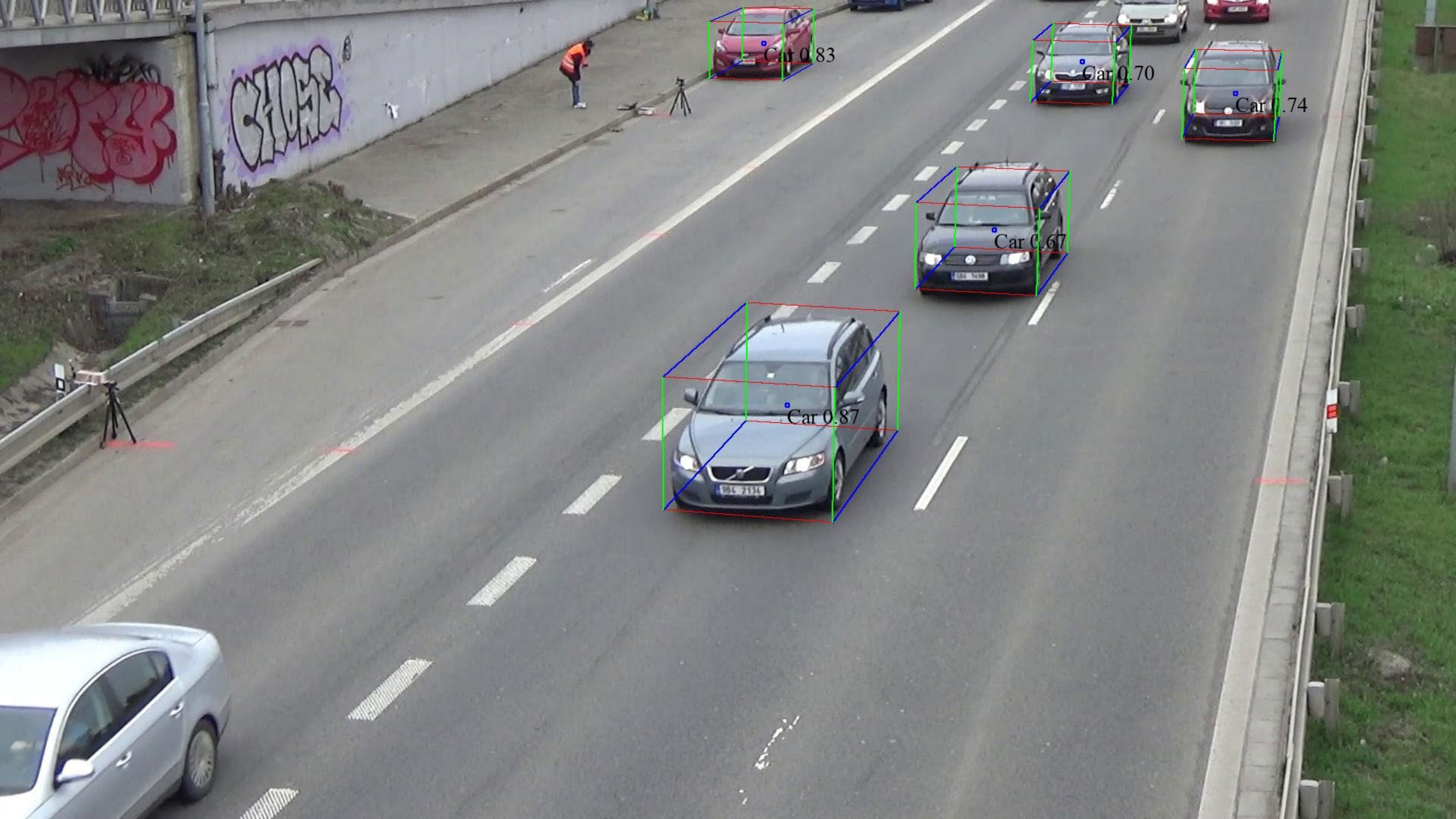}%
		\includegraphics[height=2.1cm]{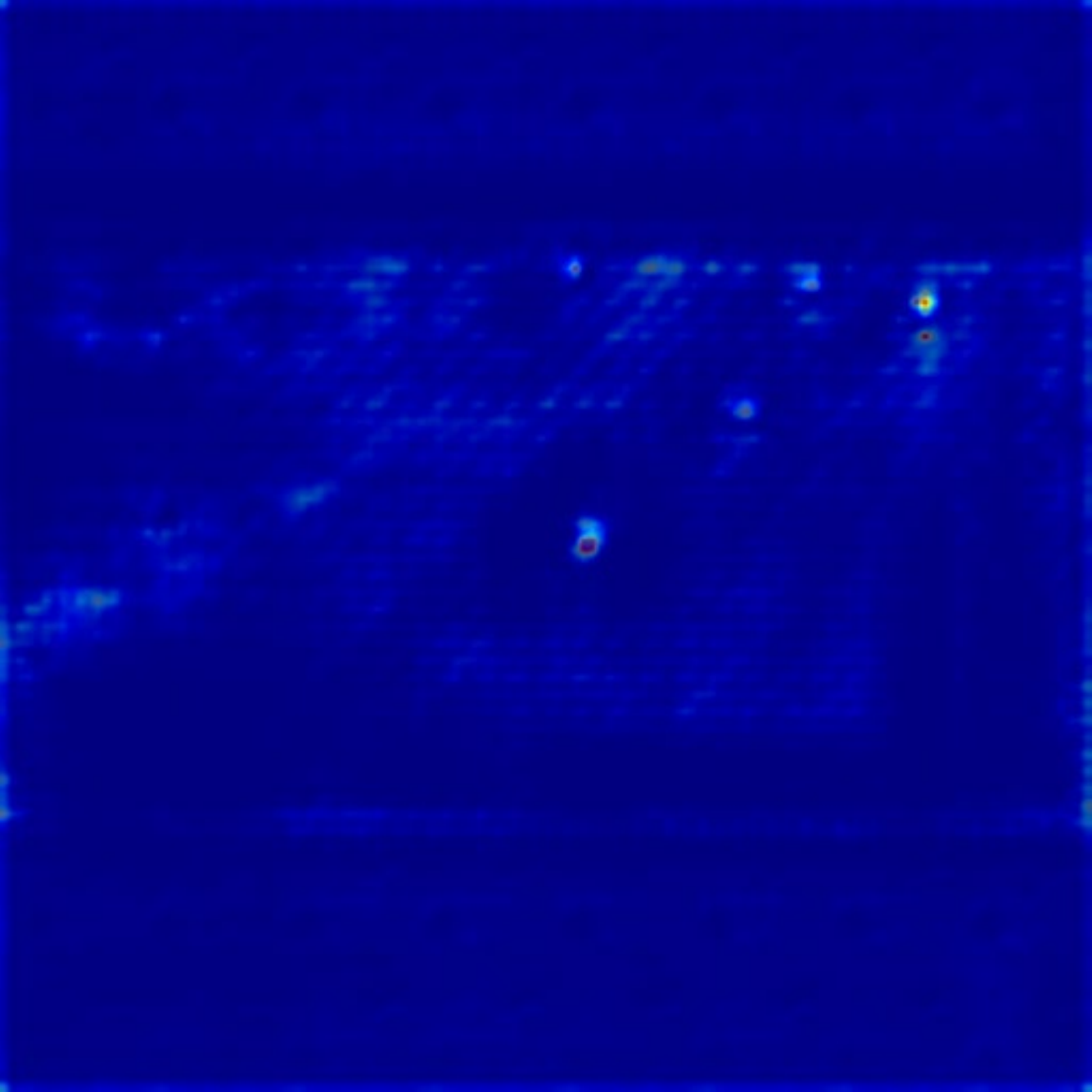}%
	}
	\subcaptionbox{\centering Scene D\label{subfig:vis_results_all_d}}
	{%
		\includegraphics[height=2.1cm]{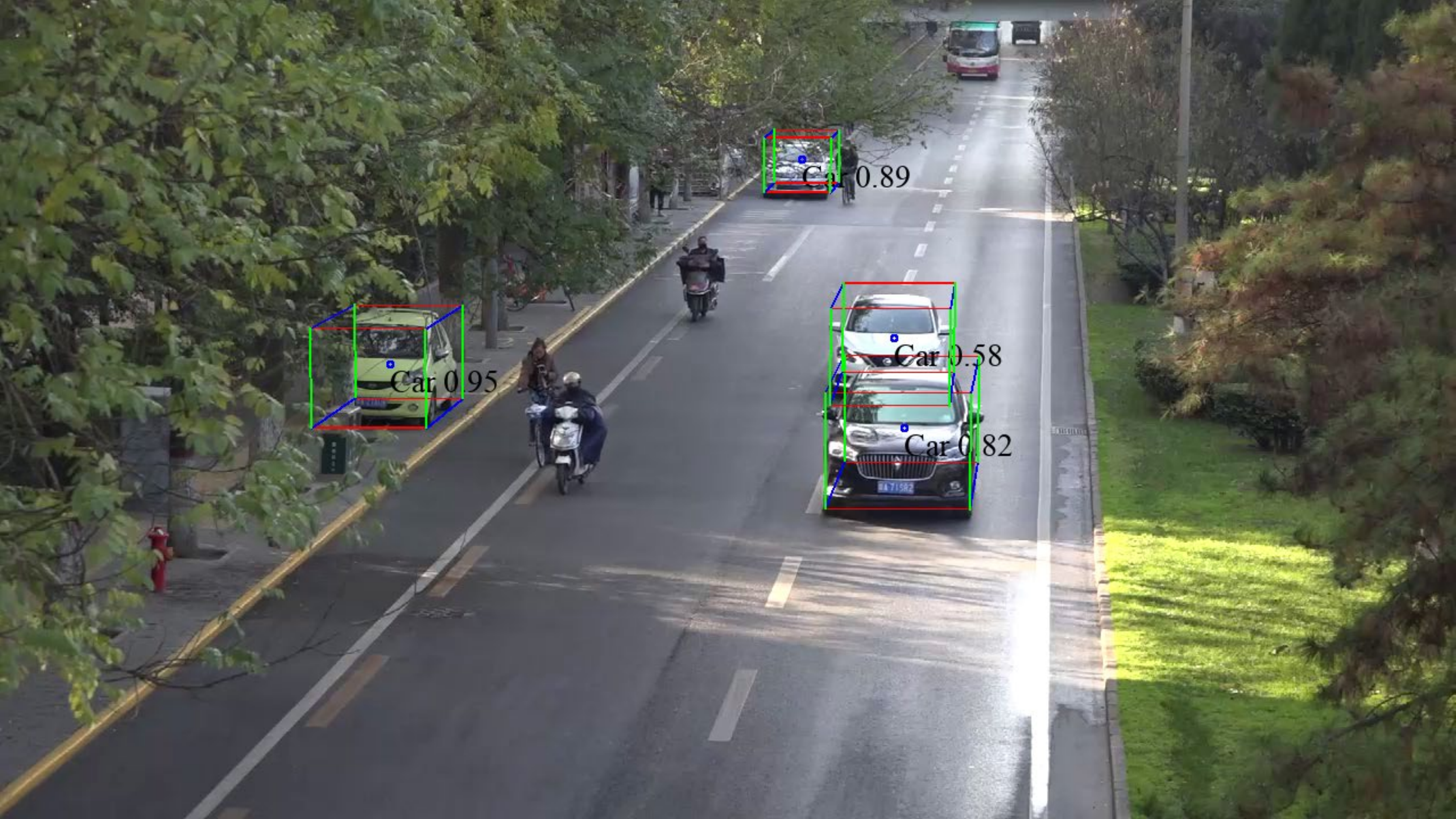}%
		\includegraphics[height=2.1cm]{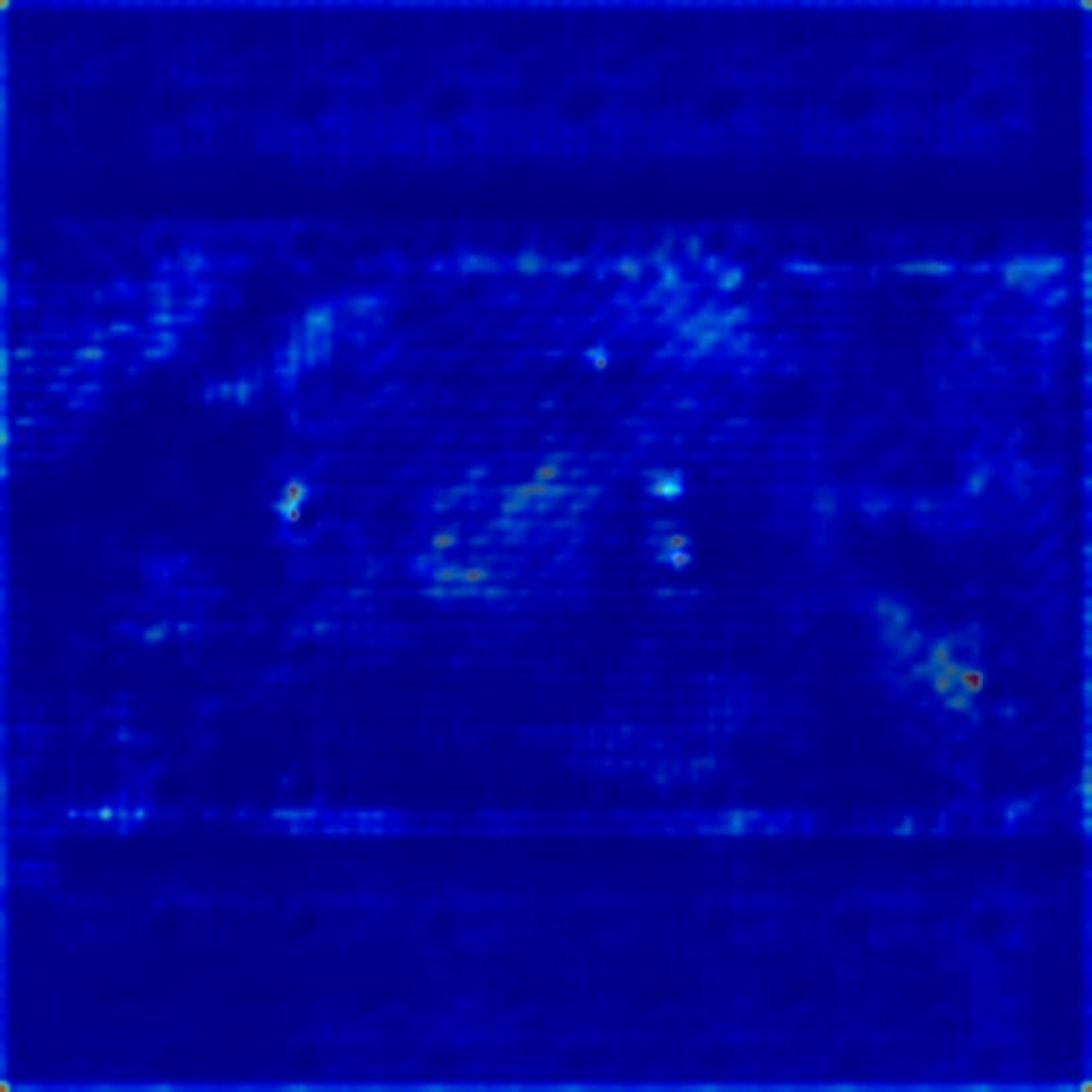}\quad
		\includegraphics[height=2.1cm]{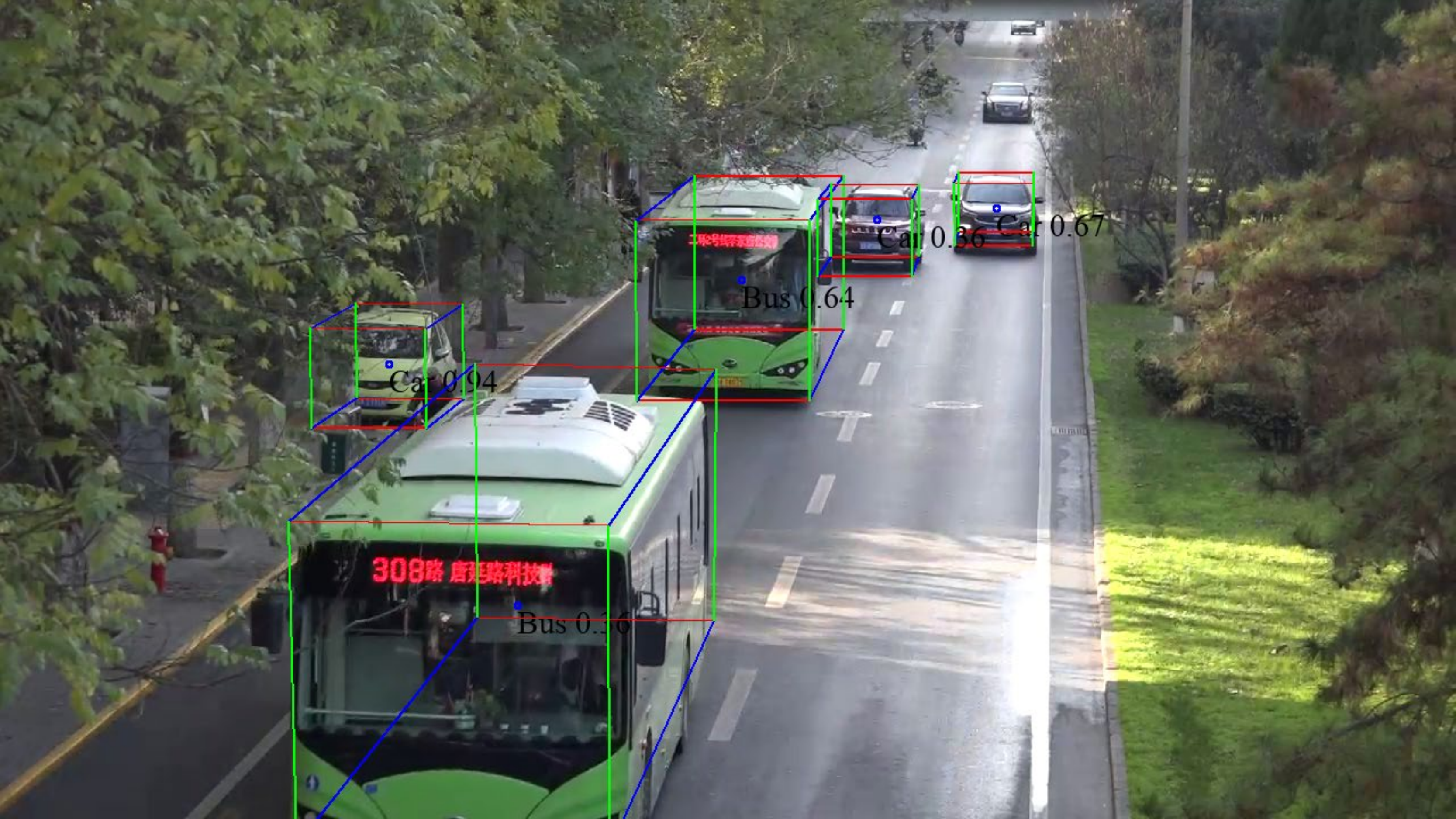}%
		\includegraphics[height=2.1cm]{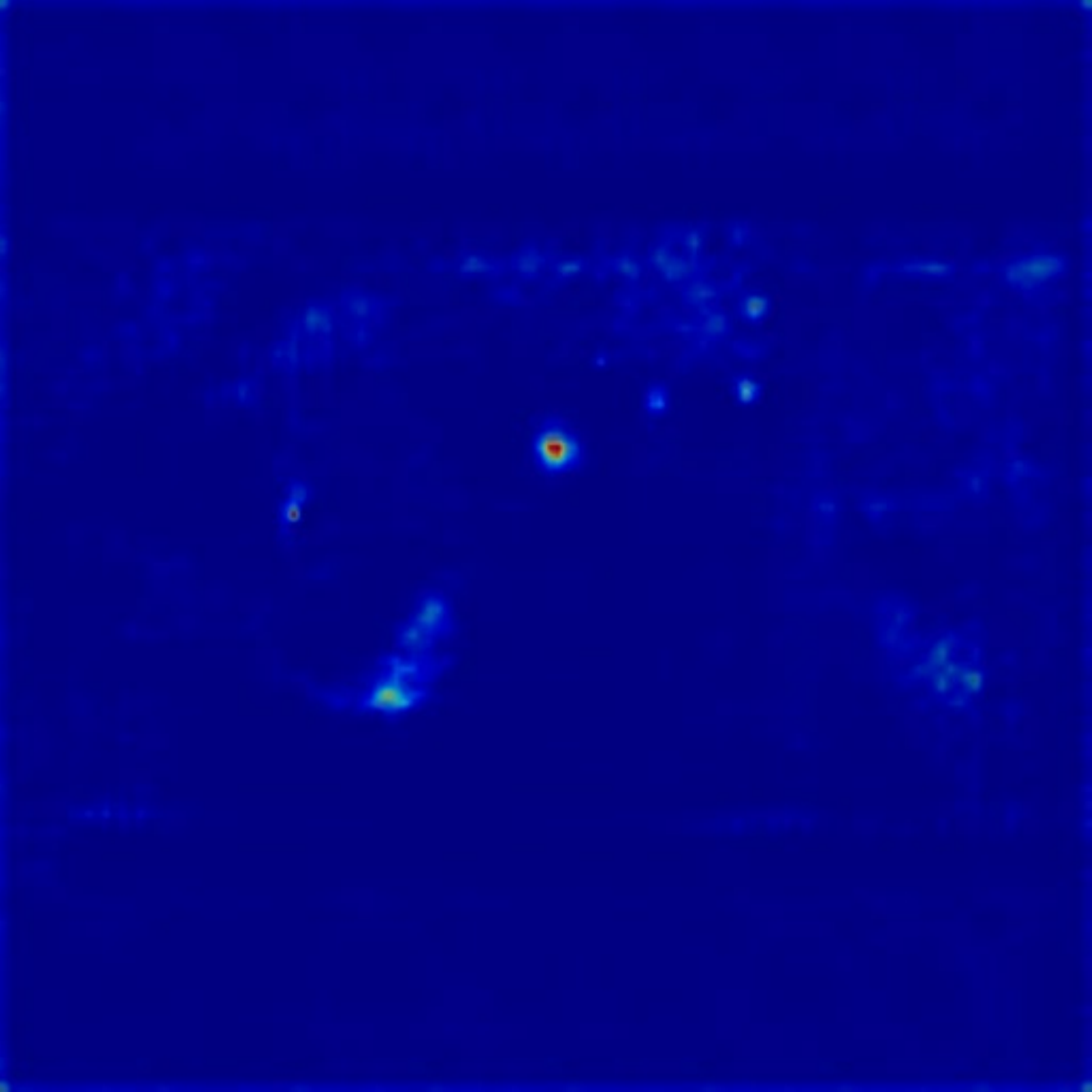}%
	}
	\subcaptionbox{\centering Scene E\label{subfig:vis_results_all_e}}
	{%
		\includegraphics[height=2.1cm]{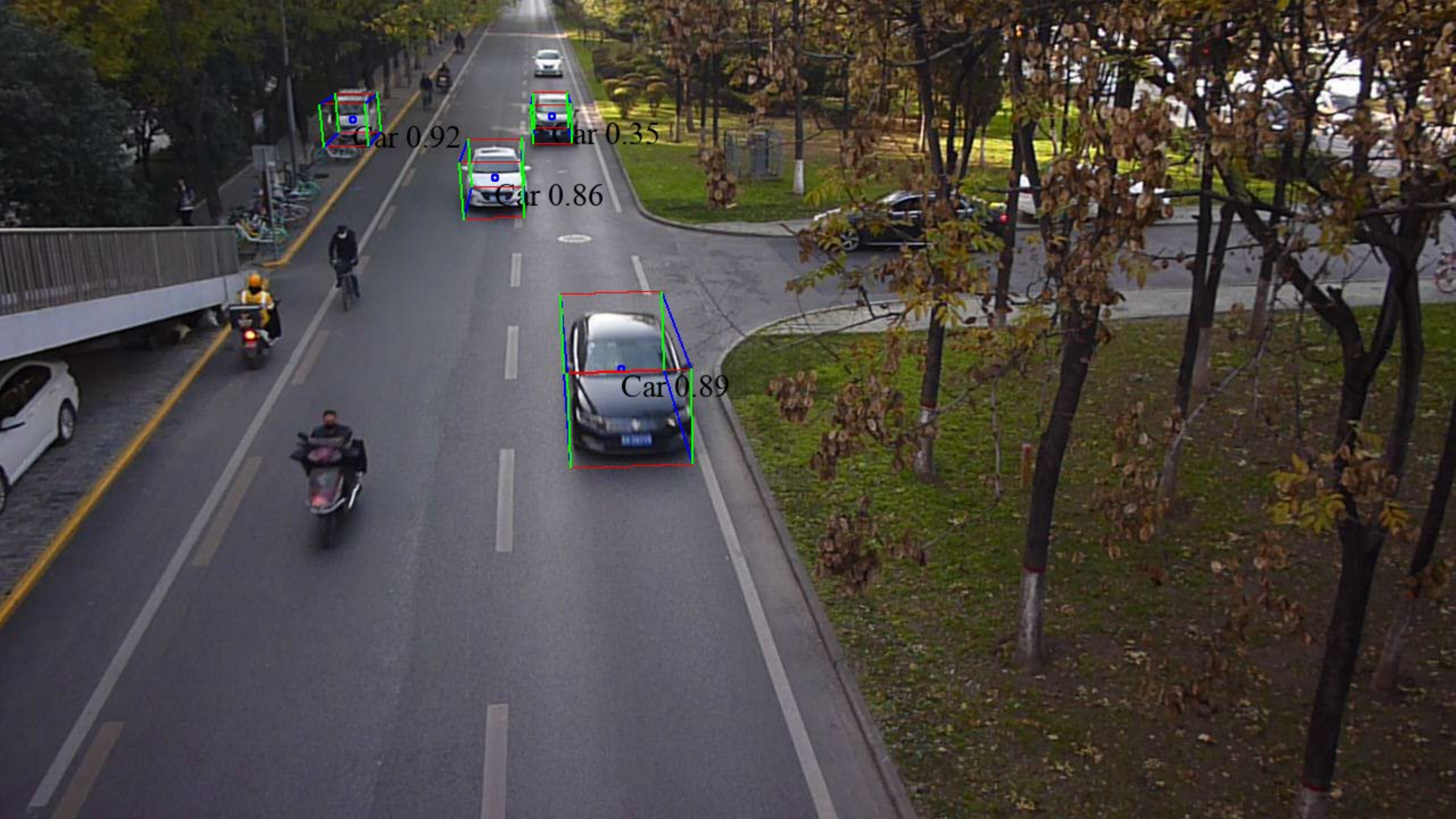}%
		\includegraphics[height=2.1cm]{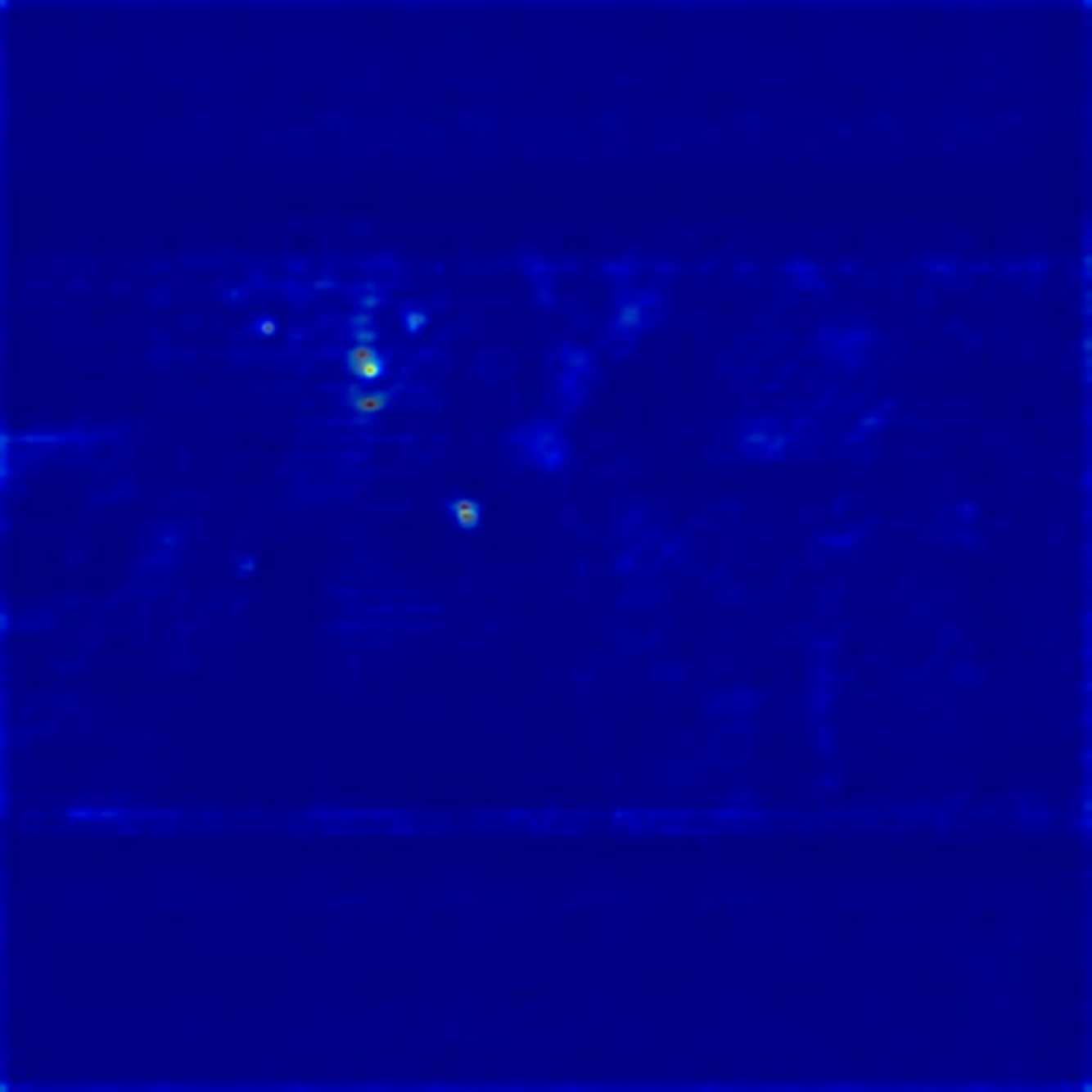}\quad
		\includegraphics[height=2.1cm]{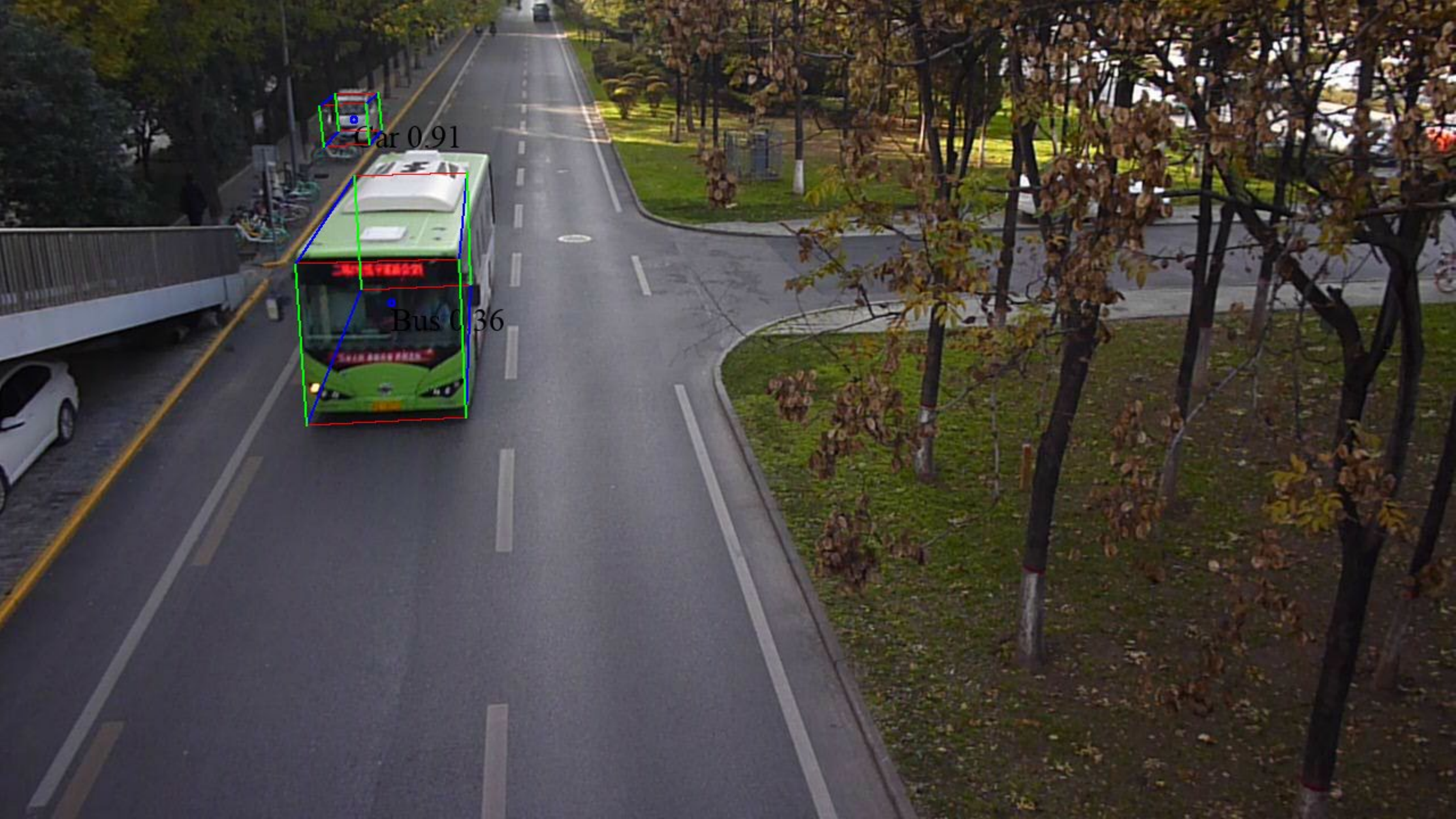}%
		\includegraphics[height=2.1cm]{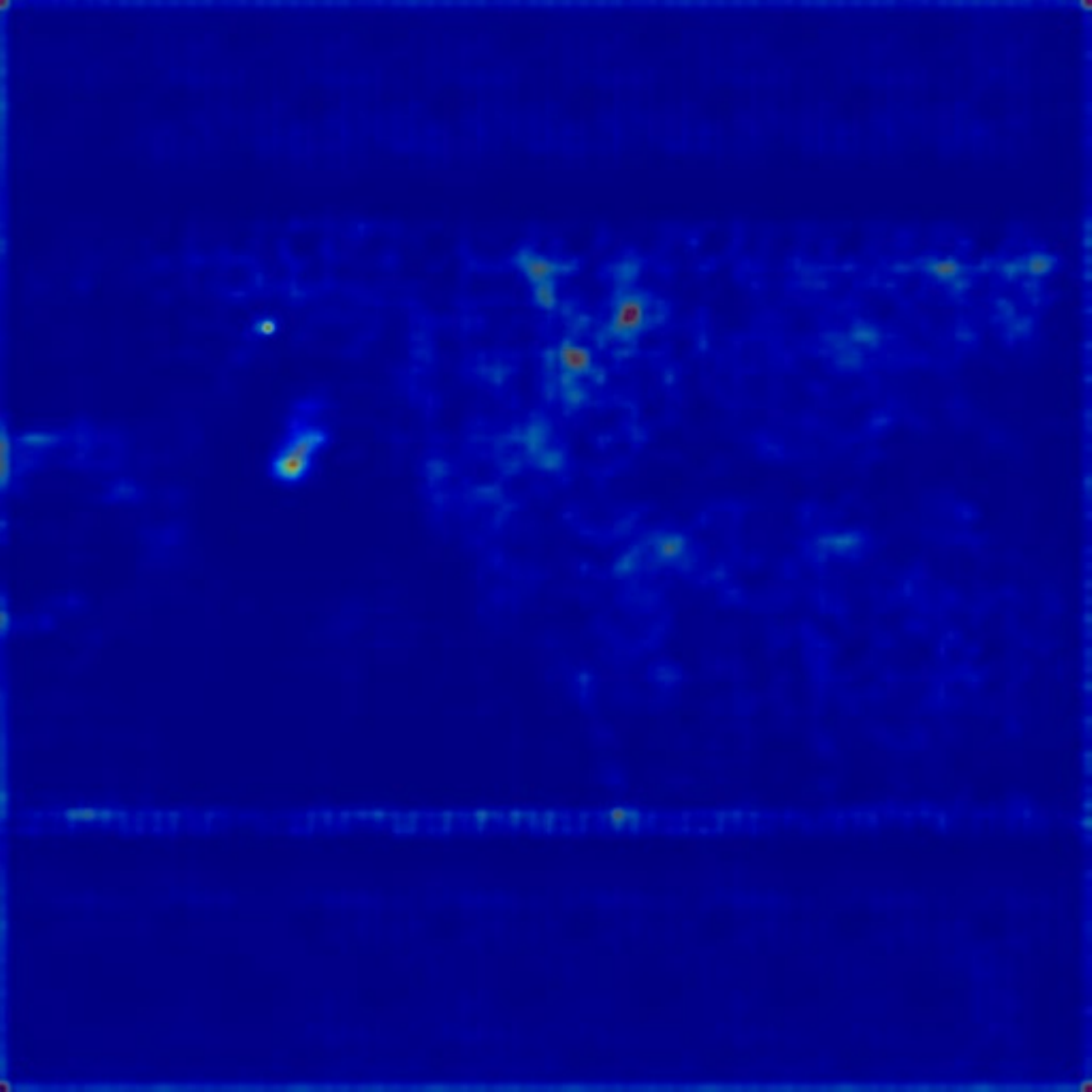}%
	}
	\caption{\leftskip=0pt \rightskip=0pt plus 0cm Visualization results on SVLD-3D test set. 3D vehicle detection results (left) are represented by color bounding boxes. Centroids of heatmaps (right) are represented in red circles. \label{fig:vis_results_all}}
\end{figure*}

Figure \ref{fig:vis_results_all} illustrates visualization results on SVLD-3D test set. Different views and types of vehicles in SVLD-3D test set are tested, with occlusion of environment and other vehicles. Vehicles are widely distributed in the scene. From Table \ref{tab:table_compare_kitti} and Figure \ref{fig:vis_results_all}, detected vehicles all response with red circle in the heatmaps. It can be seen that our network achieves real-time 3D vehicle detection and is adaptive to occlusion and small vehicles.

3D vehicle information which can be obtained by different methods is compared in Table \ref{tab:table_compare_func}. It can be seen that our method obtains not only 3D bounding boxes but also 3D centroids, dimensions, and 3D locations, with real-time performance.

\begin{table}[!h]
	\centering
	\caption{Comparison of 3D vehicle information obtained by different methods.}
	\label{tab:table_compare_func}
	\begin{tabular}{ccccc}
		\toprule
		Method                                          & BBox                   & Centroid               & Dimension              & Location           \\
		\midrule
		MonoGRNet \cite{2019monogrnet} & \checkmark &                           & \checkmark & \checkmark \\
		Deep3DBox \cite{2017deep3dbox} & \checkmark &                           &                           &                           \\
		GS3D \cite{2019gs3d}           & \checkmark &                           & \checkmark &                           \\
		RTM3D \cite{2020rtm3d}         & \checkmark & \checkmark & \checkmark &                           \\
		SMOKE \cite{2020SMOKE}         & \checkmark & \checkmark & \checkmark &                           \\
		KM3D \cite{2021km3d}           & \checkmark &                           & \checkmark &                           \\
		Lite-FPN \cite{2021litefpn}    & \checkmark &                           & \checkmark &                           \\
		Ours                                            & \checkmark & \checkmark & \checkmark & \checkmark \\
		\bottomrule
	\end{tabular}
\end{table}

\subsection{3D Vehicle Localization Precision and Error of CenterLoc3D}
\label{subsec:3dlocpe of centerloc3d}

\begin{figure*}[htbp]
	\centering
	\subfloat[\centering ]{\includegraphics[width=0.28\linewidth]{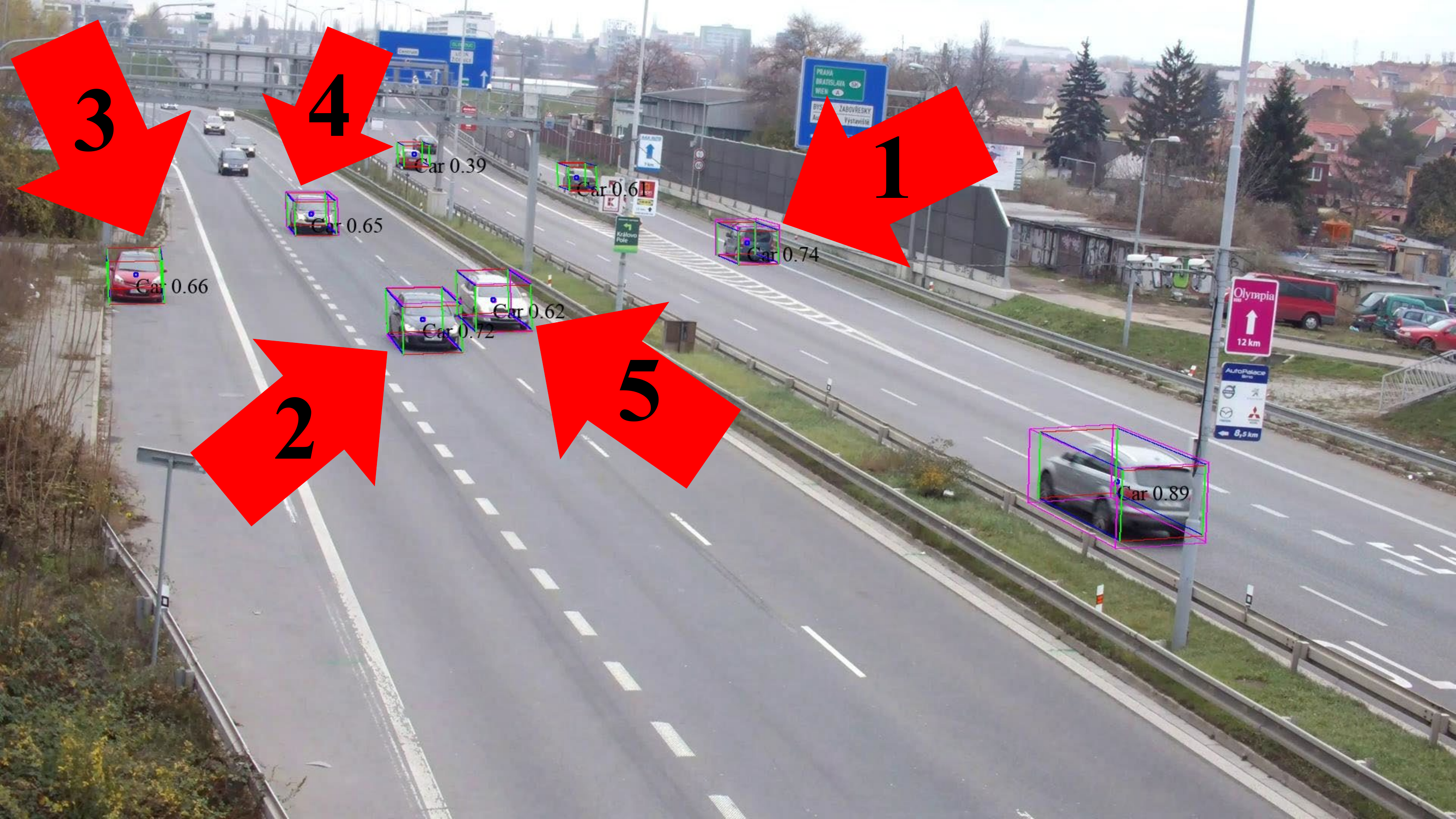}%
		\label{fig:vis_loc_results_a}}
	\hfil
	\subfloat[\centering ]{\includegraphics[width=0.28\linewidth]{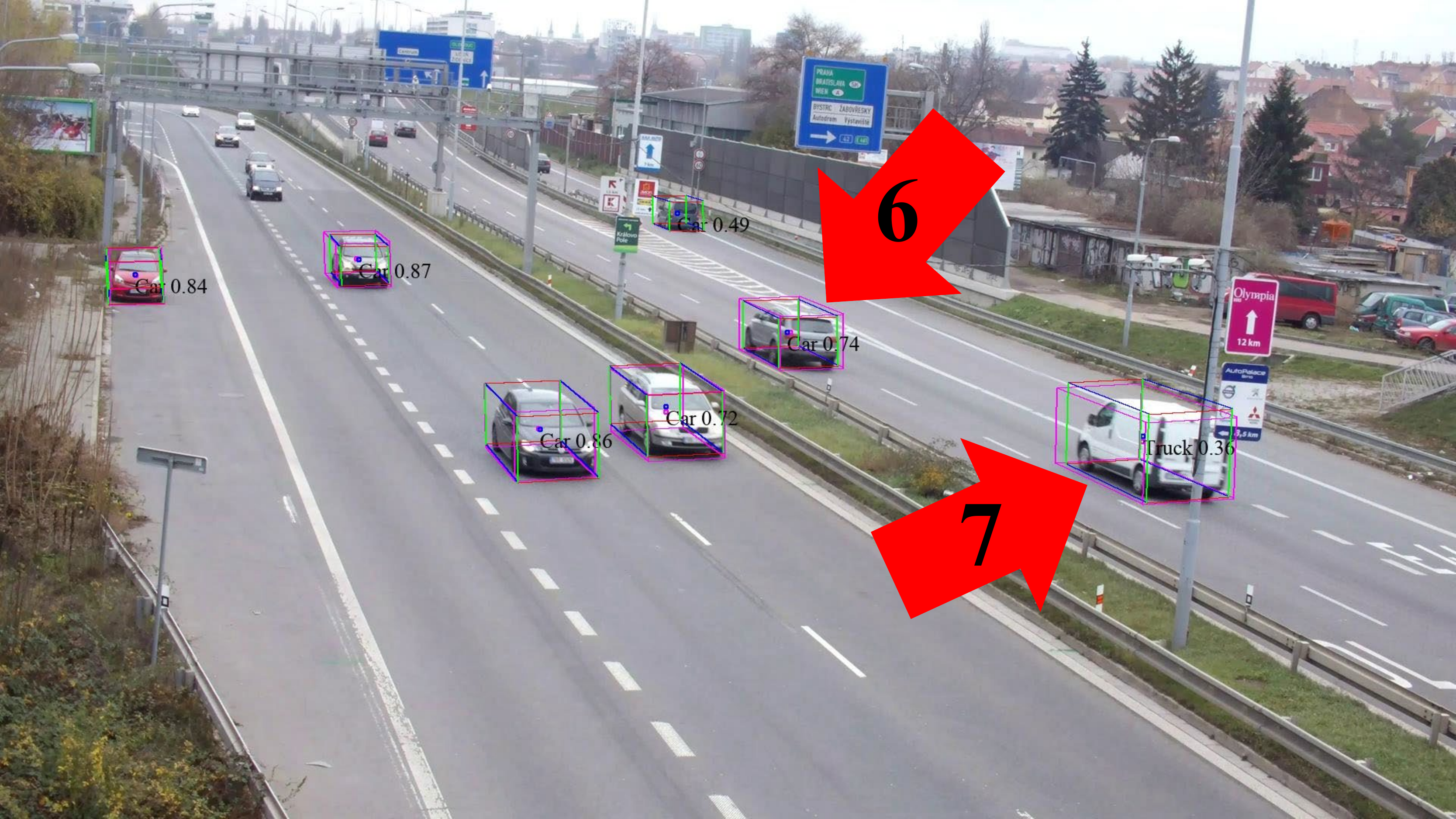}%
		\label{fig:vis_loc_results_b}}
	\hfil
	\subfloat[\centering ]{\includegraphics[width=0.28\linewidth]{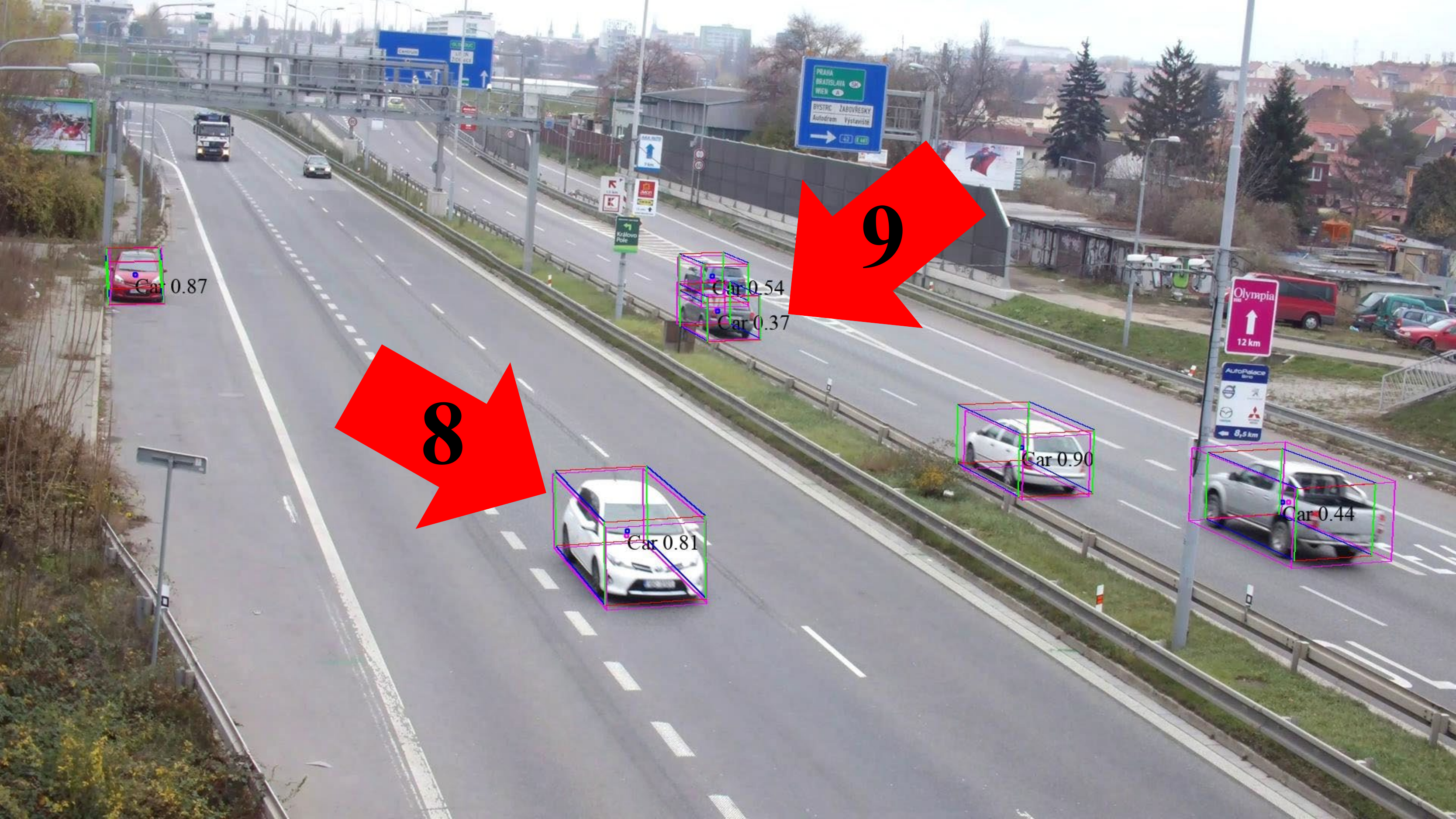}%
		\label{fig:vis_loc_results_c}}
	\newline
	\subfloat[\centering ]{\includegraphics[width=0.28\linewidth]{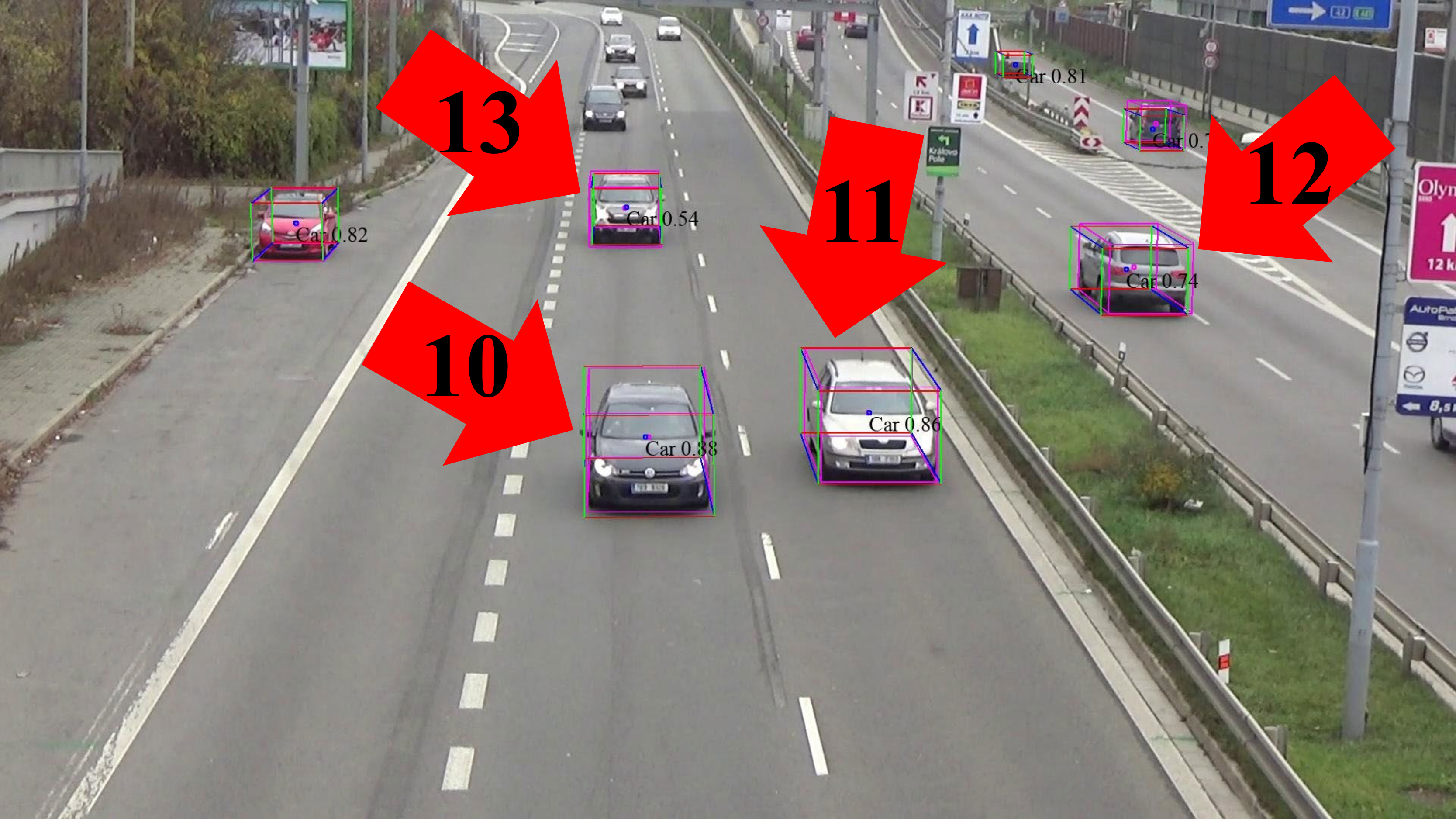}%
		\label{fig:vis_loc_results_d}}
	\hfil
	\subfloat[\centering ]{\includegraphics[width=0.28\linewidth]{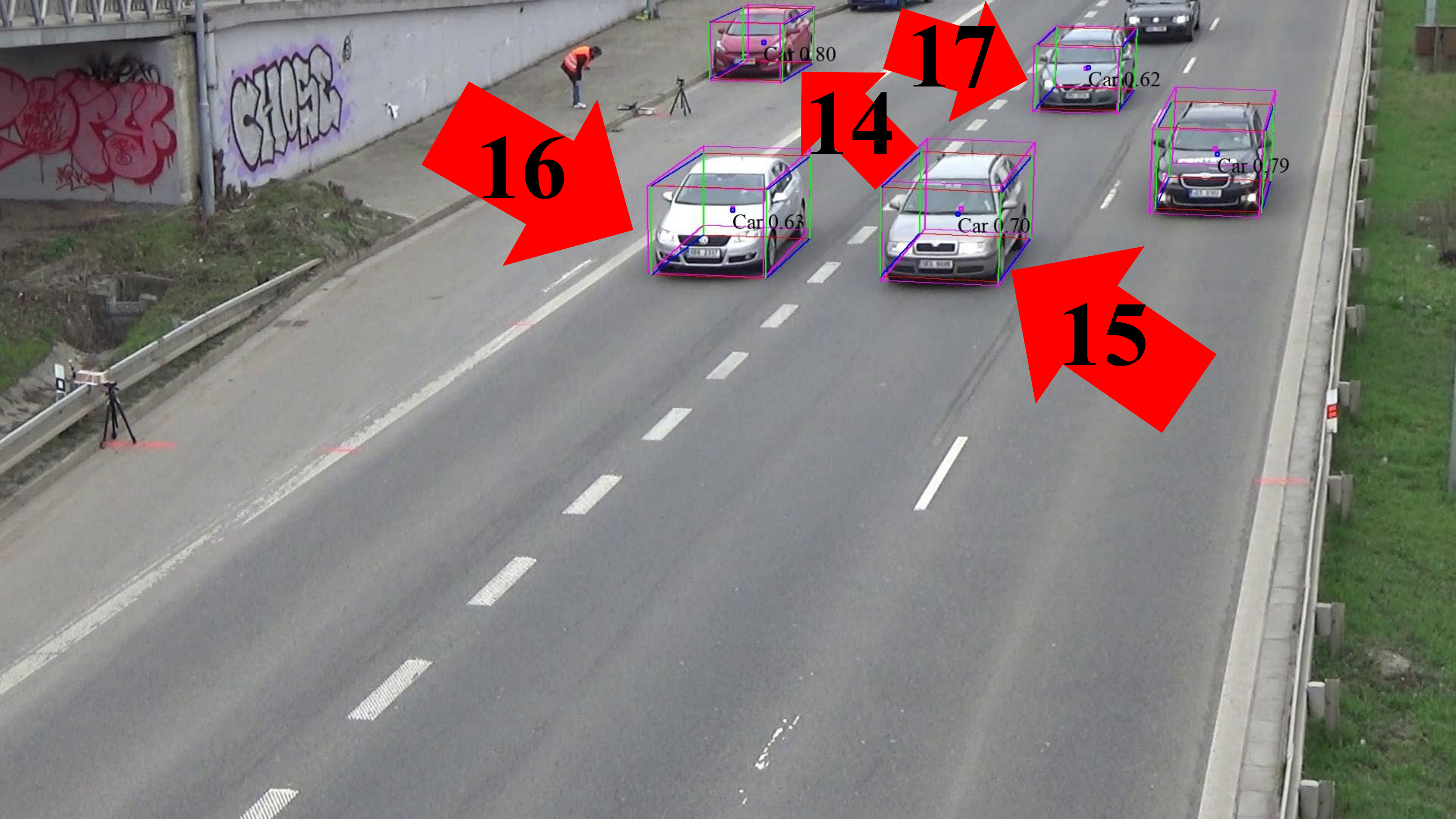}%
		\label{fig:vis_loc_results_e}}
	\hfil
	\subfloat[\centering ]{\includegraphics[width=0.28\linewidth]{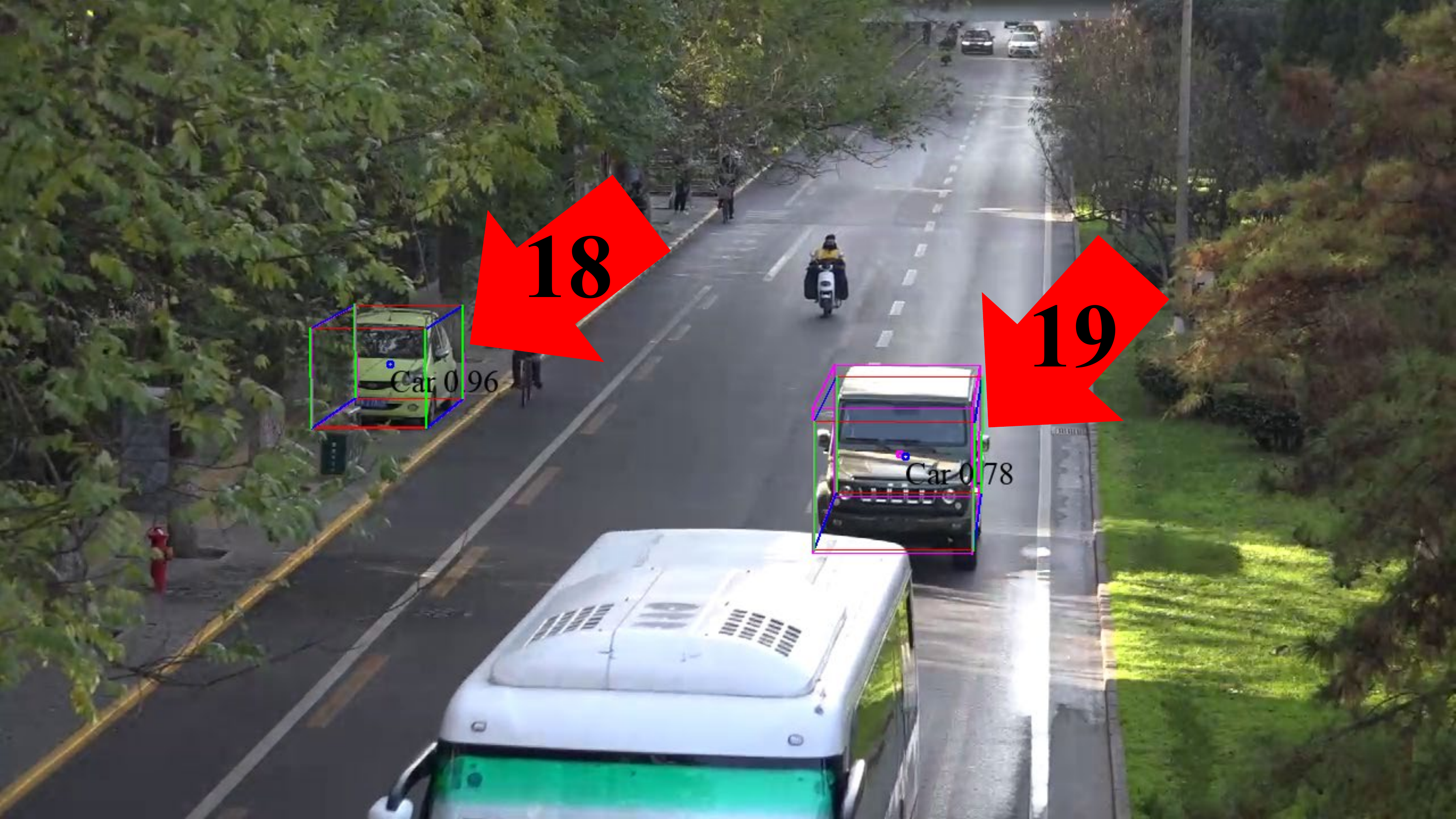}%
		\label{fig:vis_loc_results_f}}
	\newline
	\subfloat[\centering ]{\includegraphics[width=0.28\linewidth]{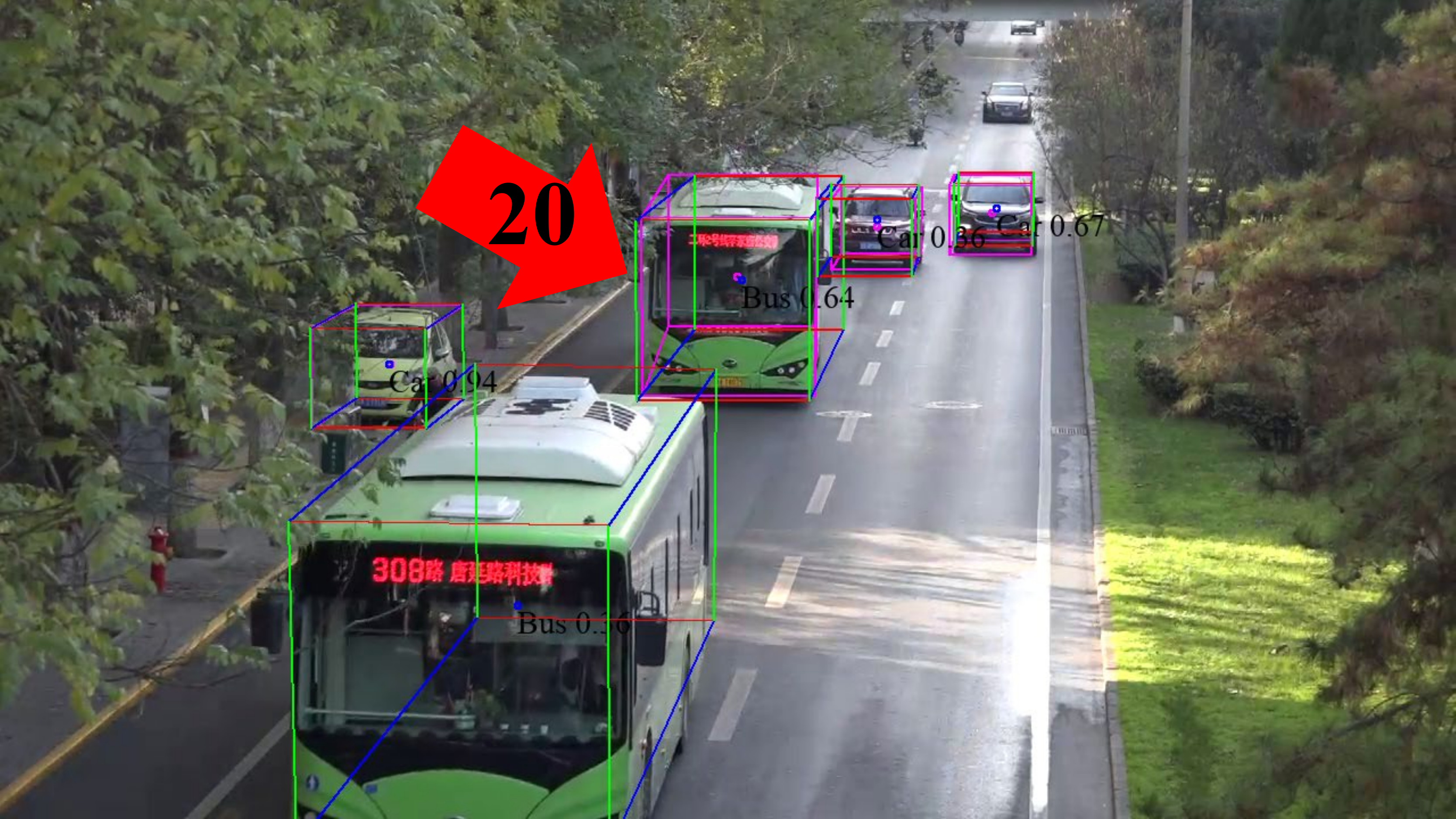}%
		\label{fig:vis_loc_results_g}}
	\hfil
	\subfloat[\centering ]{\includegraphics[width=0.28\linewidth]{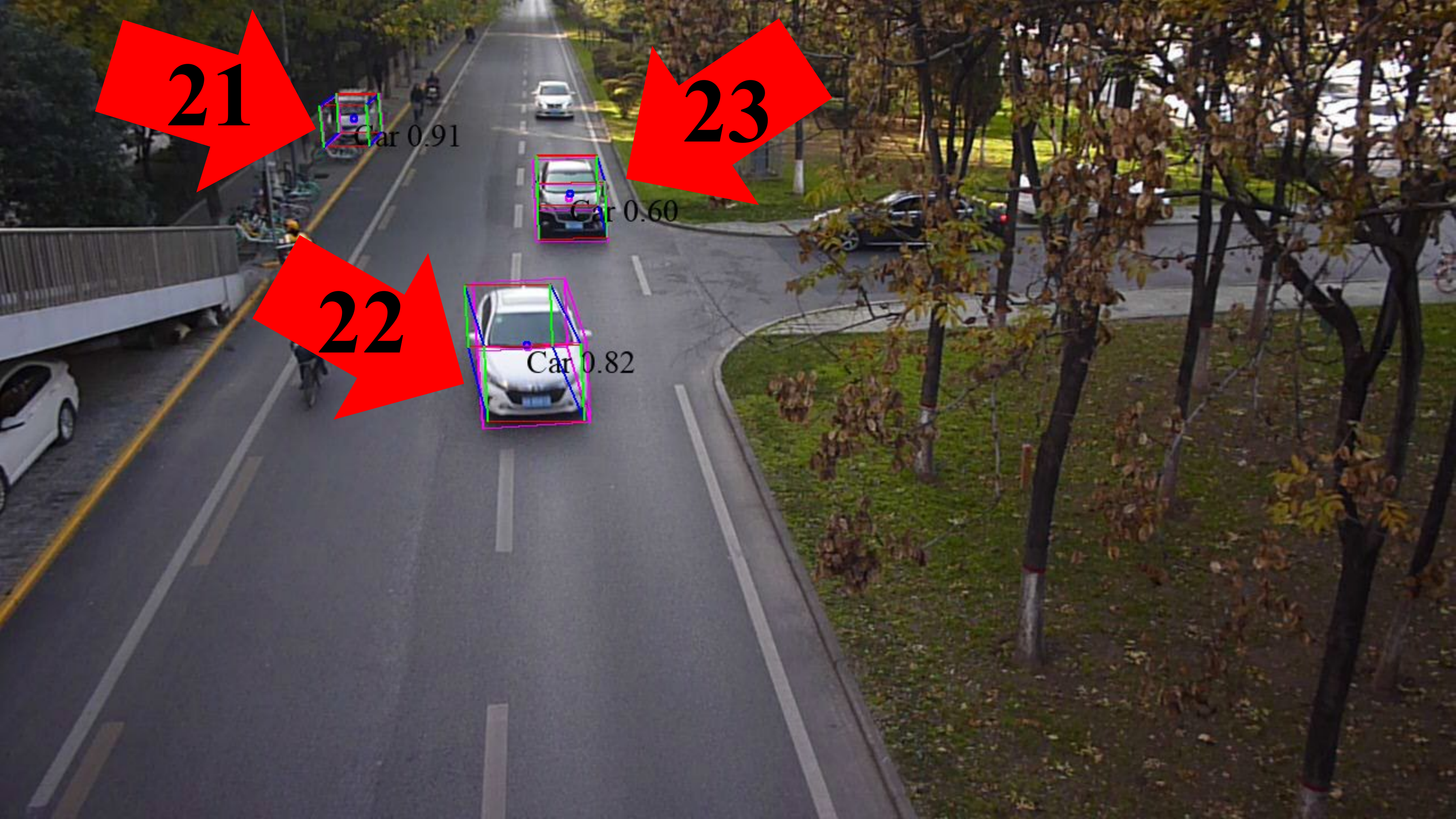}%
		\label{fig:vis_loc_results_h}}
	\hfil
	\subfloat[\centering ]{\includegraphics[width=0.28\linewidth]{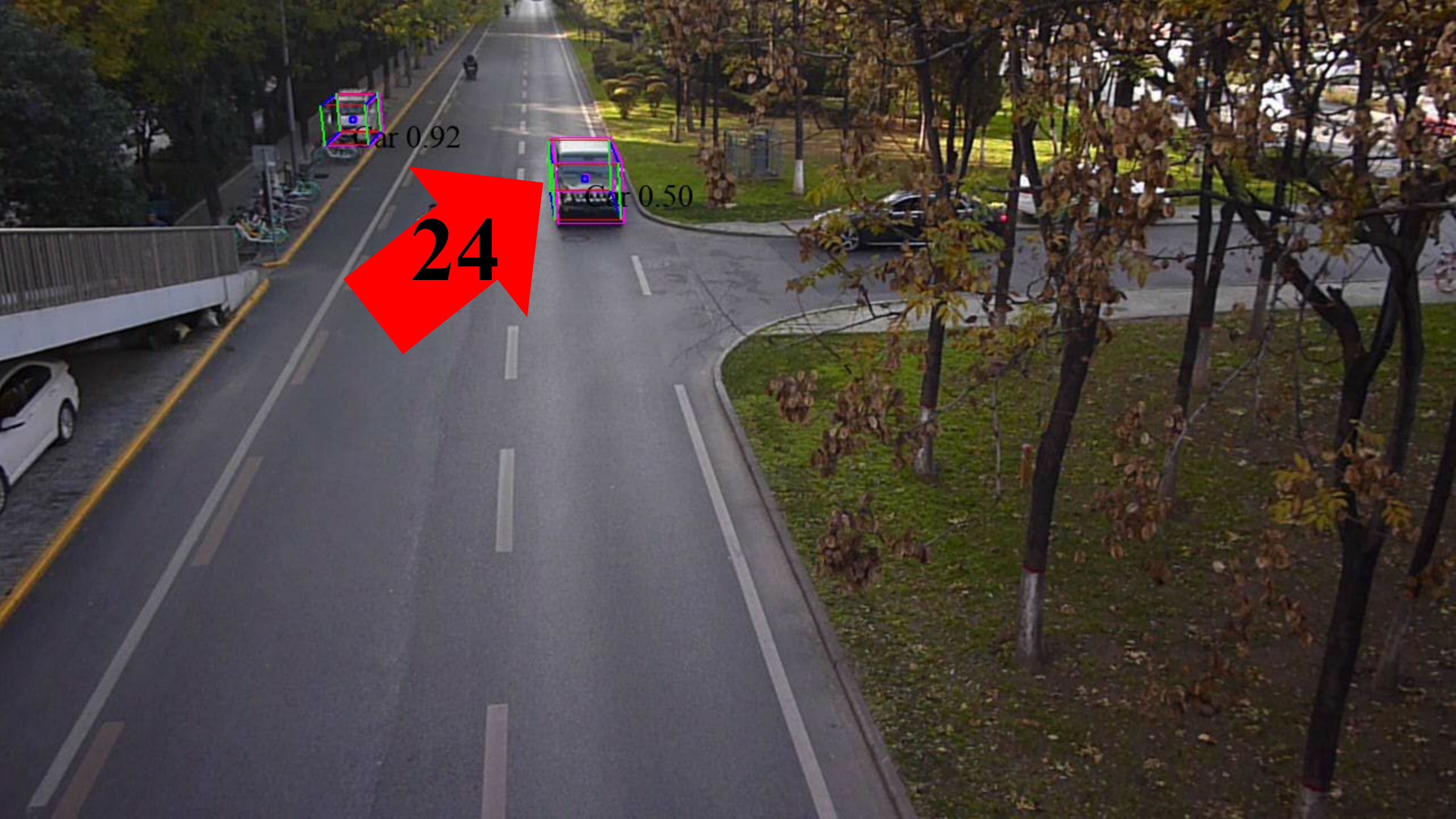}%
		\label{fig:vis_loc_results_i}}
	\newline
	\caption{\leftskip=0pt \rightskip=0pt plus 0cm Visualization results of 3D vehicle localization on SVLD-3D test set. Color boxes indicate the predicted results. Pink boxes indicate the ground truth results. Numbers on the arrow pointing to vehicles are consistent with numbers in Table \ref{tab:table_loc_results}.}
	\label{fig:vis_loc_results}
\end{figure*}

Table \ref{tab:table_loc_results} and Figure \ref{fig:vis_loc_results} show 3D vehicle localization results and precision of different scenes and types in SVLD-3D test set, which is calculated by Equation \ref{equa_ploc}. It can be seen that our network can also achieve good results in 3D vehicle localization, with an average precision of 98\%.

\begin{figure}[htbp]
	\centering
	\subfloat[\centering Scene A-4093]{\includegraphics[width=0.15\linewidth]{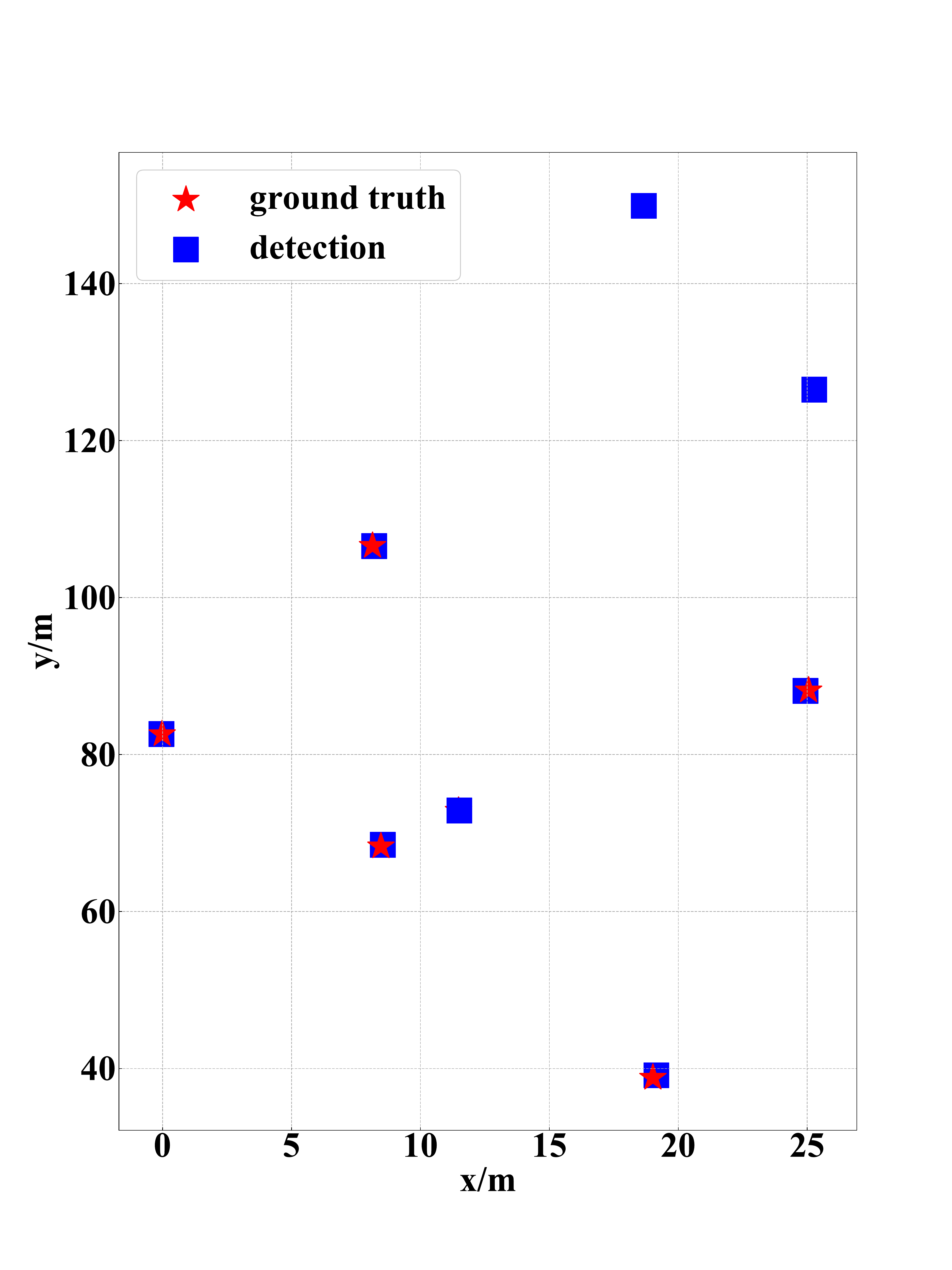}%
		\label{fig:vis_loc_results_top_a}}
	\hfil
	\subfloat[\centering Scene A-4172]{\includegraphics[width=0.15\linewidth]{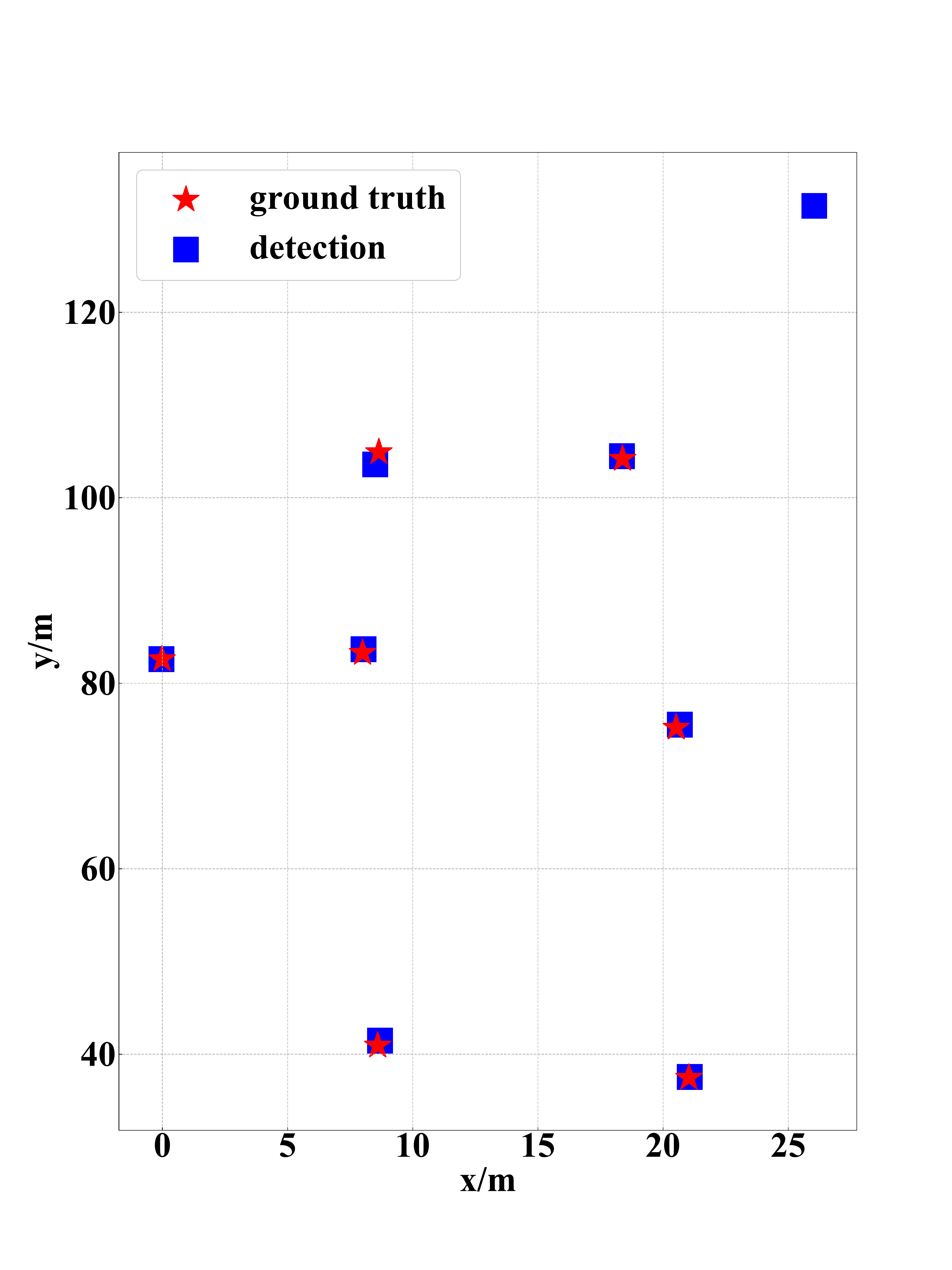}%
		\label{fig:vis_loc_results_top_b}}
	\hfil
	\subfloat[\centering Scene A-4391]{\includegraphics[width=0.15\linewidth]{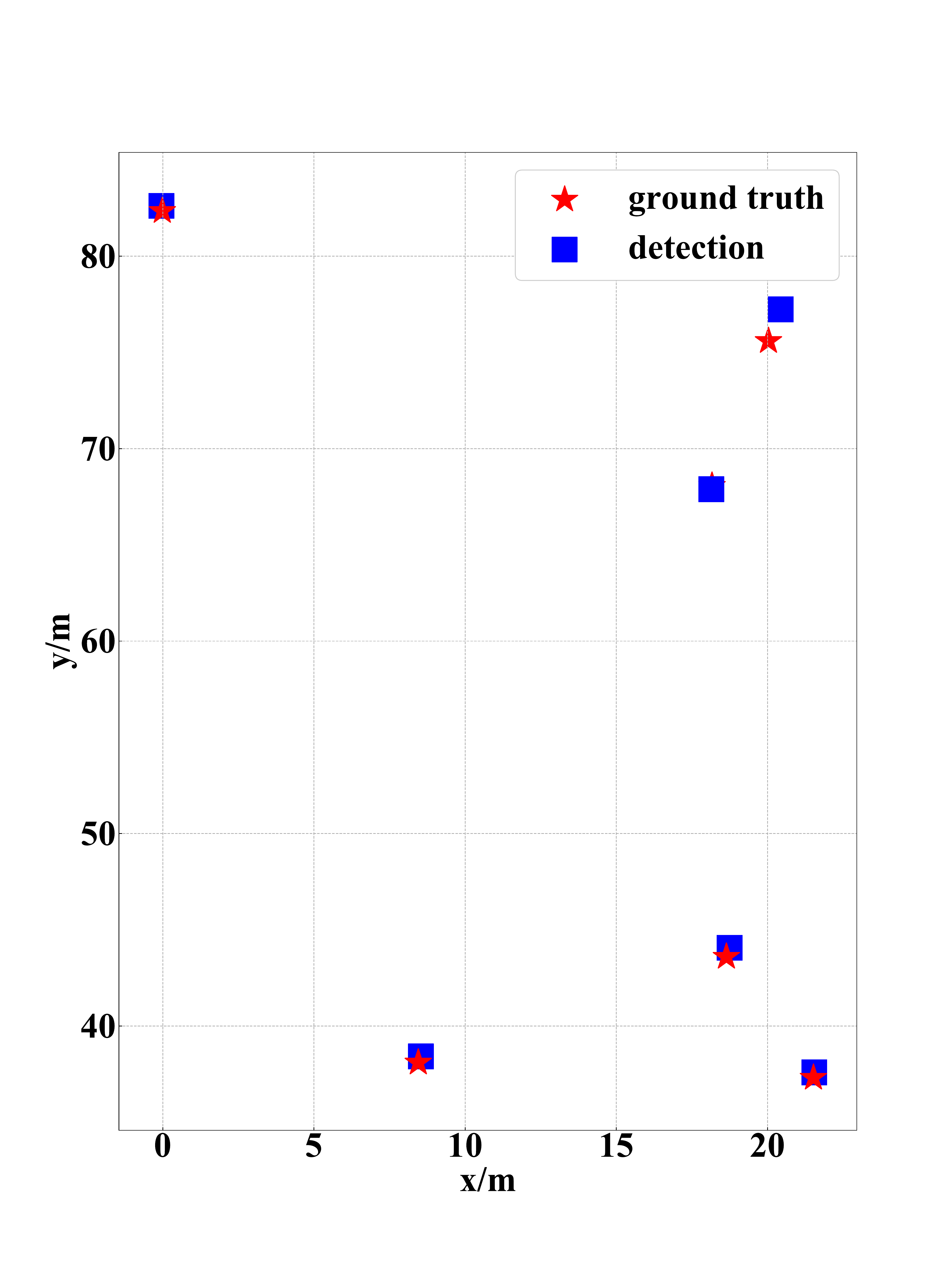}%
		\label{fig:vis_loc_results_top_c}}
	\newline
	\subfloat[\centering Scene B-6647]{\includegraphics[width=0.15\linewidth]{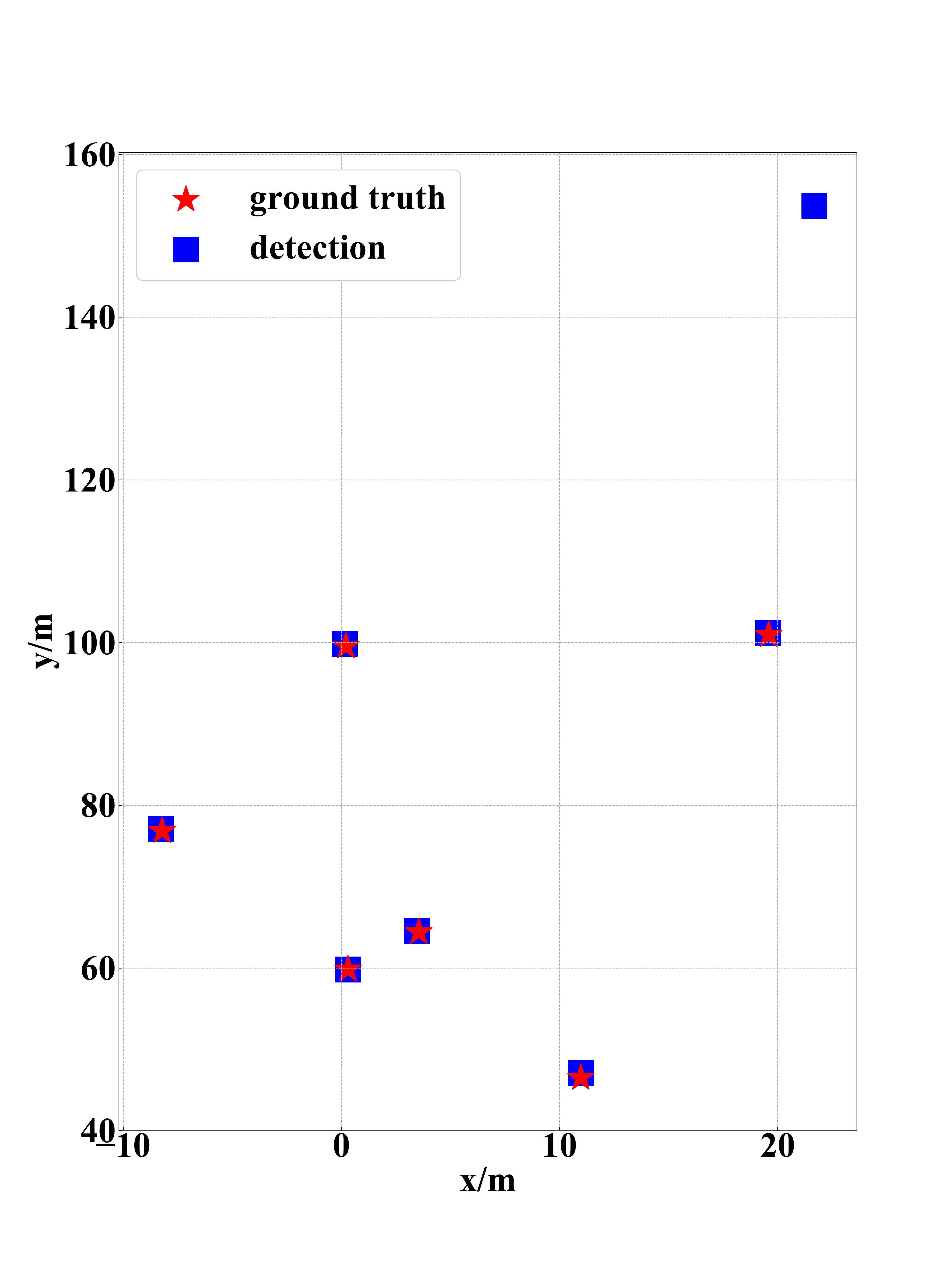}%
		\label{fig:vis_loc_results_top_d}}
	\hfil
	\subfloat[\centering Scene B-6751]{\includegraphics[width=0.15\linewidth]{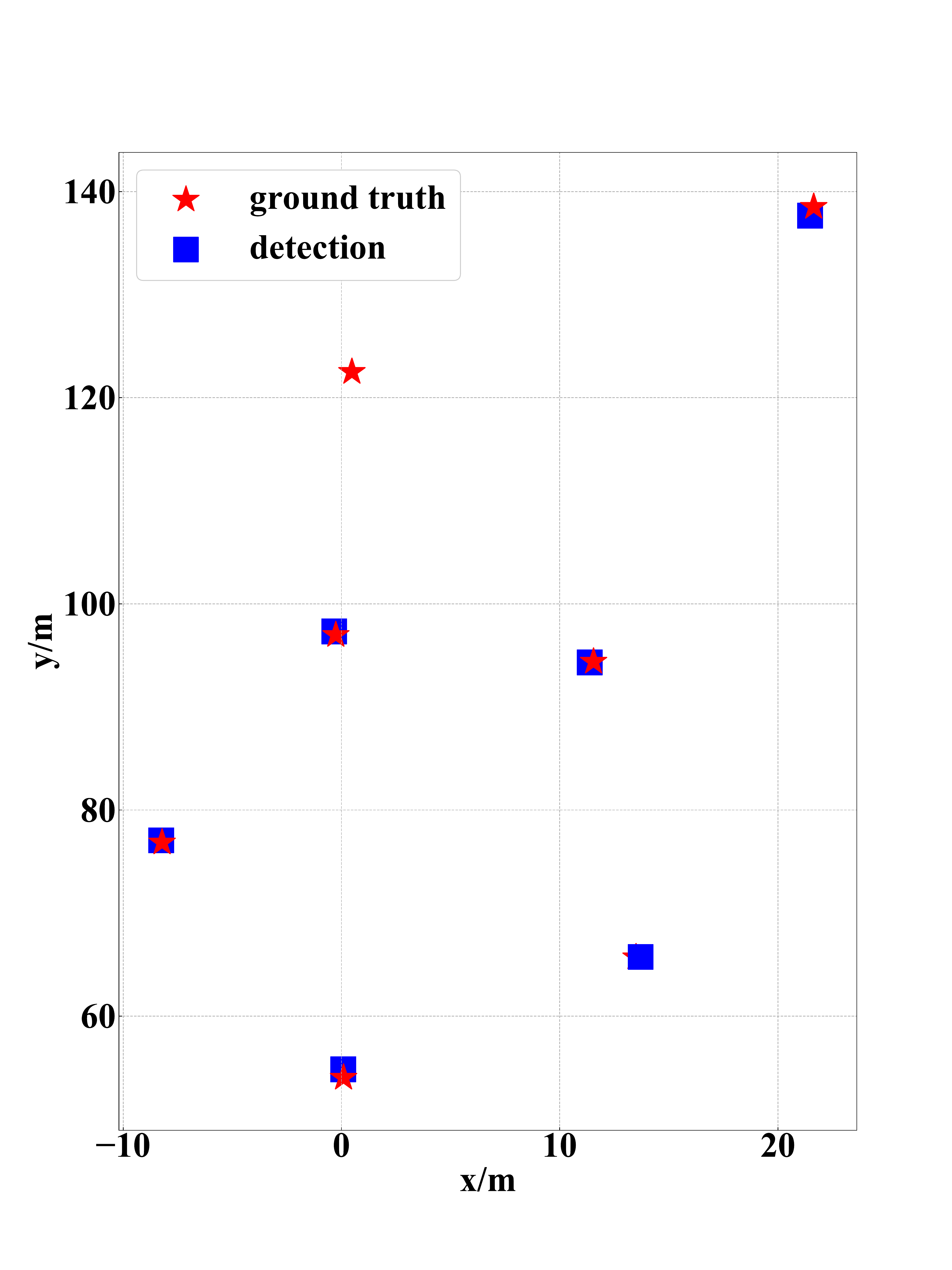}%
		\label{fig:vis_loc_results_top_e}}
	\hfil
	\subfloat[\centering Scene B-8001]{\includegraphics[width=0.15\linewidth]{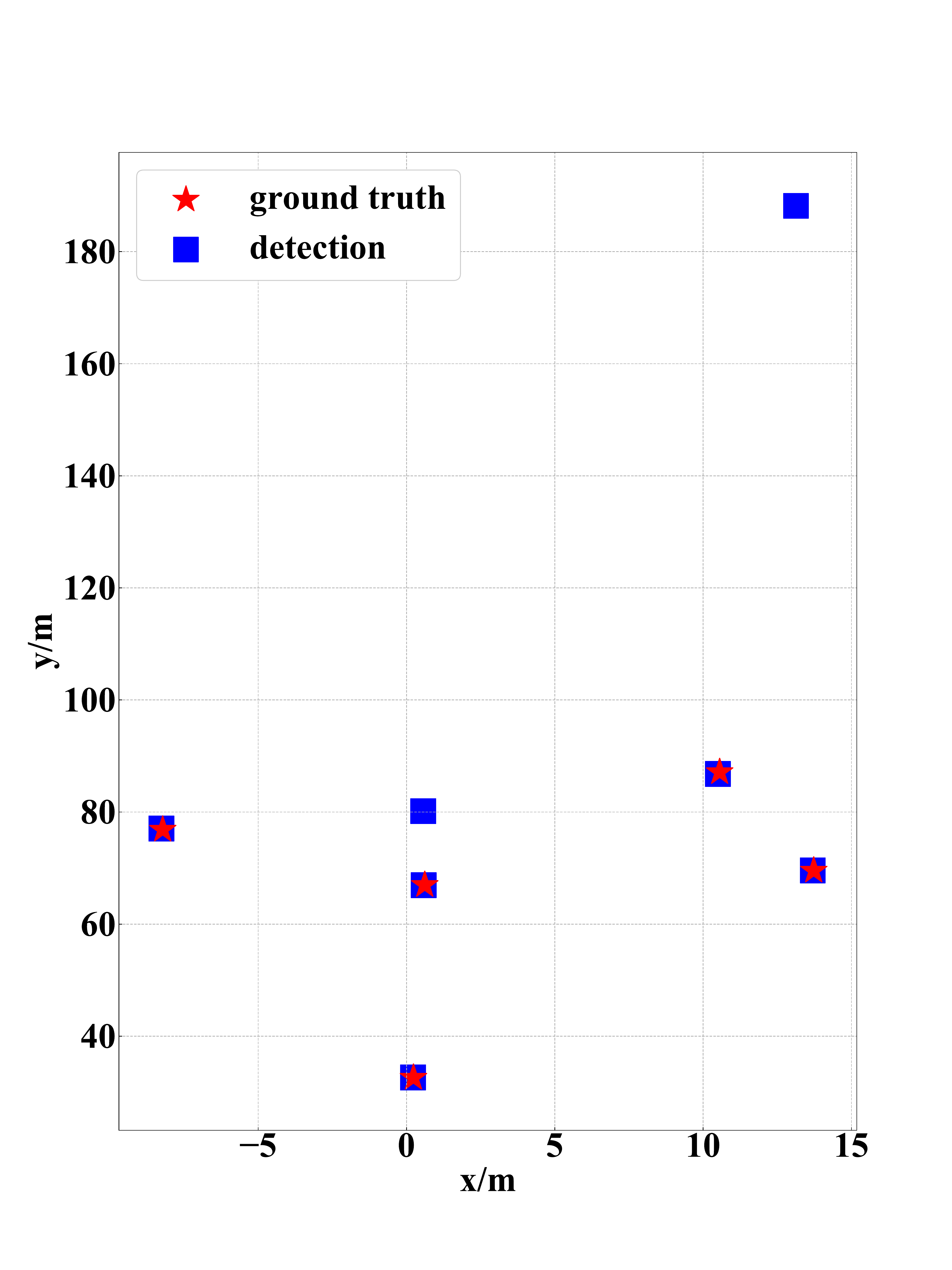}%
		\label{fig:vis_loc_results_top_f}}
	\newline
	\subfloat[\centering Scene C-8241]{\includegraphics[width=0.15\linewidth]{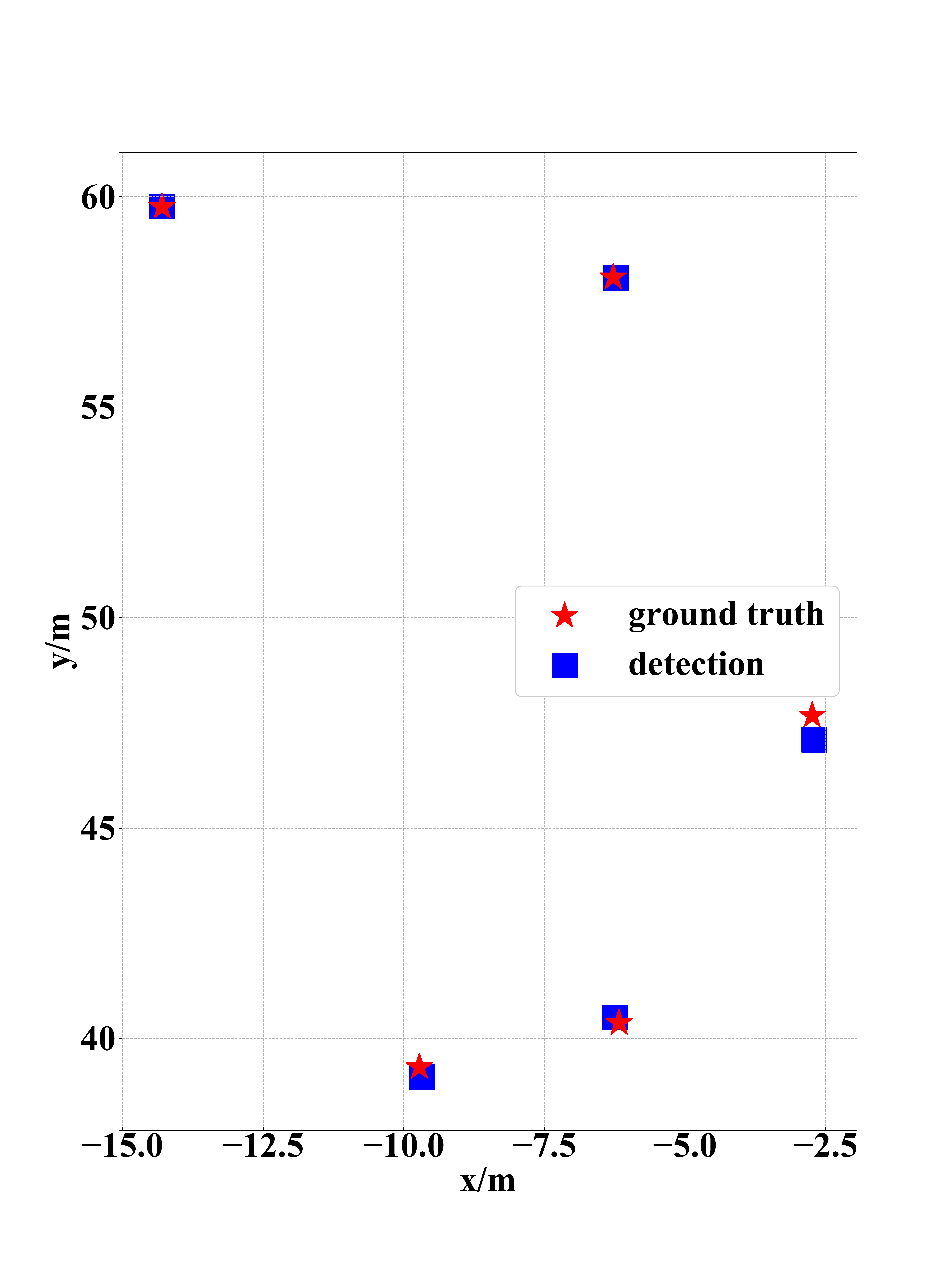}%
		\label{fig:vis_loc_results_top_g}}
	\hfil
	\subfloat[\centering Scene C-8287]{\includegraphics[width=0.15\linewidth]{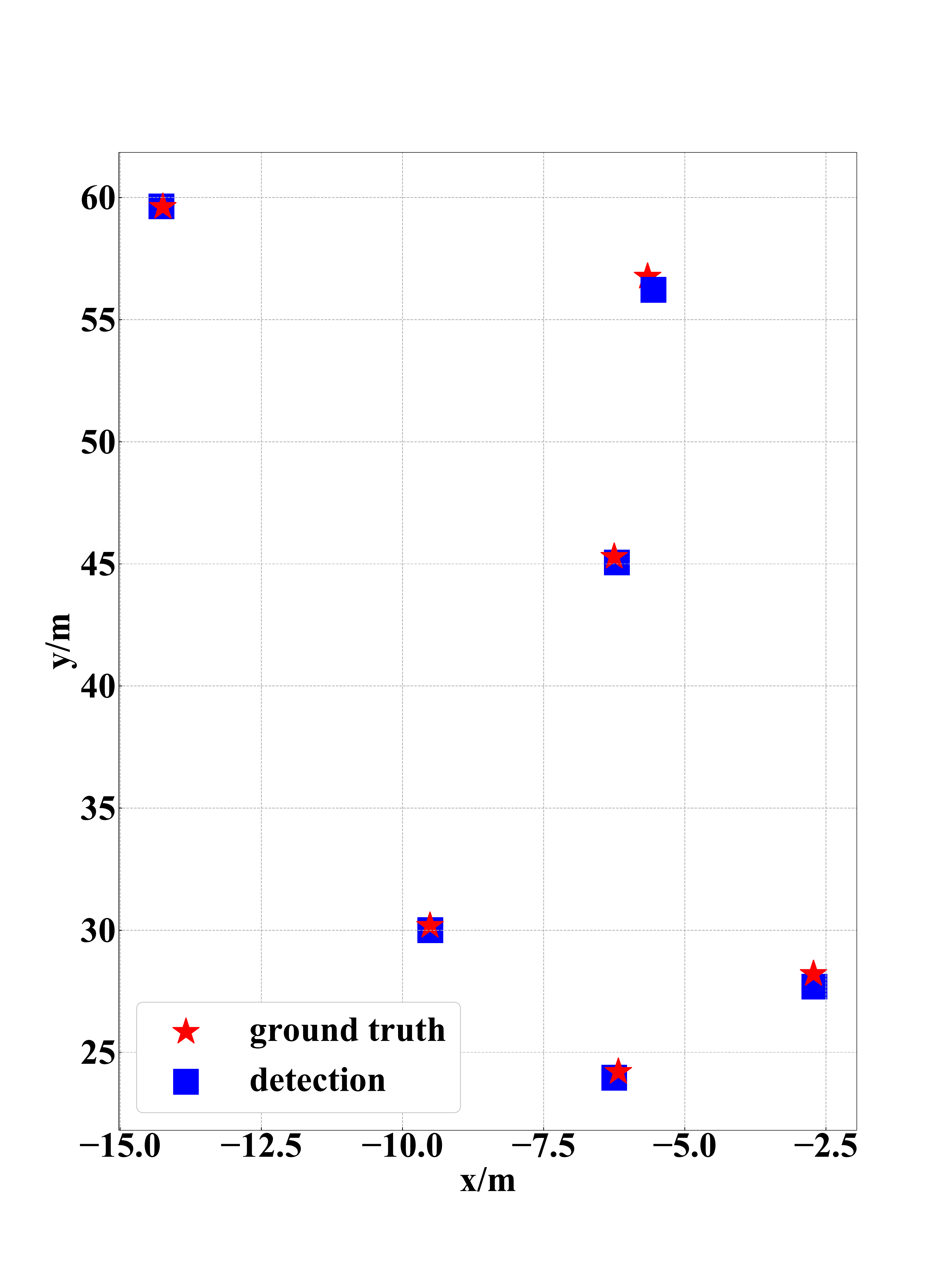}%
		\label{fig:vis_loc_results_top_h}}
	\hfil
	\subfloat[\centering Scene C-8351]{\includegraphics[width=0.15\linewidth]{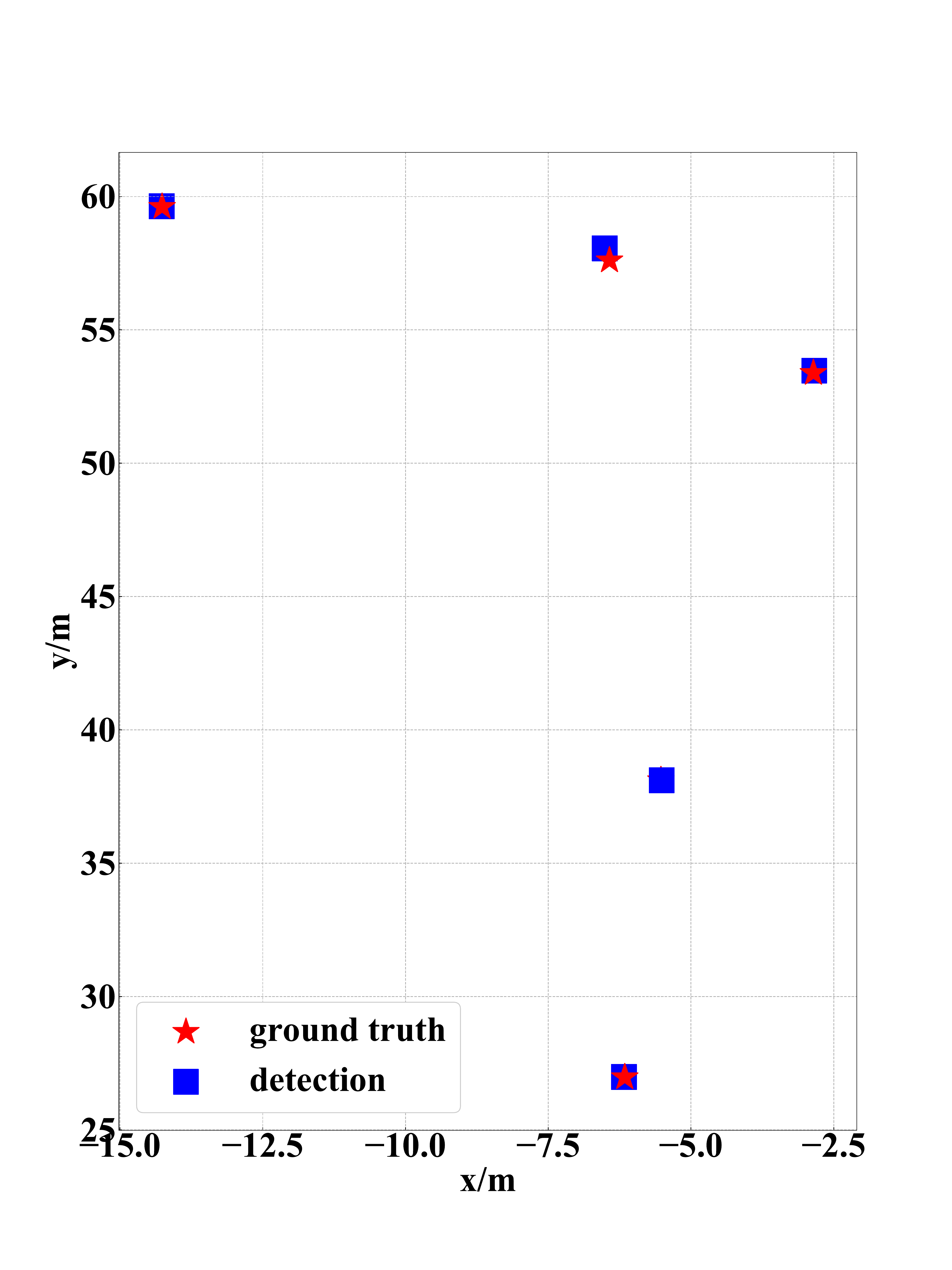}%
		\label{fig:vis_loc_results_top_i}}
	\newline
	\subfloat[\centering Scene D-4752]{\includegraphics[width=0.15\linewidth]{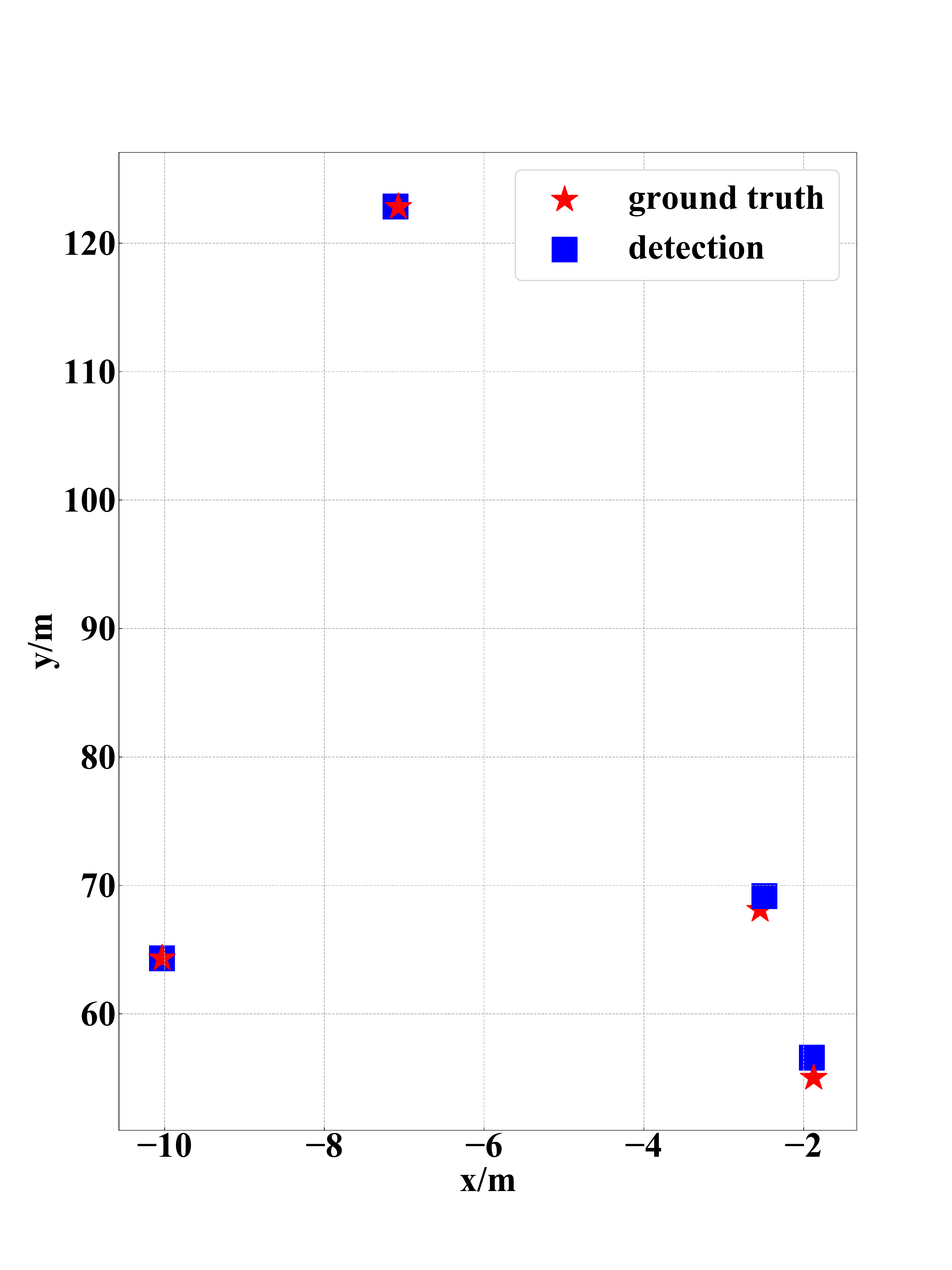}%
		\label{fig:vis_loc_results_top_j}}
	\hfil
	\subfloat[\centering Scene D-5743]{\includegraphics[width=0.15\linewidth]{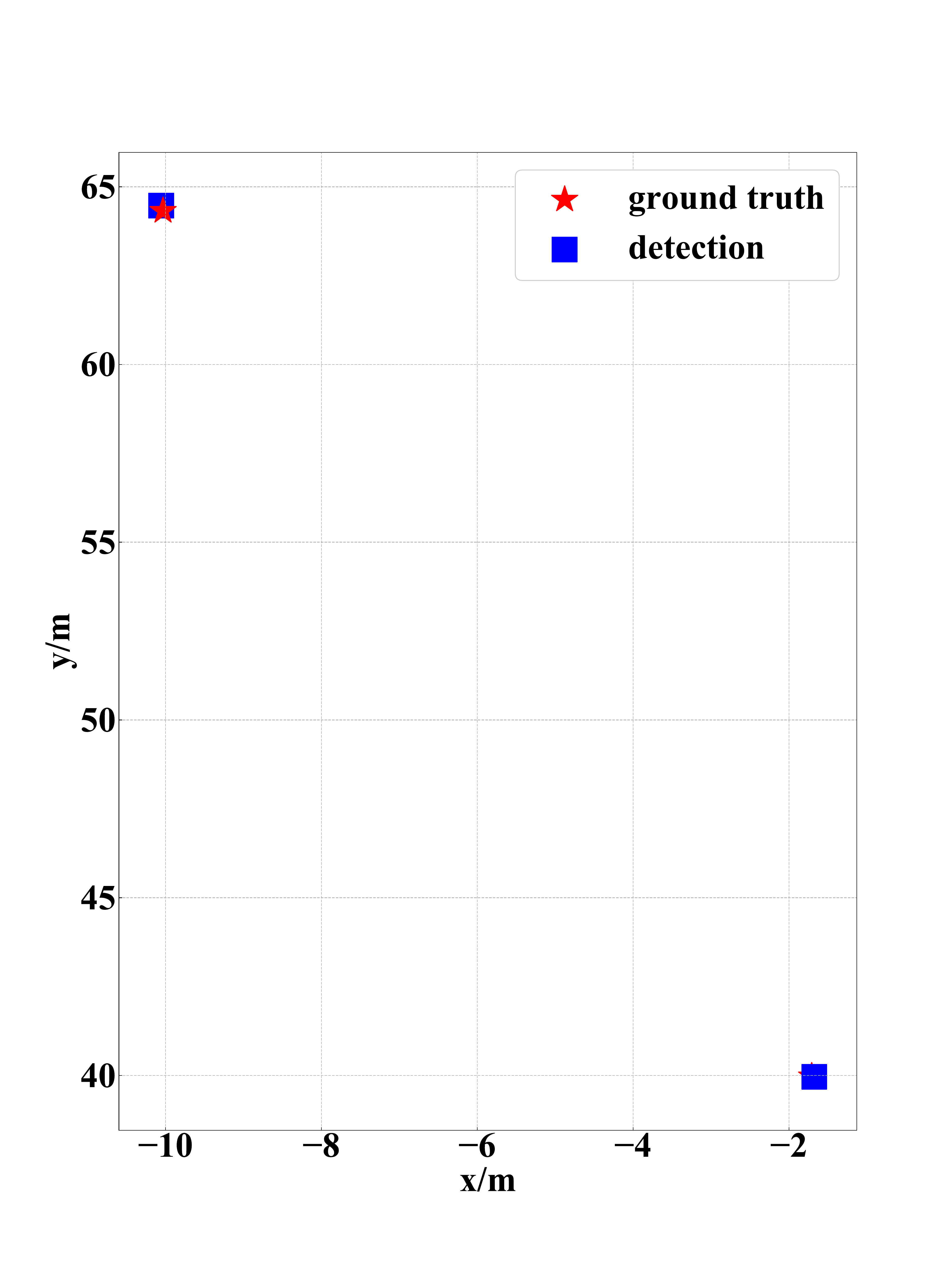}%
		\label{fig:vis_loc_results_top_k}}
	\hfil
	\subfloat[\centering Scene D-7930]{\includegraphics[width=0.15\linewidth]{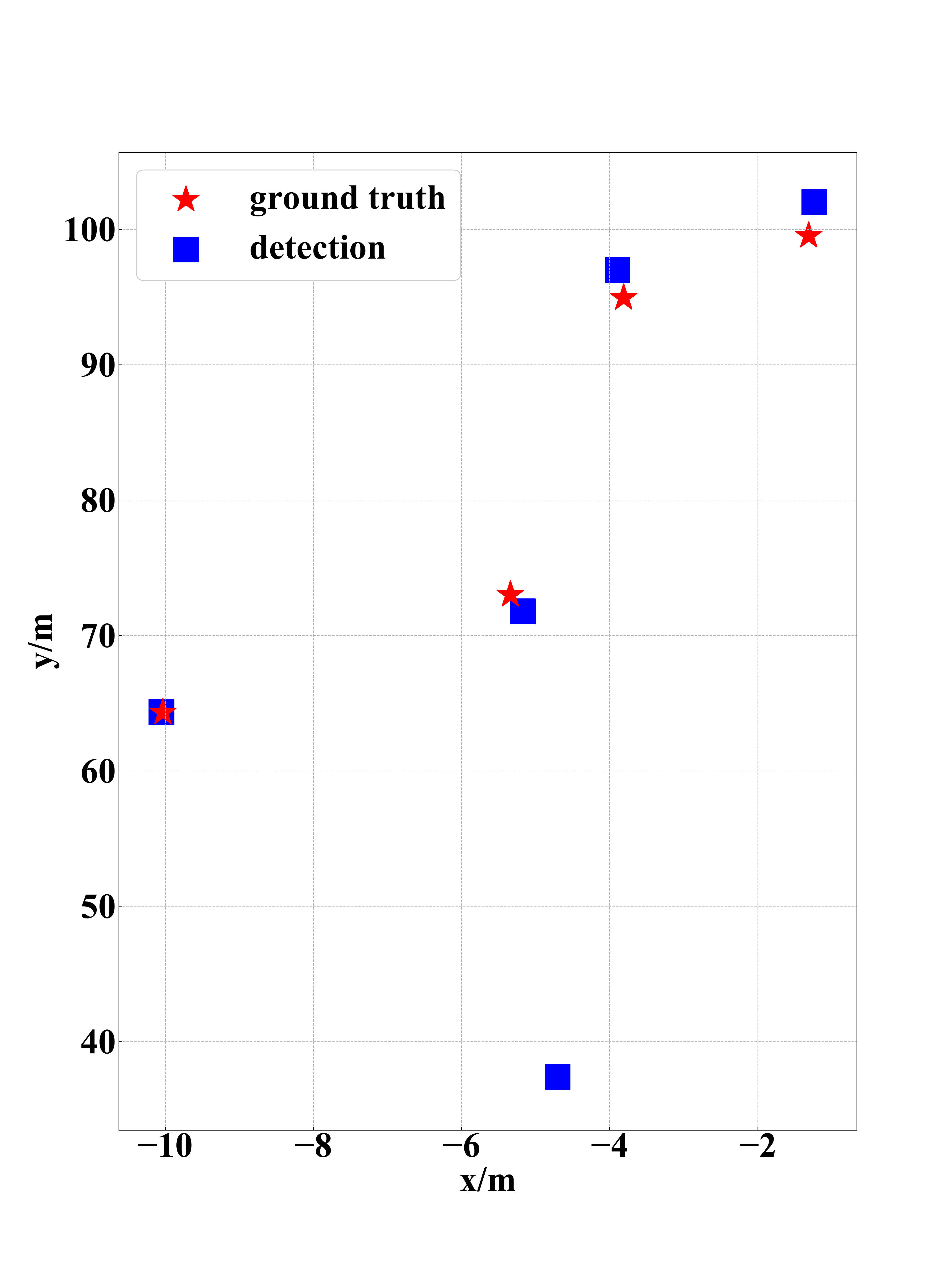}%
		\label{fig:vis_loc_results_top_l}}
	\newline
	\subfloat[\centering Scene E-4060]{\includegraphics[width=0.15\linewidth]{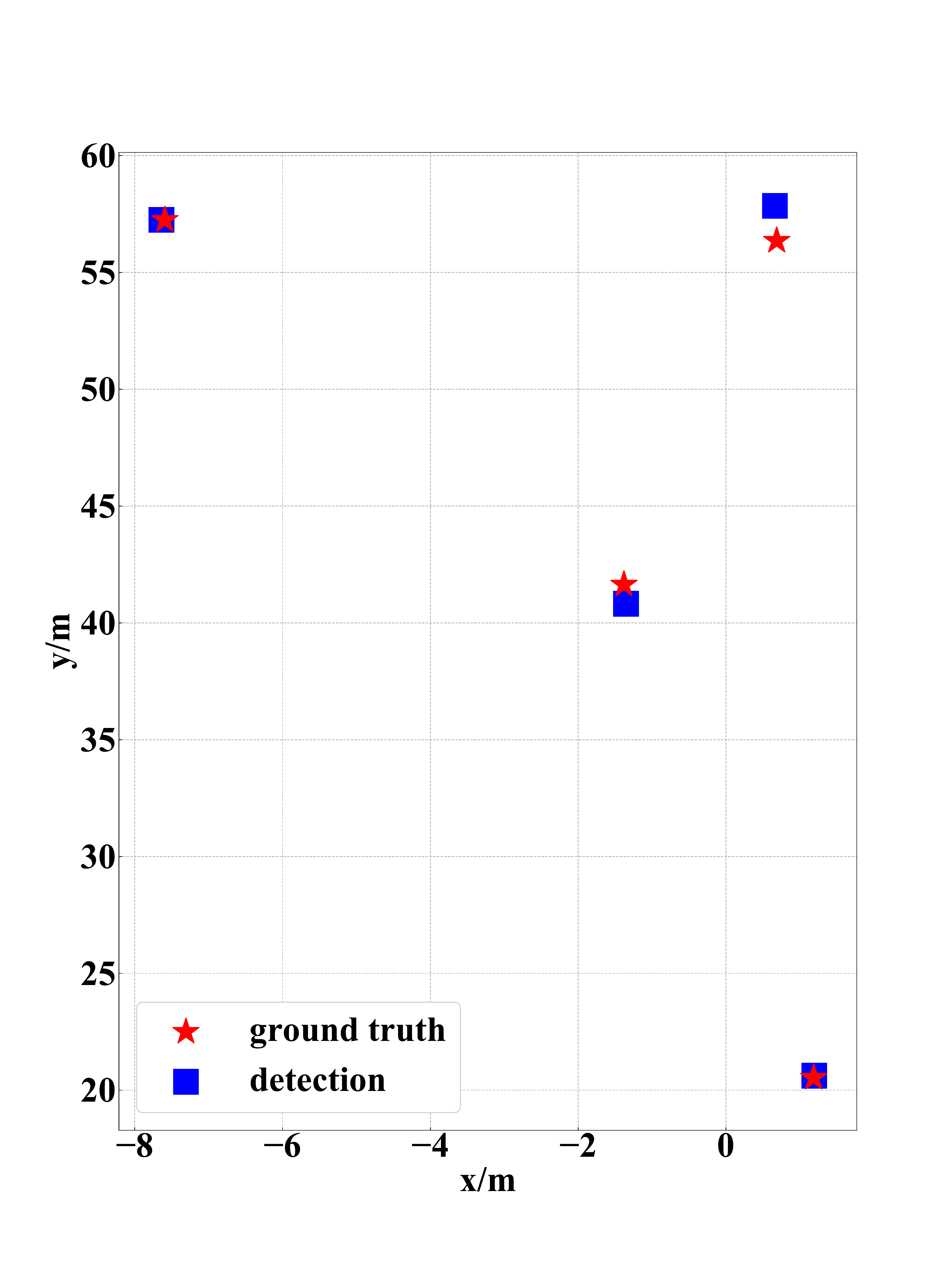}%
		\label{fig:vis_loc_results_top_m}}
	\hfil
	\subfloat[\centering Scene E-5116]{\includegraphics[width=0.15\linewidth]{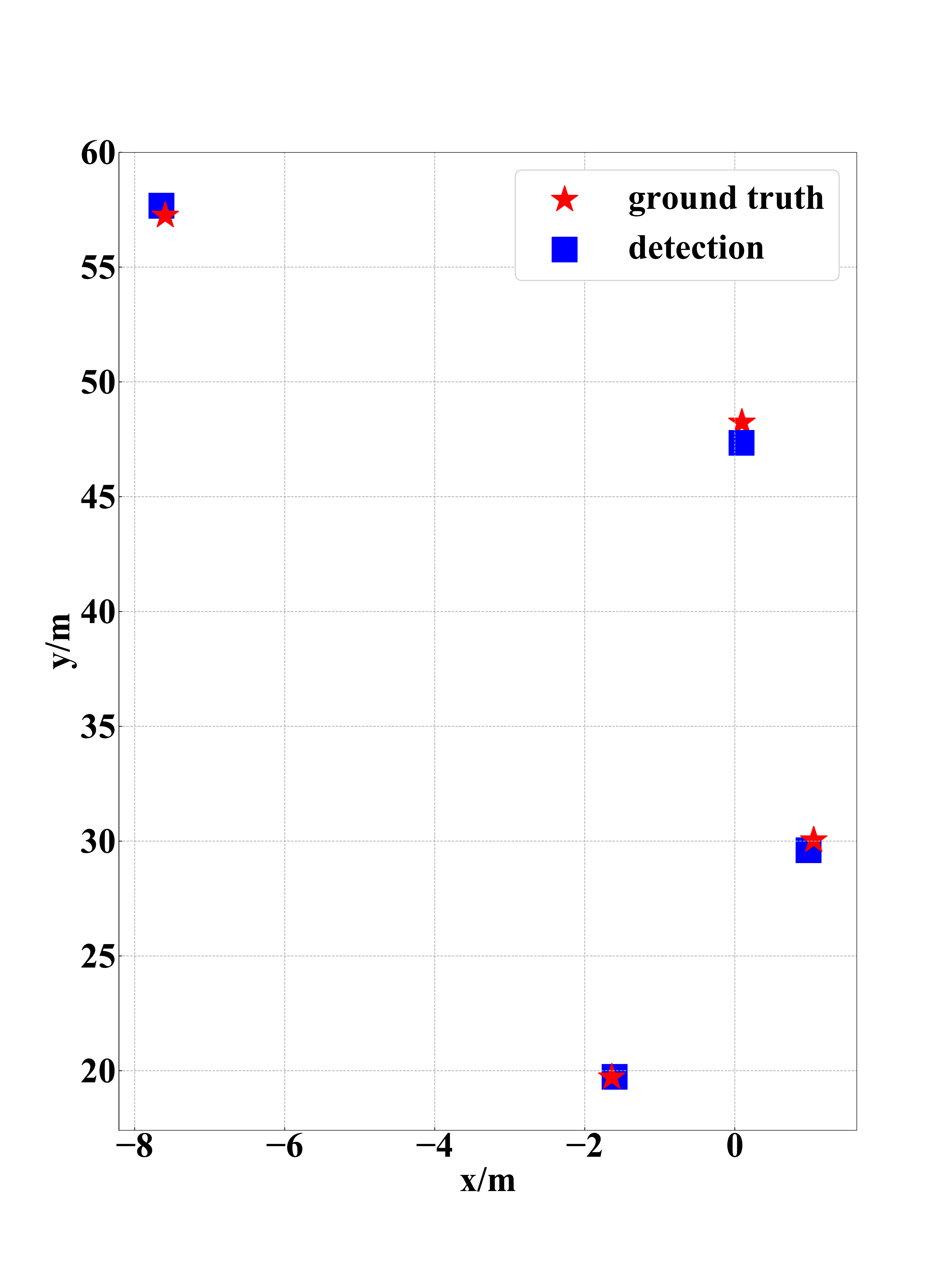}%
		\label{fig:vis_loc_results_top_n}}
	\hfil
	\subfloat[\centering Scene E-6712]{\includegraphics[width=0.15\linewidth]{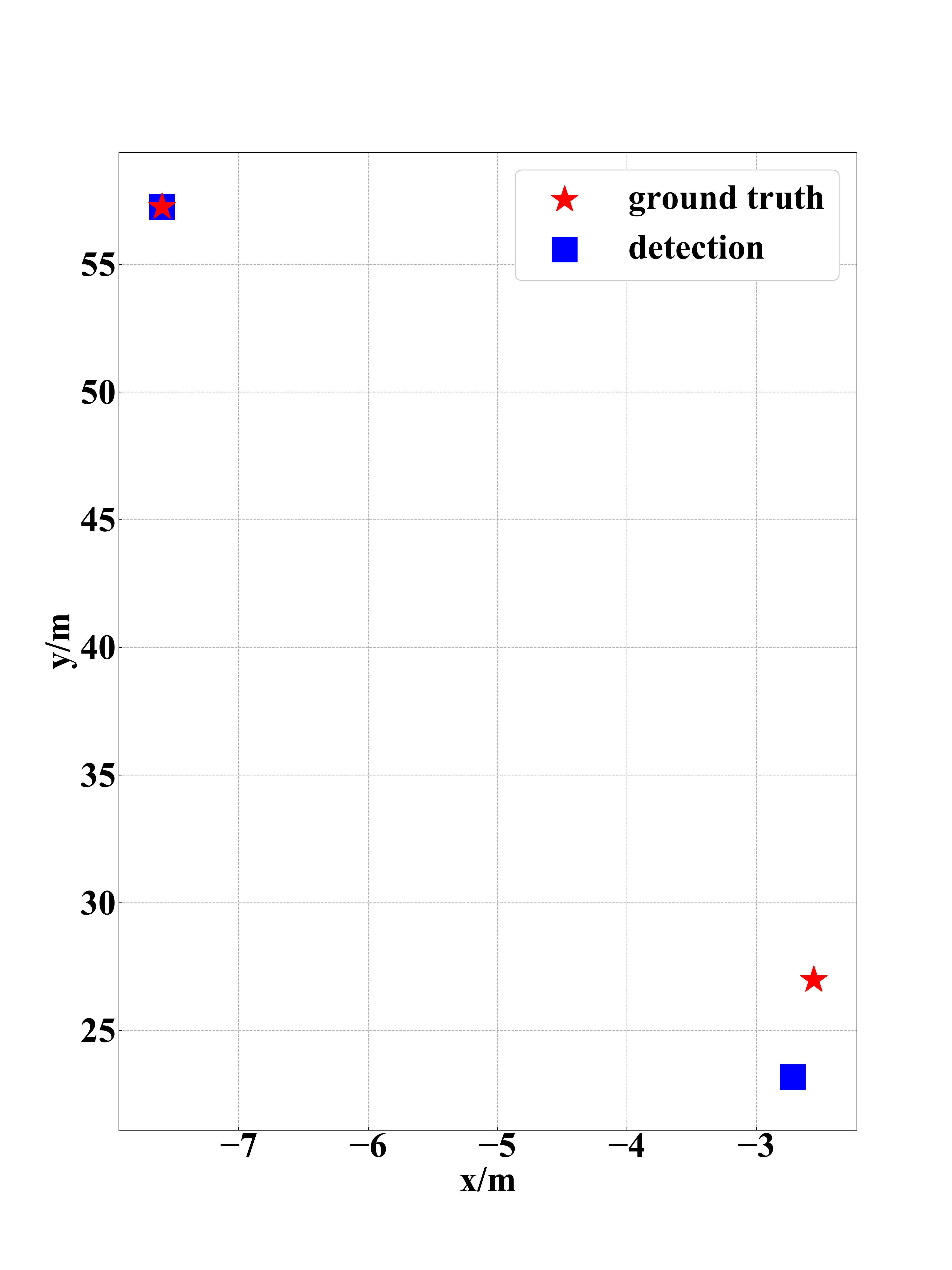}%
		\label{fig:vis_loc_results_top_o}}
	\hfil
	\newline
	\caption{\leftskip=0pt \rightskip=0pt plus 0cm Top views of 3D vehicle localization of different frames in SVLD-3D test set. The red star indicates ground truth 3D vehicle locations. The blue square indicates predicted 3D vehicle locations. The number below each sub-figure indicates the frame number.}
	\label{fig:vis_loc_results_top}
\end{figure}

Top views of 3D vehicle localization of different frames in SVLD-3D test set are shown in Figure \ref{fig:vis_loc_results_top}, which also shows that our network has high precision in 3D vehicle localization. Each sub-figure in Figure \ref{fig:vis_loc_results_top} represents different frames in different scenes. At the same time, vehicles far from the roadside camera can also be detected and localized.

\begin{table}[htbp]
	\centering
	\caption{3D vehicle localization results and precision on SVLD-3D test set.}
	\label{tab:table_loc_results}
	\resizebox{0.7\textwidth}{!}{%
		\begin{tabular}{ccccc}
			\toprule
			Vehicle & Type & ${P_{centroid}}$            & ${\widetilde P_{centorid}}$ & Precision \\
			\midrule
			1      & Car & 24.927, 88.430, 0.673 & 24.952, 88.519, 0.760 & 0.996     \\
			2      & Car & 8.611, 68.119, 0.717 & 8.585, 67.729, 0.770  & 0.991     \\
			3      & Car & -0.015, 82.595, 0.755  & -0.015, 82.595, 0.755  & 0.999     \\
			4      & Car & 8.188, 105.823, 0.780 & 8.217, 105.714, 0.825 & 0.996     \\
			5      & Car & 11.508, 72.225, 0.731 & 11.434, 71.608, 0.785 & 0.984     \\
			6      & Car & 18.829, 62.322, 0.727 & 18.791, 61.883, 0.790 & 0.990     \\
			7      & Truck & 21.315, 43.322, 0.958  & 21.219, 43.156, 0.875  & 0.990     \\
			8      & Car & 8.538, 38.425, 0.735 & 8.451, 38.100, 0.730 & 0.988     \\
			9      & Car & 18.144, 67.893, 0.674  & 18.165, 68.067, 0.700  & 0.995     \\
			10     & Car & 0.336, 43.784, 0.730 & 0.382, 43.959, 0.700 & 0.993     \\
			11     & Car & 3.812, 46.123, 0.697  & 3.794, 46.043, 0.710  & 0.997     \\
			12     & Car & 11.035, 65.584, 0.730 & 11.261, 65.935, 0.740 & 0.976     \\
			13     & Car & 0.142, 81.101, 0.721 & 0.100, 80.344, 0.770 & 0.984     \\
			14     & Car & -14.298, 59.759, 0.662 & -14.297, 59.758, 0.680 & 0.999     \\
			15     & Car & -6.232, 39.097, 0.672 & -6.193, 39.138, 0.750 & 0.993     \\
			16      & Car & -9.671, 38.371, 0.705 & -9.754, 38.703, 0.665 & 0.978     \\
			17      & Car & -6.249, 56.957, 0.681 & -6.275, 56.747, 0.690  & 0.989     \\
			18      & Car & -10.033, 64.324, 0.740  & -10.032, 64.321, 0.740  & 0.999     \\
			19      & Car & -1.770, 53.300, 0.756 & -1.820, 52.683, 0.860 & 0.975     \\
			20      & Bus & -5.174, 71.789, 1.452 & -5.341, 73.016, 1.410 & 0.936     \\
			21      & Car & -7.645, 57.673, 0.800 & -7.593, 57.247, 0.800 & 0.975     \\
			22      & Car & 0.356, 22.313, 0.739  & 0.340, 22.388, 0.670  & 0.994     \\
			23      & Car & 0.862, 37.990, 0.735 & 0.804, 37.053, 0.765 & 0.957     \\
			24      & Car & 1.376, 40.059, 0.825  & 1.412, 40.053, 0.860  & 0.993     \\
			\bottomrule
		\end{tabular}
	}
\end{table}

\begin{figure*}[htbp]
	\centering
	\subfloat[\centering Scene A]{\includegraphics[width=0.32\linewidth]{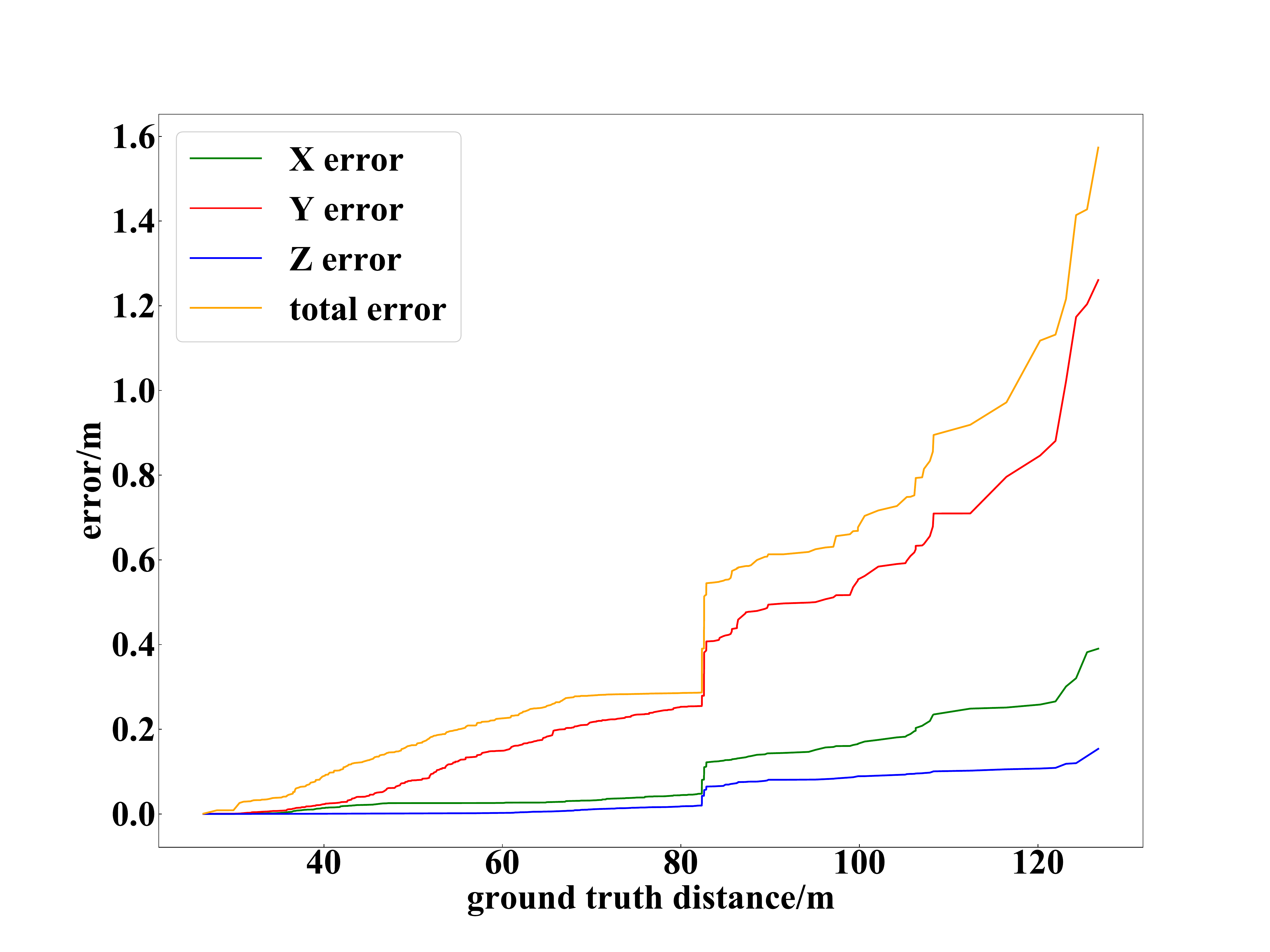}%
		\label{fig:error_loc_curves_a}}
	\subfloat[\centering Scene B]{\includegraphics[width=0.32\linewidth]{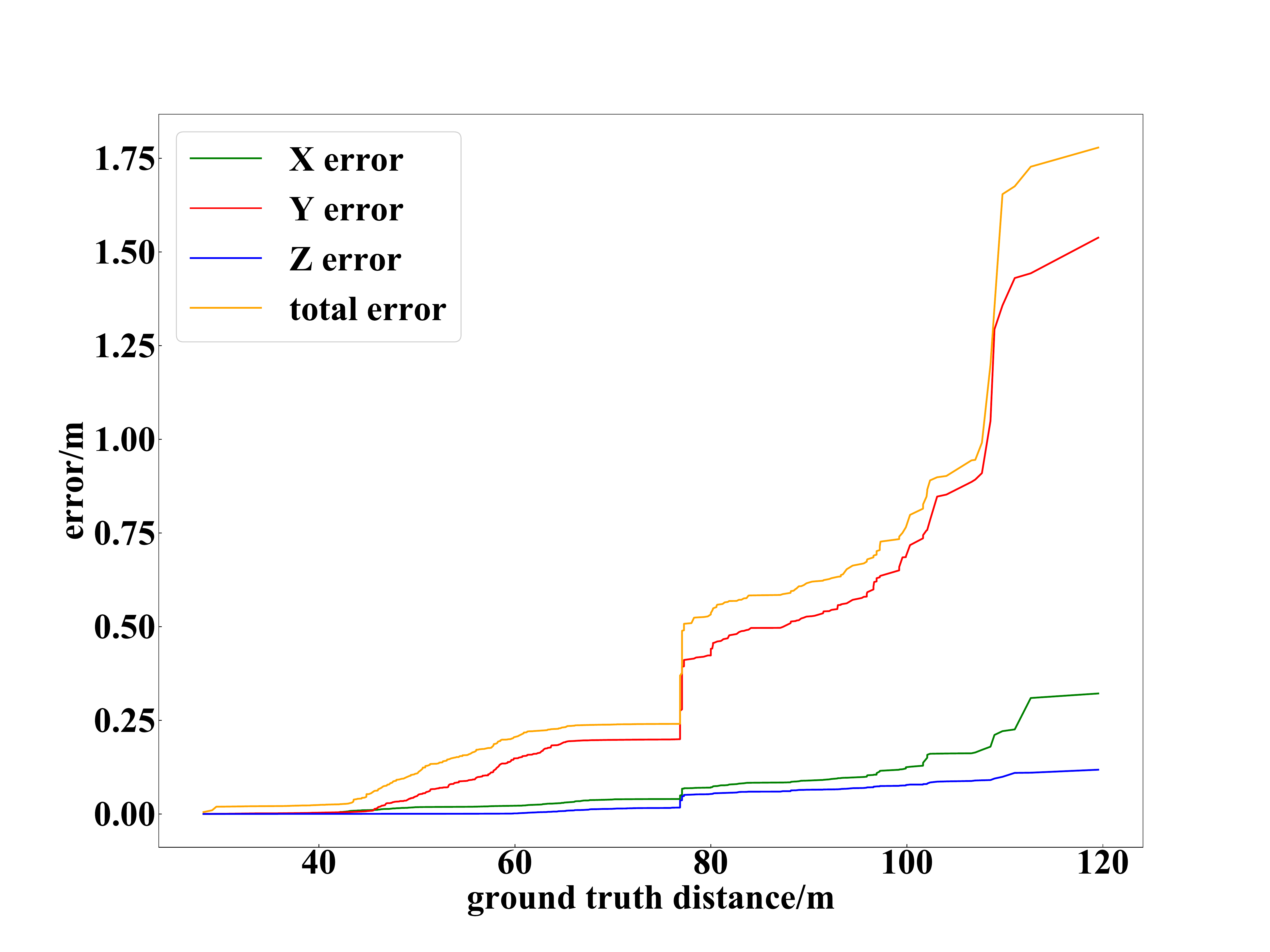}%
		\label{fig:error_loc_curves_b}}
	\subfloat[\centering Scene C]{\includegraphics[width=0.32\linewidth]{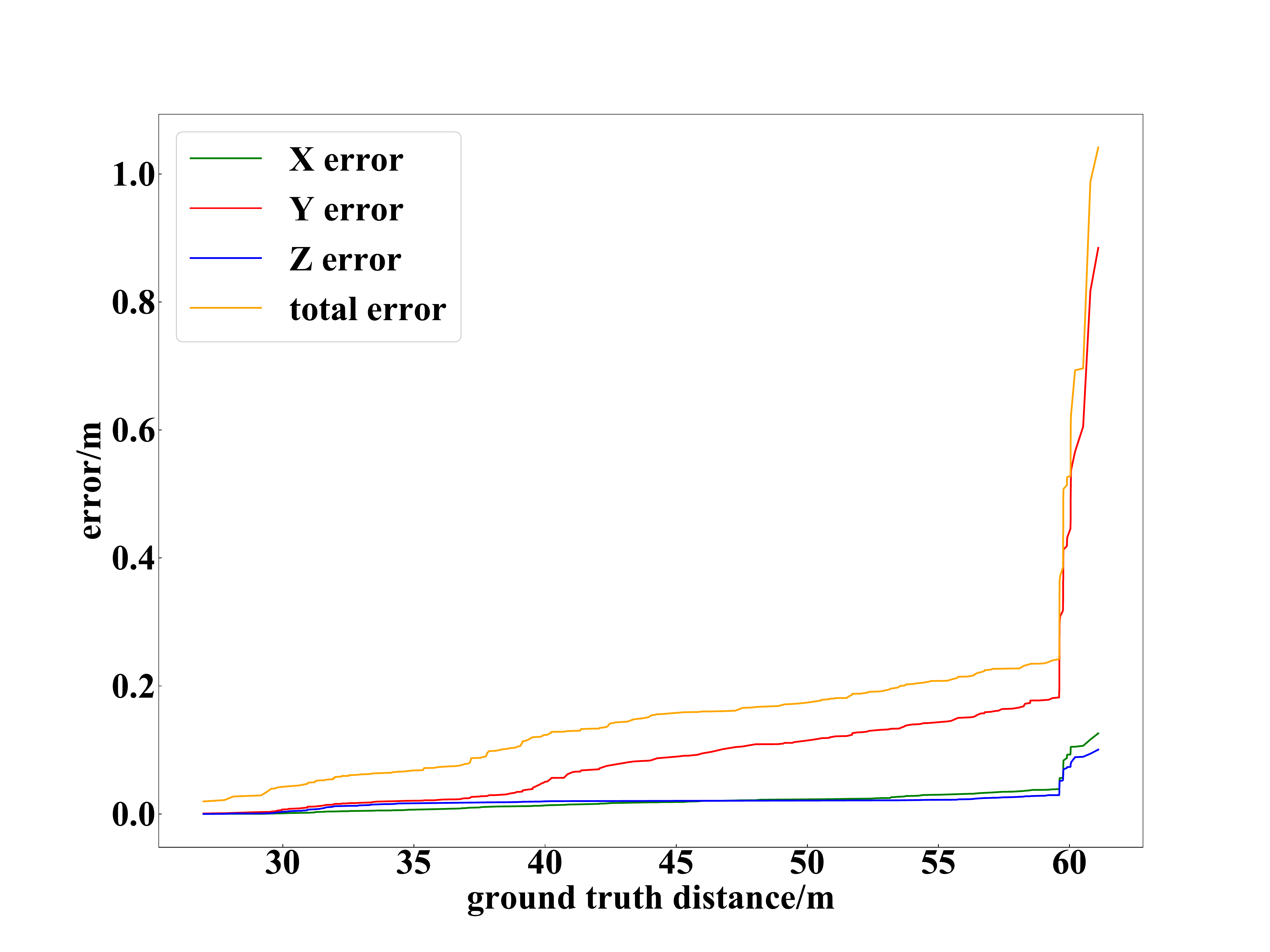}%
		\label{fig:error_loc_curves_c}}
	\newline
	\begin{center}
		\subfloat[\centering Scene D]{\includegraphics[width=0.32\linewidth]{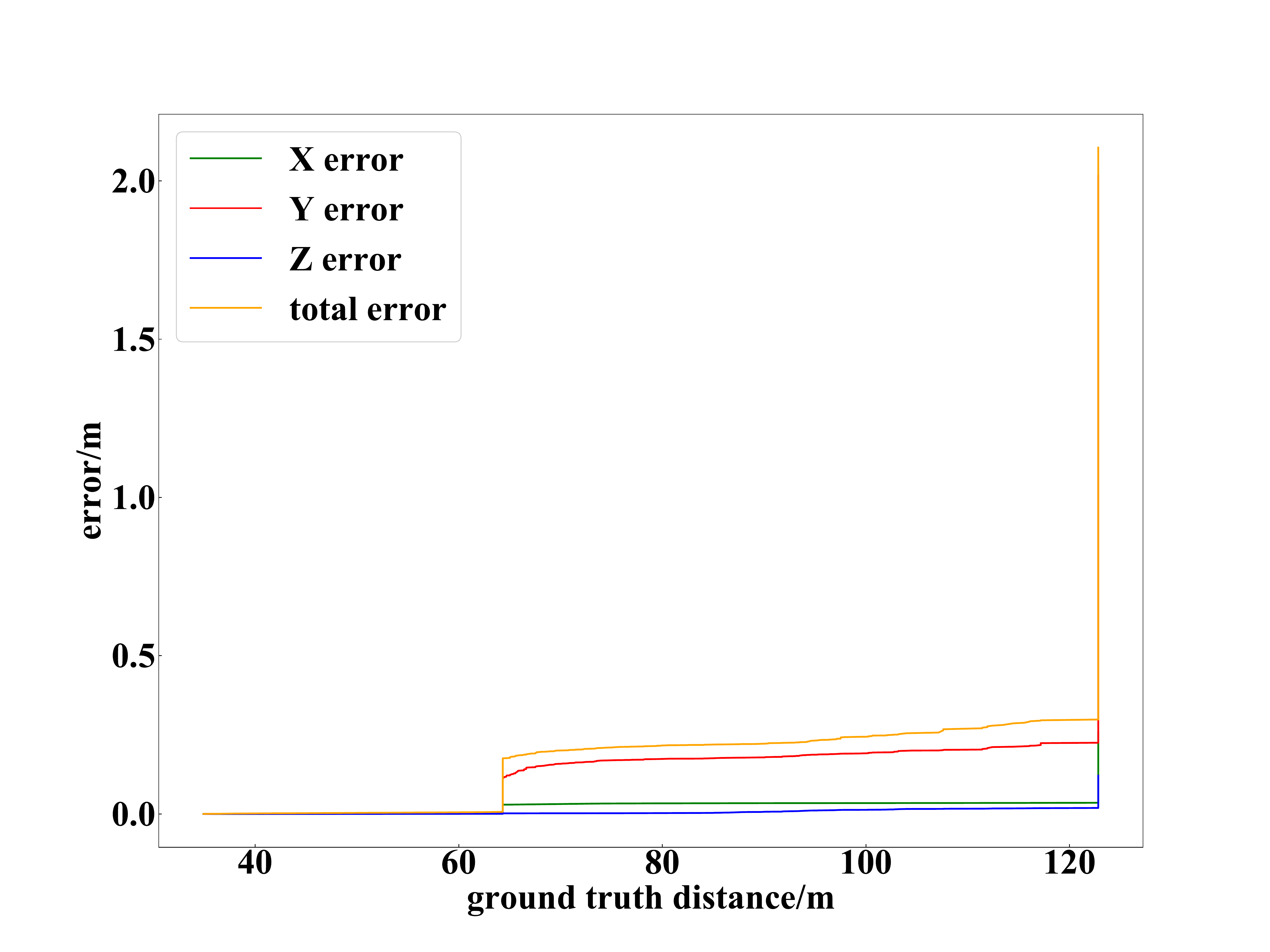}%
			\label{fig:error_loc_curves_d}}
		\subfloat[\centering Scene E]{\includegraphics[width=0.32\linewidth]{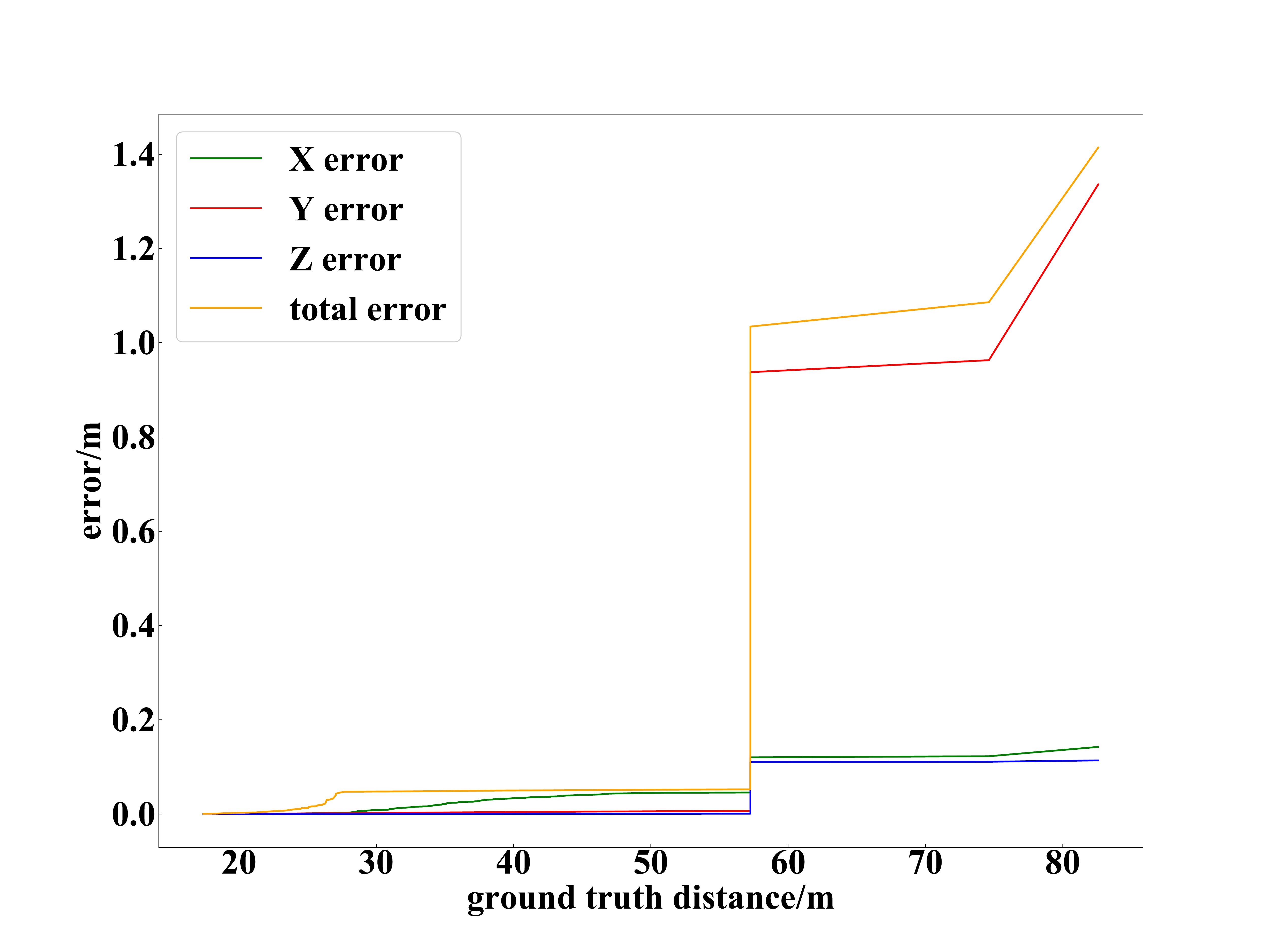}%
			\label{fig:error_loc_curves_e}}
	\end{center}
	\caption{\leftskip=0pt \rightskip=0pt plus 0cm 3D vehicle localization error of X- (green), Y- (red), Z- (blue) axis, and total error (orange) according to the distances between vehicles and roadside cameras in different scenes of SVLD-3D test set.}
	\label{fig:error_loc_curves}
\end{figure*}

3D vehicle localization error is calculated by Equation \ref{equa_eloc}. In Figure \ref{fig:error_loc_curves}, we can see that 3D vehicle localization error increases as the distance grows between vehicles and roadside cameras of different scenes in SVLD-3D test set. The Y-axis error is the largest among the X-, Y-, and Z-axis. The largest error in scene B is larger than that in other scenes. This is due to the fact that the camera pan angle in scene B is closer to 0° than in the other scenes, which leads to incomplete vehicle feature learning along the vehicle length direction.

\subsection{3D Vehicle Dimension Precision and Error of CenterLoc3D}
\label{subsec:3dsizepe of centerloc3d}

\begin{figure*}[htbp]
	\centering
	\subfloat[\centering ]{\includegraphics[width=0.28\linewidth]{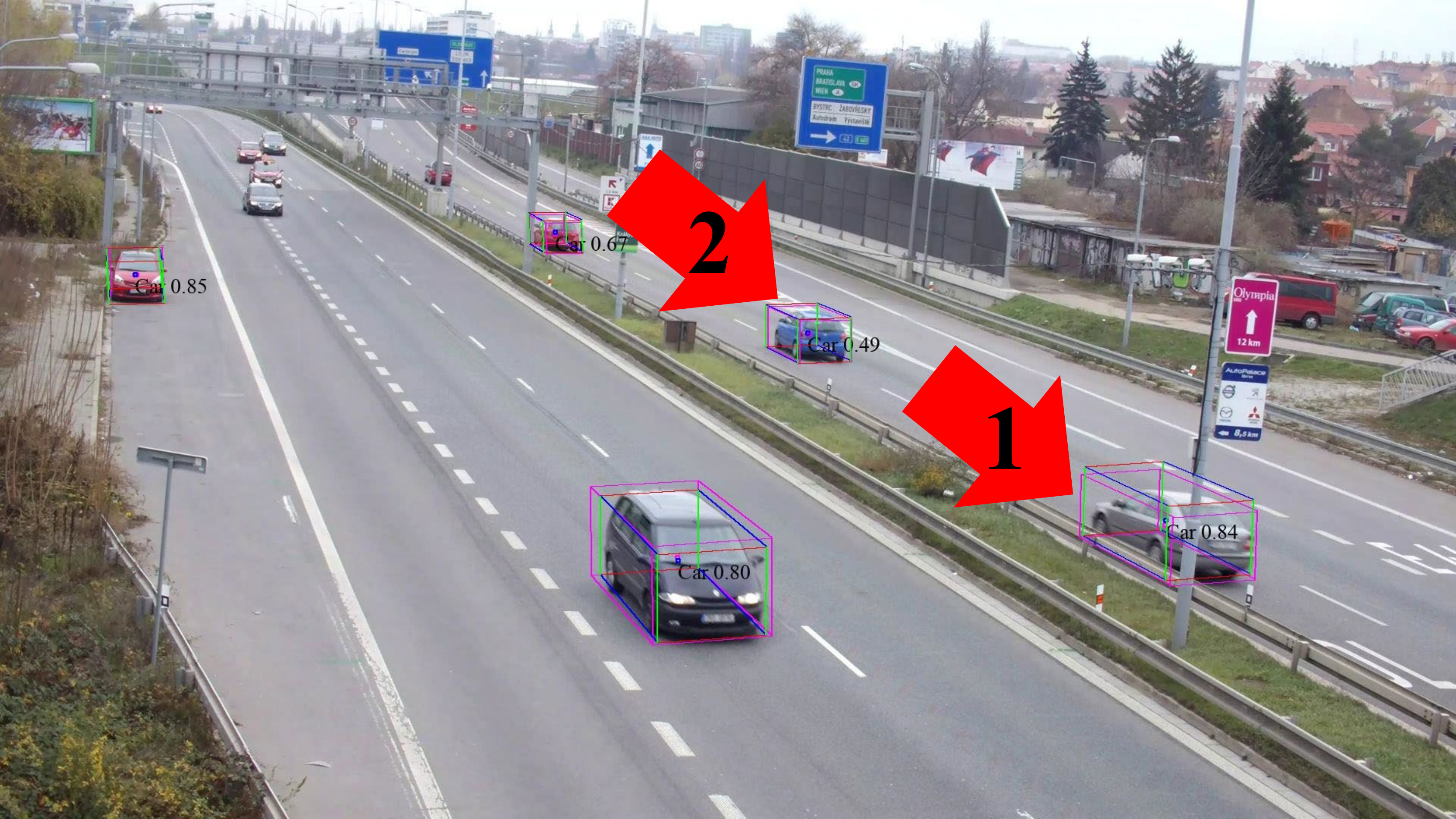}%
		\label{fig:vis_size_results_a}}
	\hfil
	\subfloat[\centering ]{\includegraphics[width=0.28\linewidth]{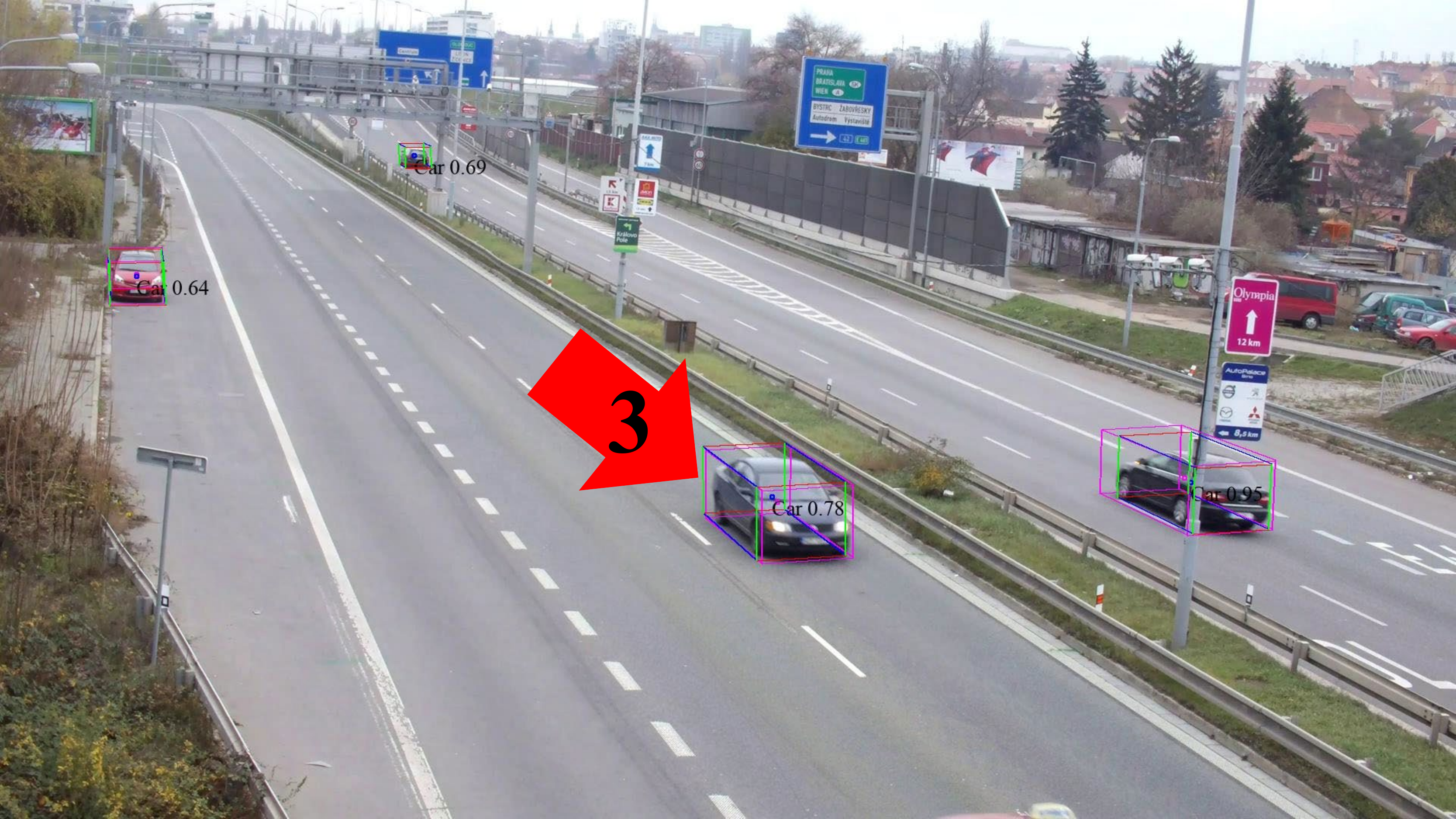}%
		\label{fig:vis_size_results_b}}
	\hfil
	\subfloat[\centering ]{\includegraphics[width=0.28\linewidth]{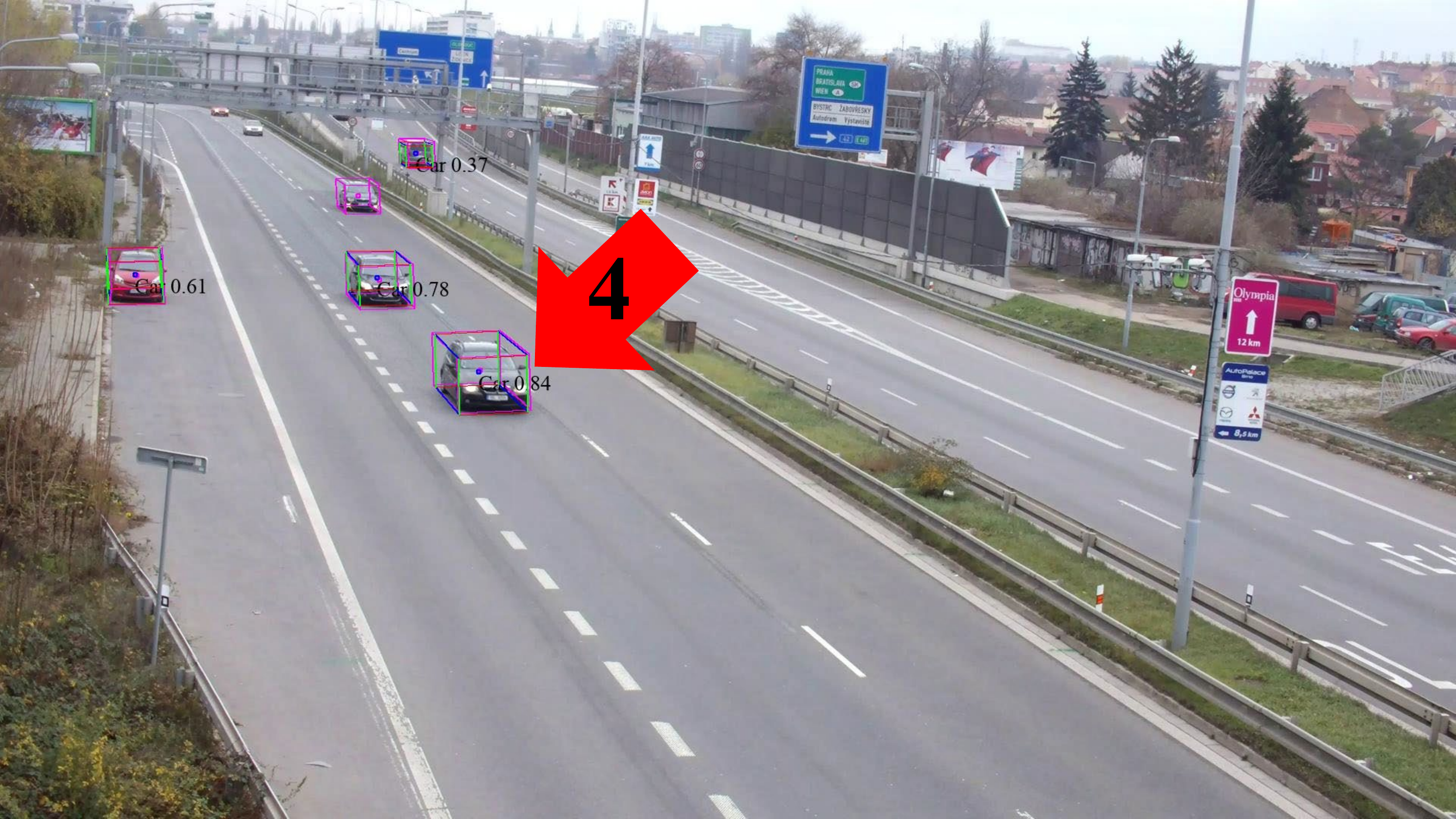}%
		\label{fig:vis_size_results_c}}
	\newline
	\subfloat[\centering ]{\includegraphics[width=0.28\linewidth]{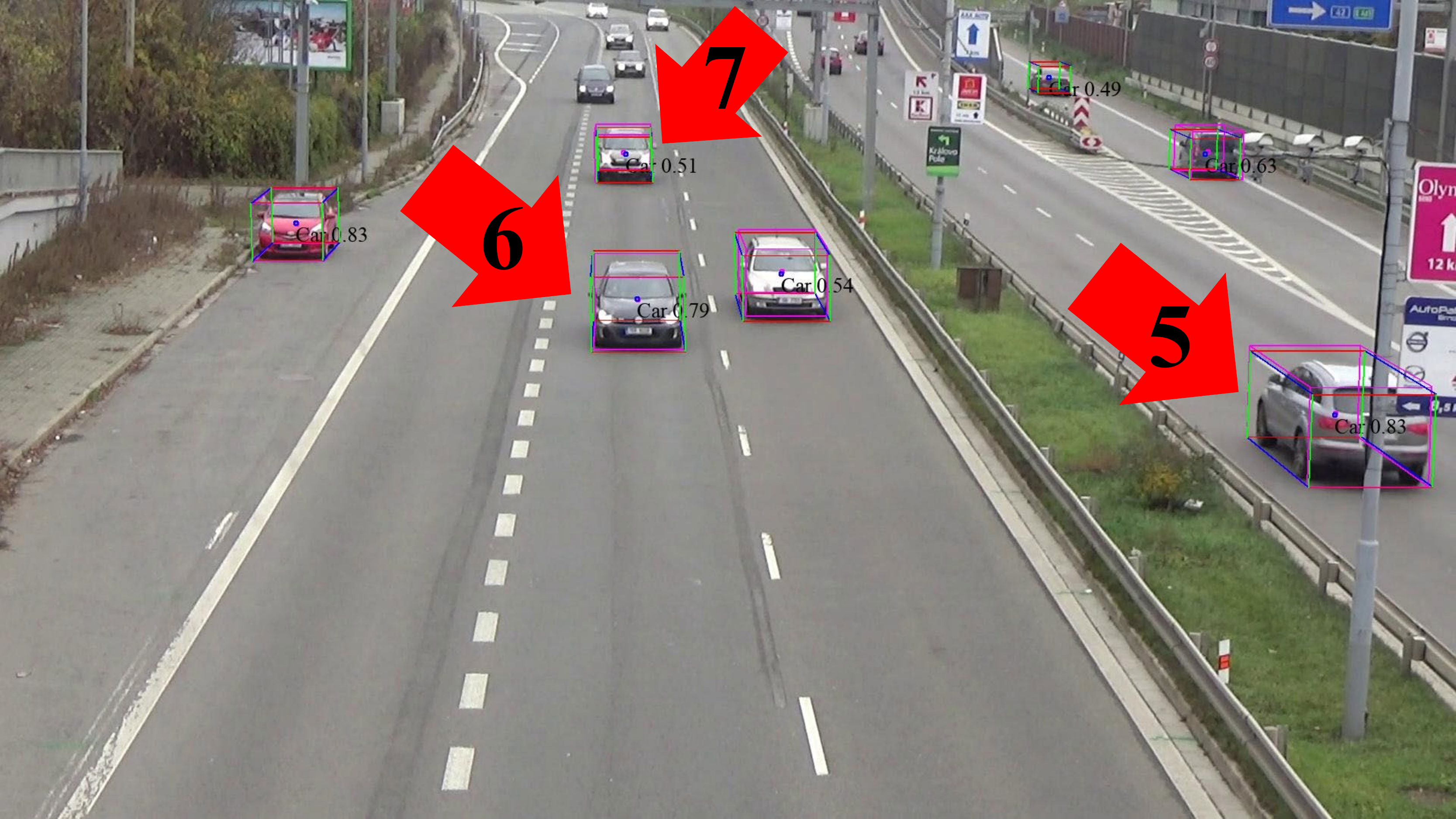}%
		\label{fig:vis_size_results_d}}
	\hfil
	\subfloat[\centering ]{\includegraphics[width=0.28\linewidth]{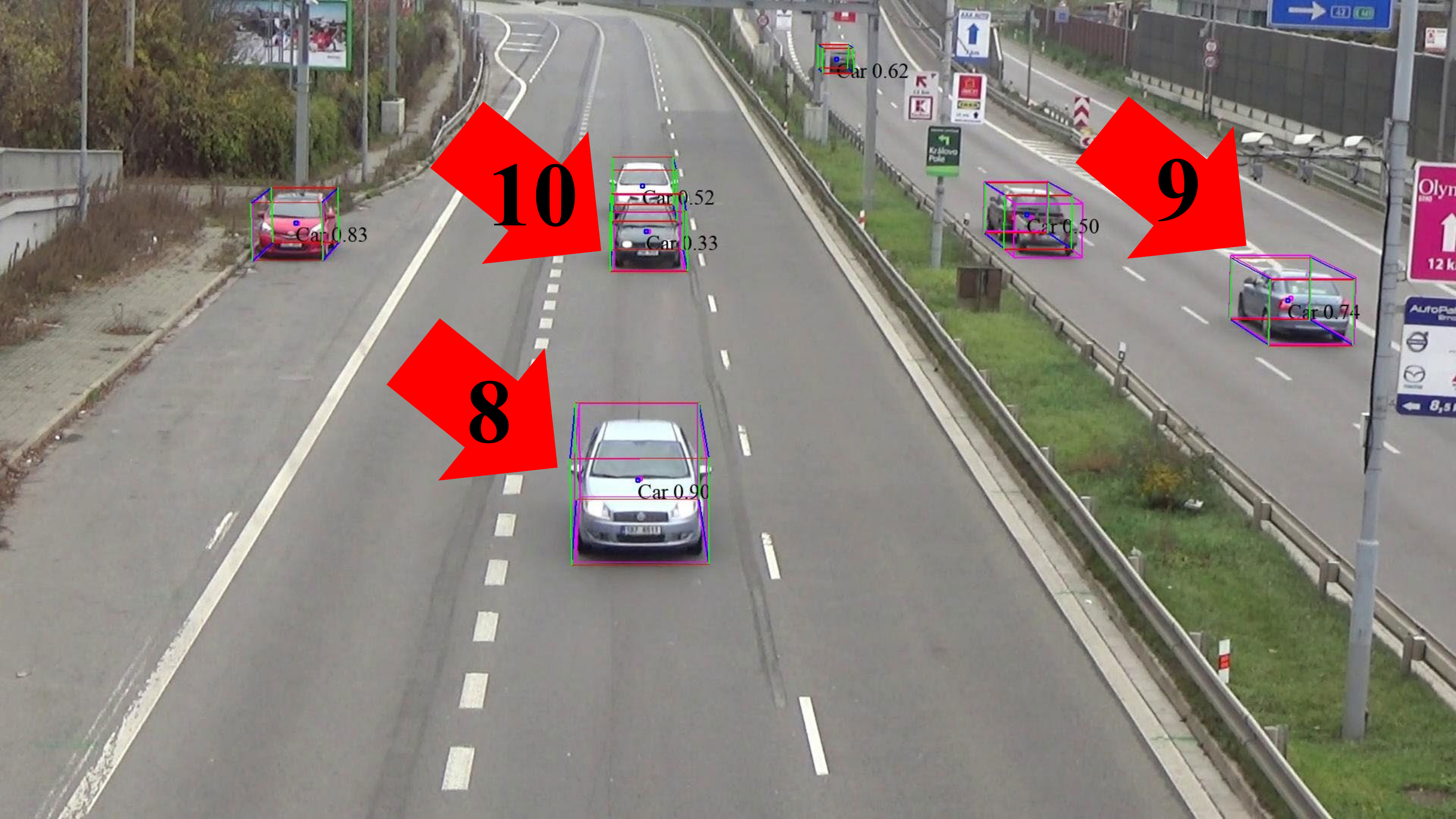}%
		\label{fig:vis_size_results_e}}
	\hfil
	\subfloat[\centering ]{\includegraphics[width=0.28\linewidth]{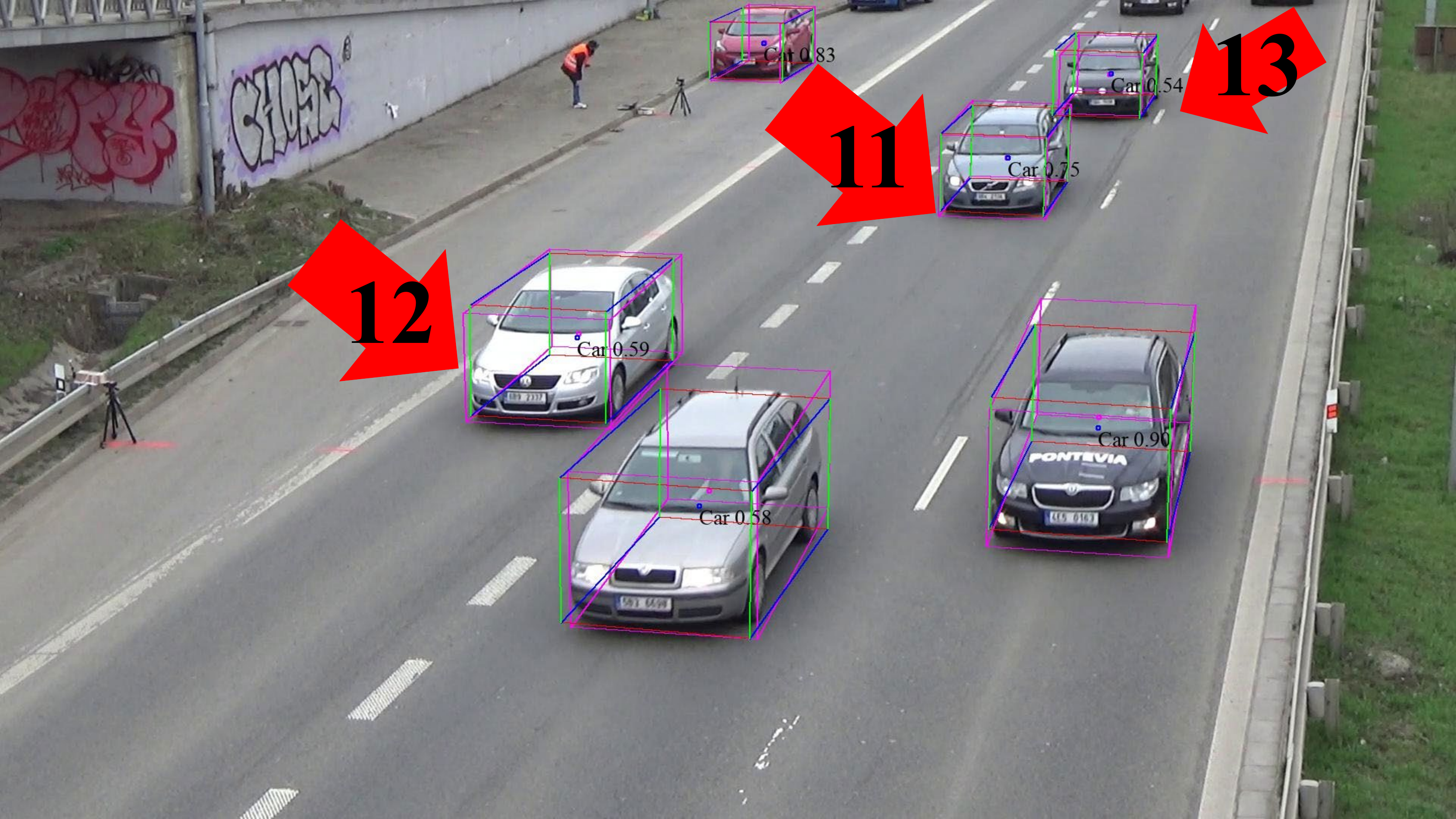}%
		\label{fig:vis_size_results_f}}
	\newline
	\subfloat[\centering ]{\includegraphics[width=0.28\linewidth]{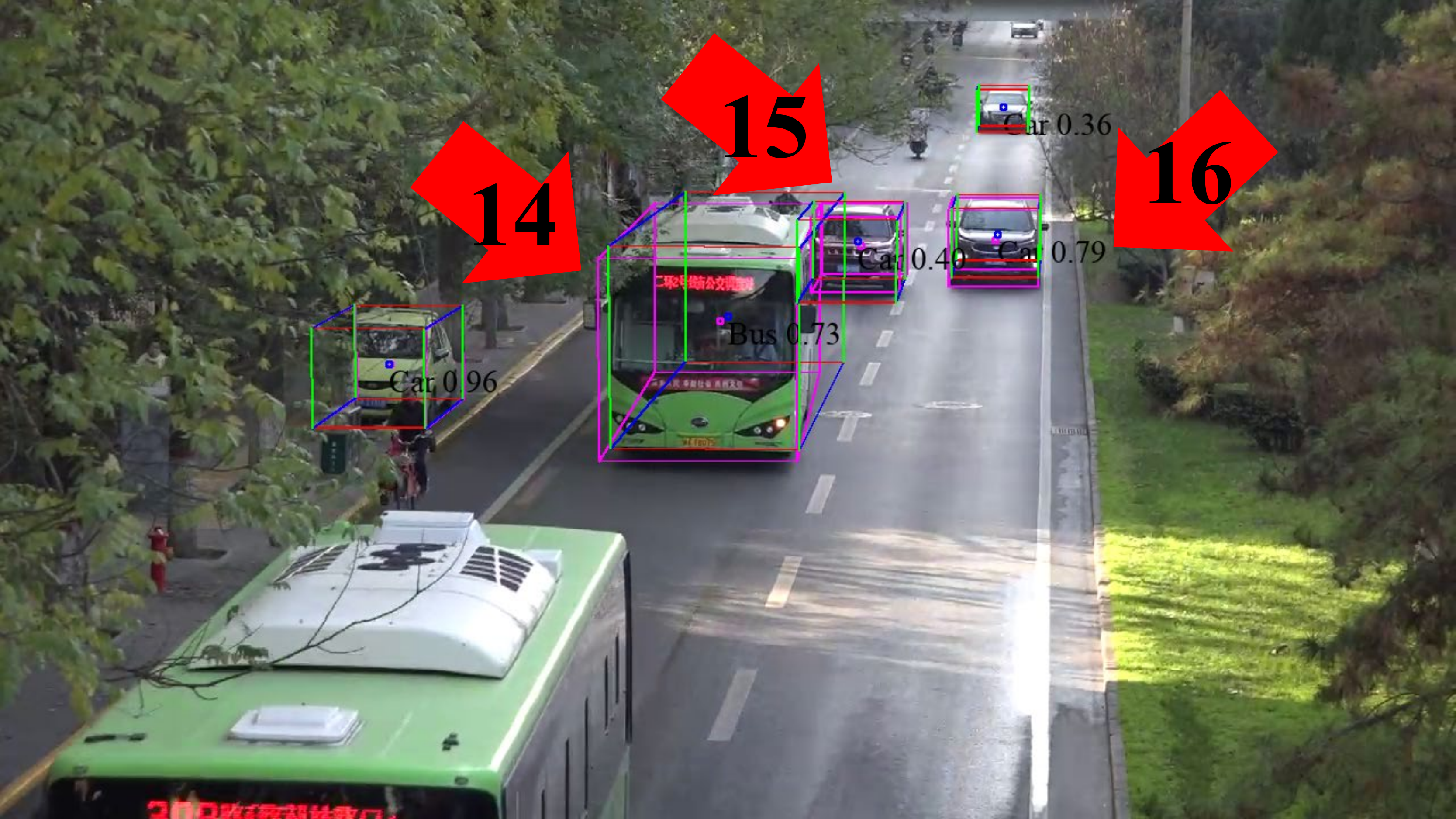}%
		\label{fig:vis_size_results_g}}
	\hfil
	\subfloat[\centering ]{\includegraphics[width=0.28\linewidth]{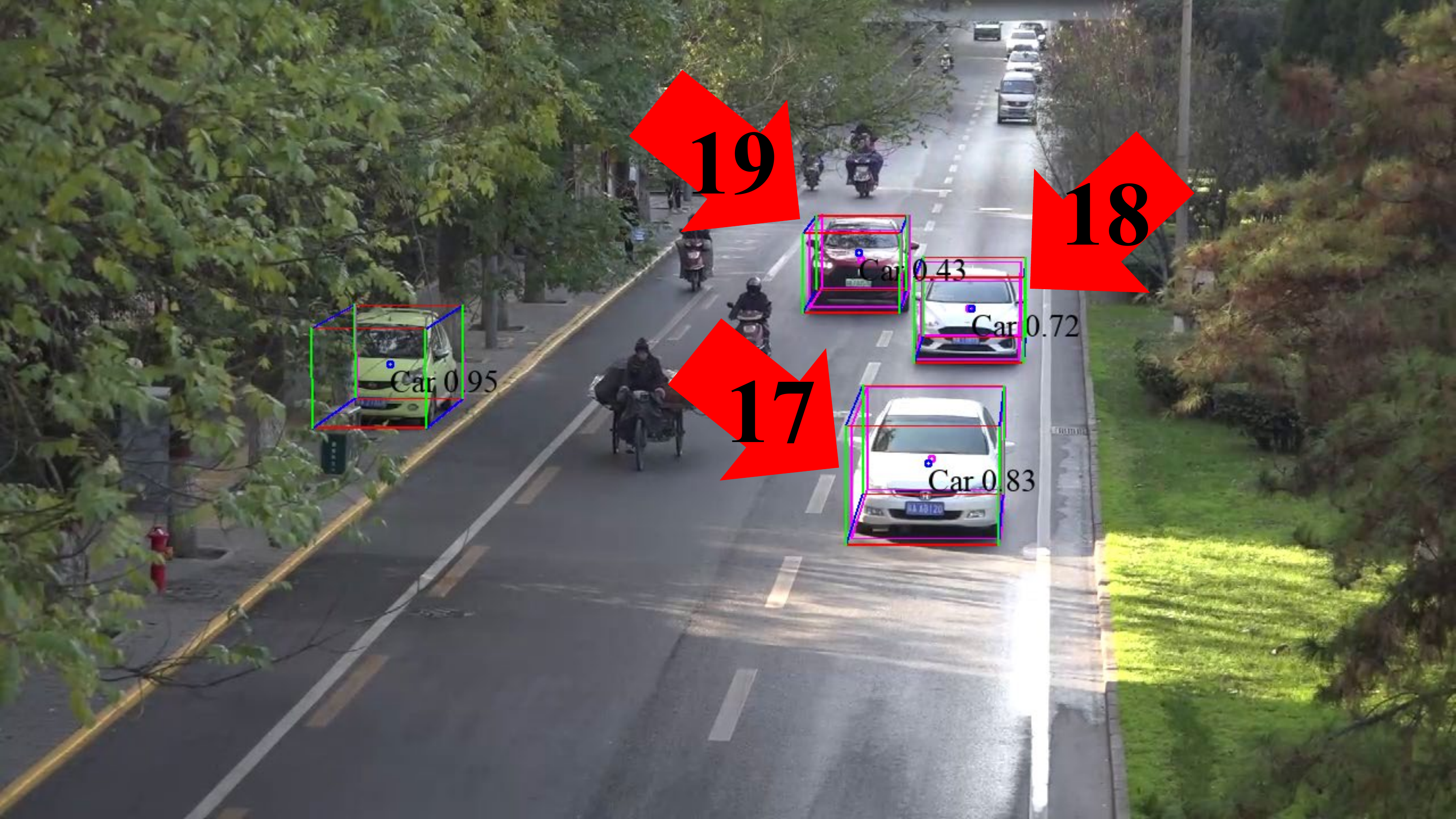}%
		\label{fig:vis_size_results_h}}
	\hfil
	\subfloat[\centering ]{\includegraphics[width=0.28\linewidth]{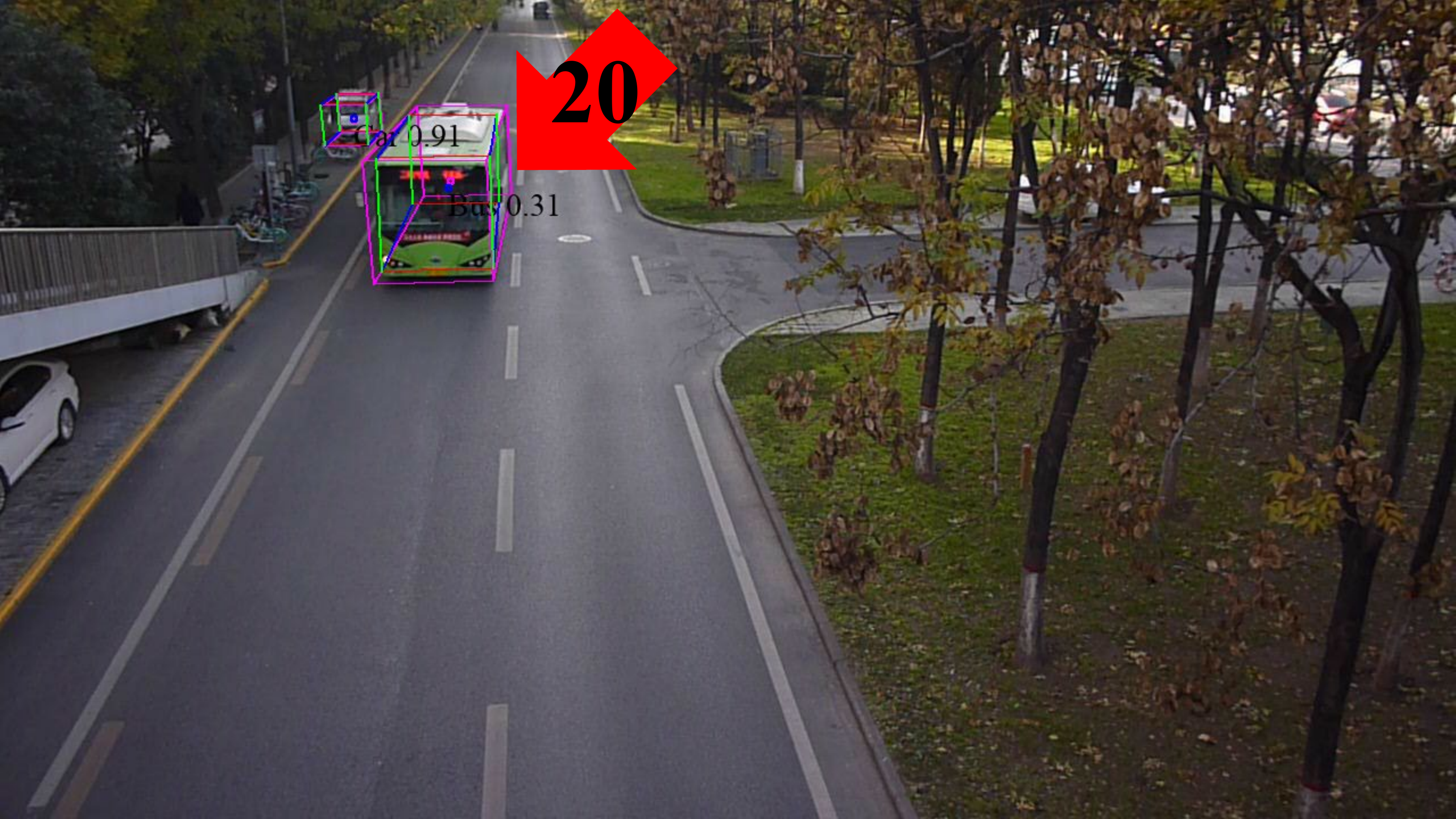}%
		\label{fig:vis_size_results_i}}
	\newline
	\caption{\leftskip=0pt \rightskip=0pt plus 0cm Visualization results of 3D vehicle dimension prediction on SVLD-3D test set. Color boxes indicate the predicted results. Pink boxes indicate the ground truth results. Numbers on the arrow pointing to vehicles are consistent with numbers in Table \ref{tab:table_size_results}.}
	\label{fig:vis_size_results}
\end{figure*}

Table \ref{tab:table_size_results} and Figure \ref{fig:vis_size_results} show 3D vehicle dimension prediction results and precision of different scenes and types in SVLD-3D test set, which is calculated by Equation \ref{equa_pdim}. It can be seen that our network can also achieve good results in 3D vehicle dimension prediction, with an average precision of 85\%.

\begin{table}[htbp]
	\centering
	\caption{3D vehicle dimension prediction results and precision on SVLD-3D test set.}
	\label{tab:table_size_results}
	\resizebox{0.7\textwidth}{!}{%
		\begin{tabular}{ccccc}
			\toprule
			Vehicle & Type & ${D_v}$          & ${\widetilde D_v}$ & Precision \\
			\midrule
			1     & Car & 3.60, 1.71, 1.37 & 3.79, 1.70, 1.27   & 0.860     \\
			2     & Car  & 3.26, 1.67, 1.31 & 3.18, 1.61, 1.25   & 0.890      \\
			3     & Car  & 4.05, 1.76, 1.40 & 3.92, 1.80, 1.40   & 0.942      \\
			4     & Car  & 4.51, 1.81, 1.47 & 4.42, 1.88, 1.46   & 0.935     \\
			5     & Car  & 4.43, 1.78, 1.37 & 4.40, 1.78, 1.48   & 0.915     \\
			6     & Car  & 4.74, 1.80, 1.46 & 4.33, 1.77, 1.40   & 0.843     \\
			7     & Car  & 4.58, 1.82, 1.45 & 4.96, 1.86, 1.54   & 0.845      \\
			8     & Car  & 4.50, 1.79, 1.40 & 4.50, 1.70, 1.36   & 0.912     \\
			9     & Car  & 3.74, 1.64, 1.27 & 4.07, 1.68, 1.30   & 0.872     \\
			10    & Car  & 4.55, 1.80, 1.42 & 4.58, 1.68, 1.40   & 0.910     \\
			11     & Car & 3.57, 1.80, 1.35 & 4.11, 1.80, 1.38   & 0.844     \\
			12     & Car  & 3.71, 1.76, 1.36 & 3.90, 1.80, 1.33   & 0.912      \\
			13     & Car  & 3.34, 1.77, 1.32 & 3.70, 1.76, 1.25   & 0.838      \\
			14     & Bus  & 12.83, 2.71, 2.75 & 12.00, 2.76, 2.82   & 0.886     \\
			15     & Car  & 4.74, 1.87, 1.48 & 4.77, 1.83, 1.53   & 0.939     \\
			16     & Car  & 5.00, 1.89, 1.48 & 4.75, 1.86, 1.56   & 0.880     \\
			17     & Car  & 4.69, 1.84, 1.44 & 4.60, 1.81, 1.37   & 0.914      \\
			18     & Car  & 4.68, 1.85, 1.43 & 4.56, 1.81, 1.34   & 0.885     \\
			19     & Car  & 4.64, 1.84, 1.45 & 4.68, 1.82, 1.50   & 0.947     \\
			20    & Bus  & 12.74, 2.68, 2.62 & 12.00, 2.76, 2.82   & 0.838     \\
			\bottomrule
		\end{tabular}
	}
\end{table}

\begin{figure*}[htbp]
	\centering
	\subfloat[\centering Scene A]{\includegraphics[width=0.32\linewidth]{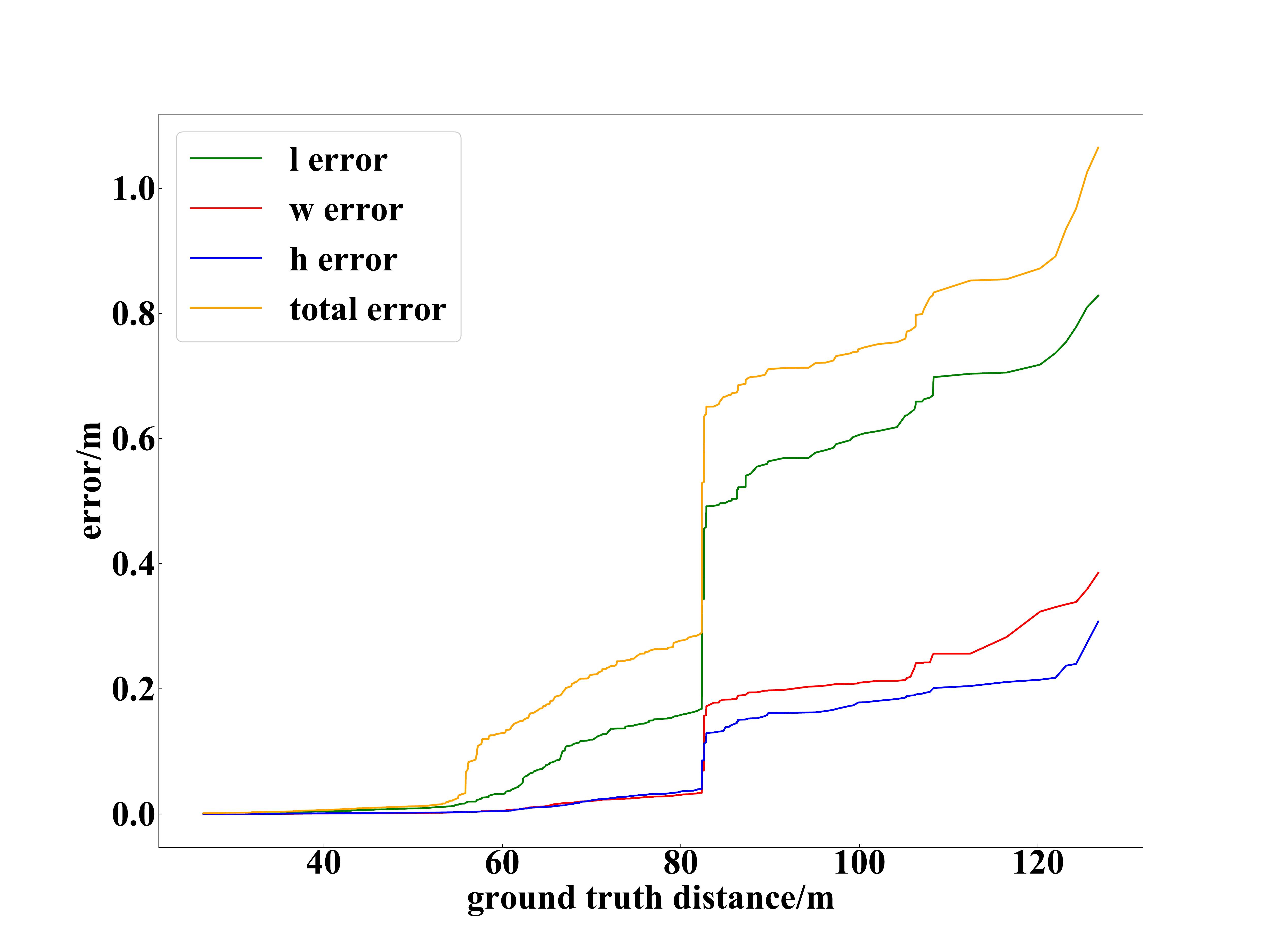}%
		\label{fig:error_size_curves_a}}
	\subfloat[\centering Scene B]{\includegraphics[width=0.32\linewidth]{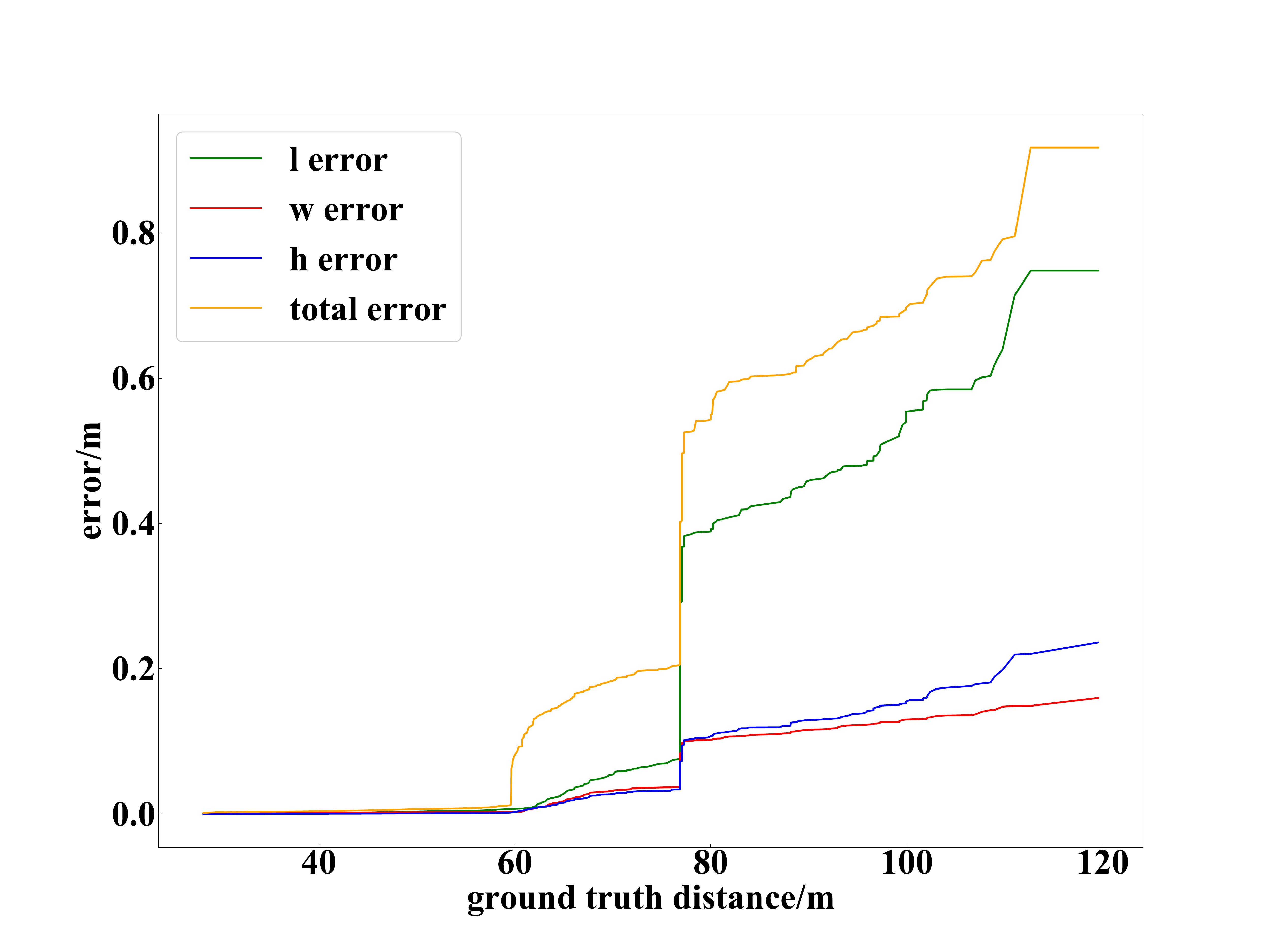}%
		\label{fig:error_size_curves_b}}
	\subfloat[\centering Scene C]{\includegraphics[width=0.32\linewidth]{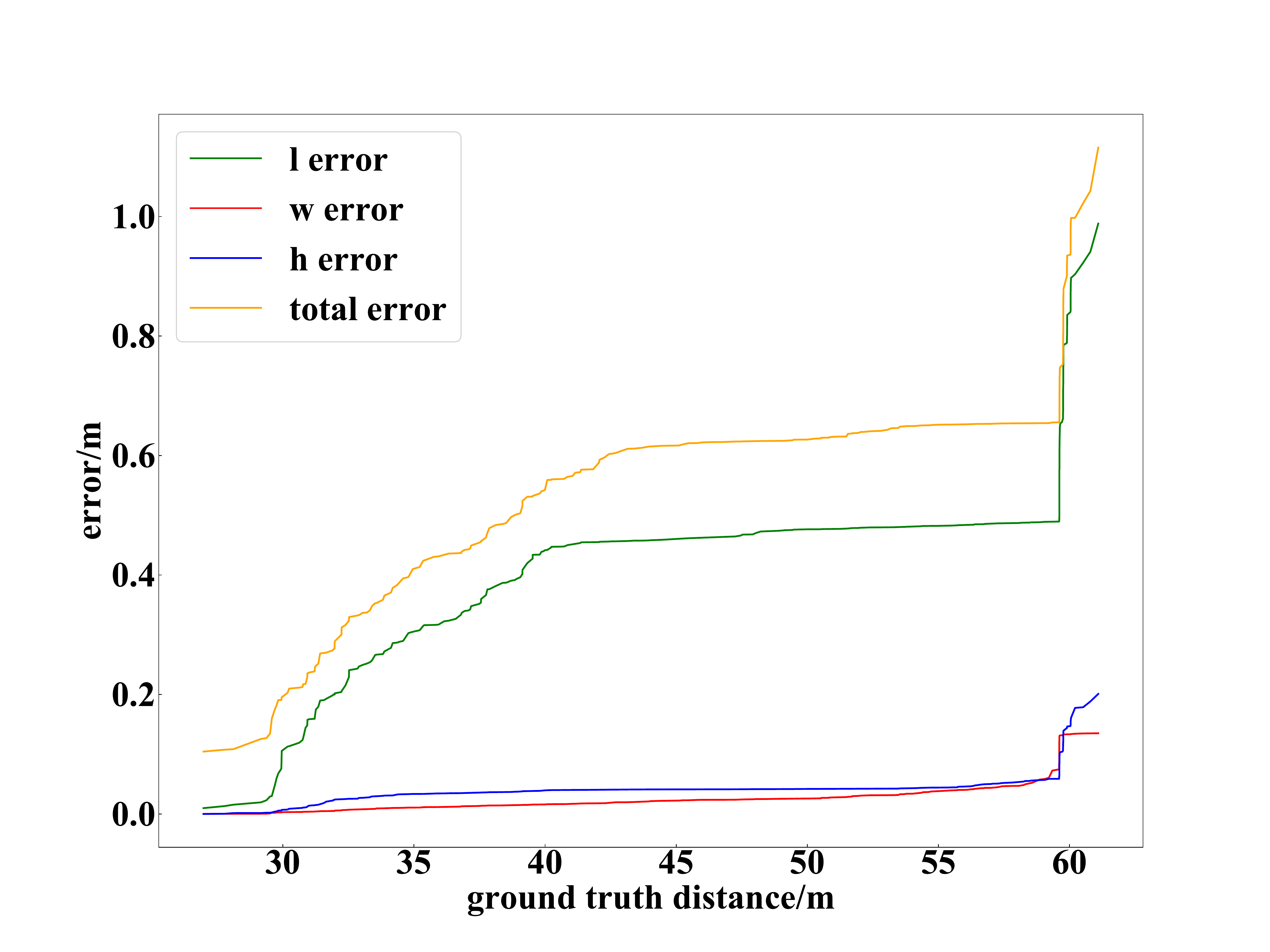}%
		\label{fig:error_size_curves_c}}
	\newline
	\begin{center}
		\subfloat[\centering Scene D]{\includegraphics[width=0.32\linewidth]{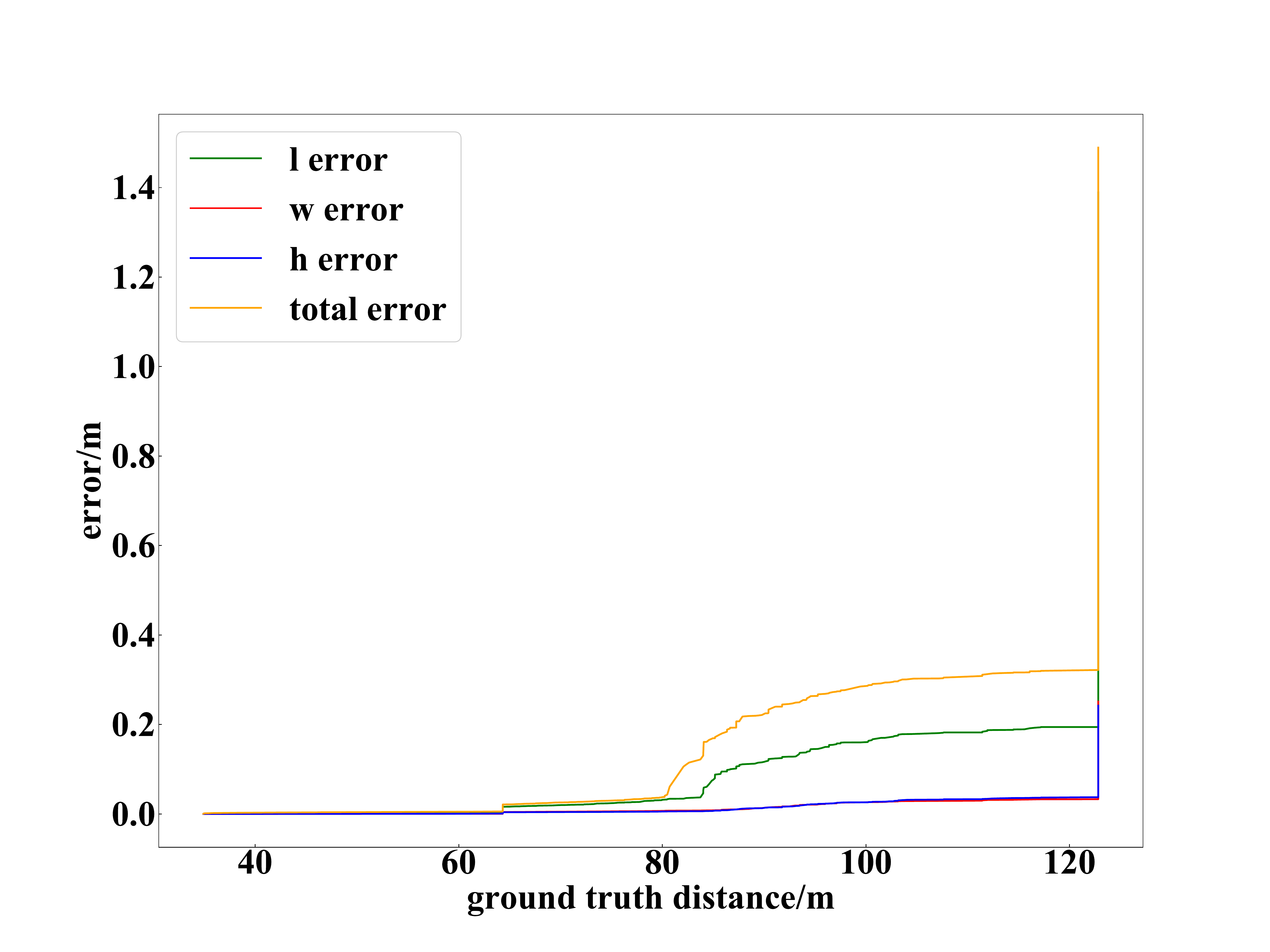}%
			\label{fig:error_size_curves_d}}
		\subfloat[\centering Scene E]{\includegraphics[width=0.32\linewidth]{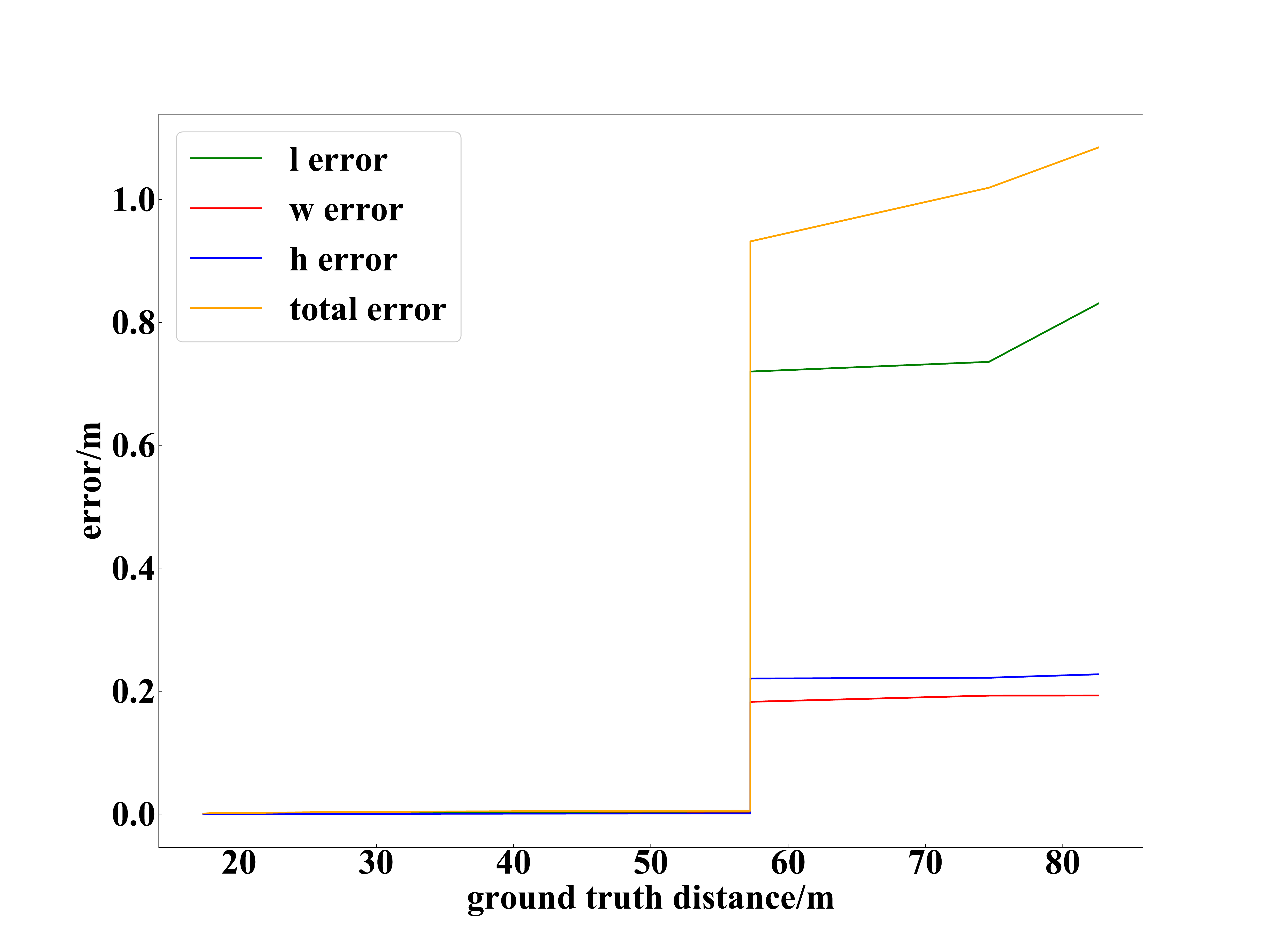}%
			\label{fig:error_size_curves_e}}
	\end{center}
	\caption{\leftskip=0pt \rightskip=0pt plus 0cm 3D vehicle dimension prediction error of length (green), width (red), height (blue), and total error (orange) according to the distances between vehicles and roadside cameras in three different scenes of SVLD-3D test set.}
	\label{fig:error_size_curves}
\end{figure*}

In Figure \ref{fig:error_size_curves}, we can see that 3D vehicle dimension error increases continuously as the distance grows between vehicles and roadside cameras of different scenes in SVLD-3D test set. The vehicle length error is the largest among length, width, and height. Since the driving directions of vehicles are usually parallel to the road direction, partial feature occlusion exists along the vehicle length direction. Therefore, the vehicle dimension prediction error in the length direction is larger than width and height.

3D vehicle dimension prediction error is calculated by Equation \ref{equa_edim}. The errors of different methods are shown in Table \ref{tab:compare_dimensions}, from which it can be seen that our network has certain advantages in 3D vehicle dimension prediction.

\begin{table}[htbp]
	\centering
	\caption{Comparison of 3D vehicle dimension prediction errors of different methods.}
	\label{tab:compare_dimensions}
	\begin{tabular}{cccc}
		\toprule
		Method        & Length/m & Width/m & Height/m \\
		\midrule
		3DOP \cite{2017stereo}    & 0.504    & 0.094   & 0.107    \\ 
		Mono3D \cite{2016mono3ddet} & 0.582    & 0.103   & 0.172    \\ 
		MonoGRNet \cite{2019monogrnet} & 0.412    & 0.084   & 0.084    \\ 
		MonoGRK \cite{2020keypoint}   & 0.403    & 0.091   & 0.101    \\
		Ours          & 0.137    & 0.031   & 0.030    \\
		\bottomrule
	\end{tabular}
\end{table}

\subsection{Ablation Study of CenterLoc3D}
\label{subsec:ablation study}

To further validate the effect of the weighted-fusion module and loss with spatial constraint embedding in our network, ablation experiments are conducted on SVLD-3D validation and test set. CenterNet \cite{2019centernet} with ResNet-50 \cite{2016resnet} is used as our baseline. We added the proposed modules one by one for validation. Ablation results are shown in Table \ref{tab:ablation_study}. For the improvement column, only $A{P_{3D}}$ of IoU threshold 0.7 on the test set is listed for comparison. In Table \ref{tab:ablation_study}, we can see that the results of adding three modules one by one outperform the former model by 6.64\%, 2.48\%, and 5.98\% in $A{P_{3D}}$ respectively. Therefore, conclusions can be summarized as follows: 1) Weighted-fusion module not only enables the network adaptive to different vehicle sizes but also increases network generalization capability. 2) Spatial constraints of camera calibration and vehicle IoU in loss help accurate 3D bounding box learning.

\begin{table*}[htbp]
	\centering
	\caption{Ablation study with different modules in CenterLoc3D on SVLD-3D validation and test set.}
	\label{tab:ablation_study}
	\resizebox{0.9\textwidth}{!}{%
		\begin{tabular}{ccccccc}
			\toprule
			\multirow{2}{*}{Model} & \multicolumn{3}{c}{Modules}                                                 & \multirow{2}{*}{$A{P_{3D}}(IOU > 0.7)$} & \multirow{2}{*}{FPS} & \multirow{2}{*}{Improvement of $A{P_{3D}}$} \\ \cline{2-4}
			& Weighted-Fusion & Reprojection & IoU &                                                          &                      &                                             \\
			\midrule
			${M_{base}}$           &                           &                              &                 & 52.52 / 36.20                                            & 46.73                 & -                                           \\
			${M_1}$                & \checkmark                          &                               &                  & 57.07 / 42.84                                            & 43.23                 & 6.64                                       \\
			${M_2}$                & \checkmark                          & \checkmark                             &                  & 68.38 / 45.32                                            & 41.31                 & 2.48                                        \\
			${M_3}$                & \checkmark                          & \checkmark                             & \checkmark                & 79.36 / 51.30                                            & 41.18                 & 5.98                                        \\
			\bottomrule
		\end{tabular}
	}
\end{table*}

\begin{figure*}[htbp]
	\centering
	\subfloat[\centering ${M_{base}}$-Scene A-0.34236]{\includegraphics[width=0.32\linewidth]{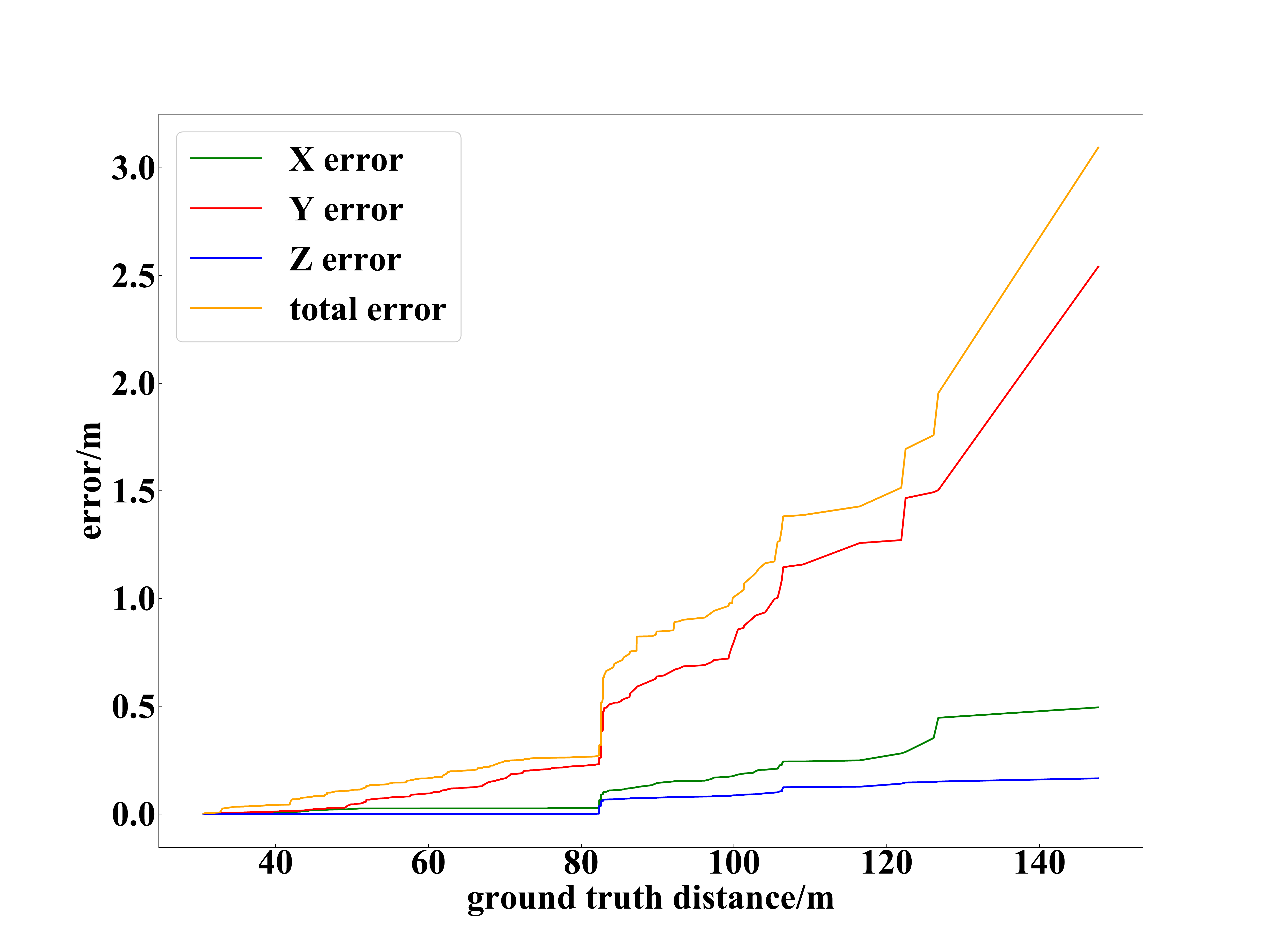}%
		\label{fig:ablation_study_loc_error_a}}
	\hfil
	\subfloat[\centering ${M_{base}}$-Scene B-0.35048]{\includegraphics[width=0.32\linewidth]{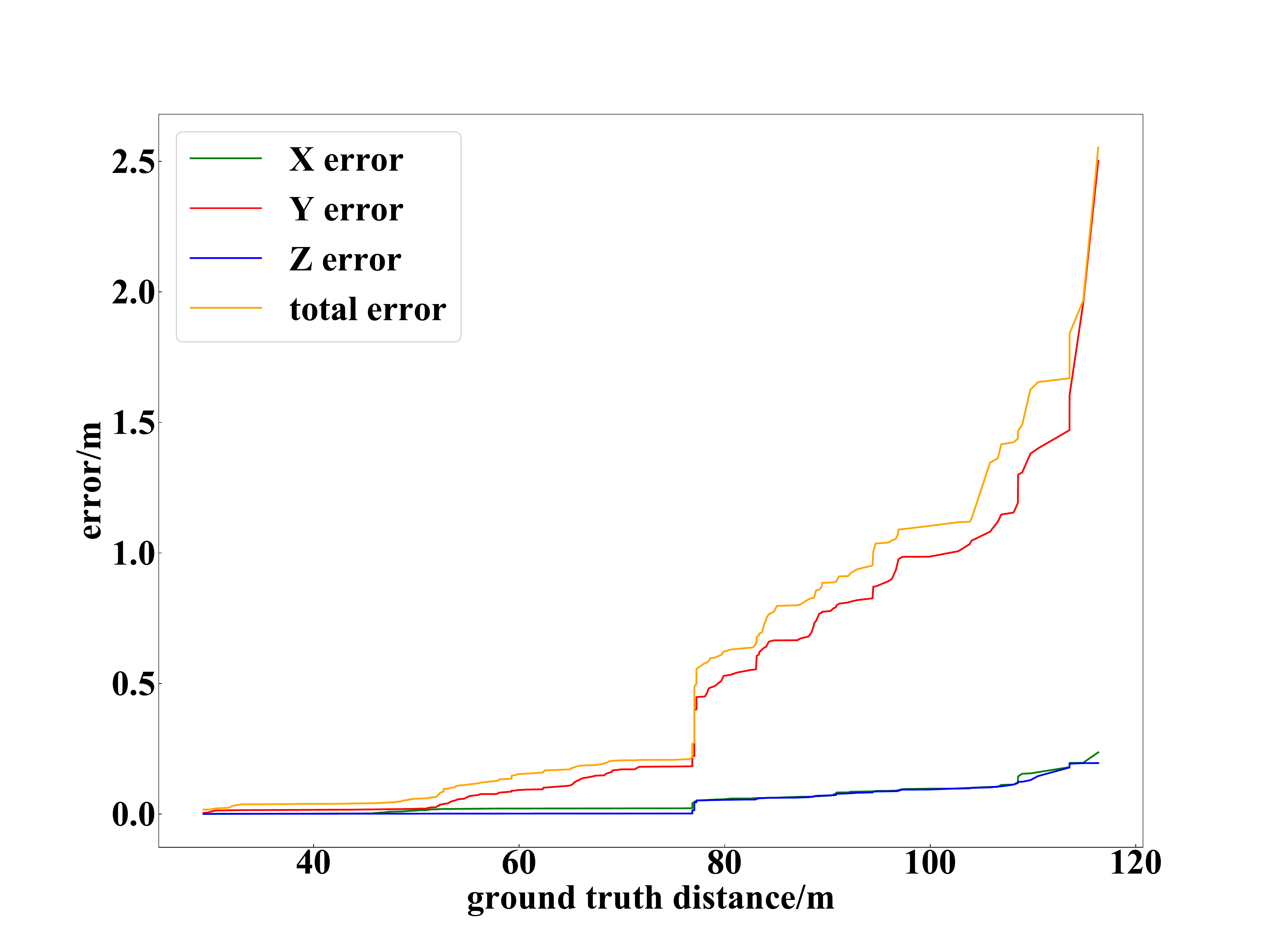}%
		\label{fig:ablation_study_loc_error_b}}
	\hfil
	\subfloat[\centering ${M_{base}}$-Scene C-0.29607]{\includegraphics[width=0.32\linewidth]{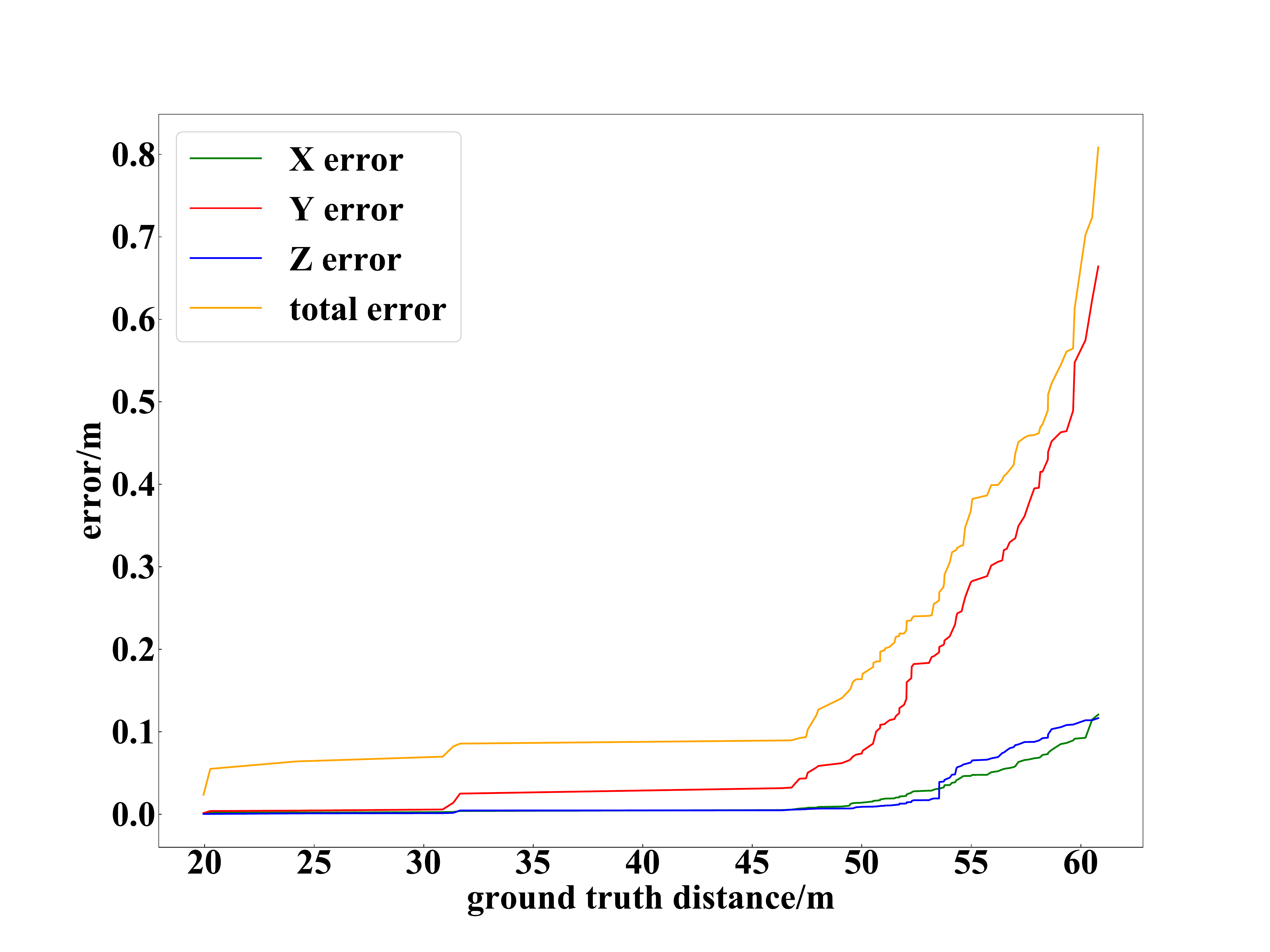}%
		\label{fig:ablation_study_loc_error_c}}
	\newline
	\subfloat[\centering ${M_1}$-Scene A-0.26307]{\includegraphics[width=0.32\linewidth]{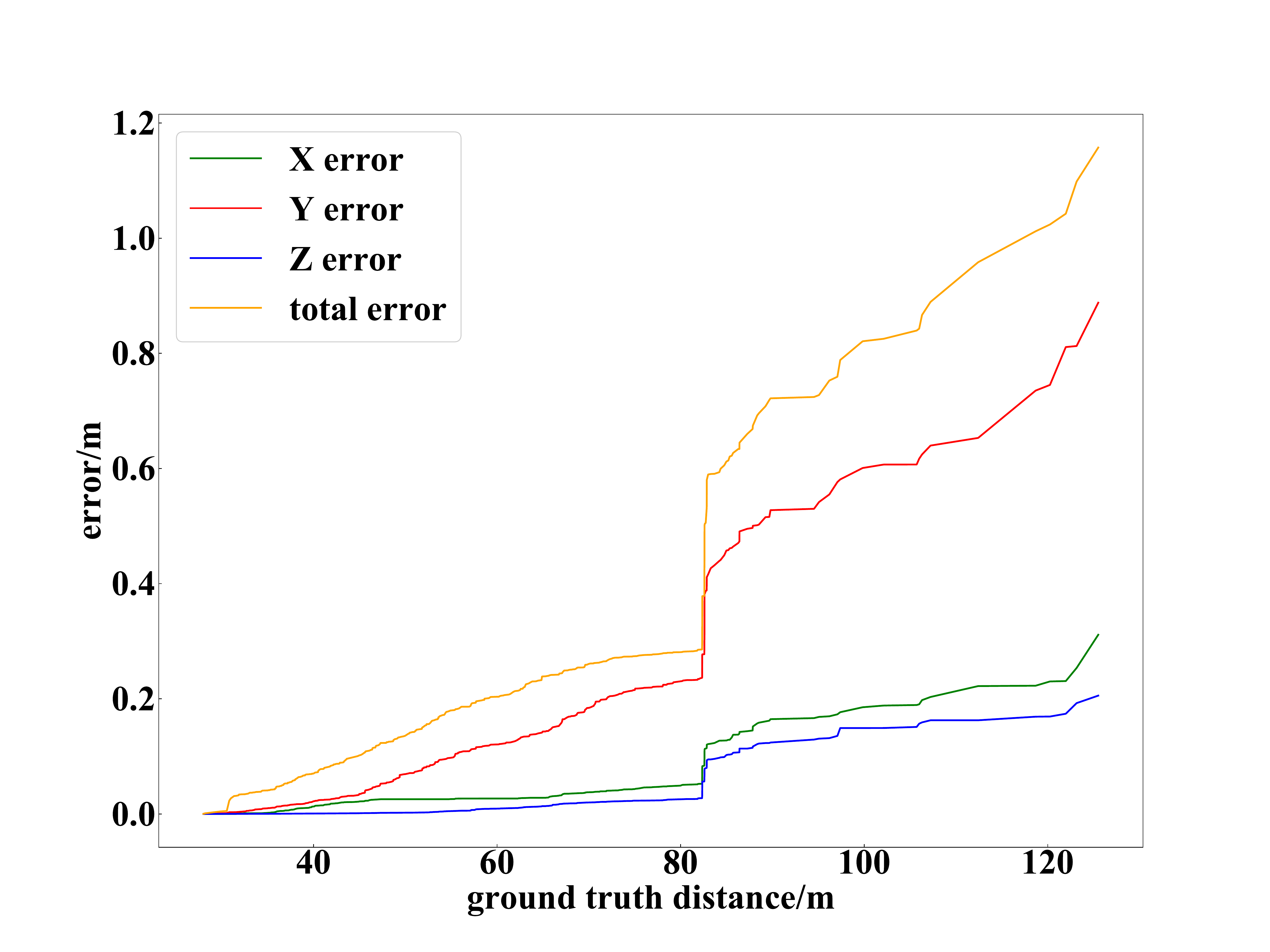}%
		\label{fig:ablation_study_loc_error_d}}
	\hfil
	\subfloat[\centering ${M_1}$-Scene B-0.26794]{\includegraphics[width=0.32\linewidth]{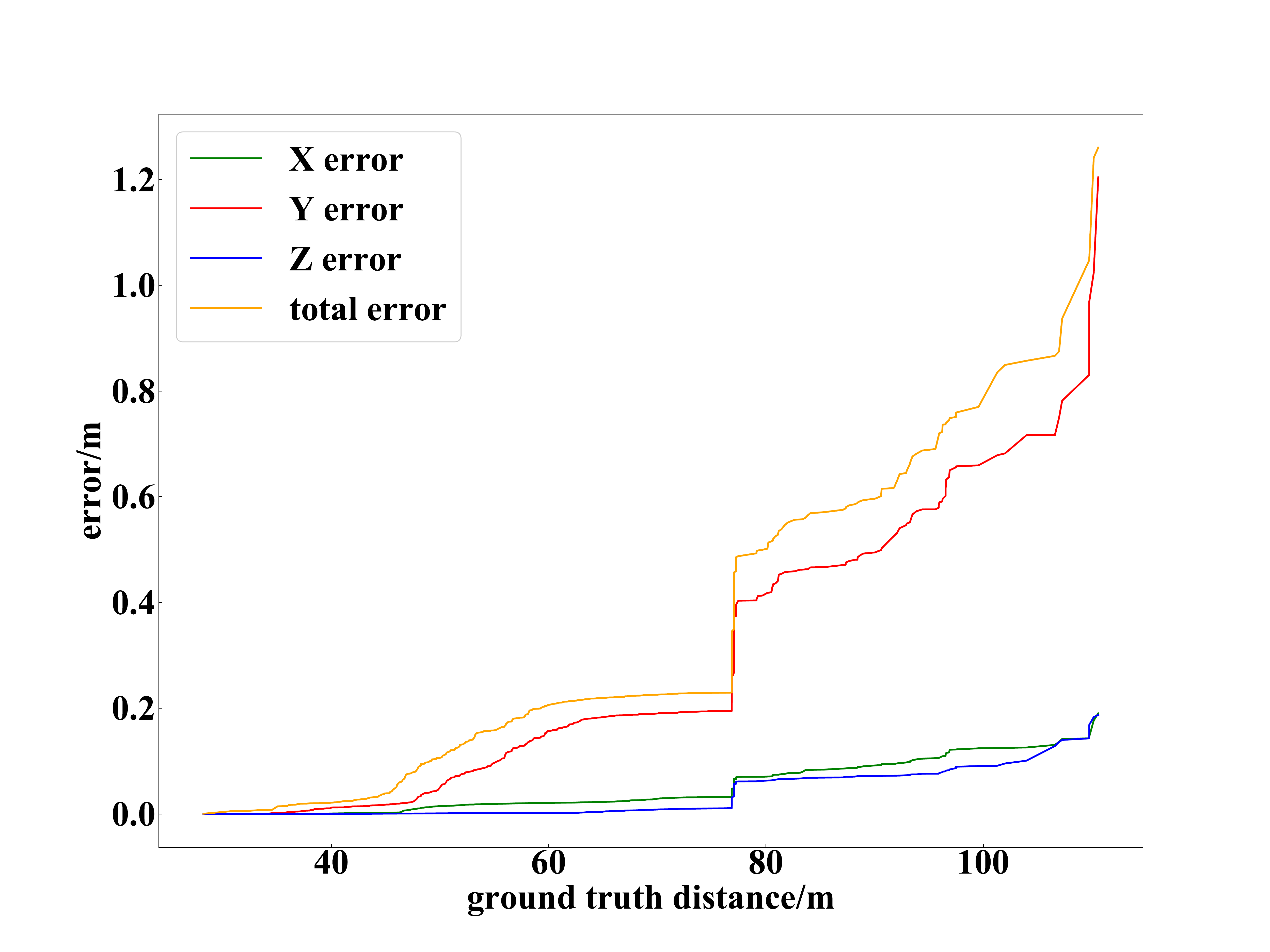}%
		\label{fig:ablation_study_loc_error_e}}
	\hfil
	\subfloat[\centering ${M_1}$-Scene C-0.25254]{\includegraphics[width=0.32\linewidth]{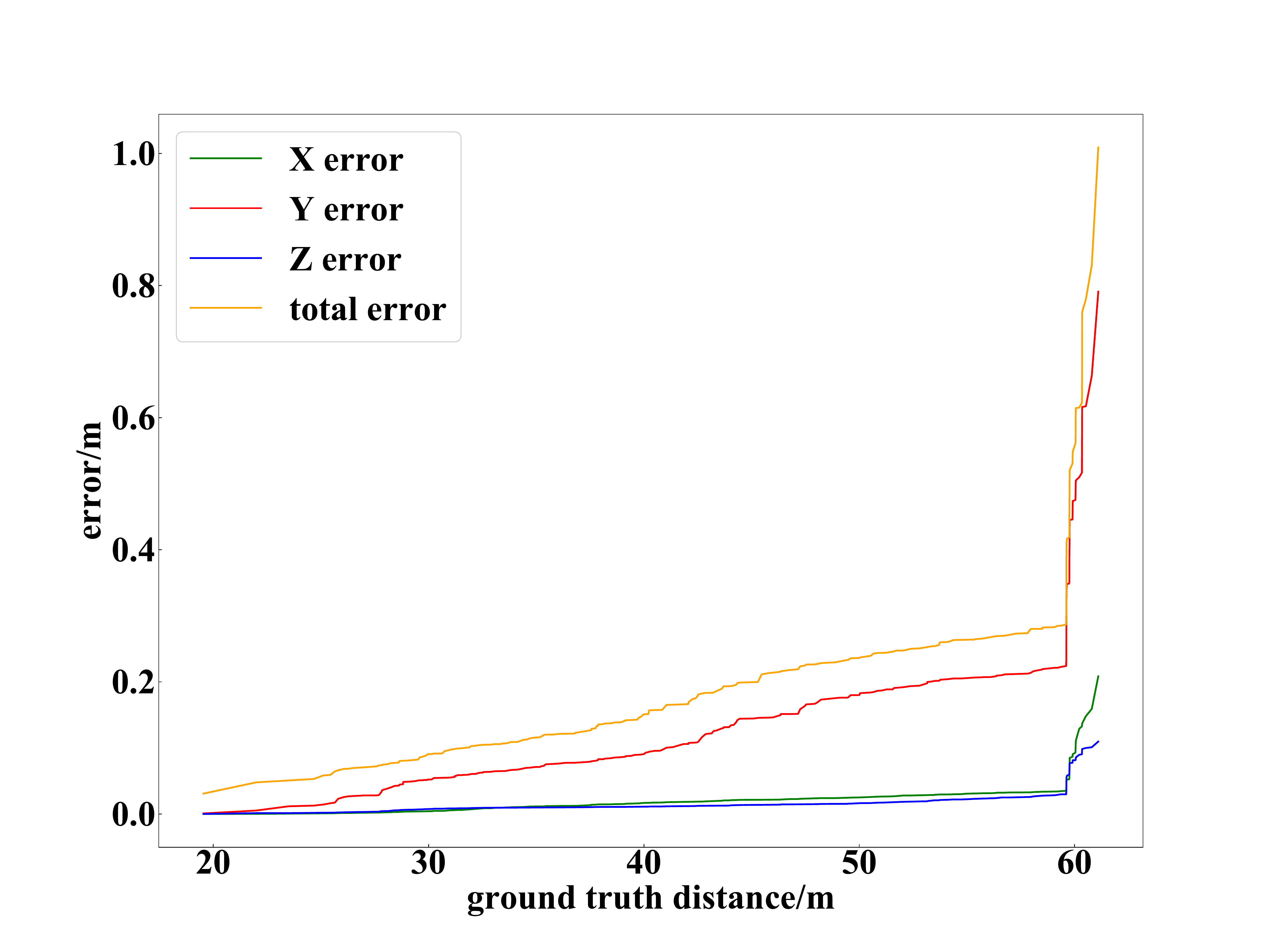}%
		\label{fig:ablation_study_loc_error_f}}
	\newline
	\subfloat[\centering ${M_2}$-Scene A-0.29456]{\includegraphics[width=0.32\linewidth]{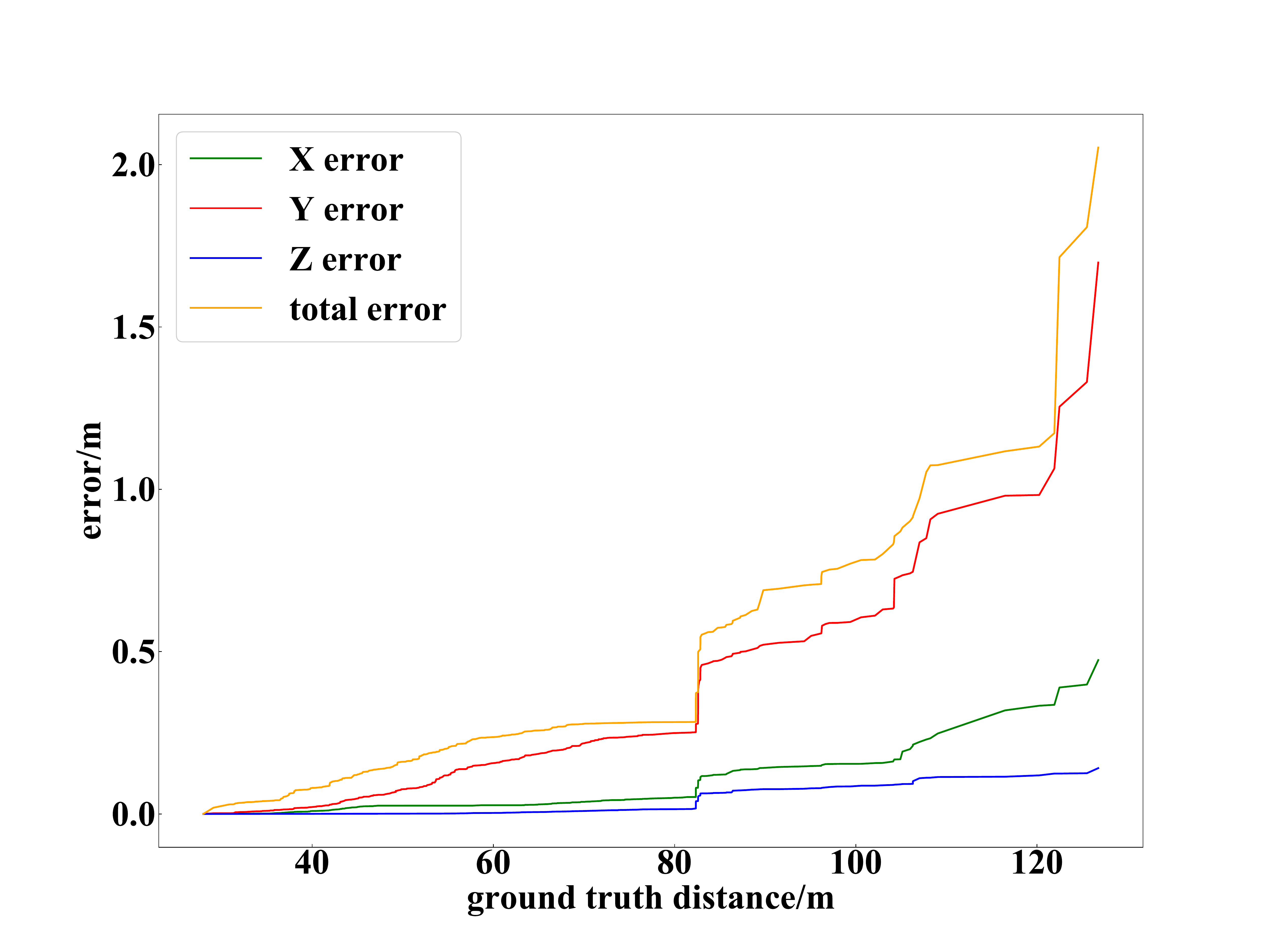}%
		\label{fig:ablation_study_loc_error_g}}
	\hfil
	\subfloat[\centering ${M_2}$-Scene B-0.30740]{\includegraphics[width=0.32\linewidth]{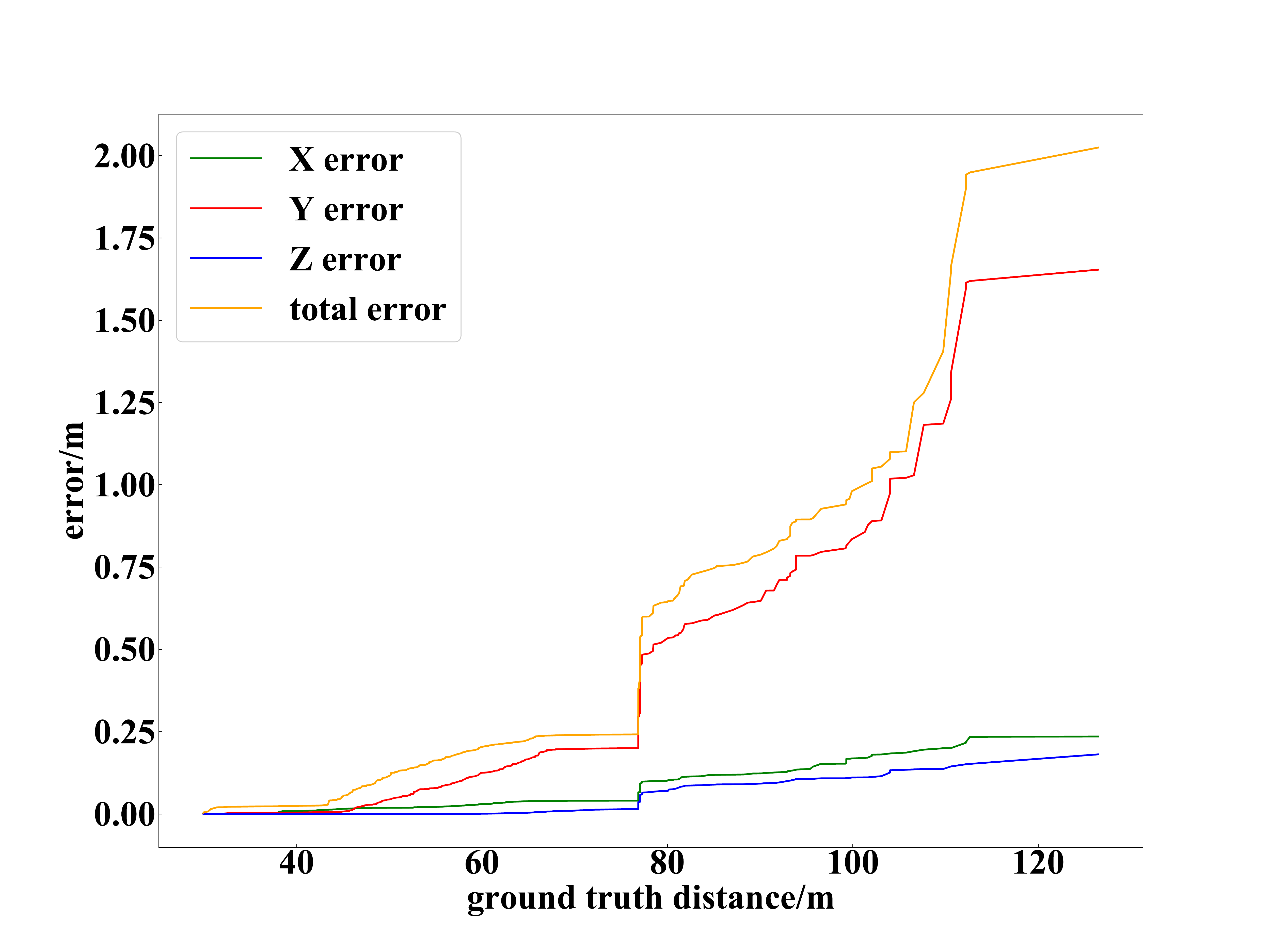}%
		\label{fig:ablation_study_loc_error_h}}
	\hfil
	\subfloat[\centering ${M_2}$-Scene C-0.24080]{\includegraphics[width=0.32\linewidth]{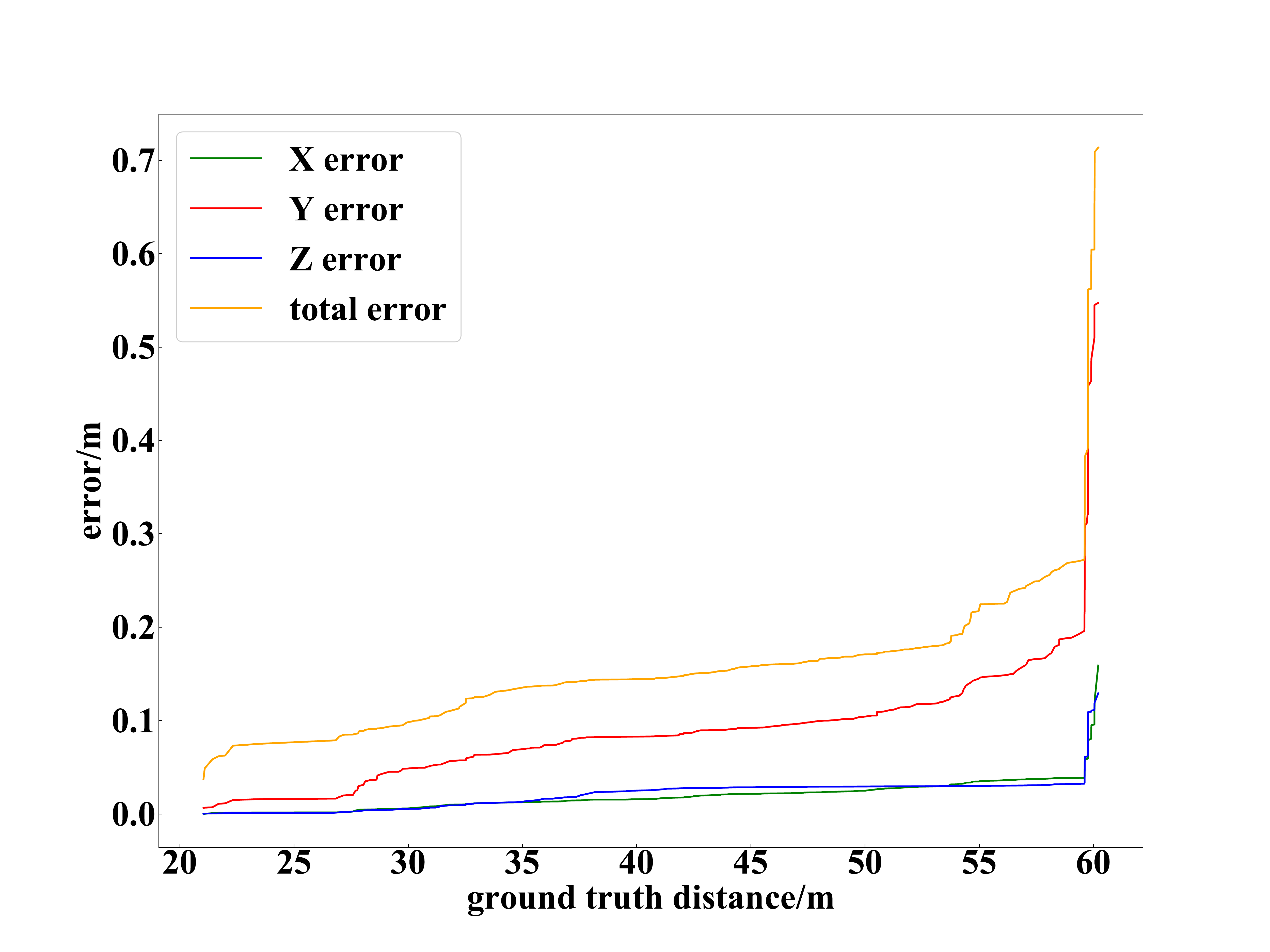}%
		\label{fig:ablation_study_loc_error_i}}
	\newline
	\subfloat[\centering ${M_3}$-Scene A-0.29415]{\includegraphics[width=0.32\linewidth]{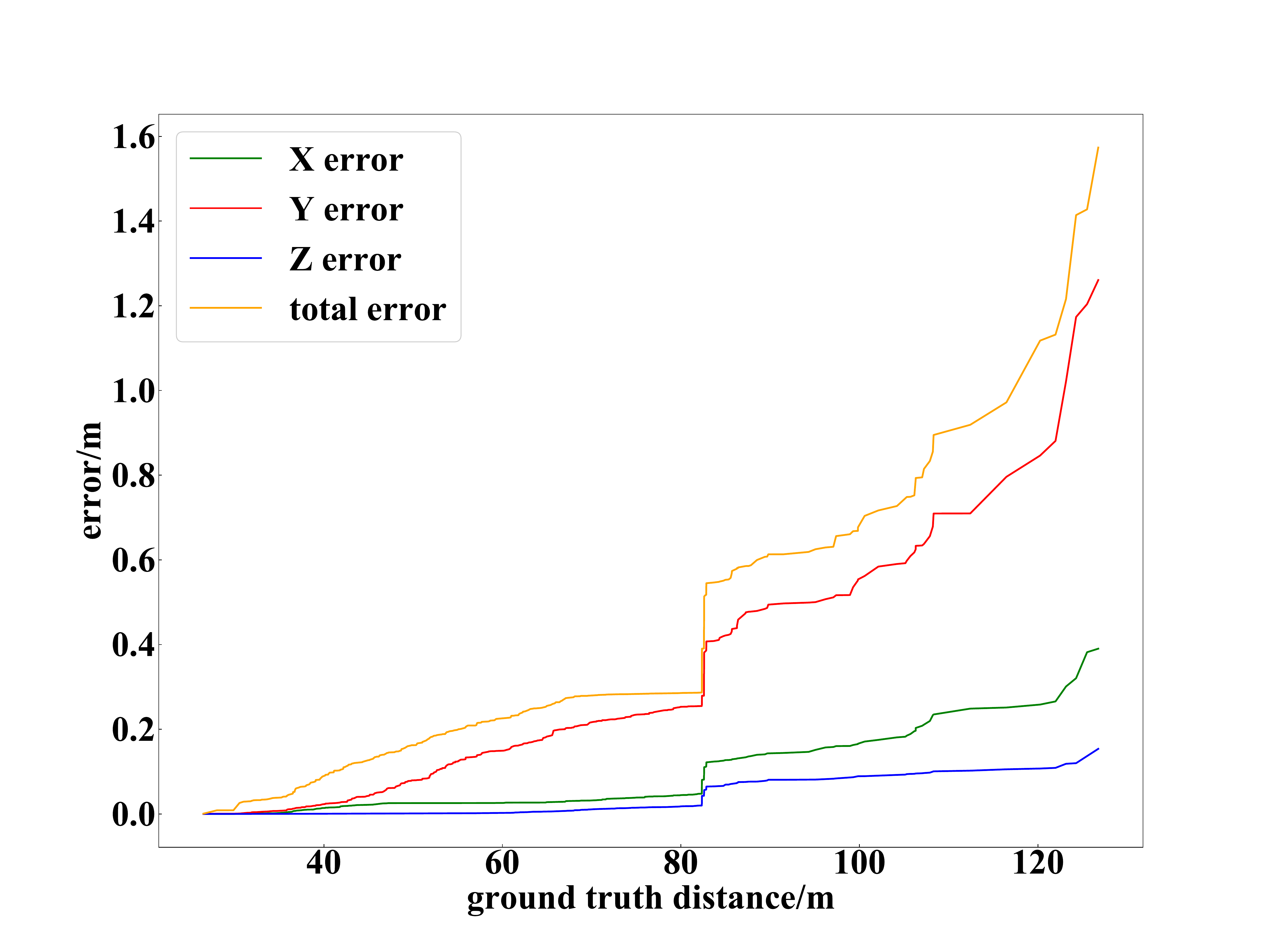}%
		\label{fig:ablation_study_loc_error_j}}
	\hfil
	\subfloat[\centering ${M_3}$-Scene B-0.29399]{\includegraphics[width=0.32\linewidth]{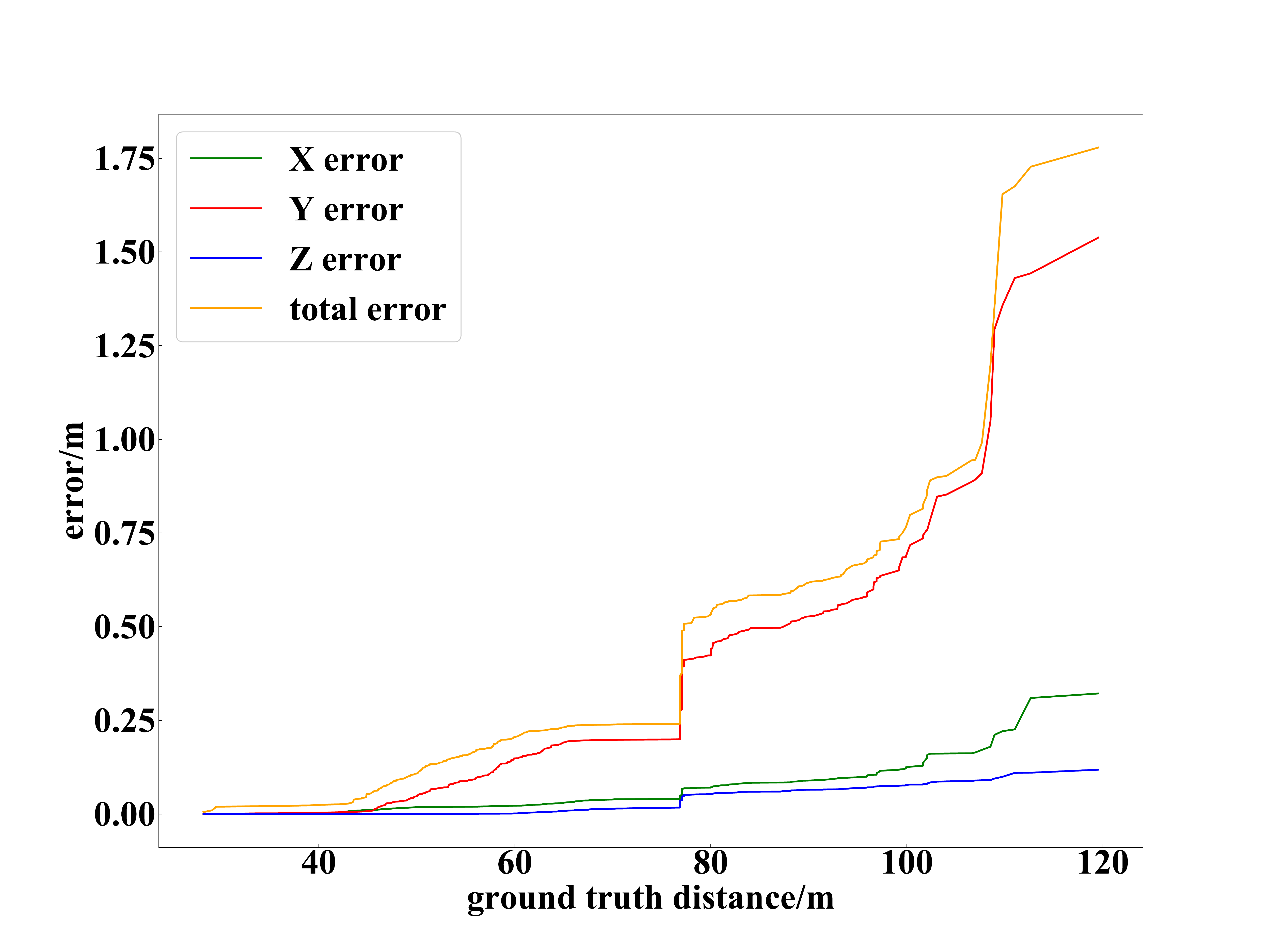}%
		\label{fig:ablation_study_loc_error_k}}
	\hfil
	\subfloat[\centering ${M_3}$-Scene C-0.22648]{\includegraphics[width=0.32\linewidth]{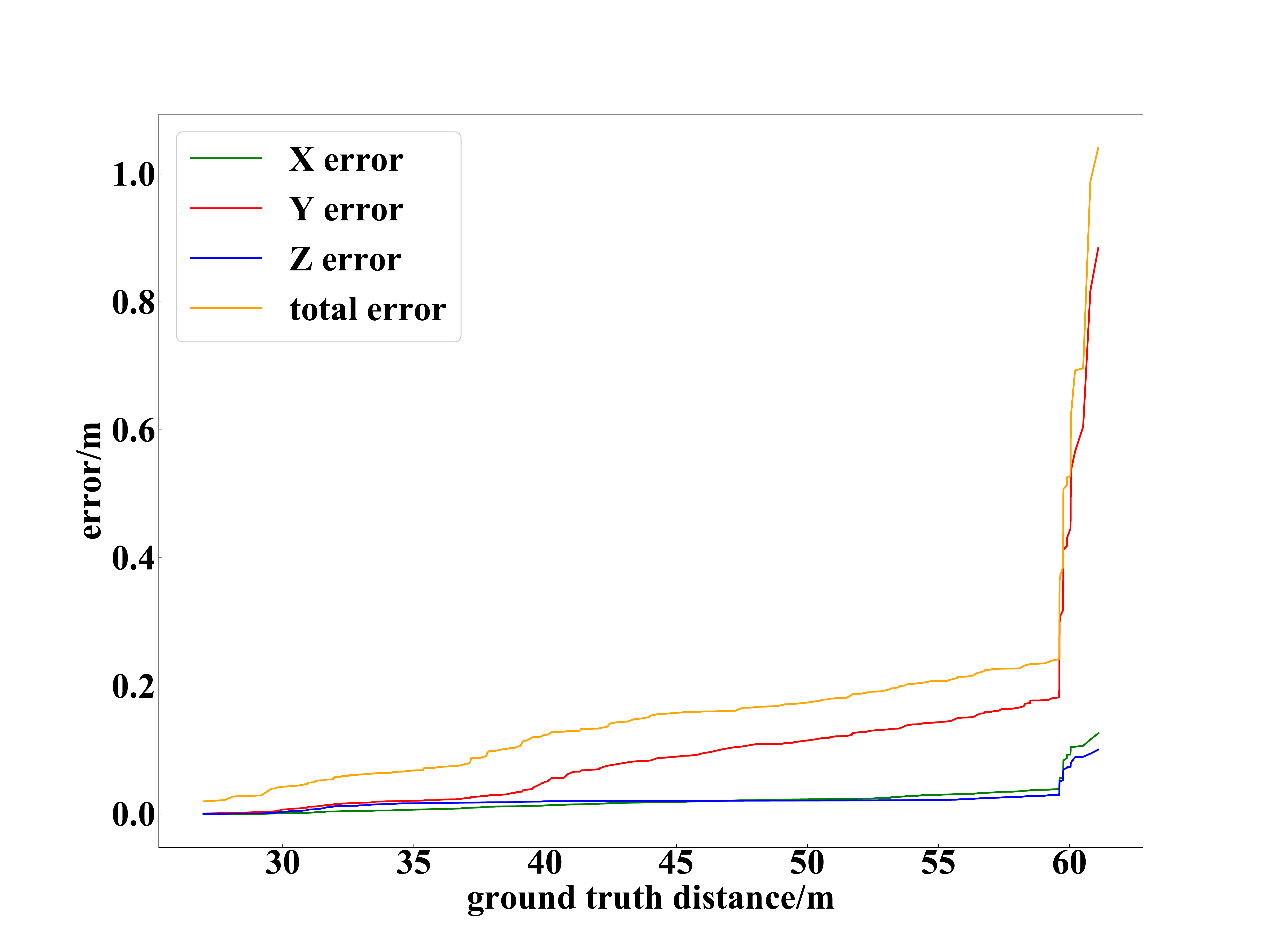}%
		\label{fig:ablation_study_loc_error_l}}
	\newline
	\caption{\leftskip=0pt \rightskip=0pt plus 0cm 3D vehicle localization error of ablation study on SVLD-3D test set. The results of ${M_{base}}$, ${M_1}$, ${M_2}$ and ${M_3}$ of different scenes are shown from the first to fourth row. 3D vehicle localization error of X-, Y-, Z-axis, and total error according to the distances between vehicles and roadside cameras on SVLD-3D test set are represented by green, red, blue, and orange lines. The number below each sub-figure indicates the average 3D vehicle localization error in meters of the corresponding scene.}
	\label{fig:ablation_study_loc_error}
\end{figure*}

Figure \ref{fig:ablation_study_loc_error} shows 3D vehicle localization error of ablation study on SVLD-3D test set. Only three views (left, middle and right) of BrnoCompSpeed scenes in SVLD-3D test set are included in this study. In this figure, it can be seen that 3D vehicle localization error from ${M_{base}}$ to ${M_3}$ in the same scene decreases gradually, which indicates that the designed modules can effectively reduce 3D vehicle localization error.

\section{Conclusion}
\label{section:conclusion}
Through experimental validation, CenterLoc3D achieves good performance on 3D vehicle detection, localization, and dimension prediction for roadside surveillance cameras, with $A{P_{3D}}$ of 51.30\%, average 3D localization precision of 98\%, average 3D dimension precision of 85\% and real-time performance with FPS of 41.18. Our contributions are as follows: 1) A 3D vehicle localization network CenterLoc3D for roadside monocular cameras is proposed, which can directly obtain 3D bounding boxes and 3D dimensions without leveraging 2D detectors. 2) A weighted-fusion strategy is proposed, which can effectively enhance feature extraction and improve generalization. 3) Loss function with constraints of camera calibration and vehicle IoU are embedded in CenterLoc3D, which reduces 3D vehicle localization error. In addition, we also propose a benchmark including a dataset, an annotation tool, and evaluation metrics, which provides a data basis for experimental validation.

However, CenterLoc3D is still needed to be improved for practical and advanced applications. When the camera pan angle is close to 0°, the features along the vehicle length direction are incomplete, leading to an increase in 3D vehicle localization error. In future work, the dataset needs to be further expanded to contain scenes with more views and more types of vehicles. In the meanwhile, more effective feature extraction modules and loss functions are needed to be designed to further improve 3D vehicle localization precision in roadside monocular traffic scenes.

\bibliography{ref}


\end{document}